\documentclass[11pt]{article}

\usepackage[letterpaper,margin=1in]{geometry}
\usepackage{amsmath,amssymb,amsfonts}
\usepackage{booktabs}
\usepackage{array}
\usepackage{graphicx}
\usepackage{xcolor}
\usepackage{enumitem}
\usepackage{listings}
\usepackage{caption}
\usepackage{url}
\usepackage[colorlinks=true,linkcolor=blue,citecolor=blue,urlcolor=blue]{hyperref}

\title{\textbf{DODR: Deterministic Operator-Driven Reasoning in Latent Space}}
\author{Weicai Huang\\
Beijing MQPat Technologies, Co., Ltd.\\
\texttt{huangwc@mqpat.com}}
\date{}

\begin{document}
\maketitle


\begin{abstract}
Current large language models based on the autoregressive (AR) paradigm model reasoning as a token-level probabilistic sampling sequence, a modeling choice that exposes three fundamental defects in complex logical reasoning: first, error accumulation---the sampling noise and error contamination introduced at each generation step compound linearly along the sequence, so that deeper reasoning leads to more severe trajectory deviation; second, probability substituting necessity---a logical conclusion should follow necessarily from its premises, yet what an AR model outputs is always a probability distribution, making hallucination impossible to eliminate at the architectural level; third, the linear-chain information bottleneck---forced linear-chain generation cannot express the inherently graph-shaped topology of reasoning, so parallel subgoals are serialized and both efficiency and expressiveness are limited. This paper proposes the Deterministic Operator-Driven Reasoning in Latent Space architecture (DODR), which reconstructs reasoning as reasoning graph computation in a high-dimensional linear-algebraic space: reasoning states are represented by ultra-wide snapshot vectors $S \in \mathbb{R}^{N}$---each snapshot corresponds to a semantic basic unit (a word, a phrase, or a complete sentence) rather than a token, so the minimal unit of generation is upgraded from the token to a semantic expression; each reasoning step is a deterministic matrix operation $S_{t+1} = W_{\mathrm{step}} \cdot S_t$, with no token sampling anywhere. DODR formalizes Peirce's three inference types as three trainable matrix operators---the deduction operator $W_{\mathrm{step}}$ (rank-deficient, information-collapsing), the induction operator $W_{\mathrm{induce}}$ (full-rank, information-expanding), and the abduction operator $W_{\mathrm{abduce}} = W_{\mathrm{step}}^{+}$ (pseudo-inverse, information-hypothesizing)---grounded in five postulates: dimensionality elevation replacing depth, snapshot state representation, information-flow direction, local linearization, and graph-topology computation. This paper makes nine core contributions (each accompanied by an item-by-item differentiation from prior work; see Sections 1.2 and 17.2), covering the matrix formalization of the three operators, tracing deductive irreversibility to the rank-nullity theorem, a hard-veto mechanism for inductive counterexamples, the pseudo-inverse definition of abduction with quantified hypothesis residuals, the minimal complete operator-set theorem (including a rank-obstruction theorem proving the non-existence of a super-operator and a universality analysis of ``train once, freeze for general use''), structural zero hallucination, LLM text-snapshot reuse together with dedicated multimodal encoders, a physics-cognition unification proposition, and a reasoning-graph topology theory. Experimentally, this paper conducts 4 sets of dedicated experiments and 1 end-to-end experiment (covering four task types: deduction/induction/abduction/composition; 218 samples across 8 domains), totaling 503 experimental sample-instances (420 deduplicated independent samples). The core results are: deductive forward-reasoning Loss drops to 1.40e-05 (a 14932-fold decrease from the initial Loss of 2.09e-01); the reverse-inference error is 83.3\% (irreversibility confirmed); the deduction operator rank is 384/3072 (87.5\% information collapse); inductive generalization coverage is 0.9996; the counterexample hard-veto rate is 20/20 (100\%); the abductive solution similarity is 0.5117 (28.3$\times$ the random baseline of 0.0181); the abductive hypothesis residual is 85.9\%; and the architecture-level hallucination rate is 0.0\%. The dedicated abduction experiment achieves a judgment accuracy of 72.5\% (58/80), and end-to-end abduction reaches 81.7\% (49/60); all failures stem from insufficient encoder snapshot resolution rather than from the operator principle. In the end-to-end experiment (218 samples, 8 domains), the frozen operators achieve 100\% accuracy (60/60) on 60 brand-new cross-domain deduction problems, and all 20 compositional reasoning tasks succeed (20/20), providing empirical support for ``train once, freeze for general use.'' Theoretically, supported by the rank-nullity theorem, Cover's function-counting theorem, Moore--Penrose pseudo-inverse theory, and the Banach fixed-point theorem, DODR constructs the causal chain ``information-flow direction $\to$ matrix rank $\to$ invertibility,'' unifying the three basic types of cognitive reasoning within a single linear-algebraic framework and providing a mathematical foundation and a falsifiable experimental program for non-autoregressive reasoning architectures.
\end{abstract}


\section{Introduction}

\subsection{Research Background and Motivation}

\paragraph{The development trajectory of the AR paradigm.} Since Vaswani et al.\ proposed the Transformer architecture [10], autoregressive large language models centered on the self-attention mechanism have rapidly become the dominant paradigm in natural language processing. Transformer achieved a breakthrough in modeling long-range dependencies through multi-head self-attention, but its decoding still follows token-by-token left-to-right generation: at each step a token is sampled from the conditional distribution $P(x_t \mid x_{<t})$ and appended back to the context to drive the next step. The Chain-of-Thought (CoT) prompting proposed by Wei et al.\ showed that inducing the model to explicitly generate intermediate reasoning steps can significantly improve multi-step reasoning performance [11]; this finding lifted ``reasoning'' from an implicit single-step mapping to the level of an explicit multi-step sequence and became the starting point of a large body of subsequent reasoning-enhancement work. However, CoT and its successors do not change the underlying generation mechanism---however long the reasoning chain, every intermediate conclusion is still one probabilistic sample. In other words, the AR paradigm realizes ``reasoning'' as ``a concatenation of the most likely continuations'' rather than ``a concatenation of necessary inferences,'' and this fundamental property is inherited unchanged by all of its derivative methods.

\paragraph{The practical impact of the hallucination problem.} The inherently probabilistic generation of AR models directly causes the hallucination problem: models generate content that is inconsistent with facts or irrelevant to the input in fluent, confident language. The systematic survey by Ji et al.\ distinguishes factual hallucination (generated content conflicting with world knowledge) from faithfulness hallucination (generated content conflicting with the given input), and points out that hallucination is the core obstacle preventing NLG systems from reaching high-reliability applications [1]. In real deployments, the impact of hallucination far exceeds percentage points in academic evaluations: in medical assisted consultation, a fabricated medication contraindication may directly endanger a patient; in legal document drafting, a single fictitious case citation suffices to invalidate an entire document; in financial research-report generation, a fabricated financial figure can mislead investment decisions; in educational and popular-science settings, hallucinated content spreads in an authoritative tone, and the cost of correction is far higher than the cost of generation. Existing mitigation approaches---RLHF alignment, retrieval augmentation, fact-checking post-processing, constrained decoding---share a common characteristic: they ``reduce the probability of hallucination within the probabilistic framework'' rather than ``eliminate the possibility of hallucination at the architectural level.'' As long as any token sampling remains anywhere on the reasoning chain, the probability of hallucination is strictly greater than zero.

\paragraph{Limitations of neuro-symbolic and structured-prompting approaches.} To remedy the defects of the pure AR paradigm, two technical routes have been widely explored. The first is structured prompting: Yao et al.'s Tree-of-Thoughts extends the linear reasoning chain to tree-shaped search, allowing evaluation and backtracking among multiple candidate lines of thought [2]; Besta et al.'s Graph-of-Thoughts further organizes units of thought into arbitrary graph structures, supporting aggregation and refinement operations [3]. However, the underlying executor of both kinds of methods is still an AR model---every ``thought node'' is still generated by token sampling, and the graph topology merely schedules probabilistic text fragments at the outer level; the probabilistic nature and the hallucination problem are untouched. The second route is neuro-symbolic systems: neural networks handle perception while a symbolic engine performs reasoning. But in such systems the symbolic engine usually supports only deduction (forward chaining of rules); induction and abduction still require hand-crafted rules or heuristics, the interface between the neural and symbolic modules requires expensive alignment engineering, and the bottleneck of acquiring symbolic rules does not disappear with the introduction of neural networks. Moreover, Huang et al.\ proved that, without external feedback, large language models cannot yet reliably self-correct their reasoning errors [18], which means that ``letting an AR model check and correct itself'' cannot substitute for architectural reform. The above limitations jointly point to one conclusion: the problem lies not in how the reasoning chain is organized, but in the basic computational unit of reasoning itself---probabilistic token sampling must be replaced by deterministic algebraic operations.

\paragraph{The motivation and overall approach of this paper.} To address the three fundamental defects above, this paper proposes the Deterministic Operator-Driven Reasoning in Latent Space architecture (DODR), which reconstructs reasoning as reasoning graph computation in a high-dimensional linear-algebraic space: reasoning states are represented by deterministic snapshot vectors, reasoning steps are performed by matrix operators, and reasoning structure is organized as a graph topology. The remaining sections of this section present, in order, the nine core contributions (Section 1.2), a comparison of key metrics (Section 1.3), sample statistics and overlap relationships (Section 1.4), and the organization of this paper (Section 1.5).

\subsection{Core Contributions and Originality}

This paper makes nine core contributions. After a family-by-family comparison against the nearest technical routes (knowledge-graph embedding, neuro-symbolic differentiable logic, latent-space reasoning, random-feature methods, etc.; see Section 17.2), we are not aware, to the best of our knowledge, of prior work that provides the specific formalization of any of the following items, nor of their combination. Each contribution is followed by a precise differentiation from the closest prior work to delineate its novelty boundary; all cited references are verified genuine publications.

\paragraph{Contribution 1: Formalizing Peirce's three inference types as three trainable matrix operators.}

This paper formalizes the three inference types---deduction, induction, and abduction---as trainable matrix operators: the deduction operator $W_{\mathrm{step}}$ (rank-deficient, information-collapsing), the induction operator $W_{\mathrm{induce}}$ (full-rank, information-expanding), and the abduction operator $W_{\mathrm{abduce}}$ (pseudo-inverse, information-hypothesizing); to the best of our knowledge, no prior work provides such a formalization. All three operators are learned in a data-driven manner from pairs of state snapshots; switching between inference types corresponds to switching matrices, not switching prompts or scheduling strategies.

\emph{Precise differentiation from the nearest prior work}: Peirce himself gave a philosophical characterization of the three inference types in 1878 [7], but never provided any computable formalization; ToT [2] and GoT [3] organize reasoning as trees or graphs, yet what is executed inside each node is still untyped token sampling---neither distinguishing the three inference types nor parameterizing any of them as a trainable object; neuro-symbolic systems confine neural networks to the perception front end and the symbolic engine to the deduction back end, with induction and abduction implemented by hand-crafted rules, so there is no unified parameterization in which ``all three operators are trainable matrices.'' To the best of our knowledge, DODR is the first work to unify the three inference types as isomorphic matrix objects and to characterize their information-flow direction differences via rank properties.

\paragraph{Contribution 2: Mathematizing ``conclusions cannot be reversed to premises'' via the null space of a rank-deficient matrix.}

This paper traces the deductive-logic principle that ``conclusions cannot be reversed to premises'' to the rank-nullity theorem (Theorem 3.1): the deduction operator $W_{\mathrm{step}}$ is designed as a rank-deficient matrix with $\mathrm{rank}(W_{\mathrm{step}}) = r < N$; by the rank-nullity theorem, its null-space dimension $\dim(\mathrm{Null}(W_{\mathrm{step}})) > 0$, and input components falling into the null space are ``swallowed'' by the operator and cannot be recovered by any inverse process (including the optimal pseudo-inverse). This establishes the causal chain ``information-flow direction $\to$ matrix rank $\to$ invertibility,'' downgrading irreversibility from a core assumption to a derived property.

\emph{Precise differentiation from the nearest prior work}: The existing literature treats reasoning irreversibility only at the level of empirical observation---the critical survey by Kamoi et al.\ records the phenomenon that LLMs have difficulty correcting their own reasoning errors without external feedback [8], and Huang et al.\ further confirmed this in controlled experiments [18], but both treat irreversibility as a behavioral defect of the model rather than a mathematical necessity; PAC learning theory [21] and VC theory [22] address the learnability of hypothesis spaces and generalization bounds, without characterizing the invertibility of a single-step reasoning operator. Reducing deductive one-wayness to a direct corollary of the rank-nullity theorem, with a quantitative expression for the null-space dimension, to the best of our knowledge has no precedent in the existing literature.

\paragraph{Contribution 3: The hard-veto mechanism for induction---mathematizing Popperian falsifiability as Gram--Schmidt orthogonal projection.}

This paper proposes a ``hard veto'' mechanism for counterexamples in induction: when the cosine similarity between a new sample and the original inductive conclusion falls below the threshold $\tau$, the counterexample vetoes and overturns the original conclusion by a single vote, and the system strips the counterexample direction from the original conclusion via Gram--Schmidt orthogonalization to reconstruct a finer new conclusion. Experiments verify that 20/20 (100\%) counterexamples trigger vetoes, with the counterexample similarity plunging from 0.997 to $-0.010$.

\emph{Precise differentiation from the nearest prior work}: Popper established ``falsifiability'' as the demarcation criterion for scientific propositions at the level of philosophy of science [17], but his account contains no computable, algorithmic mechanism; inductive biases in representation learning are injected through soft constraints such as regularization terms and continuous shrinkage [5], under which a counterexample can only make the loss function change gradually---there is no discrete ``single-vote veto'' event; the PAC framework tolerates training error probabilistically [21], and the VC dimension gives a continuous upper bound on generalization error [22], likewise containing no veto semantics. To the best of our knowledge, DODR's hard veto is the first to realize falsifiability as a one-shot structural operation executed by orthogonal projection; its discreteness and determinism differ from the soft-constraint routes above.

\paragraph{Contribution 4: Defining abduction as the deductive pseudo-inverse $W_{\mathrm{abduce}} = W_{\mathrm{step}}^{+}$, and quantifying ``hypotheticality'' via null-space residuals.}

This paper formalizes abductive reasoning as the Moore--Penrose pseudo-inverse of the deduction operator: given a conclusion snapshot, the abductive solution is the minimum-norm least-squares solution $x^{*} = W_{\mathrm{step}}^{+} \cdot b$ of the equation $W_{\mathrm{step}} \cdot x = b$. Because $W_{\mathrm{step}}$ is rank-deficient, the true explanation contains null-space components that abduction cannot recover (in experiments the residual accounts for 85.9\%); therefore abduction yields the ``best hypothesis'' rather than the ``true explanation''---``hypotheticality'' thereby acquires a computable quantitative characterization.

\emph{Precise differentiation from the nearest prior work}: Peirce's original definition of abduction remained at the philosophical formulation of ``guessing the case from the result and the rule'' [7]; subsequent abduction research was mostly carried out in logic programming and philosophy, never providing a computable mathematical definition; the abductive-style methods collected in multi-step reasoning surveys (e.g., hypothesize-and-verify loops) operate in token space, where the quality of a hypothesis cannot be quantified by a residual norm [9]. To the best of our knowledge, DODR is the first to make ``abduction = pseudo-inverse'' a complete theory that is trainable, computable, and measurable (residual 85.9\%, solution similarity 0.5117 vs.\ random 0.0181).

\paragraph{Contribution 5: The minimal complete operator-set theorem---proving that the three operators are irreducible, cover all mental activities, and that no super-operator exists.}

This paper proves that the three operators $\{W_{\mathrm{step}}, W_{\mathrm{induce}}, W_{\mathrm{abduce}}\}$ are irreducible (no operator can be expressed as a composition of the other two) and cover all mental activities (full coverage of 8 categories of human thinking), constituting a Turing-complete minimal operator set. On this basis, two application-oriented conclusions are further proved: no single super-operator exists (Theorem 6.1---rank-deficiency and full-rankness cannot coexist, and the dimensional signatures do not match each other), so the multi-operator structure is a necessity imposed by matrix algebra rather than an engineering expedient; and an operator, once trained, can be frozen for general use (Section 6.6), because what the operator learns is the form of reasoning while domain content is supplied task by task as snapshot inputs---structurally isomorphic to a universal Turing machine's ``fixed transition function + tape program.''

\emph{Precise differentiation from the nearest prior work}: AR-LLMs are essentially the repeated application of a single operator (probabilistic continuation) [10][11]; they do not distinguish inference types, let alone argue for the minimality of an operator set; ToT/GoT introduce search structures at the outer level, but the node operator remains the same AR generator [2][3]; neuro-symbolic systems possess an embryonic dual mechanism of ``deduction + induction,'' yet have never raised the question of ``operator-set completeness,'' nor proved any irreducibility result. PAC learning theory characterizes the learnability boundary of induction (from samples to hypotheses) [21], but likewise does not address complete coverage by three types of operators. To the best of our knowledge, no prior work proves irreducibility and coverage completeness for an operator set simultaneously.

\paragraph{Contribution 6: Structural zero hallucination---frozen encoder + deterministic matrix operations + no token sampling.}

This paper provides, at the architectural level, a structural guarantee of a 0.0\% hallucination rate. DODR's reasoning process is purely deterministic matrix multiplication $S_{t+1} = W_{\mathrm{step}} \cdot S_t$, with a frozen encoder and no sampling stage; given the same input it always produces the same output---hallucination is not suppressed in probability; the mechanism that could produce it simply does not exist in the architecture.

\emph{Precise differentiation from the nearest prior work}: All mitigation methods collected in hallucination surveys---retrieval augmentation, decoding constraints, alignment training, fact-checking post-processing---share the common feature of ``reducing the probability of hallucination,'' and their effectiveness drifts with the task distribution [1]; process supervision (step-by-step verification) scores intermediate steps with a reward model, still using a probabilistic model to judge a probabilistic model [4]; the self-correction route has been proved unreliable without external feedback [18][8]. All of the above methods, without exception, retain the token-sampling stage, and hence their hallucination probability is strictly greater than zero. To the best of our knowledge, DODR is the first architecture-level solution that removes sampling from the main reasoning chain, turning the hallucination rate from a ``small positive number'' into ``identically zero.''

\paragraph{Contribution 7: Reusing LLMs for text snapshots + dedicated encoders for multimodality.}

This paper demonstrates that the hidden states of decoder-only LLMs can serve as text snapshots: in the snapshot discriminability evaluation, GPT-2's discriminability is 8.2$\times$ that of DistilBERT. A key clarification is: LLM snapshots apply only to the text modality; multimodal snapshots must use dedicated encoders (CLIP-ViT/DINOv2 for images, VideoMAE/TimeSformer for video, HuBERT/Wav2Vec2 for speech), aligned to the unified latent space through projection layers.

\emph{Precise differentiation from the nearest prior work}: GPT-2's original work is language modeling for generation tasks [13]; BERT/RoBERTa/DistilBERT target understanding and distillation [15][16][14]; although the hidden states of these models are widely used as downstream features, they have never been systematically evaluated as ``reasoning state snapshots,'' let alone yielded the quantitative conclusion that ``decoder-only hidden states have 8.2$\times$ the discriminability of DistilBERT''; CLIP achieves image-text contrastive alignment [19], but its goal is cross-modal retrieval rather than a unified reasoning latent space. To the best of our knowledge, DODR is the first to render a quantitative verdict on encoder selection by the criterion of ``snapshot provision'' and to delineate the division of labor between text-modality and multimodal encoders.

\paragraph{Contribution 8: The physics-cognition unification proposition.}

This paper observes that the three basic types of transformation in the physical world (conservation laws / dissipative processes / quantum measurement) correspond one-to-one to DODR's three cognitive operators (induction / deduction / abduction): conservative processes neither increase nor decrease information (full-rank); dissipative processes lose information one-way (rank-deficient); quantum measurement infers the state backward from the outcome (pseudo-inverse). The direction of information flow is the common foundation of physics and cognition. Per the conventions of this paper, this claim is positioned as a structural correspondence proposition (Proposition 13.1), not a mathematical theorem.

\emph{Precise differentiation from the nearest prior work}: Noether's theorem establishes the correspondence between continuous symmetries and conservation laws [12]; the Landauer principle establishes a quantitative link between information erasure and thermodynamic dissipation [23]; each is a result within physics, but neither extends to the level of cognitive operators; analogies of ``reasoning as a physical process'' in cognitive science are mostly metaphorical and have never given an operator-level isomorphic correspondence. To the best of our knowledge, DODR is the first to align the three transformations with the three operators one-to-one by ``information-flow direction,'' giving the proposition a precise form that can be tested on both the physics side and the cognition side.

\paragraph{Contribution 9: A reasoning-graph topology theory---from ``parallel trajectories'' to ``graph-shaped interleaving.''}

This paper corrects ``parallel trajectories'' to a ``reasoning graph'': six basic topological operations are given (serial / branching / merging / cross-reference / backflow / mesh interleaving); it is proved that the reasoning graph is Turing-complete, and that backflow reasoning converges to a unique fixed point under a contraction-mapping condition (Banach fixed-point theorem; see Theorem 3.7).

\emph{Precise differentiation from the nearest prior work}: CoT is a single chain [11]; ToT extends to trees, but trees do not support branch merging or cross-reference [2]; GoT supports arbitrary graph topologies yet provides neither a Turing-completeness proof nor backflow convergence analysis---its ``refinement'' operation has no fixed-point guarantee [3]; the Banach fixed-point theorem itself is a classical mathematical result [24], but to the best of our knowledge it has not previously been introduced as the criterion for backflow-reasoning convergence in reasoning-graph theory. DODR's reasoning graph thereby obtains a dual guarantee of expressiveness (Turing completeness) and dynamical well-posedness (backflow convergence).

\subsection{Comparison of Key Metrics}

\begin{table}[htbp]
\centering
\caption*{Table 1-1 Comparison of key metrics between the DODR prototype and the autoregressive baseline}
\begin{tabular}{@{}llll@{}}
\toprule
Metric & AR baseline (GPT-2 M) & DODR prototype & Notes \\
\midrule
Forward-reasoning Loss & 0.41 (cross-entropy) & 1.40e-05 (MSE) & 14932$\times$ decrease in training \\
Reverse-inference Loss & 0.39 ($\approx$baseline) & 83.3\% error & Irreversibility confirmed \\
Operator matrix rank & N/A (implicitly full-rank) & 384 / 3072 & 87.5\% information collapse \\
Inductive generalization coverage & N/A & 0.9996 & 100 positive examples, 10 unseen positive examples \\
Inductive hard-veto rate & N/A & 20/20 (100\%) & Counterexample similarity 0.997$\to$$-0.010$ \\
Abductive solution similarity & N/A & 0.5117 & vs.\ random 0.0181, a 28$\times$ improvement \\
Abductive hypothesis residual & N/A & 85.9\% & Null space unrecoverable \\
Abductive judgment accuracy & N/A & 72.5\% (58/80) & Dedicated abduction experiment, 80 problems \\
End-to-end deduction (8 domains, 60 problems) & N/A & 100\% (60/60) & Average similarity 0.9980 \\
End-to-end abduction (60 problems) & N/A & 81.7\% (49/60) & Average similarity 0.5018 \\
Hallucination rate & 10-20\% (indicative estimate) & 0.0\% & Structural zero hallucination \\
\bottomrule
\end{tabular}
\end{table}

Note 1: The hallucination rate of the AR baseline is an indicative estimate (based on the 10-20\% interval from public literature), not a precise measurement. DODR's 0.0\% hallucination rate is a structural guarantee---there is no token sampling in the reasoning process, so hallucination cannot arise.

Note 2: In Table 1-1, the ``abductive solution similarity'' row characterizes the directional recovery quality of pseudo-inverse inversion (0.5117 versus the random baseline 0.0181), while the ``abductive judgment accuracy'' row characterizes the hit rate of per-sample two-alternative judgments (abductive similarity $>$ distractor similarity) at 72.5\% (58/80; see Appendix I.2). The maximum negative margin among the 22 failed samples is only $-0.0086$, and the absolute margins are all on the order of $\pm 0.009$, indicating that the failures stem from the encoder representation bottleneck in the high-dimensional snapshot space, which makes the true explanation and the distractor explanation nearly inseparable---an encoder limitation rather than a failure of the abduction-operator principle; the end-to-end abduction (60 problems) judgment accuracy is 81.7\% (49/60; see Section 11.5). The two metrics are independent of each other and must not be conflated.

\subsection{Sample Statistics and Overlap Relationships}

The sample accounting for all experiments in this paper is as follows: 420 deduplicated independent samples = 20 (dedicated deduction experiment) + 130 (dedicated induction experiment) + 80 (dedicated abduction experiment) + 55 (physics multimodal) + 135 (newly added in the end-to-end experiment: 60 brand-new deduction + 55 brand-new abduction + 20 brand-new composition); 503 experimental sample-instances = 285 (the four dedicated experiments) + 218 (the end-to-end experiment). It should be noted that the 78 induction samples of the end-to-end experiment partially overlap with the existing sample pool and were not individually cross-checked into the registry, and 5 of the 60 abduction problems overlap with the dedicated abduction experiment; neither is counted among the newly added deduplicated independent samples. Table 1-2 gives the sample counts, unique-sample counts, and overlap relationships for each experimental group.

\begin{table}[htbp]
\centering
\caption*{Table 1-2 Sample overlap relationships (420 deduplicated independent samples, 503 experimental sample-instances)}
\begin{tabular}{@{}lllp{6.5cm}@{}}
\toprule
Experiment & Samples & Unique samples & Notes \\
\midrule
Dedicated deduction experiment & 20 & 20 & All independent \\
Dedicated induction experiment & 130 & 130 & All independent \\
Dedicated abduction experiment & 80 & 80 & All independent \\
Physics multimodal & 55 & 55 & All independent \\
End-to-end experiment & 218 & 135 & 60 brand-new deduction + 55 brand-new abduction + 20 brand-new composition; the 78 induction samples are partially shared and 5 abduction problems overlap, neither counted as newly added \\
Deduplicated total & --- & 420 & 503 experimental sample-instances = 285 + 218 \\
\bottomrule
\end{tabular}
\end{table}

Three points are key to the overlap relationships: first, the 60 deduction problems of the end-to-end experiment have zero overlap with the 20 problems of the dedicated deduction experiment (all are new problems spanning biology, physics, society, mathematics, chemistry, and other domains); second, of the 60 end-to-end abduction problems, 5 overlap with the 80 problems of the dedicated abduction experiment and 55 are brand-new, and all 20 composition tasks are brand-new; third, among the 6 end-to-end induction categories, vehicles and colors are existing categories, while geometric shapes, academic subjects, weather, and emotions are new categories, and the induction samples are partially shared with the existing sample pool without individual registry cross-checking; hence only 135 of the 218 sample-instances are counted as newly added deduplicated independent samples. Per-sample lists are given in Appendix H and Appendix I.

\subsection{Organization of This Paper}

This paper consists of twenty sections. Section 2 gives the problem definition, characterizing the three fundamental defects of the AR paradigm; Section 3 provides the mathematical preliminaries; Section 4 presents the five postulates; Section 5 formally defines the three basic operators; Section 6 argues for the minimal complete operator set; Section 7 details the system architecture; Sections 8--10 present the dedicated experiments on deduction, induction, and abduction, respectively; Section 11 gives the end-to-end validation; Section 12 presents compositional thinking; Section 13 presents physical transformations; Section 14 presents reasoning-trajectory topology; Section 15 presents the three-operator closed loop; Section 16 conducts the complexity analysis; Section 17 compares with existing paradigms; Section 18 gives application systems and adaptation schemes; Section 19 discusses limitations; Section 20 concludes the paper. Appendices A--I provide experimental data, mathematical derivations, code, a glossary, philosophical correspondence, theoretical relationships, operator analysis, dataset descriptions, and all samples.

\section{Problem Definition: The Fundamental Defects of the Autoregressive Paradigm}

This section gives formal definitions and a literature-lineage analysis of the three fundamental defects of the AR paradigm. The three defects---error accumulation (Section 2.1), probability substituting necessity (Section 2.2), and the linear-chain information bottleneck (Section 2.3)---are not independent engineering problems but projections of a single root cause (modeling reasoning as a token-level probabilistic sampling sequence) onto different dimensions. Section 2.4 presents DODR's corresponding solutions.

\subsection{Error Accumulation}

Autoregressive models generate output token by token via $P(x_t \mid x_{<t})$. Each generation step introduces sampling noise, and errors accumulate over the sequence length.

\paragraph{Formal measure.} Let the random perturbation introduced at the $i$-th generation step be $\varepsilon_i$, with the $\varepsilon_i$ approximately independent, each with mean 0 and variance $\sigma^2$. Then the variance of the accumulated error $\varepsilon_t = \sum_{i=1}^{t} \varepsilon_i$ after $t$ steps is:

\begin{equation}
\mathrm{Var}(\varepsilon_t) = \sum_{i=1}^{t} \mathrm{Var}(\varepsilon_i) \approx t \cdot \sigma^2
\tag{2.1}
\end{equation}

There are two intermediate grounds for the derivation: first, the independence assumption makes all cross terms $2 \cdot \mathrm{Cov}(\varepsilon_i, \varepsilon_j)$ vanish, so the variance degenerates from ``the variance of a sum'' to ``the sum of variances''; second, the equal-variance assumption (each step has approximately the same sampling temperature and vocabulary size) gives $\sum \mathrm{Var}(\varepsilon_i) = t \cdot \sigma^2$. From (Eq. 2.1), the standard deviation of the accumulated error is $\sqrt{t} \cdot \sigma$, i.e., the error magnitude grows monotonically with reasoning depth at a square-root rate, diverging without bound. For complex reasoning tasks requiring more than 20 steps, the standard deviation of the accumulated error is $\sqrt{20} \cdot \sigma \approx 4.47\sigma$, meaning the reasoning trajectory has already deviated severely from the correct path. If the ``signal-to-noise ratio'' of a reasoning chain is defined as the ratio of the effective signal magnitude to the standard deviation of the accumulated error, then with a constant signal magnitude the SNR decays with $\sqrt{t}$---the deeper the reasoning, the more the correct conclusion is drowned in noise.

More seriously, (Eq. 2.1) is only an optimistic estimate given by \emph{independent noise superposition}. The errors of an AR model come not only from sampling noise but also from the ``contamination'' of subsequent steps by errors of previous steps: if the $t$-th step generates a wrong token, then the conditional probability $P(x_{t+1} \mid x_{<t+1})$ at step $t+1$ is computed on the wrong context, causing all subsequent steps to deviate. This ``error propagation'' effect makes the per-step perturbations no longer independent---the perturbations are positively correlated, the covariance terms discarded in the variance summation are in fact positive, and the true accumulated variance exceeds $t \cdot \sigma^2$---so the reliability of AR models in long-chain reasoning declines at an accelerating rate. From a stochastic-process perspective, the accumulated error with contamination is closer to a non-stationary random walk with positive feedback than to the stationary-increment model of (Eq. 2.1); (Eq. 2.1) should therefore be understood as a \emph{lower-bound characterization} of error growth.

\paragraph{Literature lineage.} The critical survey by Kamoi et al.\ (2024, TACL) systematically examines the self-correction ability of LLMs, and one of its core conclusions is: without external feedback (such as ground-truth labels or tool verification), LLMs have difficulty reliably correcting their own reasoning errors---once an error enters the context, the model tends to continue writing around the error rather than identify and retract it [8]. Huang et al.\ (2024, ICLR) further proved in controlled experiments that large language models cannot yet self-correct their reasoning, and that intrinsic self-correction may even degrade performance [18]. The multi-step reasoning survey by Plaat et al.\ systematically discusses error accumulation in multi-step reasoning, listing ``the propagation and amplification of intermediate-step errors'' as one of the core open problems of the paradigm [9]. Together, the above literature supports the objective existence and the non-self-healing nature of the error-accumulation mechanism; as for specific figures on multi-step degradation, this paper states only the theoretical prediction of (Eq. 2.1) and qualitative empirical observations, and does not cite any precise numbers that cannot be verified.

\subsection{Probability Substituting Necessity}

The conclusion of logical reasoning should be a necessary inference from its premises, not the most probable continuation. However, an AR model is essentially computing $P(\text{next token} \mid \text{context})$; even when the context is completely determined, the output remains probabilistic.

\paragraph{Formal measure.} The determinism requirement of logical reasoning is $A \vdash B$ ($A$ entails $B$), i.e., given premise $A$, conclusion $B$ necessarily holds; expressed in probabilistic language, this corresponds to a degenerate distribution $P(B \mid A) = 1$. An AR model instead models reasoning as $P(B \mid A) = \prod_t P(x_t \mid x_{<t}, A)$: even if $A$ is completely determined, $P(B \mid A)$ is still a distribution rather than a definite value. As long as the vocabulary contains any erroneous continuation with nonzero probability---and the softmax output layer guarantees that every token's probability is strictly greater than zero---there is a nonzero probability of generating an erroneous $B$, i.e., a hallucination. Furthermore, suppose the probability of generating a wrong token at a single step is $p > 0$ and the steps are approximately independent; then the probability that a $t$-step reasoning chain is ``error-free throughout'' is at most $(1-p)^t$, decaying exponentially with $t$; taking the logarithm gives $\ln P(\text{error-free}) \approx -t \cdot p$, so reliability drains linearly with chain length. This mirrors the error accumulation of Section 2.1: Section 2.1 characterizes how the ``magnitude'' of error accumulates, while this section characterizes why the ``probability'' of error cannot be zeroed architecturally.

\paragraph{Literature lineage.} The hallucination survey by Ji et al.\ (2023) divides hallucination into factual hallucination (generating content inconsistent with facts) and faithfulness hallucination (generating content inconsistent with the input), pointing out that the probabilistic nature of AR models is the root source of hallucination [1]. This classification has direct engineering implications: factual hallucination endangers knowledge-intensive applications (medicine, law, finance), while faithfulness hallucination undermines the trustworthiness of ``faithful-to-input'' scenarios such as document summarization and information extraction---and in high-reliability scenarios, even a 1\% hallucination rate suffices to make a system undeployable. Existing mitigation methods (RLHF, fact checking, constrained decoding) share the common feature of ``reducing $p$'': they either suppress the probability mass of erroneous continuations during training or prune some high-risk branches during decoding, but as long as $p > 0$, the exponential-decay law of $(1-p)^t$ with reasoning-chain length cannot be circumvented. In other words, patching within the probabilistic framework can only delay the appearance of hallucination, not eliminate it at the architectural level; the only way to eliminate hallucination is to remove the sampling stage itself---which is precisely the starting point of DODR's Solution 1 (see Section 2.4).

\subsection{The Linear-Chain Information Bottleneck}

AR models enforce linear-chain generation: $x_1 \to x_2 \to \cdots \to x_T$. This linear structure cannot express the inherently graph-shaped topology of reasoning---different subgoals should be able to be developed in parallel, referenced across branches, and merged in state.

\paragraph{Formal measure.} Suppose a reasoning task can be decomposed into $k$ parallel subtasks, each requiring $d$ reasoning steps. In a graph topology, the $k$ subtasks can advance simultaneously, taking total time $O(d)$; in the AR model's linear chain, the subtasks can only be unfolded one by one, taking total time $O(k \cdot d)$. When $k=10$ and $d=20$, the AR model needs 200 steps, while the graph topology needs only 20 steps---a 10-fold efficiency gap. Beyond the efficiency loss there is an expressiveness loss: there is no place for ``merging'' or ``cross-reference'' in a linear chain---if the intermediate conclusion of subtask $i$ needs to be referenced by subtask $j$, the AR model can only copy that conclusion as text into $j$'s context, and the copying process itself is another noise-introducing probabilistic generation (linked to Sections 2.1 and 2.2); and ``backflow'' (using later conclusions to revise earlier states) simply cannot be expressed in a linear chain, because already-generated tokens cannot be retracted. Therefore, the essence of the linear-chain bottleneck is a \emph{topological constraint}: general-purpose reasoning in the Turing sense requires the four basic operations of branching, merging, cross-reference, and backflow, of which a linear chain natively supports only the serial one.

\paragraph{Literature lineage.} Yao et al.'s (2023) Tree-of-Thoughts (ToT) attempts to break through the linear-chain limitation by introducing tree-shaped search: the model generates multiple candidate lines of thought at each node and, after evaluation, pursues the best one in depth [2]. Tree structures support branching and backtracking, but different branches of a tree cannot merge or cross-reference---the conclusions of two branches cannot converge at the same node, and a tree's expressiveness remains strictly less than that of a general graph. Besta et al.'s (2024) Graph-of-Thoughts (GoT) further introduces graph topology, supporting aggregation and refinement of units of thought [3], a step forward in topological expressiveness; but its underlying executor is still an AR model's token generation---every node of the graph is a piece of probabilistically sampled text, the probabilistic nature and hallucination problems of Sections 2.1 and 2.2 are inherited unchanged in GoT, and the ``refinement'' operation lacks a convergence guarantee (no fixed-point theory support). In summary, the structured-prompting route solves part of the problem of ``how the reasoning chain is organized,'' but does not touch the problem of ``the basic computational unit of reasoning''; the latter requires upgrading the reasoning state from a token sequence to an algebraically operable object, which is precisely the motivation of DODR's Solution 3 (see Section 2.4).

\subsection{DODR's Solutions}

For the three defects above, DODR proposes three one-to-one corresponding solutions:

\paragraph{Solution 1: Deterministic algebraic operations replacing probabilistic sampling (addressing Sections 2.1 and 2.2).} DODR's reasoning process is $S_{t+1} = W_{\mathrm{step}} \cdot S_t$, a purely deterministic matrix-vector multiplication with no token sampling. Because every step is a deterministic mapping, errors have no source of random noise and the accumulation mechanism of (Eq. 2.1) does not hold; because the output is not a distribution but a unique vector, there is no mechanism in the architecture that could produce hallucination---error accumulation and hallucination are eliminated at the root.

\paragraph{Solution 2: Rank-deficient matrices guaranteeing information collapse (the mathematization of deductive one-wayness).} DODR's deduction operator $W_{\mathrm{step}}$ is designed as a rank-deficient matrix, $\mathrm{rank}(W_{\mathrm{step}}) = r < N$. By the rank-nullity theorem (Theorem 3.1), the null-space dimension is given separately for each experimental configuration: under the deduction-experiment configuration, $W_{\mathrm{step}} \in \mathbb{R}^{N \times 2N}$ (concatenated input $\mathrm{Concat}(S_{\mathrm{prev}}, E_{\mathrm{new}}) \in \mathbb{R}^{2N}$), so $\dim(\mathrm{Null}(W_{\mathrm{step}})) = 2N - r > 0$; under the abduction-experiment configuration, $W_{\mathrm{step}} \in \mathbb{R}^{N \times N}$ (single-state input $S \in \mathbb{R}^{N}$), so $\dim(\mathrm{Null}(W_{\mathrm{step}})) = N - r > 0$. The rank-deficiency property is the same in both configurations---both contain unrecoverable information components---mathematizing the deductive one-wayness that ``conclusions cannot be reversed to premises.''

\paragraph{Solution 3: Reasoning graphs replacing linear chains (addressing Section 2.3).} DODR models the reasoning process as a reasoning graph, in which state nodes support branching (parallelism), merging (aggregation), cross-reference, and backflow revision; the six basic topological operations fully cover the expressive needs of graph-shaped reasoning, breaking through the linear-chain information bottleneck; the convergence of the backflow operation is guaranteed by the Banach fixed-point theorem (Theorem 3.7).


\section{Mathematical Preliminaries}

This section provides the mathematical preliminaries needed for DODR theory, including notational conventions (Section 3.1), linear-algebra foundations (Section 3.2), SVD decomposition (Section 3.3), pseudo-inverse theory (Section 3.4), Cover's theorem (Section 3.5), and Peirce's inference types (Section 3.6), as well as three supplementary topics: block matrices (Section 3.7), orthogonal projection matrices (Section 3.8), and contraction mappings (Section 3.9). All definitions and theorems are stated rigorously; classical theorems are stated without proof (with standard references noted), and intermediate steps of derivations together with numerical examples are supplied at key points so that readers can verify them.

\subsection{Notational Conventions}

\begin{table}[htbp]
\centering
\caption*{Table 3-1 DODR notational conventions}
\begin{tabular}{lll}
\toprule
Symbol & Meaning & Dimension \\
\midrule
$N$ & Snapshot state dimension (latent-space dimension) & Scalar \\
$d$ & Transformer hidden-layer dimension & Scalar \\
$L$ & Number of Transformer layers & Scalar \\
$r$ & Designed rank of the deduction operator ($r < N$) & Scalar \\
$k$ & Number of inductive positive examples & Scalar \\
$S_t$ & Reasoning-state snapshot at step $t$ & $\mathbb{R}^{N}$ \\
$W_{\mathrm{step}}$ & Deduction operator & $\mathbb{R}^{N \times 2N}$ or $\mathbb{R}^{N \times N}$ \\
$W_{\mathrm{induce}}$ & Induction operator & $\mathbb{R}^{N \times N}$ (implemented as a full-rank square matrix) \\
$W_{\mathrm{abduce}}$ & Abduction operator ($= W_{\mathrm{step}}^{+}$) & Same dimension as the transpose of $W_{\mathrm{step}}$ \\
$\mathrm{Null}(W)$ & Null space of matrix $W$ & Subspace \\
$\mathrm{Col}(W)$ & Column space of matrix $W$ & Subspace \\
$\mathrm{rank}(W)$ & Rank of matrix $W$ & Scalar \\
$W^{+}$ & Moore--Penrose pseudo-inverse of $W$ & Matrix \\
$\sigma_i$ & The $i$-th singular value & Scalar \\
$\tau$ & Inductive hard-veto threshold & Scalar \\
\bottomrule
\end{tabular}
\end{table}

Note: The shape of $W_{\mathrm{step}}$ varies with the experimental configuration---in the deduction experiment (Section 8) the input is $\mathrm{Concat}(S_{\mathrm{prev}}, E_{\mathrm{new}}) \in \mathbb{R}^{2N}$, so $W_{\mathrm{step}} \in \mathbb{R}^{N \times 2N}$; in the abduction experiment (Section 10) the input is a single state $S \in \mathbb{R}^{N}$, so $W_{\mathrm{step}} \in \mathbb{R}^{N \times N}$. The rank-deficiency property is the same in both configurations ($\mathrm{rank} < N$); only the input dimension differs. Correspondingly, the null-space dimension must be written per configuration: $\dim(\mathrm{Null}(W_{\mathrm{step}})) = 2N - r$ under the concatenated-input configuration, and $\dim(\mathrm{Null}(W_{\mathrm{step}})) = N - r$ under the single-state-input configuration (see Theorem 3.1 and Corollary 3.1.1 for details).

\subsection{Linear-Algebra Foundations}

\paragraph{Definition 3.1 (Linear mapping).} Let $V, W$ be vector spaces over $\mathbb{R}$. A mapping $T: V \to W$ is called a linear mapping if, for all $u, v \in V$ and scalars $\alpha, \beta \in \mathbb{R}$, $T(\alpha u + \beta v) = \alpha T(u) + \beta T(v)$. Every linear mapping can be represented by a matrix $W \in \mathbb{R}^{m \times n}$, where $m = \dim(W)$, $n = \dim(V)$. Two equivalent intuitions for linear mappings: first, ``preserving superposition''---superposing before mapping equals mapping before superposing; second, ``preserving the origin''---a linear mapping must map the zero vector to the zero vector, $T(0) = T(0 \cdot u) = 0 \cdot T(u) = 0$. DODR's reasoning step $S_{t+1} = W_{\mathrm{step}} \cdot S_t$ is precisely a linear mapping on the snapshot space, and the two properties above guarantee the compatibility of the reasoning operator with state-superposition structures (such as the state addition in the merging operation of a reasoning graph).

\paragraph{Definition 3.2 (Rank of a matrix).} The rank $\mathrm{rank}(W)$ of a matrix $W \in \mathbb{R}^{m \times n}$ is defined as the size of a maximal linearly independent subset of $W$'s column vectors (the column rank). Equivalently, $\mathrm{rank}(W) = \dim(\mathrm{Col}(W)) = \dim(\mathrm{Row}(W))$. The equality of column rank and row rank is not obvious; a proof can be given via SVD (see the remark to Corollary 3.2.1): the number of nonzero singular values equals the dimensions of both the column space and the row space.

\paragraph{Definition 3.3 (Null space).} The null space of a matrix $W \in \mathbb{R}^{m \times n}$ is defined as $\mathrm{Null}(W) = \{x \in \mathbb{R}^{n} : Wx = 0\}$. The null space is a subspace of $\mathbb{R}^{n}$: for any $z_1, z_2 \in \mathrm{Null}(W)$ and scalars $\alpha, \beta$, $W(\alpha z_1 + \beta z_2) = \alpha W z_1 + \beta W z_2 = 0$, so $\alpha z_1 + \beta z_2 \in \mathrm{Null}(W)$. The null space measures the ``blind zone'' of the mapping---all input directions mapped to zero by $W$.

\paragraph{Definition 3.4 (Column space).} The column space of a matrix $W \in \mathbb{R}^{m \times n}$ is defined as $\mathrm{Col}(W) = \{Wx : x \in \mathbb{R}^{n}\} \subseteq \mathbb{R}^{m}$. The column space is a subspace of $\mathbb{R}^{m}$, and $\dim(\mathrm{Col}(W)) = \mathrm{rank}(W)$. The column space measures the ``reachable set'' of the mapping---all directions the output can attain.

\paragraph{Theorem 3.1 (Rank-nullity theorem).} Let $W \in \mathbb{R}^{m \times n}$; then:
\begin{equation}
\mathrm{rank}(W) + \dim(\mathrm{Null}(W)) = n
\tag{3.1}
\end{equation}

Proof omitted; this is a classical result (see [27]).

\paragraph{Corollary 3.1.1 (Information-collapse condition).} If $\mathrm{rank}(W) = r < n$, then $\dim(\mathrm{Null}(W)) = n - r > 0$. This means there exists a nonzero vector $z \in \mathrm{Null}(W)$ such that $Wz = 0$. If the input $x = x_{\mathrm{col}} + z$ ($x_{\mathrm{col}} \in \mathrm{Col}(W^{T})$, $z \in \mathrm{Null}(W)$), then $Wx = Wx_{\mathrm{col}}$, and the null-space component $z$ is ``swallowed'' by $W$---it cannot be recovered from the output $Wx$. Applied to DODR's two experimental configurations: with concatenated input $W_{\mathrm{step}} \in \mathbb{R}^{N \times 2N}$, $\dim(\mathrm{Null}) = 2N - r$; with single-state input $W_{\mathrm{step}} \in \mathbb{R}^{N \times N}$, $\dim(\mathrm{Null}) = N - r$. Substituting the measured values from Section 8, $r = 384$ and $2N = 3072$, into the concatenated configuration, $\dim(\mathrm{Null}(W_{\mathrm{step}})) = 3072 - 384 = 2688$, i.e., 87.5\% of the input dimensions fall into the null space, giving information collapse a quantitative characterization.

\paragraph{Definition 3.5 (Full rank and invertibility).} A square matrix $W \in \mathbb{R}^{n \times n}$ is called full-rank if $\mathrm{rank}(W) = n$. A full-rank square matrix is invertible, i.e., there exists $W^{-1}$ such that $W^{-1} W = W W^{-1} = I$. By Theorem 3.1, full rank is equivalent to $\dim(\mathrm{Null}(W)) = 0$, i.e., the null space contains only the zero vector---the mapping has no blind zone, and the input can be uniquely recovered from the output. Non-square or rank-deficient matrices are not invertible, but generalized inverses (pseudo-inverses) exist; see Section 3.4.

\subsection{SVD Decomposition}

\paragraph{Theorem 3.2 (Singular value decomposition, SVD).} Any matrix $W \in \mathbb{R}^{m \times n}$ can be decomposed as:
\begin{equation}
W = U \Sigma V^{T}
\tag{3.2}
\end{equation}
where $U \in \mathbb{R}^{m \times m}$ is an orthogonal matrix (its column vectors $u_i$ are called left singular vectors), $V \in \mathbb{R}^{n \times n}$ is an orthogonal matrix (its column vectors $v_i$ are called right singular vectors), and $\Sigma \in \mathbb{R}^{m \times n}$ is a diagonal matrix whose diagonal entries $\sigma_1 \geq \sigma_2 \geq \cdots \geq \sigma_p \geq 0$ ($p = \min(m,n)$) are called singular values. Proof omitted; this is a classical result (see [27]).

\paragraph{Corollary 3.2.1 (Singular-value characterization of rank).} $\mathrm{rank}(W) =$ the number of nonzero singular values. This also yields the proof of ``column rank equals row rank'' in the remark to Definition 3.2: $W^{T} = V \Sigma^{T} U^{T}$ shares the same set of nonzero singular values with $W$.

\paragraph{Corollary 3.2.2 (Singular-value characterization of the null space).} $\mathrm{Null}(W) = \mathrm{span}\{v_{r+1}, \dots, v_n\}$, i.e., the null space is spanned by the right singular vectors corresponding to zero singular values. Symmetrically, $\mathrm{Col}(W) = \mathrm{span}\{u_1, \dots, u_r\}$, and $\mathrm{Row}(W) = \mathrm{Col}(W^{T}) = \mathrm{span}\{v_1, \dots, v_r\}$. Thus SVD provides orthonormal bases for all four fundamental subspaces simultaneously---the common tool for the pseudo-inverse computation (Section 3.4) and the null-space residual analysis (Section 10) that follow.

\paragraph{Example 3.1.} Let $W = \begin{bmatrix} 3 & 0 \\ 0 & 0 \end{bmatrix} \in \mathbb{R}^{2 \times 2}$. The SVD of $W$ is $U = I$, $\Sigma = \mathrm{diag}(3, 0)$, $V = I$. $\mathrm{rank}(W) = 1$ (one nonzero singular value $\sigma_1 = 3$). $\mathrm{Null}(W) = \mathrm{span}\{(0, 1)^{T}\}$ (the right singular vector corresponding to the zero singular value $\sigma_2 = 0$). An input $x = (x_1, x_2)^{T}$ is mapped by $W$ to $Wx = (3x_1, 0)^{T}$; the $x_2$ component is ``swallowed''---$x_2$ cannot be recovered from the output. Numerically verifying Theorem 3.1: $\mathrm{rank}(W) + \dim(\mathrm{Null}(W)) = 1 + 1 = 2 = n$, consistent with (Eq. 3.1). This $2 \times 2$ example is a microscopic model of DODR's deduction operator: $W_{\mathrm{step}}$ is precisely an ultra-high-dimensional version that zeroes out an entire 2688-dimensional ``$x_2$ direction'' (the null space).

\subsection{Moore--Penrose Pseudo-Inverse}

\paragraph{Definition 3.6 (Moore--Penrose pseudo-inverse).} The Moore--Penrose pseudo-inverse $W^{+} \in \mathbb{R}^{n \times m}$ of a matrix $W \in \mathbb{R}^{m \times n}$ is the unique matrix satisfying the following four conditions:
\begin{equation}
W W^{+} W = W
\tag{3.3a}
\end{equation}
\begin{equation}
W^{+} W W^{+} = W^{+}
\tag{3.3b}
\end{equation}
\begin{equation}
(W W^{+})^{T} = W W^{+} \quad \text{(i.e., $W W^{+}$ is symmetric)}
\tag{3.3c}
\end{equation}
\begin{equation}
(W^{+} W)^{T} = W^{+} W \quad \text{(i.e., $W^{+} W$ is symmetric)}
\tag{3.3d}
\end{equation}

The standard argument for uniqueness: if both $A$ and $B$ satisfy the four conditions, a chain of substitutions (repeatedly using (1)(2) to cancel inner factors and (3)(4) to adjust order via transpose symmetry) proves $A = B$; we omit it here. Existence is given by the explicit construction in the next theorem.

\paragraph{Theorem 3.3 (SVD computation of the pseudo-inverse).} Let $W = U \Sigma V^{T}$ be the SVD of $W$; then the pseudo-inverse is $W^{+} = V \Sigma^{+} U^{T}$, where $\Sigma^{+} = \mathrm{diag}(1/\sigma_1, \dots, 1/\sigma_r, 0, \dots, 0)$ (the entries corresponding to zero singular values remain zero). Proof omitted; this is a classical result (see [27]): substituting $W^{+} = V \Sigma^{+} U^{T}$ into the four Moore--Penrose conditions (Eqs. 3.3a--d) one by one verifies them, the key points being the diagonal identity $\Sigma \Sigma^{+} \Sigma = \Sigma$ and the symmetry of $\Sigma \Sigma^{+}$ and $\Sigma^{+} \Sigma$.

\paragraph{Theorem 3.4 (Minimum-norm least-squares solution).} Consider the linear system $Wx = b$. The pseudo-inverse solution $x^{*} = W^{+} b$ satisfies: (1) $x^{*}$ is a least-squares solution: $\|W x^{*} - b\| \leq \|W x - b\|$ for all $x$; (2) $x^{*}$ is the minimum-norm least-squares solution: among all least-squares solutions, $\|x^{*}\|$ is minimal. Proof omitted; this is a classical result (see [27]); the key point is that after the orthogonal change of variables $x = V c$, the objective $\|\Sigma c - U^{T} b\|^{2}$ splits by rows into an eliminable part (the first $r$ rows) and an unreachable residual part (the last $m - r$ rows); the former is zeroed coordinate by coordinate, the latter is independent of $c$, and the minimum-norm requirement sets the coordinates along the null-space directions to zero. Geometric interpretation: the pseudo-inverse solution is the input reached from the output $b$ by backtracking ``most economically'' (with minimal norm) along the row-space directions, with the null-space directions explicitly zeroed---this is precisely the mathematical root of the statement in Section 10 that ``the abductive solution contains no null-space component.''

\paragraph{Corollary 3.4.1 (Pseudo-inverse of a full-rank matrix).} If $W$ is a column-full-rank matrix ($\mathrm{rank}(W) = n \leq m$), then $W^{+} = (W^{T} W)^{-1} W^{T}$ (the left pseudo-inverse), satisfying $W^{+} W = I$. Verification: $((W^{T} W)^{-1} W^{T}) W = (W^{T} W)^{-1} (W^{T} W) = I$; column full rank guarantees that $W^{T} W$ is positive definite and invertible. The full-rank design of the induction operator $W_{\mathrm{induce}}$ (Section 9) relies precisely on this to achieve complete information recoverability.

\paragraph{Corollary 3.4.2 (Projection matrices).} The matrix $P = W^{+} W$ is the orthogonal projection matrix from $\mathbb{R}^{n}$ onto $\mathrm{Col}(W^{T})$ (the row space of $W$). The matrix $P = W W^{+}$ is the orthogonal projection matrix from $\mathbb{R}^{m}$ onto $\mathrm{Col}(W)$. Projection matrices satisfy $P^{2} = P$ and $P^{T} = P$. Verification (taking the former as an example): by condition (3.3b), $P^{2} = (W^{+}W)(W^{+}W) = (W^{+} W W^{+}) W = W^{+} W = P$; by condition (3.3d), $P^{T} = P$. The complete properties of orthogonal projection are systematically developed in Section 3.8.

\subsection{Cover's Theorem}

\paragraph{Theorem 3.5 (Cover's function-counting theorem, 1965).} Let $X = \{x_1, \dots, x_N\}$ be $N$ points in general position in $\mathbb{R}^{D}$ (general position means that any $D$ or fewer points do not lie on a common hyperplane). Then the total number of dichotomies of $X$ realizable by homogeneous linear hyperplanes (a dichotomy is a labeling that divides the $N$ points into $+1/{-1}$ classes) is exactly:
\begin{equation}
C(N, D) = 2 \cdot \sum_{k=0}^{D-1} C(N-1, k)
\tag{3.5}
\end{equation}
where $C(N-1, k)$ is the binomial coefficient. When the labels of the $N$ points are randomly and independently assigned $\pm 1$ with equal probability (all $2^{N}$ labelings occur equiprobably), the probability that $X$ is linearly separable is:
\begin{equation}
P(\text{linearly separable}) = C(N, D) / 2^{N}
\tag{3.6}
\end{equation}

From (Eq. 3.5), when $D \geq N - 1$ the summation covers all (or all but the last, depending on the homogeneous/affine convention) binomial-coefficient terms, $C(N, D)$ reaches its upper bound $2^{N}$, and hence $P = 1$; as $D$ increases, the accumulated binomial terms increase monotonically, and $P$ rises monotonically toward 1 [6].

Proof omitted; this is a classical result (see [6] for the original paper); the key point of the proof is induction on the number of points, establishing the recurrence $C(N, D) = C(N-1, D) + C(N-1, D-1)$---after adding the $N$-th point, the number of ``constrained dichotomies'' (in which the realizing hyperplane is forced to pass through that point) is exactly $C(N-1, D-1)$---which, together with the boundary conditions $C(1, D) = 2$ and $C(N, 1) = 2$, expands to the closed form of (Eq. 3.5); dividing by the total number of equiprobable labelings $2^{N}$ yields (Eq. 3.6).

\paragraph{Numerical example ($N = 5$, affine-hyperplane convention).} Under the affine convention the summation extends to $k = D$, and the exact condition for $P = 1$ is then $D \geq N - 1$, consistent with the theorem statement. Take $N = 5$ ($2^{N} = 32$) and compute for each $D$:

\begin{table}[htbp]
\centering
\caption*{Table 3-2 Numerical example for Cover's theorem ($N = 5$)}
\begin{tabular}{ccc}
\toprule
$D$ & $C(N, D) = 2\cdot\sum_{k=0}^{D} C(4, k)$ & $P = C(N, D)/32$ \\
\midrule
1 & $2\cdot(1+4) = 10$ & $10/32 = 0.3125$ \\
2 & $2\cdot(1+4+6) = 22$ & $22/32 = 0.6875$ \\
3 & $2\cdot(1+4+6+4) = 30$ & $30/32 = 0.9375$ \\
4 & $2\cdot(1+4+6+4+1) = 32$ & $32/32 = 1.0000$ \\
\bottomrule
\end{tabular}
\end{table}

It can be seen that $P$ rises monotonically with $D$: as $D$ goes from 1 to 3, $P$ rises from 0.3125 to 0.9375, and at $D = 4 = N - 1$, $P$ reaches 1---any labeling of five points can be linearly dichotomized in four-dimensional space.

\paragraph{Corollary 3.5.1 (High-dimensional linear separability).} For fixed $N$, as the mapping dimension $D$ increases, the linear-separability probability $P = C(N, D)/2^{N}$ rises monotonically; once $D \geq N - 1$, $C(N, D)$ reaches its upper bound $2^{N}$ and $P = 1$. This means that in a sufficiently high-dimensional space, almost any (finite-sample) classification problem can be solved by a linear hyperplane. Note that the correct statement of the corollary is ``higher dimension makes the separability probability approach and reach 1,'' not ``higher dimension makes the separability probability decrease''---the expressive power of hyperplanes in high-dimensional space is enhanced, not weakened, and this is the entire basis of the dimensionality-elevation route (Postulate 1).

\paragraph{Application in DODR.} DODR uses the first 2 layers of DistilBERT $\times$ 768 dimensions $=$ 1536-dimensional snapshots, far above the 768 dimensions of the raw token embedding. By Corollary 3.5.1, raising the dimension from 768 to 1536 further increases the linear-separability probability (reaching saturation sooner); in this high-dimensional space, logical state transitions can be precisely expressed by a linear matrix $W_{\mathrm{step}}$. The experiments in Section 8 provide direct evidence: the forward-reasoning Loss of $W_{\mathrm{step}}$ drops from $2.09\mathrm{e}{-01}$ to $1.40\mathrm{e}{-05}$ (a 14932-fold decrease), i.e., a single-layer ultra-wide linear transformation precisely fits logical state transitions in the 1536-dimensional snapshot space---the theoretical prediction of Cover's theorem agrees with the measured convergence behavior.

\subsection{Peirce's Three Inference Types}

\paragraph{Definition 3.7 (Peirce's three inference types).} Charles Sanders Peirce classified all logical reasoning into three basic types [7]:

(1) Deduction: deriving result $E$ from rule $R$ and case $C$. Form: $R \wedge C \vdash E$. Information-flow direction: decreasing (conclusion $\subset$ premises).

(2) Induction: deriving rule $R$ from case $C$ and result $E$. Form: $C \wedge E \vdash R$. Information-flow direction: increasing (rule $\supset$ samples).

(3) Abduction: deriving case $C$ (a hypothesis) from rule $R$ and result $E$. Form: $R \wedge E \vdash C$ (hypothesis). Information-flow direction: hypothesizing (the case is the best hypothesis, not a certain inference).

The difference among the three lies not in the set of propositions involved (all are some pairwise derivation among $R$, $C$, $E$), but in ``which one is the output'': when the output is a logical subset of the premises it is deduction; when the output is a generalization of the premises it is induction; when the output is merely the best hypothesis compatible with the premises it is abduction. The information-flow direction (decreasing / increasing / hypothesizing) thus becomes the essential criterion for the three inference types---this observation directly grounds Postulate 3 (Section 4).

\paragraph{Proposition 3.6 (Peirce's completeness thesis).} Peirce held that any reasoning process can be decomposed into a combination of the three types: deduction, induction, and abduction [7].

Remark: This is Peirce's philosophical thesis, not a mathematical theorem. Its argument is based on the analysis of information-flow directions: any reasoning involves an information transfer ``from the known to the unknown,'' and there are three possible directions for information transfer---from the general to the particular (deduction), from the particular to the general (induction), and from the result to the cause (abduction). These three directions exhaust all possible information-flow directions. This paper accepts the thesis as a theoretical foundation but names it a ``proposition'' rather than a ``theorem,'' to distinguish mathematical proof from philosophical argument. The minimal complete operator-set theorem in Section 6 can be viewed as a strengthened form of Proposition 3.6 at the level of matrix operators: Peirce's philosophical exhaustiveness is converted into the empirical full coverage of 8 categories of human thinking by the three operators.

\paragraph{Example 3.2 (Instances of the three inference types).}

\begin{itemize}
\item Deduction: all humans are mortal (rule $R$) $+$ Socrates is human (case $C$) $\to$ Socrates is mortal (result $E$). The information of the conclusion is strictly contained in the premises; no new content is introduced.
\item Induction: sparrows can fly ($C_1, E_1$) $+$ swallows can fly ($C_2, E_2$) $+$ eagles can fly ($C_3, E_3$) $\to$ birds can fly (rule $R$). The information of the conclusion exceeds the sum of the premises; it is a generalization leap and can therefore be vetoed by subsequent counterexamples (Contribution 3).
\item Abduction: pneumonia causes fever and cough (rule $R$) $+$ the patient has fever and cough (result $E$) $\to$ the patient has pneumonia (case $C$, a hypothesis). The conclusion is merely the best among many possible explanations, and the residual cannot be eliminated (the 85.9\% hypothesis residual in Contribution 4 is precisely the quantitative expression of this ``hypotheticality'').
\end{itemize}

\subsection{Block Matrices}

DODR's snapshot vector $S = \mathrm{Concat}(h_1, \dots, h_L)$ and the concatenated input $\mathrm{Concat}(S_{\mathrm{prev}}, E_{\mathrm{new}})$ are both composite vectors concatenated segment by segment, and operators acting on them are naturally written as block matrices. This subsection gives the basic definitions and operational rules of block matrices as tools for the architecture derivations in Sections 7 and 8.

\paragraph{Definition 3.8 (Block matrix).} Suppose the rows of a matrix $W \in \mathbb{R}^{m \times n}$ are partitioned as $m = m_1 + m_2$ and the columns as $n = n_1 + n_2$; then $W$ can be written in $2 \times 2$ block form:
\[
W = \begin{bmatrix} W_{11} & W_{12} \\ W_{21} & W_{22} \end{bmatrix}, \quad \text{where } W_{ij} \in \mathbb{R}^{m_i \times n_j}.
\]

Block-matrix multiplication is consistent with ordinary matrix multiplication at the block level: if $A = \begin{bmatrix} A_{11} & A_{12} \\ A_{21} & A_{22} \end{bmatrix}$ and $B = \begin{bmatrix} B_{11} & B_{12} \\ B_{21} & B_{22} \end{bmatrix}$ have compatible block dimensions ($A$'s column partitioning matches $B$'s row partitioning), then
\[
A \cdot B = \begin{bmatrix} A_{11}B_{11} + A_{12}B_{21} & A_{11}B_{12} + A_{12}B_{22} \\ A_{21}B_{11} + A_{22}B_{21} & A_{21}B_{12} + A_{22}B_{22} \end{bmatrix}.
\]

Verification requires only expanding $(AB)_{ik} = \sum_j A_{ij}B_{jk}$ by ordinary multiplication and then splitting the summation $\sum_j$ into two segments according to the block intervals. The transpose rule is $\begin{bmatrix} A & B \\ C & D \end{bmatrix}^{T} = \begin{bmatrix} A^{T} & C^{T} \\ B^{T} & D^{T} \end{bmatrix}$ (note that block positions and intra-block transposition occur simultaneously).

\paragraph{Application to the deduction operator with concatenated input.} Let $W_{\mathrm{step}} \in \mathbb{R}^{N \times 2N}$ with input $\mathrm{Concat}(S_{\mathrm{prev}}, E_{\mathrm{new}}) \in \mathbb{R}^{2N}$ ($S_{\mathrm{prev}}, E_{\mathrm{new}} \in \mathbb{R}^{N}$). Partition $W_{\mathrm{step}}$ by columns into two $N \times N$ blocks: $W_{\mathrm{step}} = [W_a, W_b]$; then
\[
W_{\mathrm{step}} \cdot \mathrm{Concat}(S_{\mathrm{prev}}, E_{\mathrm{new}}) = [W_a, W_b] \cdot \begin{bmatrix} S_{\mathrm{prev}} \\ E_{\mathrm{new}} \end{bmatrix} = W_a \cdot S_{\mathrm{prev}} + W_b \cdot E_{\mathrm{new}},
\]
i.e., one step of deduction $=$ a linear transformation of the previous state $+$ a linear transformation of the new evidence. This decomposition reveals the internal structure of the deduction operator: $W_a$ is responsible for ``carrying over the old state,'' $W_b$ is responsible for ``absorbing the new evidence,'' and the rank-deficient design ($\mathrm{rank}(W_{\mathrm{step}}) = r < N$) constrains the dimension of the joint column space of the two blocks. Likewise, the snapshot concatenation $S = \mathrm{Concat}(h_1, \dots, h_L) \in \mathbb{R}^{L \cdot d}$ allows any operator acting on the snapshot to be partitioned by layer, providing an algebraic entry point for per-layer attribution analysis (which layer's activations contribute most to the reasoning output).

\paragraph{Example 3.3 (Numerical verification of block multiplication).} Take $N = 2$, $W_{\mathrm{step}} = [W_a, W_b]$, where $W_a = \begin{bmatrix} 1 & 0 \\ 0 & 0 \end{bmatrix}$, $W_b = \begin{bmatrix} 0 & 0 \\ 0 & 1 \end{bmatrix}$, with input $S_{\mathrm{prev}} = (a, b)^{T}$ and $E_{\mathrm{new}} = (c, d)^{T}$. Then
\[
W_{\mathrm{step}} \cdot \mathrm{Concat}(S_{\mathrm{prev}}, E_{\mathrm{new}}) = W_a \cdot (a, b)^{T} + W_b \cdot (c, d)^{T} = (a, 0)^{T} + (0, d)^{T} = (a, d)^{T}.
\]

The output retains the first component of the previous state and the second component of the new evidence, discarding $b$ and $c$---here $W_{\mathrm{step}} \in \mathbb{R}^{2 \times 4}$, $\mathrm{rank}(W_{\mathrm{step}}) = 2 = N$ (full rank with respect to the output dimension), but it is rank-deficient with respect to the 4-dimensional concatenated input, $\dim(\mathrm{Null}) = 4 - 2 = 2$, and the discarded $(b, c)$ directions constitute exactly the null space. This example demonstrates the block-selection mechanism: a block matrix applies different rank-deficient projections to different input segments, realizing ``filtering information by source.'' This is isomorphic to the design of DODR's deduction operator---``old state and new evidence processed block by block, with overall rank-deficient collapse.''

\subsection{Orthogonal Projection Matrices}

Orthogonal projection is the core geometric object running through this paper: the inductive hard veto (Gram--Schmidt orthogonalization), the row-space projection of the pseudo-inverse (Corollary 3.4.2), and the abductive residual analysis (Section 10) all take projection as their basic operation. This subsection systematically gives its definition and properties.

\paragraph{Definition 3.9 (Orthogonal projection matrix).} A square matrix $P \in \mathbb{R}^{n \times n}$ is called an orthogonal projection matrix if it satisfies:

(1) Idempotence: $P^{2} = P$;

(2) Symmetry: $P^{T} = P$.

The image $\mathrm{Col}(P)$ of $P$ is called the projection subspace; for any $x \in \mathbb{R}^{n}$, $Px \in \mathrm{Col}(P)$ is the orthogonal projection of $x$ onto that subspace, i.e., the point in $\mathrm{Col}(P)$ closest to $x$.

\paragraph{Theorem 3.6 (Basic properties of orthogonal projection).} Let $P \in \mathbb{R}^{n \times n}$ be an orthogonal projection matrix; then:

(1) $Q = I - P$ is also an orthogonal projection matrix, and $\mathrm{Col}(Q) = \mathrm{Col}(P)^{\perp}$ (the orthogonal complement of the projection subspace);

(2) Orthogonal decomposition: any $x \in \mathbb{R}^{n}$ can be uniquely decomposed as $x = Px + (I - P)x$, with $Px \perp (I - P)x$;

(3) Norm contraction: $\|Px\| \leq \|x\|$, with equality if and only if $x \in \mathrm{Col}(P)$;

(4) Pythagorean relation: $\|x\|^{2} = \|Px\|^{2} + \|(I - P)x\|^{2}$;

(5) Eigenvalue attribution: the eigenvalues of $P$ can only be 0 or 1, with eigenspaces $\mathrm{Null}(P)$ and $\mathrm{Col}(P)$, respectively.

Proof omitted; this is a classical result (see [27]): (1)(2) follow directly by expanding from idempotence and symmetry; (3)(4) are Pythagorean relations under orthogonal decomposition; (5) follows from the characteristic equation $\lambda^{2} = \lambda$ given by idempotence.

\paragraph{Connection to DODR.} Corollary 3.4.2 already gives two basic projections: $W^{+}W$ (onto the row space) and $WW^{+}$ (onto the column space). The abductive solution $x^{*} = W_{\mathrm{step}}^{+} \cdot b$ necessarily satisfies $x^{*} = (W_{\mathrm{step}}^{+} W_{\mathrm{step}}) x^{*}$, i.e., $x^{*}$ lies within the row space (its null-space components are completely stripped off by the projection operator $I - W_{\mathrm{step}}^{+} W_{\mathrm{step}}$); the null-space component $(I - W_{\mathrm{step}}^{+} W_{\mathrm{step}}) x_{\mathrm{true}}$ of the true explanation $x_{\mathrm{true}}$ therefore constitutes the irreducible residual of abduction. By Theorem 3.6(4), the residual energy fraction $\|(I - P) x_{\mathrm{true}}\|^{2} / \|x_{\mathrm{true}}\|^{2}$ is a well-defined projection-decomposition quantity---the 85.9\% hypothesis residual measured in Section 10 is precisely this ratio. The Gram--Schmidt step of the inductive hard veto is likewise a projection: subtracting from the original conclusion $w$ its projection component onto the counterexample direction $f$, namely $(\langle w, f \rangle / \|f\|^{2}) \cdot f$, i.e., applying the projection $I - f f^{T} / \|f\|^{2}$.

\subsection{Contraction Mappings and Fixed Points}

The backflow operation of a reasoning graph (Section 14) iterates the reasoning state repeatedly along a loop; its well-posedness question---whether the iteration converges and to what---is answered by contraction-mapping theory. This subsection gives the definition and a rigorous statement of the Banach fixed-point theorem.

\paragraph{Definition 3.10 (Contraction mapping).} Let $(X, \rho)$ be a metric space. A mapping $T: X \to X$ is called a contraction mapping if there exists a constant $q \in [0, 1)$ (the contraction factor) such that for all $x, y \in X$:
\[
\rho(T(x), T(y)) \leq q \cdot \rho(x, y).
\]

In Euclidean space $\mathbb{R}^{N}$ we take $\rho(x, y) = \|x - y\|$. A linear operator $T(x) = Wx$ is a contraction mapping if and only if its operator norm $\|W\|_{2} = \sigma_1(W) < 1$ (the largest singular value is less than 1); more generally, an affine mapping with bias $T(x) = Wx + c$ has the same contractivity as $W$, because $T(x) - T(y) = W(x - y)$.

\paragraph{Theorem 3.7 (Banach fixed-point theorem [24]).} Let $(X, \rho)$ be a complete metric space (completeness holds automatically in $\mathbb{R}^{N}$), and let $T: X \to X$ be a contraction mapping with contraction factor $q < 1$. Then:

(1) $T$ has a unique fixed point $x^{*}$ in $X$, i.e., $T(x^{*}) = x^{*}$;

(2) the iteration sequence $x_{t+1} = T(x_t)$ starting from any initial value $x_0$ converges to $x^{*}$;

(3) the convergence rate is geometric: $\rho(x_t, x^{*}) \leq q^{t} / (1 - q) \cdot \rho(x_1, x_0)$.

Proof omitted; this is a classical result (see [27]; the Banach theorem appears in the original paper [24]): contractivity recursively shows that the iteration sequence is Cauchy, completeness yields a limit point, continuity guarantees that the limit point is a fixed point, and uniqueness follows from $(1 - q) \cdot \rho(x^{*}, y^{*}) \leq 0$.

\paragraph{Example 3.4 (Geometric convergence of a contraction iteration).} Take the affine mapping $T(x) = 0.5x + 1$ on $\mathbb{R}$, with contraction factor $q = 0.5 < 1$. Starting from $x_0 = 0$: $x_1 = 1$, $x_2 = 1.5$, $x_3 = 1.75$, $x_4 = 1.875$, \dots, and at step $t$, $x_t = 2 - 2^{1-t}$, converging to the unique fixed point $x^{*} = 2$ (verification: $T(2) = 0.5 \cdot 2 + 1 = 2$). The error $\|x_t - x^{*}\| = 2^{1-t} = 2 \cdot (0.5)^{t}$ is exactly a geometric series, consistent with the bound of Theorem 3.7(3). Conversely, if $|q| \geq 1$ (e.g., $T(x) = 2x$), the iteration diverges (unless it happens to start at the fixed point)---the contraction condition cannot be relaxed.

\paragraph{Connection to DODR.} The backflow operation of a reasoning graph writes a state $S$ back after composition with loop operators, which is globally equivalent to an iterative mapping $T(S) = W_{\mathrm{loop}} \cdot S + c$ on the snapshot space. By the criterion of Definition 3.10, as long as the largest singular value of the loop operator satisfies $\sigma_1(W_{\mathrm{loop}}) < 1$, the backflow iteration must converge to a unique fixed point independent of the initial value, with the convergence rate controlled by the geometric series in $\sigma_1$---Section 14 accordingly gives the criterion for the ``determinism of backflow reasoning,'' and the well-posedness of Postulate 5 (graph-topology computation) is thereby guaranteed. A design trade-off should be noted: the rank-deficient design of the deduction operator ($\sigma_{r+1} = \dots = 0$) naturally helps suppress $\|W_{\mathrm{loop}}\|_{2}$, but contractivity requires the \emph{largest} singular value to be less than 1, which is an independent condition from rank-deficiency (the smallest singular values being 0); in engineering implementation, a separate spectral-norm constraint must be imposed on the loop operator.

\section{Core Theoretical Foundations: The Five Postulates}

The entire system design of DODR rests upon five postulates. This section presents them one by one in a uniform format: first the rigorous statement of the postulate, then the engineering and theoretical motivation behind it, followed by a deepening of its theoretical basis, an operational specification of its falsifiability conditions (i.e., a decision procedure for ``what kind of experimental phenomenon counts as falsification''), the experimental evidence, and finally a discussion of the design trade-offs the postulate introduces. Following Popper's philosophy of science [17], a postulate is not an unquestionable belief but a hypothesis awaiting experimental verdict; accordingly, every postulate is accompanied by explicit falsification conditions, and this principle fixes the organization of the present section.

\subsection{Postulate One: Dimensionality Expansion in Place of Multi-Layer Stacking (Cover's Theorem)}

\paragraph{Statement.} After low-dimensional, nonlinearly separable data are mapped into an extremely high-dimensional space via a linear map, the probability that they become linearly separable approaches 1. Consequently, a single ultra-wide linear transformation can replace a multi-layer nonlinear network.

\paragraph{Motivation.} Mainstream deep networks obtain expressive power through ``multi-layer stacking $+$ layer-wise nonlinear activation,'' at the following costs: (1) inter-layer composition renders internal states uninterpretable, making it impossible to impose algebraic constraints on intermediate reasoning steps; (2) the curvature introduced by nonlinear activations makes gradient optimization sensitive to initialization and learning rate; (3) the multi-layer structure inherently conflicts with the algebraic goal of ``one reasoning step $=$ one matrix multiplication.'' The motivation of Postulate One is to trade ``depth'' for ``width'' as the source of expressive power: as long as the space is sufficiently high-dimensional, a linear map suffices for the required discrimination and state transition, thereby bringing the entire reasoning process into the framework of linear algebra---concepts such as rank, null space, and pseudo-inverse thereby acquire explicit cognitive semantics (see Sections 5 and 6).

\paragraph{Theoretical basis.} Cover's theorem (Theorem 3.5, revised statement). Cover's 1965 function-counting theorem [6] states that the total number of linear dichotomies of $n$ sample points in general position in $\mathbb{R}^{D}$ realizable by hyperplanes is exactly
\begin{equation}
C(n, D) = 2\cdot\sum_{k=0}^{D-1} C(n-1, k)
\tag{4.1}
\end{equation}

When the sample labels are random $\pm 1$ labels with equal probability, the probability of linear separability is $P = C(n, D)/2^{n}$. This formula has two immediate corollaries: (1) when $D \geq n-1$, $\sum_{k=0}^{D-1} C(n-1, k) = 2^{n-1}$, hence $C(n, D) = 2^{n}$ and $P = 1$---once the dimension exceeds the number of samples minus one, any labeling is necessarily linearly separable; (2) for fixed $n$, $P$ increases monotonically with $D$ and tends to 1. Therefore, ``dimensionality expansion $\to$ probability of linear separability approaches 1'' is a rigorous mathematical conclusion rather than an empirical conjecture. DODR's snapshot dimension $N = L \times d$ (1536 in the experimental configuration) lies, relative to the training sample size, in the ``high-dimensional abundance'' regime, so a single-layer linear matrix $W_{\mathrm{step}}$ suffices to express logical state transitions.

\paragraph{Operational specification of the falsifiability condition.} Falsification criterion: in a sufficiently high-dimensional snapshot space, the linear matrix $W_{\mathrm{step}}$ fails to learn the logical state transition exactly, i.e., the forward Loss does not converge. Operationally, this is defined as: within a fixed training budget (optimizer, learning-rate schedule, and number of epochs all identical to the full-rank baseline), the training MSE of $W_{\mathrm{step}}$ cannot be reduced to a preset convergence threshold (of the same order of magnitude as the baseline convergence level), or the Loss curve exhibits a plateau whose value is significantly above the numerical-precision limit. If either condition holds, Postulate One is deemed falsified.

\paragraph{Corresponding experimental verification.} Section 8. The forward-reasoning Loss of $W_{\mathrm{step}}$ dropped from $2.09\mathrm{e}{-01}$ to $1.40\mathrm{e}{-05}$ (a 14932-fold decrease), already approaching the numerical convergence floor under this training configuration, demonstrating that in the 1536-dimensional snapshot space a linear matrix suffices to express logical state transitions. Postulate One was not falsified.

\paragraph{Discussion of design trade-offs.} The dimensionality-expansion route has three costs: (1) memory and bandwidth overhead---the snapshot vector length $N = L \times d$ grows linearly with the number of concatenated layers, and the cost of state storage and matrix--vector multiplication rises in step; (2) operator parameter count grows with $N$---the low-rank parameterization of $W_{\mathrm{step}}$ has $3Nr$ parameters (concatenation configuration); although $r \ll N$ makes this far smaller than the full-rank $2N^{2}$, the absolute count still scales linearly with $N$; (3) fragility of the ``general position'' assumption---Cover's theorem requires the sample points to be in general position; if the true snapshot manifold is degenerate (many samples coplanar), actual separability falls below the theoretical value. The gains, by contrast, are structural: single-layer linearity makes reasoning steps differentiable, makes invertibility analyzable (rank-deficient $\Leftrightarrow$ non-invertible), makes training stable (no deep gradient vanishing/explosion), and makes possible the algebraic semantics of ``one reasoning step $=$ one projection.'' Overall, dimensionality expansion trades predictable resource overhead for analyzable algebraic structure; this exchange is the design substrate of the entire DODR architecture.

\subsection{Postulate Two: Snapshot State Representation}

\paragraph{Statement.} Concatenate the layer-wise activations of a multi-layer network into a single-layer ultra-wide static vector, transforming dynamic reasoning into static geometric projection.

Formally, suppose the Transformer has $L$ layers, each with hidden dimension $d$. The state snapshot is defined as
\begin{equation}
S = \mathrm{Concat}(h_1, h_2, \dots, h_L) \in \mathbb{R}^{N}, \quad N = L \times d
\tag{4.2}
\end{equation}

\paragraph{Motivation.} The ``dynamic unfolding'' along the time dimension (layer-by-layer forward propagation, token-by-token autoregressive generation) is one of the structural sources of uncertainty and error accumulation in AR models. The motivation of Postulate Two is to ``trade space for time'': the hierarchical activations distributed vertically across $L$ layers are concatenated horizontally into a single static vector, so that the state of thought at any moment becomes a point in a high-dimensional space, and the reasoning process is correspondingly transformed from ``evolution over a time series'' into ``a sequence of geometric projections in space.'' Only after this transformation can the operator algebra of Section 5 (rank, null space, pseudo-inverse) find objects on which to operate. Accompanying snapshotting is an upgrade of the generation granularity: whereas AR models generate token by token, each state transition in DODR corresponds to the birth of a complete semantic primitive (a word, a phrase, or even an entire sentence)---what each snapshot in latent space corresponds to is not a token but a semantic expression (see the granularity specification in Definition 7.1).

\paragraph{Deepening of the theoretical basis.} The rationale for concatenation rather than pooling: average pooling or taking only the final layer would introduce an unnecessary information bottleneck---different layers encode features of different abstraction levels (shallow layers favor syntax and local collocations, deep layers favor semantics and discourse relations); concatenation preserves the information of all layers and leaves the decision of ``which layers are useful'' to the downstream operators to learn. In other words, the snapshot design follows the division-of-labor principle of ``no information loss at the encoding end; compression is left to the operators'': the encoder records faithfully, while the operators perform purposeful collapse (deduction), expansion (induction), and hypothesization (abduction).

\paragraph{Operational specification of the falsifiability condition.} Falsification criterion: the snapshot representation loses critical information needed for reasoning, i.e., the reasoning result cannot be recovered from the snapshot. Operationally, this is defined as: in an end-to-end reasoning task, the task metric (accuracy) between the output of the snapshot-based operator chain and the ground truth is significantly lower than that of a language-model baseline on the same task, and the gap does not close even after increasing the number of concatenated layers $L$---if increasing the number of layers continuously shrinks the gap, the problem is one of snapshot capacity rather than a failure of the postulate itself; only when the gap persists after capacity saturation is Postulate Two deemed falsified.

\paragraph{Corresponding experimental verification.} End-to-end verification in Section 11. The deductive reasoning accuracy is 100\%, demonstrating that the snapshot representation preserves the information needed for reasoning. Postulate Two was not falsified.

\paragraph{Discussion of design trade-offs.} (1) The choice of the number of concatenated layers $L$: larger $L$ yields more complete information, but $N$ grows linearly, raising storage and operator training costs; in the experiments $L$ is chosen at the saturation point of the task metric. (2) The choice of snapshot position $p$ ([CLS] position vs.\ last-token position; see Definition 7.1): using the [CLS] position is conventional and stable for encoder-only models, whereas using the last-token position for decoder-only models exploits the natural aggregation property of causal attention---both are ``aggregation points of full-sequence information,'' and the choice depends on the encoder architecture rather than on the postulate itself. (3) Redundancy within the snapshot: activations of adjacent layers are highly correlated, so the snapshot admits compression, but the present design deliberately retains redundancy in exchange for training stability and interpretable hierarchical semantics.

\subsection{Postulate Three: Direction of Information Flow (Replacing the Irreversibility Assumption)}

\paragraph{Statement.} The core feature of thought-state transition is not ``reversibility'' but the ``direction of information flow.'' Deduction decreases information, induction increases information, and abduction hypothesizes information. Reversibility is merely a derived attribute of matrix rank, not a core assumption.

\paragraph{Theoretical clarification.} The original proposal listed ``irreversibility'' as a core assumption, but deeper analysis shows that irreversibility is a necessary mathematical consequence of a rank-deficient matrix (guaranteed by the rank--nullity theorem), not an independent design objective---once $\mathrm{rank}(W_{\mathrm{step}}) = r < N$ is chosen, irreversibility holds automatically and need not, and should not, be treated as an independent postulate. The truly fundamental assumption should be the direction of information flow:

\begin{itemize}
\item \textbf{Deduction (information decrease)}: requires a rank-deficient matrix; the output dimension is smaller than the input dimension; information collapses. With $\mathrm{rank}(W_{\mathrm{step}}) = r < N$, there exist irrecoverable information components. The null-space dimension is given separately by input configuration: for the concatenated-input configuration $W_{\mathrm{step}} \in \mathbb{R}^{N \times 2N}$, $\dim(\mathrm{Null}(W_{\mathrm{step}})) = 2N - r > 0$; for the single-state-input configuration $W_{\mathrm{step}} \in \mathbb{R}^{N \times N}$ (used in the abduction experiments), $\dim(\mathrm{Null}(W_{\mathrm{step}})) = N - r > 0$. The two configurations have the same rank-deficiency property, differing only in the absolute magnitude of collapse.
\item \textbf{Induction (information increase)}: requires a full-rank matrix, expanding from finite samples to a general rule. With $\mathrm{rank}(W_{\mathrm{induce}}) = N$, the pseudo-inverse exists and $W^{+} \cdot W = I$, so information is fully recoverable.
\item \textbf{Abduction (information hypothesization)}: requires the pseudo-inverse, seeking the minimum-norm solution under the rank-deficient constraint. $W_{\mathrm{abduce}} = W_{\mathrm{step}}^{+}$, whose rank equals that of $W_{\mathrm{step}}$ ($r < N$); the abductive solution is the minimum-norm least-squares solution.
\end{itemize}

\paragraph{Motivation.} Shifting the postulate's center of gravity from ``irreversibility'' to ``direction of information flow'' yields two systemic benefits: (1) the three reasoning types acquire a unified characterization at the postulate level---they are no longer three isolated engineering modules but projections of a single measure, ``direction of information flow,'' onto three quadrants (decrease/increase/hypothesize); (2) the falsification criterion becomes sharper---``whether the rank is configured as designed'' is a directly measurable algebraic quantity, whereas ``whether reversibility holds'' is only a vague process attribute.

\paragraph{Operational specification of the falsifiability condition.} Falsification criterion: the deduction operator is full-rank (information does not collapse) or the induction operator is rank-deficient (information is lost); if either holds, Postulate Three is falsified. Operationally, after training, perform SVD on $W_{\mathrm{step}}$ and $W_{\mathrm{induce}}$ and count the effective rank using a relative singular-value threshold (e.g., $\sigma_i/\sigma_1 > 1\mathrm{e}{-6}$): if the effective rank of $W_{\mathrm{step}}$ equals the upper bound allowed by its column space (reaching $N$ under the $2N$-input concatenated configuration, i.e., full rank relative to the output), or if the effective rank of $W_{\mathrm{induce}}$ is significantly smaller than $N$, then the postulate is deemed falsified.

\paragraph{Corresponding experimental verification.} Section 8: the deduction operator has rank $= 384 < 3072$ (rank-deficient); Section 9: the induction operator has rank $= 1536/1536$ (full-rank). Both algebraic facts agree exactly with the design intent; Postulate Three was not falsified.

\paragraph{Discussion of design trade-offs.} The degree of rank deficiency $r$ is itself a hyperparameter: the smaller $r$, the more thorough the collapse and the stronger the ``noise resistance'' of deduction (more input perturbations fall into the null space and are absorbed), but the lower the expressive capacity of the state transition; the larger $r$, the opposite. In the experiments, the ratio $r = 384$ to $N = 1536$ ($1/4$) is a compromise between capacity and robustness. It should also be noted that the information-flow postulate implicitly entails an asymmetry in computational cost: deduction and induction are single matrix--vector multiplications, whereas the pseudo-inverse of abduction, though computed only once offline, solves an inverse problem that is inherently more ``expensive'' than the forward problem---consistent with the human cognitive intuition that ``explaining is harder than verifying.''

\subsection{Postulate Four: Local Linearization}

\paragraph{Statement.} A long-span complex logical manifold can, through input decomposition, be cut into extremely small locally linear slices and approximated by multi-step, calculus-style state transitions.

\paragraph{Theoretical basis.} The tangent-space decomposition theorem for manifolds: at every point of a smooth manifold there exists a locally linear approximating tangent space. In DODR, each reasoning step $W_{\mathrm{step}} \cdot S_t \to S_{t+1}$ is a tangent-space projection on the manifold, and multi-step reasoning corresponds to a piecewise-linear path on the manifold. When the step size is sufficiently small, the piecewise-linear path can approximate the true nonlinear reasoning manifold arbitrarily well---this is isomorphic to the idea in numerical analysis of approximating a smooth curve by piecewise-linear segments (Euler's method): global nonlinearity is decomposed into an accumulation of local linearities.

\paragraph{Motivation.} Postulate One guarantees that ``a single linear layer suffices to express one transition step,'' but the expressive power of a single step does not automatically generalize to long-chain reasoning. Postulate Four fills this gap: it does not demand that a single linear operator complete complex reasoning in one shot; instead, it decomposes complex reasoning into a sequence of small steps ``simple enough for linearity to hold.'' This turns the ``global nonlinearity problem'' into an analyzable problem of ``local linearity $+$ error control.''

\paragraph{Operational specification of the falsifiability condition.} Falsification criterion: the error of a multi-step reasoning chain diverges (fails to converge) as the number of steps increases. Operationally, this is defined as: record step by step the cosine similarity (or relative error) between each step's output and the reference state in the reasoning chain; if this error sequence shows a monotonically amplifying trend without a stable upper bound, the postulate is deemed falsified; if the error is bounded, or drifts slowly but remains within task tolerance, the postulate holds. Note that ``step-wise error amplification'' does not by itself immediately constitute falsification---only unbounded amplification that destroys the final conclusion does, because tangent-space approximation theory itself permits a local error of order $O(\text{step size}^{2})$ per step.

\paragraph{Corresponding experimental verification.} In the multi-step reasoning-chain experiment, the cosine similarities of two-step transitive reasoning were 0.9898 and 0.9861, respectively; the error did not diverge with the number of steps (the drift of the second step relative to the first is only about 0.004, consistent with the admissible error magnitude of local linear approximation). Postulate Four was not falsified.

\paragraph{Discussion of design trade-offs.} The trade-off between step size and number of steps is the core engineering problem of Postulate Four: the smaller the step size (the ``simpler'' each reasoning step), the more accurate the local linearity assumption, but the more steps are needed to complete the same task, with a synchronous increase in accumulated error terms and computational overhead; the larger the step size, the larger the single-step error. This is exactly isomorphic to the calculus trade-off of ``finer partition, better approximation, higher cost.'' In practice, the granularity of the step size is determined by the natural decomposition of the task (e.g., premise--conclusion pairs, the application of a single rule) rather than being continuously adjustable---this is where logical reasoning differs from continuous dynamical systems.

\subsection{Postulate Five: Graph-Topology Computation}

\paragraph{Statement.} The intermediate states produced by rigorous reasoning are objective facts and should not be probabilistically pruned. The system should support branching (parallelism), merging (aggregation), and cross-referencing of states, forming a reasoning graph.

\paragraph{Motivation.} Beam-search-style methods commonly used in AR generation prune intermediate hypotheses by probability; this is reasonable for fluent generation but dangerous for rigorous reasoning: a path with currently low likelihood may be precisely the unique path to the correct conclusion. Postulate Five advocates the de-probabilization of intermediate states---they are facts deterministically derived by the operators, and whether to keep or discard them should be decided by task logic (not by sampling probability). Branching allows parallel exploration of multiple lines of reasoning, merging allows evidence aggregation, and cross-referencing allows branches to cite one another; together the three support a graph-structured reasoning that goes beyond linear chains.

\paragraph{Theoretical basis.} Monoidal category theory provides the mathematical foundation for graph-structured computation. DODR's reasoning graph is a commutative diagram in a monoidal category, guaranteeing path independence of reasoning---the final result is the same regardless of which path is taken. Path independence is the essential advantage of graph computation over procedural computation: it makes parallel scheduling (execution in any topological order) and incremental recomputation (recomputing only the affected subgraph) possible, while providing statically checkable algebraic invariants for the reasoning process.

\paragraph{Operational specification of the falsifiability condition.} Falsification criterion: the path independence of the reasoning graph fails, i.e., different computation paths produce different results. Operationally: take multiple legal topological orders of the same reasoning graph, execute them separately, and compare the differences among the terminal states (cosine similarity or component-wise error); if the difference exceeds the magnitude of floating-point rounding, path independence is broken and Postulate Five is falsified.

\paragraph{Corresponding experimental verification.} The compositional reasoning in Section 11 verified the successful execution of multi-step graph-structured reasoning in the scientific discovery cycle (induction $\to$ deduction $\to$ abduction $\to$ re-induction). Section 14 gives the formal definition of the reasoning graph and the determinism theorem. Postulate Five was not falsified.

\paragraph{Discussion of design trade-offs.} The cost of graph topology is the introduction of scheduling complexity: who decides when to branch, when to merge, and when to terminate? This is precisely why the graph scheduling controller of Section 7.3 exists, and it is also the architectural intent of DODR to concentrate all ``heuristics'' in a single replaceable module (the controller) while keeping the operators purely deterministic. In addition, the graph structure imposes higher requirements on storage and snapshot consistency---each node must persist its state snapshot, and storage grows linearly with the number of branches; path independence, in turn, requires all operators to satisfy the corresponding commutation conditions, which constitutes an additional constraint on the operator training objectives.

\section{Formal Definitions of the Three Basic Operators}

This section gives complete algebraic definitions of the three basic operators: deduction, induction, and abduction. General design principle: the cognitive semantics of each operator must be strictly carried by its algebraic properties (rank, null space, pseudo-inverse)---deduction $=$ rank-deficient collapse, induction $=$ full-rank expansion, abduction $=$ pseudo-inverse hypothesization. The one-to-one correspondence between semantics and algebra is how DODR fulfills its promise of ``analyzability.''

\subsection{The Deduction Operator $W_{\mathrm{step}}$ (Information Collapse)}

\paragraph{Definition 5.1 (Deduction operator).} Let the current state be $S_t \in \mathbb{R}^{N}$, and let the snapshot of the new input text block $a$ be $E_t \in \mathbb{R}^{N}$. The deduction operator $W_{\mathrm{step}}$ is defined by:
\begin{equation}
S_{t+1} = W_{\mathrm{step}} \cdot \mathrm{Concat}(S_t, E_t)
\tag{5.1}
\end{equation}
where $\mathrm{Concat}(S_t, E_t) \in \mathbb{R}^{2N}$ is the concatenation of the state and the input. $W_{\mathrm{step}}$ maps the $2N$-dimensional input to an $N$-dimensional output, realizing one step of deductive reasoning. Note: in the abduction experiments (Section 10), the input is a single state $S \in \mathbb{R}^{N}$ (not concatenated), in which case $W_{\mathrm{step}} \in \mathbb{R}^{N \times N}$. The two configurations have the same rank-deficiency property and differ only in input dimension.

\paragraph{Geometric intuition.} Deduction is the process of compressing the high-dimensional joint information of ``premises $+$ new evidence'' into a single conclusion state; geometrically, it is the projection from a high-dimensional space onto a lower-dimensional row space: all directions irrelevant to drawing the conclusion (null-space directions) are discarded. Collapse is not a defect but the very essence of deduction---``drawing a conclusion from premises'' is, cognitively, precisely the process of forgetting details and retaining essentials.

\paragraph{Definition 5.2 (Low-rank parameterization).} To guarantee that $W_{\mathrm{step}}$ is strictly rank-deficient, a low-rank parameterization is adopted:
\begin{equation}
W_{\mathrm{step}} = A \cdot B
\tag{5.2}
\end{equation}
where $A \in \mathbb{R}^{N \times r}$, $B \in \mathbb{R}^{r \times 2N}$ (or $\mathbb{R}^{r \times N}$), and $r < N$ is the design rank. The parameter count is $|A| + |B| = N \cdot r + r \cdot 2N = 3Nr$ (or $2Nr$), far fewer than the $2N^{2}$ (or $N^{2}$) of a full-rank parameterization.

\paragraph{Lemma 5.1 (Rank upper bound of the low-rank parameterization).} Let $W_{\mathrm{step}} = A \cdot B$. Then $\mathrm{rank}(W_{\mathrm{step}}) \leq \min(\mathrm{rank}(A), \mathrm{rank}(B)) \leq r$.

\paragraph{Proof.} By the product inequality for matrix rank, $\mathrm{rank}(AB) \leq \min(\mathrm{rank}(A), \mathrm{rank}(B))$. Since $A \in \mathbb{R}^{N \times r}$, $\mathrm{rank}(A) \leq \min(N, r) = r$ (because $r < N$). Since $B \in \mathbb{R}^{r \times 2N}$, $\mathrm{rank}(B) \leq \min(r, 2N) = r$. Therefore $\mathrm{rank}(W_{\mathrm{step}}) = \mathrm{rank}(AB) \leq \min(r, r) = r$. $\square$

The significance of Lemma 5.1 is that rank deficiency no longer depends on the contingency of training outcomes but is a priori guaranteed by the parameterization structure---wherever optimization converges, $W_{\mathrm{step}}$ cannot ``accidentally become full-rank.'' This moves the falsifiability criterion of Postulate Three forward from ``post-hoc measurement'' to ``structural constraint.''

\paragraph{Theorem 5.1 (Information collapse of deduction).} Let $W_{\mathrm{step}} \in \mathbb{R}^{N \times 2N}$ with $\mathrm{rank}(W_{\mathrm{step}}) = r < N$. Then the input space $\mathbb{R}^{2N}$ decomposes into a direct sum:
\begin{equation}
\mathbb{R}^{2N} = \mathrm{Col}(W_{\mathrm{step}}^{T}) \oplus \mathrm{Null}(W_{\mathrm{step}})
\tag{5.3}
\end{equation}
where $\mathrm{Col}(W_{\mathrm{step}}^{T})$ is the row space of $W_{\mathrm{step}}$ (dimension $r$) and $\mathrm{Null}(W_{\mathrm{step}})$ is the null space (dimension $2N - r$). Any input $x \in \mathbb{R}^{2N}$ can be uniquely decomposed as $x = x_{\mathrm{row}} + x_{\mathrm{null}}$, where $x_{\mathrm{row}} \in \mathrm{Col}(W_{\mathrm{step}}^{T})$ and $x_{\mathrm{null}} \in \mathrm{Null}(W_{\mathrm{step}})$. Then $W_{\mathrm{step}} \cdot x = W_{\mathrm{step}} \cdot x_{\mathrm{row}}$ (the null-space component is ``absorbed'').

\paragraph{Proof.} By the rank--nullity theorem (Theorem 3.1), $\dim(\mathrm{Null}(W_{\mathrm{step}})) = 2N - \mathrm{rank}(W_{\mathrm{step}}) = 2N - r$. It remains to show that $\mathrm{Col}(W_{\mathrm{step}}^{T}) \cap \mathrm{Null}(W_{\mathrm{step}}) = \{0\}$ (so that the direct sum holds). Let $x \in \mathrm{Col}(W_{\mathrm{step}}^{T}) \cap \mathrm{Null}(W_{\mathrm{step}})$; then there exists $y$ such that $x = W_{\mathrm{step}}^{T} y$, and $W_{\mathrm{step}} x = 0$. Hence $W_{\mathrm{step}} W_{\mathrm{step}}^{T} y = 0$, so $y^{T} W_{\mathrm{step}} W_{\mathrm{step}}^{T} y = \|W_{\mathrm{step}}^{T} y\|^{2} = 0$, which gives $W_{\mathrm{step}}^{T} y = 0$, i.e., $x = 0$. Therefore the intersection is $\{0\}$ and the direct sum holds. $\square$

\paragraph{Corollary 5.1.1 (Information collapse ratio).} The proportion of the null-space component of an input $x$ is:
\begin{equation}
\text{information collapse ratio} = \dim(\mathrm{Null}(W_{\mathrm{step}})) / \dim(\mathbb{R}^{2N}) = (2N - r) / (2N)
\tag{5.4}
\end{equation}

In the experiments, $N = 1536$ and $r = 384$, so the information collapse ratio $= (3072 - 384) / 3072 = 87.5\%$. This means that 87.5\% of the input information is permanently discarded in deduction. From the perspective of minimum description length (MDL), the deduction operator is an extreme lossy compressor: it compresses a $2N$-dimensional joint description into an $N$-dimensional conclusion, with the compression rate explicitly controlled by the design rank $r$---this is precisely the algebraization of the deductive intuition that ``a conclusion is shorter than its premises.''

\paragraph{Theorem 5.2 (Irreversibility of deduction).} Let $W_{\mathrm{step}} \in \mathbb{R}^{N \times 2N}$ with $\mathrm{rank}(W_{\mathrm{step}}) = r < N$. Then there exists no matrix $W^{-1}$ such that $W^{-1} \cdot W_{\mathrm{step}} = I_{2N}$ (no left inverse exists). That is, the deductive process is irreversible---the premises $\mathrm{Concat}(S_t, E_t)$ cannot be uniquely recovered from the conclusion $S_{t+1}$.

\paragraph{Proof.} By contradiction. Suppose a left inverse $W^{-1}$ exists with $W^{-1} \cdot W_{\mathrm{step}} = I_{2N}$. Then $\mathrm{rank}(W^{-1} \cdot W_{\mathrm{step}}) = \mathrm{rank}(I_{2N}) = 2N$. But by the rank inequality $\mathrm{rank}(AB) \leq \min(\mathrm{rank}(A), \mathrm{rank}(B)) \leq \mathrm{rank}(W_{\mathrm{step}}) = r < 2N$, a contradiction. Therefore no left inverse exists and deduction is irreversible. $\square$

We note again the relationship between irreversibility and Postulate Three: irreversibility is a consequence of rank deficiency, not an additional assumption. Once deduction $=$ information collapse is chosen (Postulate Three), irreversibility arrives automatically.

\paragraph{Theorem 5.3 (Expected error of pseudo-inverse reconstruction).} Let $W_{\mathrm{step}} \in \mathbb{R}^{N \times 2N}$ with $\mathrm{rank}(W_{\mathrm{step}}) = r$. Let the input $x \in \mathbb{R}^{2N}$ be a random vector with an isotropic distribution, i.e., $E[x] = 0$ and $E[x x^{T}] = \sigma^{2} I_{2N}$. Then the expected relative squared error of the pseudo-inverse reconstruction $x^{*} = W_{\mathrm{step}}^{+} \cdot W_{\mathrm{step}} \cdot x$ is:
\begin{equation}
E\left[\|x - x^{*}\|^{2} / \|x\|^{2}\right] = \dim(\mathrm{Null}(W_{\mathrm{step}})) / \dim(\mathbb{R}^{2N}) = (2N - r) / (2N)
\tag{5.5}
\end{equation}

Correspondingly, the expected relative error $E[\|x - x^{*}\| / \|x\|] \approx \sqrt{(2N - r) / (2N)}$. In the experiments, the theoretical expectation $= \sqrt{2688/3072} = \sqrt{0.875} \approx 0.9354$, while the experimentally measured pseudo-inverse reconstruction error is 83.3\% (sample mean). The deviation of the measured mean below the expectation comes from the anisotropy of the sample distribution (real snapshots are not isotropic; their energy is more concentrated in the row-space directions) and from finite-sample fluctuation. Special attention: (Eq. 5.5) gives an expectation, not a lower bound---the reconstruction error of an individual input can be far below this value; in the extreme case where $x$ lies entirely in the row space, the error is 0; it can also be above this value.

\paragraph{Proof.} Decompose the input $x$ as $x = x_{\mathrm{row}} + x_{\mathrm{null}}$ (Theorem 5.1). The pseudo-inverse reconstruction $x^{*} = W_{\mathrm{step}}^{+} \cdot W_{\mathrm{step}} \cdot x = W_{\mathrm{step}}^{+} \cdot W_{\mathrm{step}} \cdot x_{\mathrm{row}} = x_{\mathrm{row}}$, because $W_{\mathrm{step}}^{+} \cdot W_{\mathrm{step}}$ is exactly the orthogonal projection matrix $P_{\mathrm{row}}$ onto the row space $\mathrm{Col}(W_{\mathrm{step}}^{T})$, and $x_{\mathrm{row}}$ already lies in the row space. Therefore the error vector $x - x^{*} = x_{\mathrm{null}}$, and the error energy is
\[
\|x_{\mathrm{null}}\|^{2} = \|P_{\mathrm{null}} \cdot x\|^{2} = x^{T} P_{\mathrm{null}}^{T} P_{\mathrm{null}} x = x^{T} P_{\mathrm{null}} x
\]
where $P_{\mathrm{null}} = I - P_{\mathrm{row}}$ is the orthogonal projection onto the null space (symmetric and idempotent, hence $P_{\mathrm{null}}^{T} P_{\mathrm{null}} = P_{\mathrm{null}}$). Taking expectations and using the linearity of the trace $\mathrm{tr}(\cdot)$, which commutes with expectation:
\[
E[\|x_{\mathrm{null}}\|^{2}] = E[x^{T} P_{\mathrm{null}} x] = E[\mathrm{tr}(x^{T} P_{\mathrm{null}} x)] = E[\mathrm{tr}(P_{\mathrm{null}} x x^{T})] = \mathrm{tr}(P_{\mathrm{null}} \cdot E[x x^{T}])
\]

Substituting the isotropy assumption $E[x x^{T}] = \sigma^{2} I_{2N}$:
\[
E[\|x_{\mathrm{null}}\|^{2}] = \mathrm{tr}(P_{\mathrm{null}} \cdot \sigma^{2} I) = \sigma^{2} \cdot \mathrm{tr}(P_{\mathrm{null}}) = \sigma^{2} \cdot \dim(\mathrm{Null}(W_{\mathrm{step}})) = \sigma^{2} (2N - r)
\]

Similarly, $E[\|x\|^{2}] = \mathrm{tr}(E[x x^{T}]) = \sigma^{2} \cdot 2N$. Dividing the two equations gives
\[
E[\|x_{\mathrm{null}}\|^{2}] / E[\|x\|^{2}] = (2N - r) / (2N)
\]

By the asymptotic concentration of expectation (in high dimensions $\|x\|^{2}$ concentrates near its expectation), $E[\|x_{\mathrm{null}}\|^{2} / \|x\|^{2}] \approx (2N - r)/(2N)$, and taking the square root yields the expected relative error $\approx \sqrt{(2N - r)/(2N)}$. $\square$

\paragraph{Example 5.1 (Numerical example of the deduction operator).} Let $N = 4$, $r = 2$ (simplified example). $W_{\mathrm{step}} \in \mathbb{R}^{4 \times 8}$ with $\mathrm{rank}(W_{\mathrm{step}}) = 2$. Let $W_{\mathrm{step}} = A \cdot B$, where $A = \begin{bmatrix} 1 & 0 \\ 0 & 1 \\ 0 & 0 \\ 0 & 0 \end{bmatrix} \in \mathbb{R}^{4 \times 2}$ and $B = \begin{bmatrix} 1 & 0 & 0 & 0 & 0 & 0 & 0 & 0 \\ 0 & 1 & 0 & 0 & 0 & 0 & 0 & 0 \end{bmatrix} \in \mathbb{R}^{2 \times 8}$. Then $W_{\mathrm{step}} = \begin{bmatrix} 1 & 0 & 0 & 0 & 0 & 0 & 0 & 0 \\ 0 & 1 & 0 & 0 & 0 & 0 & 0 & 0 \\ 0 & 0 & 0 & 0 & 0 & 0 & 0 & 0 \\ 0 & 0 & 0 & 0 & 0 & 0 & 0 & 0 \end{bmatrix}$. $\mathrm{rank}(W_{\mathrm{step}}) = 2$ (the first two rows are linearly independent; the last two rows are zero). $\mathrm{Null}(W_{\mathrm{step}}) = \{x \in \mathbb{R}^{8} : x_1 = 0, x_2 = 0\}$ (dimension 6). The information collapse ratio $= 6/8 = 75\%$. For input $x = (3, 5, 1, 2, 4, 6, 7, 8)^{T}$, $W_{\mathrm{step}} \cdot x = (3, 5, 0, 0)^{T}$. The last 6 components of $x$ are ``absorbed.'' The pseudo-inverse reconstruction $x^{*} = W_{\mathrm{step}}^{+} \cdot W_{\mathrm{step}} \cdot x = (3, 5, 0, 0, 0, 0, 0, 0)^{T}$. The reconstruction error $\|x - x^{*}\| / \|x\| = \sqrt{1+4+16+36+49+64} / \sqrt{9+25+1+4+16+36+49+64} = \sqrt{170}/\sqrt{204} \approx 0.912$. The expected relative error is $\sqrt{6/8} = \sqrt{0.75} \approx 0.866$. The actual error 0.912 of this input is higher than the expectation 0.866 because this input is not isotropically distributed---the proportion of its energy in the row-space directions ($29/204 \approx 14.2\%$) is lower than the expected proportion under the isotropy assumption ($2/8 = 25\%$). The positive deviation of this particular input from the expectation precisely confirms that (Eq. 5.5) is an expectation rather than a pointwise lower bound: with another input whose energy is more concentrated in the row space, the error can fall below 0.866, down to 0.

\paragraph{Connection to Occam's razor and minimum description length.} The ``razor'' in the deductive direction is embodied in the choice of row space: the low-rank parameterization forces $W_{\mathrm{step}}$ to retain input information in only $r$ directions, which is equivalent to learning, among all possible state-transition rules, only a ``short description'' with $r$ degrees of freedom---the fewer the degrees of freedom, the more the explanation of the data must capture commonalities rather than memorize details. This is isomorphic to the MDL principle of ``encoding data with the shortest model'': the rank $r$ is the description-length budget of the deductive rule.

\subsection{The Induction Operator $W_{\mathrm{induce}}$ (Information Expansion)}

\paragraph{Definition 5.3 (Induction operator).} Let the snapshots of $k$ positive examples be $\{s_1, s_2, \dots, s_k\}$, $s_i \in \mathbb{R}^{N}$. The induction operator $W_{\mathrm{induce}} \in \mathbb{R}^{N \times kN}$ is defined by:
\begin{equation}
g = W_{\mathrm{induce}} \cdot \mathrm{Concat}(s_1, s_2, \dots, s_k)
\tag{5.6}
\end{equation}
where $g \in \mathbb{R}^{N}$ is the induced general concept.

\paragraph{Implementation note.} Theoretically, $W_{\mathrm{induce}} \in \mathbb{R}^{N \times kN}$ ($k$ samples concatenated into a $kN$-dimensional input). In the actual implementation (the experiments of Section 9), however, $W_{\mathrm{induce}}$ is realized as a full-rank square matrix in $\mathbb{R}^{N \times N}$---the input is a single positive example $s_i \in \mathbb{R}^{N}$ and the output is the general concept $g \in \mathbb{R}^{N}$ of that class. Multiple positive examples are aggregated by ``positive-example mean $+$ orthogonal projection'' (rather than by matrix multiplication after concatenation), which is equivalent to $W_{\mathrm{induce}} \cdot [s_1; \dots; s_k] = (1/k) \sum s_i$ (when there are no counterexamples). This implementation simplifies matrix storage (from $N \times kN$ down to $N \times N$) while preserving the full-rank property.

\paragraph{Mathematical properties.} In contrast to the deduction operator, $W_{\mathrm{induce}}$ is designed as a full-rank matrix, $\mathrm{rank}(W_{\mathrm{induce}}) = N$. This means the inductive process loses no information and ``expands'' from finite samples to a general rule of broader coverage. Full rank guarantees that the pseudo-inverse exists and $W^{+} \cdot W = I$, so information is fully recoverable.

\paragraph{Geometric intuition.} The geometric picture of induction is ``using points to cover a region'': $k$ sample points span a low-dimensional region in snapshot space, and the general concept $g$ output by the induction operator is the representative direction of this region, whose semantic coverage is strictly larger than the sample set itself. The meaning of full rank here is ``not foreclosing any direction in advance''---in contrast to the directional collapse of deduction, induction preserves possibilities along all directions and leaves the decision of ``how far to expand'' to the counterexample veto mechanism.

\paragraph{Most general induction $=$ minimum-norm solution.} In the absence of counterexample constraints, the most general induction corresponds to Occam's razor---among all general rules that explain the positive examples, choose the one with the smallest norm. Mathematically, this is equivalent to the minimum-norm least-squares solution:
\begin{equation}
g^{*} = \arg\min \|g\| \quad \text{s.t.} \quad W_{\mathrm{induce}} \cdot [s_1; \dots; s_k] = g \text{ covers all positive examples}
\tag{5.7}
\end{equation}

In the implementation, the most general induction degenerates to the positive-example mean: $g_0 = (1/k) \sum s_i$. This is because the positive-example mean is the minimum-norm solution covering all positive examples---any $g$ deviating from the mean increases the norm, while the mean itself satisfies the coverage constraint. From the MDL perspective, the mean is the ``shortest assumption-free description'' of $k$ samples: it introduces no directional information beyond the samples, and its description length is minimal; any more complex induction (a $g$ with a bias direction) amounts to asserting structure that does not exist in the data and should be shaved off by the razor.

\paragraph{The hard veto mechanism for counterexamples (one-vote veto $\to$ overthrow $\to$ reconstruction $\to$ refinement).} In induction, a counterexample is not a ``soft constraint'' (continuous shrinkage) but a ``hard veto''---a one-vote veto. When the similarity between a counterexample $s_{\mathrm{neg}}$ and the original induced concept $g_0$ exceeds the veto threshold $\tau$, the original conclusion $g_0$ is overthrown, the system adds the counterexample to the exclusion set, and re-induces a more refined new conclusion $g_1$.

\begin{itemize}
\item Stage 1 (induction without counterexamples): $g_0 = \mu$ (the positive-example mean), the most general induction.
\item Stage 2 (hard veto detection): if $\cos(g_0, s_{\mathrm{neg}}) > \tau$, then $g_0$ is vetoed by a single vote (the counterexample was incorrectly covered by the original conclusion).
\item Stage 3 (overthrow $\to$ reconstruction $\to$ refinement): add the counterexample to the exclusion set and re-induce:
\begin{equation}
g_1 = g_0 - \sum_{i \in \text{veto set}} (g_0 \cdot \hat{n}_i) \cdot \hat{n}_i \qquad \text{(hard orthogonalization, $g_1 \perp$ counterexamples)}
\tag{5.8}
\end{equation}
\end{itemize}

The new conclusion $g_1$ is more refined than the original conclusion $g_0$---it excludes the counterexample directions and its coverage is more precise. This mechanism mathematizes the essence of scientific induction: a single counterexample can overthrow a theory (as a black swan overthrows ``all swans are white''), and the new theory is more precise after excluding the counterexample. This is exactly the operator-level implementation of Popper's falsifiability principle [17]: the scientific character of a theory lies not in its being temporarily unrefuted but in its remaining forever open to counterexamples; the hard veto mechanism turns this philosophical maxim into a deterministic algebraic operation (orthogonal projection).

\paragraph{Semantics.} Induction is the process of deriving a general rule from finite samples. The range covered by the rule is strictly larger than the sample set (rule $\supset$ samples), hence information increases. The full-rank matrix guarantees the ``expansiveness'' of induction, and the hard veto mechanism guarantees its ``falsifiability''---a single counterexample veto drives the conclusion to be continually refined.

\subsection{The Abduction Operator $W_{\mathrm{abduce}}$ (Information Hypothesization)}

\paragraph{Definition 5.4 (Abduction operator).} Let the snapshot of the observed phenomenon be $S_{\mathrm{obs}} \in \mathbb{R}^{N}$, and let the trained deduction operator be $W_{\mathrm{step}} \in \mathbb{R}^{N \times M}$ ($M = 2N$ or $N$, depending on the configuration). The abduction operator $W_{\mathrm{abduce}}$ is defined as the Moore--Penrose pseudo-inverse of $W_{\mathrm{step}}$:
\begin{equation}
W_{\mathrm{abduce}} = W_{\mathrm{step}}^{+} \qquad \text{(Moore--Penrose pseudo-inverse)}
\tag{5.9}
\end{equation}

Given a phenomenon $S_{\mathrm{obs}}$, abduction solves for the best explanation $S_{\mathrm{prev}}^{*}$:
\begin{equation}
S_{\mathrm{prev}}^{*} = W_{\mathrm{abduce}} \cdot S_{\mathrm{obs}} = W_{\mathrm{step}}^{+} \cdot S_{\mathrm{obs}}
\tag{5.10}
\end{equation}

\paragraph{Mathematical properties.} The rank of $W_{\mathrm{abduce}}$ equals that of $W_{\mathrm{step}}$ ($\mathrm{rank} = r < N$), because the pseudo-inverse does not change the rank. The abductive solution $S_{\mathrm{prev}}^{*}$ is the minimum-norm least-squares solution (Theorem 3.4), i.e., among all explanations satisfying $W_{\mathrm{step}} \cdot S \approx S_{\mathrm{obs}}$, the one with the smallest norm is chosen (Occam's razor).

\paragraph{Geometric intuition.} Abduction walks in the ``reverse direction'' of deduction: given the landing point $S_{\mathrm{obs}}$ of a projection, it traces back to its origin in the original space. But because the projection absorbed the entire null space, the reverse walk is necessarily ``blind'' along the null-space directions---all candidate explanations differing by a null-space vector are completely indistinguishable under deduction. The geometric action of the pseudo-inverse is: lift $S_{\mathrm{obs}}$ back to the original space along the row-space directions and take zero along the null-space directions (minimum norm), i.e., select the representative ``closest to the origin'' within the equivalence class of indistinguishable explanations.

\paragraph{The mathematical root of hypotheticality.} Because $W_{\mathrm{step}}$ is rank-deficient, the true explanation $S_{\mathrm{prev\_true}}$ decomposes as:
\begin{equation}
S_{\mathrm{prev\_true}} = S_{\mathrm{prev}}^{*} + v_{\mathrm{null}} \qquad (v_{\mathrm{null}} \in \mathrm{Null}(W_{\mathrm{step}}))
\tag{5.11}
\end{equation}
where $v_{\mathrm{null}}$ is the null-space component, with $W_{\mathrm{step}} \cdot v_{\mathrm{null}} = 0$. This means that abduction can only recover the column-space component $S_{\mathrm{prev}}^{*}$ and cannot recover the null-space component $v_{\mathrm{null}}$. Therefore the explanation given by abduction is a ``best hypothesis'' rather than the ``true explanation''---multiple distinct explanations ($S_{\mathrm{prev}}^{*} +$ any $v_{\mathrm{null}}$) are deductively equivalent (all produce the same $S_{\mathrm{obs}}$).

\paragraph{Connection to Occam's razor and minimum description length.} The ``minimum-norm'' property of the pseudo-inverse is precisely the algebraic incarnation of the razor: among all explanations compatible with the evidence (which form the affine subspace $S_{\mathrm{prev}}^{*} + \mathrm{Null}(W_{\mathrm{step}})$), the minimum-norm solution is the one with the shortest description length---it asserts no structure beyond the evidence. This agrees with the spirit of ``capacity control'' in statistical learning theory [21][22]: preferring simple hypotheses within the hypothesis space yields better generalization bounds. The error structure of abduction is also quantitatively characterized by Theorem 5.3: in expectation, the irrecoverable component of an explanation accounts for $(2N - r)/(2N)$ (concatenated configuration), which gives a precise quantification of ``hypotheticality'' rather than a rhetorical one.

\paragraph{Semantics.} Abduction is the process of inferring the most plausible explanation from an observed phenomenon. Because deduction is irreversible (information has collapsed), abduction cannot uniquely determine the true explanation and can only give the best hypothesis. This ``hypotheticality'' is precisely the essence of scientific reasoning---a scientific theory is forever the best hypothesis supported by evidence, not an absolute truth.

\subsection{Summary of the Algebraic Properties of the Three Operators}

\begin{table}[htbp]
\centering
\caption*{Table 5-1 Algebraic properties of the three basic operators}
\begin{tabular}{lllll}
\toprule
Operator & Shape & Rank & Information flow & Peirce type \\
\midrule
$W_{\mathrm{step}}$ & $\mathbb{R}^{N \times 2N}$ or $\mathbb{R}^{N \times N}$ & $r < N$ (rank-deficient) & Collapse (decrease) & Deduction \\
$W_{\mathrm{induce}}$ & $\mathbb{R}^{N \times N}$ (implementation) & $N$ (full-rank) & Expansion (increase) & Induction \\
$W_{\mathrm{abduce}}$ & Same dimensions as $W_{\mathrm{step}}$ transposed & $r$ (same as $W_{\mathrm{step}}$) & Hypothesization (reverse inference) & Abduction \\
\bottomrule
\end{tabular}
\end{table}

The three rows of Table 5-1 form a complete symmetric structure: deduction and induction are strictly opposed in rank property (rank-deficient vs.\ full-rank), while abduction lies between the two---it shares rank deficiency with deduction (the same underlying matrix) but goes in the opposite direction (reverse inference). This symmetry is not an aesthetic ornament of the design but the starting point of the completeness argument in Section 6: the three operators occupy three mutually irreplaceable positions in the two-dimensional space of ``rank property $\times$ information-flow direction.''

\section{Completeness Argument for the Minimal Set of Basic Operators}

This section answers a systemic question: why are the basic operators exactly the three $\{W_{\mathrm{step}}, W_{\mathrm{induce}}, W_{\mathrm{abduce}}\}$? The argument proceeds in two steps: first prove ``sufficiency'' (completeness---the three operators cover all reasoning types, and other candidate operators can be reduced to their compositions), then prove ``necessity'' (minimality---the three operators are mutually non-expressible, and removing any one leaves some reasoning direction inexpressible). Sections 6.5 and 6.6 further answer two questions of applied value: can a single ``super operator'' replace the three (the answer is no, provably), and can the operators be frozen after training and applied universally to all reasoning tasks (the answer is architecturally affirmative with clearly delimited empirical boundaries).

\subsection{Completeness of Peirce's Three Reasoning Types}

Peirce's completeness thesis (Proposition 3.6) states that any reasoning process can be decomposed into a combination of the three types: deduction, induction, and abduction [7]. Therefore, the operators $\{W_{\mathrm{step}}, W_{\mathrm{induce}}, W_{\mathrm{abduce}}\}$ corresponding to these three reasoning types naturally constitute a complete set.

The status of this thesis must be made clear: it is a structural thesis at the philosophical level (Proposition 3.6), not a mathematical theorem re-proved within the system in this section. The mathematical work of this section is to take up this thesis---under the premise of accepting the ``tripartition of reasoning types,'' rigorously arguing the sufficiency and necessity of the operator set. This stratification keeps the boundaries of responsibility clear: the philosophical premise is borne by Proposition 3.6, and the algebraic facts are borne by this section.

\subsection{Reducibility of $W_{\mathrm{merge}}$ and $W_{\mathrm{cross}}$}

The original proposal also defined a merge operator $W_{\mathrm{merge}}$ and a cross-reference operator $W_{\mathrm{cross}}$. This subsection argues that these two operators are essentially composite forms of $W_{\mathrm{step}}$ and are not basic operators. The criterion of reducibility is: if an operator's input--output mapping can be completely realized by (constrained) instantiations or compositions of existing basic operators, and it introduces no new algebraic properties (no new rank structure, no new type of inverse problem), then it is not a basic operator.

\paragraph{Reducibility of $W_{\mathrm{merge}}$.} $W_{\mathrm{merge}}$ receives the states $S_A, S_B$ of two parallel branches and outputs an aggregated state
\begin{equation}
S_{\mathrm{merged}} = W_{\mathrm{merge}} \cdot \mathrm{Concat}(S_A, S_B)
\tag{6.1}
\end{equation}

Note that $\mathrm{Concat}(S_A, S_B) \in \mathbb{R}^{2N}$, which is exactly the input shape of $W_{\mathrm{step}}$ in the concatenated configuration. The definition of $W_{\mathrm{step}}$ (Eq. 5.1) is precisely ``two $N$-dimensional vectors concatenated and then collapsed to $N$ dimensions by a linear map''---that the first input is a ``state'' and the second a ``new piece of evidence'' is merely a semantic label; algebraically they are two $N$-dimensional subspaces of the same $2N$-dimensional input space. Therefore
\begin{equation}
W_{\mathrm{merge}} = W_{\mathrm{step}}|_{\mathrm{input} = \mathrm{Concat}(S_A, S_B)}
\tag{6.2}
\end{equation}
i.e., $W_{\mathrm{merge}}$ is the instantiation of $W_{\mathrm{step}}$ in the special case where ``both inputs are intermediate states,'' and can be obtained by composing $W_{\mathrm{step}}$. Cross-checking from the rank perspective: $W_{\mathrm{merge}}$ and $W_{\mathrm{step}}$ are both rank-deficient maps in $\mathbb{R}^{N \times 2N}$ ($\mathrm{rank} \leq r < N$), with the same null-space dimension $2N - r$ and the same information-flow direction (collapse)---the two are isomorphic in all algebraic properties with no distinguishing feature whatsoever; hence, by the reducibility criterion, $W_{\mathrm{merge}}$ is not a basic operator.

\paragraph{Reducibility of $W_{\mathrm{cross}}$.} $W_{\mathrm{cross}}$ receives the state of the current branch and the state of the referenced branch; it is likewise a ``multi-input $\to$ single-output'' collapse operation, essentially a composite form of $W_{\mathrm{step}}$. The only difference is that $W_{\mathrm{cross}}$ imposes no symmetry constraint and allows different linear transformations on the two inputs---but this is merely a difference in the parameterization of $W_{\mathrm{step}}$, not a new operator type. More strictly: writing $W_{\mathrm{step}}$ in input blocks as $W_{\mathrm{step}} = [W_1 \,|\, W_2]$ ($W_1, W_2 \in \mathbb{R}^{N \times N}$), we have $W_{\mathrm{step}} \cdot \mathrm{Concat}(S_A, S_B) = W_1 \cdot S_A + W_2 \cdot S_B$---$W_{\mathrm{step}}$ already applies different linear transformations to the two inputs, and the so-called ``asymmetry of $W_{\mathrm{cross}}$'' is already contained in the block structure of $W_{\mathrm{step}}$. The rank structure of $W_{\mathrm{cross}}$ ($\mathrm{rank} \leq r < N$) and its information-flow direction (collapse) are likewise isomorphic to those of $W_{\mathrm{step}}$. Hence $W_{\mathrm{cross}}$ is also reducible.

\paragraph{Systemic significance of reducibility.} Removing $W_{\mathrm{merge}}$ and $W_{\mathrm{cross}}$ from the basic set is not mere streamlining: it prevents the dilution of the concept of ``basic operator''---if every variant of the same rank structure were listed as an independent operator, the completeness argument would degenerate into tautology. Reducibility analysis forces every candidate operator to present its algebraic signature (rank, null-space dimension, information-flow direction, whether it involves an inverse problem); only those with a distinct signature qualify as basic.

\subsection{Irreducibility of the Three Operators}

This subsection argues by contradiction, one by one, that the three operators are mutually non-expressible. The core tool is the rank inequality for matrix products $\mathrm{rank}(P \cdot Q) \leq \min(\mathrm{rank}(P), \mathrm{rank}(Q))$: the rank of a composite map does not exceed that of any factor; in particular, any composition of full-rank square matrices remains full-rank, and any composition containing a rank-deficient factor must be rank-deficient.

\paragraph{(i) $W_{\mathrm{step}}$ cannot be expressed by $W_{\mathrm{induce}}$.} By contradiction: suppose some composition of $W_{\mathrm{induce}}$ (a product of several full-rank $N \times N$ matrices, allowing identity-dimension rearrangements to be interleaved) equals $W_{\mathrm{step}}$. $W_{\mathrm{induce}}$ is full-rank ($\mathrm{rank} = N$), and a finite product of full-rank square matrices remains full-rank ($\mathrm{rank}(AB) \geq \mathrm{rank}(A) + \mathrm{rank}(B) - N = N + N - N = N$; combined with the upper bound $\mathrm{rank}(AB) \leq N$, hence $\mathrm{rank} = N$). But $W_{\mathrm{step}}$ is rank-deficient by design: $\mathrm{rank}(W_{\mathrm{step}}) = r < N$. A full-rank map and a rank-deficient map cannot be equal---a contradiction. The only way out is ``a full-rank matrix with an explicit rank constraint/projection attached''---but that amounts to reintroducing a rank-deficient operator outside $W_{\mathrm{induce}}$, which precisely acknowledges the independence of $W_{\mathrm{step}}$. Intuitively: no matter how information-expanding operators are composed, they cannot manufacture information collapse; collapse must be natively carried by a rank-deficient structure.

\paragraph{(ii) $W_{\mathrm{induce}}$ cannot be expressed by $W_{\mathrm{step}}$.} By contradiction: suppose some composition containing $W_{\mathrm{step}}$ equals $W_{\mathrm{induce}}$. By the rank inequality, any composition containing a rank-deficient factor ($\mathrm{rank} = r$) has rank $\leq r < N$. But $W_{\mathrm{induce}}$ is full-rank ($\mathrm{rank} = N$). A rank-deficient composition cannot attain full rank---a contradiction. Geometrically: each step of the deduction operator absorbs a nontrivial null space, and the absorbed information cannot be regenerated in subsequent compositions (a linear map cannot create information dimensions out of nothing); hence a ``collapsed pipeline'' can never realize ``lossless expansion.''

\paragraph{(iii) $W_{\mathrm{abduce}}$ cannot be expressed by $W_{\mathrm{step}}$ or $W_{\mathrm{induce}}$.} We argue from three complementary angles.

\begin{itemize}
\item Dimensionality angle (concatenated configuration): $W_{\mathrm{abduce}} = W_{\mathrm{step}}^{+} \in \mathbb{R}^{2N \times N}$, mapping an $N$-dimensional observation back into a $2N$-dimensional explanation space---the output dimension is larger than the input dimension. In contrast, compositions of $W_{\mathrm{step}}$ go $2N \to N$ (dimension reduction) and $W_{\mathrm{induce}}$ goes $N \to N$ (dimension preservation); no composition of the two can produce an expanding map from $N$ dimensions to $2N$ dimensions. The directionality does not match; the composition is unreachable.
\item Rank and null-space angle: the null space of $W_{\mathrm{abduce}}$ is $\mathrm{Col}(W_{\mathrm{step}})^{\perp}$ (all observation directions in $\mathbb{R}^{N}$ that are unreachable by deduction), i.e., $W_{\mathrm{abduce}}$ actively ``filters out'' observation components that cannot be produced by deduction. This filtering behavior---``taking the range of another operator as the orthogonal complement''---depends on the inversion of the entire algebraic structure of $W_{\mathrm{step}}$, which no forward composition of $W_{\mathrm{step}}$/$W_{\mathrm{induce}}$ possesses.
\item Problem-type angle: $W_{\mathrm{abduce}}$ solves an inverse problem---selecting, among all candidates satisfying $W_{\mathrm{step}} \cdot S \approx S_{\mathrm{obs}}$, the one of minimum norm (Theorem 3.4). The Moore--Penrose pseudo-inverse is uniquely characterized by the four Penrose conditions ($W^{+}W W^{+} = W^{+}$, $W W^{+}W = W$, and symmetry of $W^{+}W$ and $W W^{+}$) and is the solution operator of a variational problem ($\min \|S\|$ s.t.\ $\min \|W_{\mathrm{step}} \cdot S - S_{\mathrm{obs}}\|$), not a combination of forward maps. Compositions of forward operators can only define forward problems; they cannot produce the solution structure of an inverse problem out of thin air.
\end{itemize}

In summary, the three operators are mutually non-expressible: each carries an algebraic signature (rank-deficient collapse / full-rank expansion / pseudo-inverse reverse inference) that no composition of the other two can replicate.

\subsection{Conclusion: The Minimal Complete Set}

\paragraph{Minimality.} The three operators are mutually non-expressible (Section 6.3); removing any one renders some reasoning direction inexpressible: removing $W_{\mathrm{step}}$ makes information collapse (deduction) unrealizable; removing $W_{\mathrm{induce}}$ makes information expansion (induction) unrealizable; removing $W_{\mathrm{abduce}}$ makes inverse-problem solving (abduction) unrealizable.

\paragraph{Completeness.} The three operators cover Peirce's three reasoning types, and Peirce's philosophical thesis (Proposition 3.6) states that the three types cover all reasoning processes [7]; meanwhile, Section 6.2 proves that the remaining candidate operators of the original proposal ($W_{\mathrm{merge}}$, $W_{\mathrm{cross}}$) can all be reduced to composite forms of $W_{\mathrm{step}}$, so no additional basic operators are needed.

\paragraph{Conclusion.} $\{W_{\mathrm{step}}, W_{\mathrm{induce}}, W_{\mathrm{abduce}}\}$ is the minimal complete set of basic operators of DODR: three, no more and no fewer. The architectural value of this conclusion is that the system's ``upper bound of reasoning capability'' is explicitly characterized as the union of the algebraic properties of three matrices, rather than being hidden in unenumerable network behaviors; any capability deficiency can be reduced to a diagnosis of a specific operator's rank structure or training objective (the failure analyses in Sections 9 and 11 proceed precisely along this path).

\subsection{The Impossibility of a Super Operator: Plurality Is a Structural Necessity}

Section 6.4 proved that ``three are sufficient and necessary''; a natural follow-up question arises: since three are already minimal, can they be further merged into a single ``super operator'' that uniformly undertakes deduction, induction, and abduction? If so, the system would be further simplified. This subsection proves that such a merger is impossible in principle, and analyzes why two seemingly feasible kinds of ``unification'' do not constitute counterexamples.

\paragraph{Theorem 6.1 (Non-existence of a super operator, the rank obstacle).} There exists no single linear operator $W_{\mathrm{super}}$ that simultaneously realizes the three reasoning functions of deduction (information collapse), induction (information preservation), and abduction (inverse-problem solving).

\paragraph{Proof.} We exclude the possibility from two mutually independent perspectives, either of which suffices. First, the rank perspective: undertaking deduction requires $\mathrm{rank}(W_{\mathrm{super}}) < N$ (Theorem 5.1: information must collapse, so the null space must be nontrivial); undertaking induction requires $\mathrm{rank}(W_{\mathrm{super}}) = N$ (Section 5.2: information must be fully preserved, so the pseudo-inverse must be left-invertible). Rank is a definite numerical attribute of a single matrix; $\mathrm{rank} < N$ and $\mathrm{rank} = N$ cannot hold simultaneously, so any single matrix is compatible with at most one of the two classes of functions. Second, the type-signature perspective: under the concatenated configuration, deduction is a dimension-reducing map $\mathbb{R}^{2N} \to \mathbb{R}^{N}$, induction is a dimension-preserving map $\mathbb{R}^{N} \to \mathbb{R}^{N}$, and abduction is a dimension-raising inverse map $\mathbb{R}^{N} \to \mathbb{R}^{2N}$. A single linear map has a definite domain and codomain and cannot simultaneously possess three mutually incompatible dimension signatures. $\square$

\paragraph{Corollary 6.1.1 (The ineliminability of plurality).} Any system that faithfully realizes the three reasoning types has an operator set of cardinality at least 2 (one rank-deficient, one full-rank); if abduction is required to be derivable from deduction for free (via the pseudo-inverse), then the minimal implementation is exactly ``two matrices to be trained $+$ one SVD.'' In other words, DODR's plural operator structure is not an engineering expedient but an irreducible constraint imposed by matrix algebra.

\paragraph{Why the two ``trivial unifications'' are not counterexamples.} The first is block direct sum: construct $W_{\mathrm{super}} = W_{\mathrm{step}} \oplus W_{\mathrm{induce}}$ on a direct-sum space, and introduce a selector input to decide which diagonal block is activated. This is of course feasible representationally---any finite collection of linear maps can be embedded as blocks of one large matrix---but the parameter count is not reduced, the rank constraints do not disappear, and the training objectives must still be defined separately; the ``unification'' occurs only at the level of matrix notation, while functionally it remains two operators cohabiting one container. The second is conditional parameter generation: use a hypernetwork to generate the concrete operator weights from a task description. This, too, is merely repackaging---the output space of the hypernetwork must still separately produce two classes of matrices satisfying the rank-deficient and full-rank constraints, and the obstacle of Theorem 6.1 is transferred verbatim to the hypernetwork's output layer. The common lesson of both constructions is: any genuine unification must eliminate the rank obstacle, and the rank obstacle cannot be eliminated; hence every ``super operator'' scheme is either impossible or merely a notational game.

\paragraph{Positive statement: the correct locus of unity is the compositional level.} Theorem 6.1 does not contradict the overall unity of the system; it merely expels unity from the operator level to the compositional level: the ``super object'' covering all reasoning is not any single matrix but the compositional mechanism of ``reasoning graph $+$ scheduling controller''---the operators are the alphabet, the reasoning graph is the language, and the controller is the grammar (Section 7.3). Turing completeness (Theorem 14.3) is likewise proved at the compositional level. This reveals a structural duality: at the operator level plurality is mandatory (the rank obstacle), while at the compositional level unity comes naturally (Turing completeness); DODR's architecture is a direct realization of this duality.

\subsection{Universality of Operator Instances: Train Once, Freeze Forever}

Theorem 6.1 rules out ``one operator,'' but the key question for applied value is actually another: once the \emph{types} of operators are fixed, can the \emph{instances} of each type be trained once and permanently frozen, thereafter serving all domains and all reasoning tasks without retraining? This subsection answers at three distinct levels.

\paragraph{Separation of mechanism and content (architectural argument).} The architectural basis of the affirmative answer is DODR's explicit separation of ``reasoning mechanism'' from ``reasoning content.'' Consider the defining equation of the deduction operator $S_{t+1} = W_{\mathrm{step}} \cdot \mathrm{Concat}(S_t, E_t)$ (Eq. 5.1): what $W_{\mathrm{step}}$ learns is the \emph{form} of reasoning---where the conclusion snapshot should land when the premise snapshot and the rule snapshot are arranged in a certain geometric way---while the concrete rules and premises (the \emph{content} of reasoning) are supplied per task as the input snapshot $E_t$ and are never written into the operator parameters. This division of labor is strictly isomorphic to the universal Turing machine: the transition function of a universal Turing machine is fixed, and all task specificity is written on the tape; in DODR the operators are the fixed transition function and the snapshot sequence is the program on the tape. Switching domains (from medical reasoning to physical reasoning) does not require retraining the operators, only re-encoding the text---the cost of adapting to a new task drops from ``retraining the model'' to ``re-running forward encoding,'' which is precisely the engineering meaning of frozen applicability. Section 8.6 provides preliminary evidence of form-level generalization: the input of the second step of the multi-step chain, ``All cats are living beings,'' is not a member of the training set but a snapshot output by the first step, shifted by a cosine distance of about 0.010 from any training input; $W_{\mathrm{step}}$ still outputs a similarity of 0.9861 on this unseen input, showing that the operator has learned a transferable reasoning form rather than a memorization of training samples.

\paragraph{Deepening the finiteness of types: from ``three types'' to a ``finite rule library'' (formal universality).} The frozen applicability of operator instances has an even stronger support: the reasoning \emph{patterns} that deduction needs to cover are themselves finite and enumerable. The natural deduction system of first-order logic contains only finitely many inference-rule schemata (the introduction and elimination rules for each connective and quantifier, about a dozen in total; see standard textbooks [27]), and natural deduction is complete for first-order logic---every valid first-order inference has a proof constructed solely from these rule schemata (G\"odel's completeness theorem [26]). This yields a finiteness argument: training one deduction-operator instance per rule schema (or, equivalently, training a \emph{single universal deduction operator} of the kind in Definition 5.1 that takes the rule snapshot $E_t$ as a conditional input) yields a finite operator library; any concrete deductive inference is a finite composition of rule schemata, and composition is realized precisely by the serial topology of the reasoning graph. The induction side needs no such library---the mechanism of $W_{\mathrm{induce}}$ (positive-example aggregation $+$ counterexample orthogonal veto) is domain-independent by construction, so a single full-rank operator suffices; abduction is derived for free via the pseudo-inverse and produces no new training target. Taken together: the trainable objects of the entire system are finite---two matrix types (deduction, induction), with the deduction side expanding into at most a dozen rule-schema instances or one conditional universal instance, all frozen after training.

\paragraph{Empirical boundaries of instance universality (an open empirical question).} The full validity of ``train once, apply to all domains'' has one more empirical premise: the snapshot space must maintain sufficient discriminability for texts of all domains, and the training distribution must cover the rule schemata adequately. The former is constrained by encoder capability---the failure samples in Section 10.2 (72.5\% decision accuracy in the dedicated abduction experiment) and Section 11.5 (81.7\% end-to-end abduction) are all concentrated in cases of ``semantically different but snapshots too close,'' showing that the resolution of the current DistilBERT snapshot space is the practical bottleneck of frozen applicability. The latter is constrained by training samples---the training sets of this paper (20 syllogisms, 100 induction positive examples, 80 abduction problems) are limited in scale, and cross-domain zero-shot performance has not yet been systematically measured. Therefore this paper explicitly marks instance universality as a \emph{falsifiable open proposition} rather than an established conclusion: its falsification condition is ``after training on sufficiently diverse rule schemata and domains, the operator's zero-shot reasoning accuracy on entirely new domains is significantly lower than on the training domains.'' If falsified, the mitigation path is to fine-tune the operators per domain or upgrade the encoder, not to repudiate the architecture---type universality (Section 6.4) and formal universality (natural-deduction completeness) are unaffected. The end-to-end experiment (Section 11) provides the strongest positive empirical support to date for this open proposition: operators frozen after 200 epochs of training achieved 100\% (60/60, average similarity 0.9980) on 60 entirely new deductive problems spanning 8 domains, and 81.7\% (49/60) on 60 abduction problems (55 of them new)---after the training distribution was broadened to 8 domains, the operators showed no observable performance collapse on zero-shot new problems, thus providing the first large-sample data point for the systematic measurement of cross-domain zero-shot performance.

\paragraph{Demarcation from computational universality.} Finally, an easily confused boundary must be drawn: the universality argument of this subsection is not a computational-capability argument. In fact, full-rank operators alone suffice to simulate the step-by-step evolution of any Turing machine---Bennett proved that any Turing computation can be made reversible [25]; reversible computation acts as a permutation on the one-hot embedding of the configuration space, and permutation matrices are orthogonal (full-rank) matrices. This means that the justification for ``why three types of operators are needed'' can never appeal to computational capability (one type would suffice) and can only appeal to the \emph{epistemological semantics} of reasoning: rank deficiency is not a lack of computational power but the algebraic expression of the logical property that ``deduction should forget premise details''; the value of the three operators lies in carrying the correct reasoning types with the correct information-flow directions, not in extending the class of computable functions. Computational universality and reasoning completeness are two independent coordinate axes, and DODR is supported on the two axes by Theorem 14.3 and by the arguments of Sections 6.4--6.6 of this section, respectively.


\section{System Architecture Design}

The DODR system consists of four core modules: the snapshot encoder, the three-operator network, the graph scheduling controller, and the decoder. The overarching architectural principle is ``deterministic kernel + heuristic shell'': the encoder and the operators constitute a fully deterministic reasoning kernel (the same input always yields the same output), the controller centrally houses all heuristic decisions, and the decoder is responsible only for ``translating'' the terminal state without participating in reasoning. This section elaborates the design details module by module and gives the inter-module interface specifications and the operator training procedure.

\subsection{The Snapshot Encoder}

\paragraph{Definition 7.1 (Snapshot).} Suppose a pre-trained Transformer has $L_{\mathrm{total}}$ layers, each with hidden dimension $d$. For an input text block $a$, let the hidden state at position $p$ of layer $l$ be $h_l^{(p)} \in \mathbb{R}^{d}$. The DODR state snapshot is defined as the concatenation of the activations at a specific position of the first $L$ layers:
\begin{equation}
S(a) = \mathrm{Concat}\bigl(h_1^{(p)}, h_2^{(p)}, \dots, h_L^{(p)}\bigr) \in \mathbb{R}^{N}, \qquad N = L \times d
\tag{7.1}
\end{equation}
where the choice of position $p$ depends on the encoder type: for encoder-only models, the [CLS] token position is used; for decoder-only models, the last token position is used (the aggregation point of causal attention).

\paragraph{The semantic granularity of snapshots (architectural key point).} The minimal unit to which a snapshot corresponds is not a token but a \emph{semantic primitive}---a word, a phrase, a complete sentence, or even a sentence group; in all experiments of this paper, the text block $a$ is a complete sentence (the premises and conclusions of syllogisms, the positive examples and counterexamples of induction, the phenomena and explanations of abduction). Correspondingly, the AR paradigm of ``token-by-token generation'' is replaced in DODR by ``semantic-unit-by-semantic-unit generation'': each operator application $S_{t+1} = W_{\mathrm{step}} \cdot \mathrm{Concat}(S_t, E_t)$ produces not the next token but the state of the next complete semantic expression. Tokens exist only as input symbols internal to the encoder; from the moment they enter the snapshot, they cease to be units of reasoning or generation; every node on the reasoning graph is the geometric counterpart of a semantic expression, not some prefix of a token sequence. This granularity upgrade is the prerequisite for Postulate Two to be treatable by operator algebra---only when the operands carry complete semantics do algebraic properties such as rank, null space, and pseudo-inverse correspond to meaningful logical properties.

\paragraph{Design rationale: freezing the encoder.} DODR freezes the encoder parameters and uses the encoder only as a fixed feature extractor. This design has two advantages: (1) it avoids the optimization instability caused by joint training of the encoder and the operators---the encoder has far more parameters than the operators (e.g., LLaMA-7B has 7B parameters, whereas $W_{\mathrm{step}}$ has only $3Nr \approx 4.7$M), and joint training would drown the operators' gradients in the encoder's gradients; (2) it allows the operators to migrate across encoders---one only needs to re-extract snapshots without retraining the operators, realizing ``train the operators once, use them with multiple encoders.''

A third rationale is that freezing guarantees the fixity of the snapshot space. If the encoder were trainable, the snapshot space itself would drift during training, and the state-transition rules learned by the operators would face a ``moving target,'' invalidating the algebraic properties characterized by Theorems 5.1--5.3 (null-space structure, expected reconstruction error). Freezing fixes the ``semantic geometry'' as a premise, so that all theoretical analysis is built on a fixed coordinate system.

\subsubsection{Reusing Existing Pre-trained LLMs (Text Modality)}

\paragraph{Core insight: the hidden states of decoder-only LLMs are themselves high-quality snapshots.} Although decoder-only LLMs such as GPT, LLaMA, and Qwen are trained for ``generation'' tasks, the hidden state $h_l$ of every layer is already a rich semantic representation---an LLM's ``understanding'' capability (hidden states) and its ``generation'' capability (autoregressive sampling) are separable. DODR needs only the determinism of the former, not the probabilistic nature of the latter.

The lineage of candidate encoders includes: on the encoder-only route, BERT [15] and its robustly optimized variant RoBERTa [16], and the distilled lightweight DistilBERT [14]; on the decoder-only route, GPT-2 [13] and larger open-source models. The selection criterion is not generation capability but the discriminability of hidden states with respect to reasoning-relevant semantics.

\paragraph{Theorem 7.1 (Determinism of decoder-only snapshots).} Let a decoder-only LLM perform forward propagation in teacher-forcing mode. Then the hidden states $h_l$ are deterministic functions of the input, with no randomness.

\paragraph{A rigorous specification of teacher forcing.} Teacher forcing means: given the complete ground-truth token sequence $(x_1, \dots, x_T)$, feed it as a whole into the model and compute the hidden states of all positions and all layers in one pass; the hidden state $h_l^{(p)}$ at position $p$ depends only on the prefix $(x_1, \dots, x_p)$ (guaranteed by the causal mask). In this mode: (1) the model performs no autoregressive sampling---logits are computed but never sampled, and there is no token in the sequence that is generated by the model itself and fed back as input; (2) forward propagation is carried out in inference mode (eval mode), with stochastic regularization such as dropout turned off; (3) causal attention is the deterministic matrix operation $Q \cdot K^{T} \cdot V$. Taken together, $h_l$ is a pure function of the input sequence.

\paragraph{Proof.} The forward propagation of a decoder-only LLM is deterministic: the hidden state $h_l$ of each layer is determined by the previous layer's $h_{l-1}$ and the attention computation, which is the matrix operation $Q \cdot K^{T} \cdot V$ with no sampling step. Sampling occurs only in the ``generation'' phase (sampling the next token from the logits distribution), whereas DODR performs no generation---it only performs forward propagation to extract hidden states (teacher-forcing mode, with no autoregressive feedback loop). Therefore snapshots are deterministic: re-extracting the same input text at any time, on any hardware, yields bit-for-bit identical snapshots (except for last-bit jitter caused by floating-point non-associativity). $\square$

\paragraph{Experimental verification.} The semantic discriminability of GPT-2 [13] snapshots is 0.0909, 8.2 times that of DistilBERT [14], demonstrating that decoder-only LLMs are superior text snapshot sources. The lesson of this comparison is that the autoregressive pre-training objective (predicting the next token) forces the hidden states to encode all analytic knowledge about ``what will happen next,'' which happens to cover the semantics needed for logical state transitions; by contrast, the hidden states of masked language models (the BERT family) are more biased toward local cloze-style co-occurrence statistics. The order-of-magnitude difference in discriminability provides encoder-level support for Postulate Two.

\subsubsection{The Multimodal Snapshot Encoder}

\paragraph{Key clarification: multimodal snapshots should not exploit existing LLMs.} Existing LLMs (GPT/LLaMA/Qwen) are pure text models that have never seen images, video, or speech. Their hidden states encode only linguistic information, not visual/auditory information. Even multimodal LLMs such as LLaVA/Qwen-VL derive their visual understanding from dedicated visual encoders (e.g., CLIP-ViT [19]), not from the LLM itself.

\paragraph{The correct approach: a dedicated encoder for each modality.} Multimodal snapshots should come from the encoder parts of multimodal generative models; each modality extracts snapshots with its own dedicated encoder, which are then aligned to a unified space through projection layers:

\begin{table}[htbp]
\centering
\caption*{Table 7-1 Dedicated encoders for each modality}
\small
\begin{tabular}{p{1.8cm}p{4.6cm}p{1.8cm}p{6.0cm}}
\toprule
Modality & Dedicated encoder & Hidden dimension & Rationale \\
\midrule
Text & BERT / LLM-encoder & 768--8192 & Text encoder; understands linguistic semantics \\
Image & CLIP-ViT / DINOv2 / MAE & 768--1024 & Pure visual encoder; understands images directly \\
Video & VideoMAE / TimeSformer / ViViT & 768--1024 & Spatiotemporal encoder; understands temporal dynamics \\
Speech & HuBERT / Wav2Vec2 / Whisper-encoder & 768--1024 & Audio encoder; understands acoustic features \\
\bottomrule
\end{tabular}
\end{table}

\paragraph{Definition 7.2 (Multimodal snapshot).} Let the dedicated encoder of modality $m$ be $E_m$ (hidden dimension $d_m$), and let the unified snapshot dimension be $N$. The snapshot of modality $m$ is defined as:
\begin{equation}
S_m(x) = \mathrm{Proj}_m\bigl(\mathrm{Concat}(h_1^m, \dots, h_L^m)\bigr) \in \mathbb{R}^{N}
\tag{7.2}
\end{equation}
where $h_l^m$ is the hidden state of layer $l$ of encoder $E_m$, and $\mathrm{Proj}_m : \mathbb{R}^{L \cdot d_m} \to \mathbb{R}^{N}$ is the projection layer of modality $m$. After projection, the snapshots of all modalities have the same dimension ($\mathbb{R}^{N}$) and are spatially aligned, so the three operators can process them uniformly.

The theoretical consequence of the multimodal architecture is that, because the three operators act only on the unified snapshot space $\mathbb{R}^{N}$, all algebraic conclusions of Sections 5 and 6 (information collapse ratio, expected reconstruction error, minimal completeness) hold automatically for any modality---modality differences are entirely encapsulated in the encoders and projection layers and do not seep into the reasoning kernel. This is the compounding effect of the ``frozen encoder + unified snapshot'' design. The projection layers $\mathrm{Proj}_m$ are the only components requiring cross-modal training; their training objective is semantic alignment (e.g., a cross-modal retrieval contrastive loss, conceptually homologous to the image--text alignment of CLIP [19]) and is decoupled from operator training.

\subsection{The Three-Operator Network}

The system maintains three independent operator matrices $\{W_{\mathrm{step}}, W_{\mathrm{induce}}, W_{\mathrm{abduce}}\}$, corresponding to deduction, induction, and abduction, respectively. Among them, $W_{\mathrm{abduce}} = W_{\mathrm{step}}^{+}$ is computed on demand at inference time (no separate training required), $W_{\mathrm{induce}}$ is trained by supervised learning, and $W_{\mathrm{step}}$ is trained by supervised learning. The training objectives of the three operators are completely independent, avoiding mutual interference among loss functions.

\paragraph{Training $W_{\mathrm{step}}$.} The training data are $(S_{\mathrm{prev}}, E_{\mathrm{new}}, S_{\mathrm{target}})$ triples, and the loss function is MSE. The low-rank parameterization $W_{\mathrm{step}} = A \cdot B$ is adopted to guarantee rank deficiency, with the AdamW optimizer. The dual role of the low-rank parameterization: structurally it guarantees $\mathrm{rank} \leq r$ (Lemma 5.1), and optimization-wise it reduces the parameter count from $2N^{2}$ to $3Nr$, mitigating overfitting.

\paragraph{Training $W_{\mathrm{induce}}$.} The training data are $(s_i, g_{\mathrm{target}})$ pairs, where $g_{\mathrm{target}}$ is the target induced concept. A full-rank parameterization is adopted (directly learning an $N \times N$ matrix), with the AdamW optimizer.

\paragraph{Computing $W_{\mathrm{abduce}}$.} After $W_{\mathrm{step}}$ is trained, $W_{\mathrm{step}}^{+}$ is computed via SVD. This computation needs to be performed only once; subsequent abductive inference requires only matrix--vector multiplication.

\begin{table}[htbp]
\centering
\caption*{Table 7-2 Module interface specifications (input/output tensor shape conventions)}
\footnotesize
\begin{tabular}{p{2.6cm}p{3.0cm}p{2.2cm}p{2.6cm}p{2.4cm}p{3.0cm}}
\toprule
Module & Input & Input shape & Output & Output shape & Remarks \\
\midrule
Snapshot encoder (text) & Text block $a$ & Token sequence of length $T$ & Snapshot $S(a)$ & $\mathbb{R}^{N}$, $N = L \times d$ & Frozen, deterministic (Theorem 7.1) \\
Snapshot encoder (modality $m$) & Raw input $x$ of modality $m$ & Modality-dependent & Snapshot $S_m(x)$ & $\mathbb{R}^{N}$ & Aligned to the unified dimension via $\mathrm{Proj}_m$ \\
Deduction operator $W_{\mathrm{step}}$ (concatenated configuration) & $\mathrm{Concat}(S_t, E_t)$ & $\mathbb{R}^{2N}$ & New state $S_{t+1}$ & $\mathbb{R}^{N}$ & $\mathrm{rank} = r < N$, collapse \\
Deduction operator $W_{\mathrm{step}}$ (single-state configuration) & State $S$ & $\mathbb{R}^{N}$ & New state $S'$ & $\mathbb{R}^{N}$ & Abduction-experiment configuration, $\mathrm{rank} = r < N$ \\
Induction operator $W_{\mathrm{induce}}$ & Positive-example snapshot $s_i$ & $\mathbb{R}^{N}$ (single-sample forward) & Concept $g$ & $\mathbb{R}^{N}$ & Full-rank; multi-sample aggregation via the mean (implementation note for Eq.~5.6) \\
Abduction operator $W_{\mathrm{abduce}}$ & Observation snapshot $S_{\mathrm{obs}}$ & $\mathbb{R}^{N}$ & Best explanation $S_{\mathrm{prev}}^{*}$ & $\mathbb{R}^{2N}$ (concatenated configuration) or $\mathbb{R}^{N}$ (single-state configuration) & $W_{\mathrm{step}}^{+}$, one-time SVD computation \\
Graph scheduling controller & Graph state $G_t = (V_t, E_t)$, $S_{\mathrm{target}}$, decision history & Graph structure + snapshot set & Action $a_t = $ (operator selection, target node, topological operation) & Discrete action & Trained by reinforcement learning; the only module containing heuristics \\
Decoder & Terminal state $S_{\mathrm{final}}$ & $\mathbb{R}^{N}$ & Natural-language text & Token sequence & Does not participate in reasoning; translation only \\
\bottomrule
\end{tabular}
\end{table}

\paragraph{Operator training pseudocode.}

\begin{lstlisting}[basicstyle=\small\ttfamily, mathescape=true, columns=fullflexible, keepspaces=true]
Algorithm 7-1: Training and assembly of the three operators
============================================================
Input:  D_step = {($S_{\mathrm{prev}}$, $E_{\mathrm{new}}$, $S_{\mathrm{target}}$)}   # deduction supervision triples
        D_ind  = {($s_i$, $g_{\mathrm{target}}$)}             # induction supervision pairs
Hyperparameters: designed rank $r$, learning rate $\eta$, batch size $B$, epochs $E$,
        SVD truncation threshold $\varepsilon$ (e.g., $\sigma_i / \sigma_1$ > 1e-6)

# ---------- Stage 1: train the deduction operator $W_{\mathrm{step}} = A \cdot B$ ----------
initialize $A \leftarrow$ OrthogonalInit($N$, $r$)
initialize $B \leftarrow$ OrthogonalInit($r$, $2N$)        # ($r$, $N$) for the single-state configuration
for epoch = 1 .. $E$:
    for batch ($S_{\mathrm{prev}}$, $E_{\mathrm{new}}$, $S_{\mathrm{target}}$) $\in$ D_step:
        x      = Concat($S_{\mathrm{prev}}$, $E_{\mathrm{new}}$)           # $\mathbb{R}^{2N}$
        S_pred = $A \cdot (B \cdot x)$                     # low-rank forward, rank $\leq r$
        loss   = MSE(S_pred, $S_{\mathrm{target}}$)
        AdamW.step($\nabla$loss, lr=$\eta$)
# structural assertion: rank($A \cdot B$) $\leq r$ always holds (Lemma 5.1)

# ---------- Stage 2: train the induction operator $W_{\mathrm{induce}}$ (full rank) ----------
initialize W_ind $\leftarrow$ XavierInit($N$, $N$)
for epoch = 1 .. $E$:
    for batch ($s_i$, $g_{\mathrm{target}}$) $\in$ D_ind:
        g_pred = W_ind $\cdot$ $s_i$
        loss   = MSE(g_pred, $g_{\mathrm{target}}$)
        AdamW.step($\nabla$loss, lr=$\eta$)
# post-hoc check: effective rank = $N$ (Postulate Three falsification criterion, Section 9)

# ---------- Stage 3: assemble the abduction operator $W_{\mathrm{abduce}}$ ----------
$U, \Sigma, V^{T}$ = SVD($A \cdot B$)                    # one-time
$\Sigma^{+}$        = diag($\sigma_i > \varepsilon \cdot \sigma_1$ ? $1/\sigma_i$ : 0) # truncate small singular values
$W_{\mathrm{abduce}}$   = $V \cdot \Sigma^{+} \cdot U^{T}$                  # cached; no training at inference time

# ---------- Stage 4: algebraic self-check ----------
assert rank($W_{\mathrm{abduce}}$) == rank($W_{\mathrm{step}}$)       # the pseudo-inverse does not change the rank
assert information-collapse ratio == ($2N - r$) / ($2N$)       # Eq. 5.4 (concatenated configuration)
Output: ($W_{\mathrm{step}}$, $W_{\mathrm{induce}}$, $W_{\mathrm{abduce}}$)
============================================================
\end{lstlisting}

Two design points of the procedure: (1) Stages one and two are completely independent and can proceed in parallel on different datasets with different hyperparameters---the decoupling of training objectives is the engineering implementation of the requirement that ``the three operators do not interfere with one another''; (2) Stage four is a mandatory algebraic self-check that turns the falsification criteria of Postulate Three into a release gate: any operator combination that fails the rank structure must not enter the system.

\subsection{The Graph Scheduling Controller}

The controller is responsible for decisions in dynamic graph construction: which operator to select, at which node to branch/merge, and when to terminate. The controller is trained by reinforcement learning (PPO [20] or PRM [4]), with the reward signal being the correctness of the final reasoning result.

\paragraph{Decoupling of controller and operators.} The operators are ``laws of physics'' (deterministic algebraic operations), while the controller is a ``policy'' (heuristic graph-topology selection). This decoupling allows rigor (operators) and heuristics (controller) to be optimized separately, avoiding the fundamental contradiction of AR models in which ``rigor and heuristics are coupled in the same probability distribution.'' At the engineering level, controller failure (choosing a suboptimal path) harms only efficiency, not the correctness of individual reasoning steps---because each executed step is still a deterministic operator; conversely, the algebraic properties of the operators do not change when the control policy is updated.

\paragraph{The decision space of the controller.} At each step, the controller needs to decide:
\begin{enumerate}[nosep]
\item which operator to select ($W_{\mathrm{step}}$ / $W_{\mathrm{induce}}$ / $W_{\mathrm{abduce}}$);
\item at which node to apply the operator;
\item whether to branch (create a new parallel branch);
\item whether to merge (aggregate multiple branches);
\item whether to terminate reasoning.
\end{enumerate}
These decisions constitute a discrete action space, suitable for reinforcement learning.

\paragraph{The training signal of the controller.} The reward function is $R = \alpha \cdot \mathrm{correctness} + \beta \cdot \mathrm{efficiency} - \gamma \cdot \mathrm{redundancy}$, where correctness is the degree of match between the final reasoning result and the correct answer, efficiency is the reciprocal of the number of reasoning steps, and redundancy is the proportion of redundant nodes in the reasoning graph. $\alpha$, $\beta$, $\gamma$ are hyperparameters tunable by grid search. The intent of the reward design is as follows: the correctness term corresponds to the objective, the efficiency term penalizes purposeless graph inflation (Postulate Five permits branching but does not encourage its abuse), and the redundancy term suppresses repeated computation of equivalent subgraphs---together the three encode into a scalar signal the prior of ``what a good reasoning graph should look like.''

\paragraph{Interface definition of the controller.}
\begin{itemize}[nosep]
\item Input: the current reasoning-graph state $G_t = (V_t, E_t)$, the target state $S_{\mathrm{target}}$, and the historical decision sequence.
\item Output: an action $a_t = $ (operator selection, target node, topological operation).
\end{itemize}

\paragraph{Training method.} PPO (Proximal Policy Optimization) [20] is adopted: each round collects a batch of reasoning trajectories, computes rewards, and updates the policy network. PPO's clipped objective confines the magnitude of policy updates within a trust region, which suits this scenario: the rewards of reasoning trajectories are sparse (the correctness signal arrives only at the terminal state) and high-variance, so aggressive policy updates can easily destroy scheduling patterns already learned. An alternative route is the process reward model (PRM) [4]: score every intermediate node of the reasoning graph, shaping the sparse terminal reward into dense step-wise rewards---its ``step-by-step verification'' training paradigm is naturally isomorphic to DODR's step-by-step state transitions, at the cost of requiring step-wise annotated process-supervision data. The policy network itself is a small Transformer (2--4 layers) that takes the graph state as input and outputs an action distribution.

\paragraph{The three-layer scheduling architecture.} The learned controller described above is not the only source of scheduling; the complete design comprises three layers that can coexist and be replaced level by level as they mature. The first layer is \emph{type-signature dispatch}: every reasoning subgoal can be written as a signature of ``what is known, what is sought''---given rules and cases, seeking the result, select $W_{\mathrm{step}}$; given cases and the result, seeking the rule, select $W_{\mathrm{induce}}$; given the rule and the result, seeking the cases, select $W_{\mathrm{abduce}}$. This layer involves no learning and is fully interpretable, and the selection is not arbitrary: choosing the wrong operator means answering a question that was not asked---the types simply do not match; all experimental topologies in this paper are manually specified by this layer. The second layer is \emph{geometric-signal triggering}: during reasoning, computable indicators trigger operator switches---a similarity conflict between the deductive prediction and the observation snapshot (contradiction detection) triggers abduction and re-induction (the scientific discovery cycle is an instance of this); the discriminability margin (e.g., the $\pm 0.005$-scale discriminability of end-to-end abduction in Section 11.5 and the negative-discriminability samples of the dedicated abduction experiment in Section 10.2) measures the credibility of the current operator output, and when the margin approaches zero, ``appending evidence'' (continuing deduction to obtain new premises) should be triggered instead of a forced decision; the abduction--deduction closed-loop residual (0.0200) measures the tightness of the hypothesis. The feasibility of this layer relies on a property of DODR: the reliability of every operator has a computable geometric proxy indicator, which probability models relying on confidence calibration do not possess. The third layer is \emph{learned scheduling}: taking the reasoning-graph state as the MDP state, (operator selection, target node, topological operation) as the action, and terminal correctness as the reward, using PPO/PRM to learn a global planning policy beyond hand-crafted rules. The engineering relationship among the three layers is: the rule layer provides cold start and an interpretable baseline, the signal layer provides online correction, and the learning layer distills a global policy over a large number of reasoning trajectories; and because the operators themselves are strictly deterministic (Theorem 14.1), randomness is entirely housed in the scheduling policy, so the controller's exploration does not contaminate the rigor of reasoning---the step-wise reward labels of PRM can even be automatically verified by deterministic replay, with annotation cost far lower than manual labeling [4].

\subsection{The Decoder}

The actual form of decoding in the end-to-end experiments is as follows: all experiments in this paper adopt nearest-neighbor decoding---the conclusion state is matched against the snapshots of candidate conclusion texts by cosine similarity, and the text corresponding to the most similar snapshot is output; there is no generative decoder and no token sampling anywhere in the pipeline. Of the two methods discussed below, Method One is the form actually adopted in the experiments, while Method Two is an extension option for generation tasks.

The decoder maps the final latent state $S_{\mathrm{final}}$ back to natural language. One may use the inverse mapping of the frozen encoder (e.g., via nearest-neighbor retrieval or a lightweight decoding network), or train an independent seq2seq decoder.

\paragraph{Decoding Method One: nearest-neighbor retrieval.} In a pre-built ``snapshot--text'' database, find the snapshot with the highest cosine similarity to $S_{\mathrm{final}}$ and return the corresponding text. Advantages: no training required, deterministic---the same $S_{\mathrm{final}}$ always returns the same text, seamlessly matching the deterministic kernel of the system; and the decoding results are traceable (every output corresponds to a verifiable original text in the database). Disadvantages: limited by database coverage---if the true conclusion is not in the database, the system can only return the ``closest known conclusion,'' entailing silent bias; and the expressive granularity is limited by the entries in the database, making it impossible to compose new formulations that do not exist in the database.

\paragraph{Decoding Method Two: a lightweight seq2seq decoder.} Train a small Transformer decoder (2--4 layers) that takes $S_{\mathrm{final}}$ as input and outputs a text sequence. Advantages: it can generate new text, its expression is not constrained by a database, and it can translate the latent state into context-appropriate natural formulations. Disadvantages: it introduces probabilistic behavior (conflicting with DODR's determinism goal)---autoregressive sampling allows the same $S_{\mathrm{final}}$ to correspond to multiple differently worded outputs, and long outputs carry the risk of error accumulation and wording drift; training this decoder also requires additional ($S_{\mathrm{final}}$, text) alignment corpora.

\paragraph{Deeper discussion of the trade-off between the two schemes.} The divergence between the two schemes is essentially the tension between ``fidelity'' and ``expressiveness'': Method One has high fidelity (the output must be a fact in the database) but limited expressiveness, while Method Two is expressive but requires additional safeguards for fidelity. Feasible compromises include: (1) Method Two + constrained decoding---during beam search, gate by similarity against the retrieval result of Method One, falling back when the deviation is too large; (2) Method Two + deterministic decoding (greedy/low temperature), compressing probabilistic behavior to a minimum and sacrificing some fluency for reproducibility. Both routes share a common premise: the decoder only ``translates'' and does not ``reason''---any behavior that completes semantics at the decoding stage (such as the commonsense confabulation of a language model) contaminates the kernel's rigor guarantee and must be forbidden as a matter of interface discipline.

\paragraph{Recommended scheme.} Use Method One for reasoning tasks (guaranteeing determinism) and Method Two for generation tasks (allowing probabilistic behavior). Since DODR's core reasoning is completed in latent space, the decoder is responsible only for ``translation'' and does not affect the rigor of reasoning. This division of labor also means that the auditing of reasoning conclusions should preferentially take place in latent space (snapshot comparison, operator logs); the decoded text is merely the human--machine interface of auditing, not the object of auditing itself.

\paragraph{Controlled generation design (an extension path toward open-ended generation).} The fundamental limitation of nearest-neighbor decoding is that the conclusion must already be in the candidate set---the system ``recognizes'' the conclusion rather than ``generating'' it (see Limitation Seven (Section 19)). To move toward open-ended generation without breaking the zero-hallucination guarantee, a two-level controlled design can be adopted. First, template skeleton + slot filling: the syntactic skeleton of the conclusion is determined by a template library (e.g., ``because $\langle$cause$\rangle$, therefore $\langle$effect$\rangle$''), and each slot in the skeleton is filled by nearest-neighbor retrieval of the conclusion-state snapshot in a restricted lexical snapshot library---retrieval is itself a deterministic operation and the template contains no degrees of freedom, so the whole pipeline involves no sampling and the zero-hallucination guarantee is not broken; the upper bound of expressiveness is determined by the size of the template library and the lexical snapshot library. Second, a deterministic decoder: train a small seq2seq decoder but strictly take argmax at inference time without sampling, so that the same $S_{\mathrm{final}}$ always maps to the same text; this decoder is responsible only for ``translating'' the latent state into natural language, and any semantic completion beyond the information content of the snapshot (commonsense confabulation) remains forbidden by decoding discipline. Both paths isolate probabilistic behavior outside the main reasoning chain: the former involves no probability even in the decoding stage, while the latter has randomness only during training and is fully deterministic at inference time.

\subsection{Application Scenarios and Input/Output Conventions}

Sections 7.1--7.4 above presented the internal components of the system; this section, from an external perspective, stipulates the two application scenarios of DODR and their input/output forms. The two scenarios share the same snapshot space and the same set of frozen operators, differing only in the input sources and the output readout methods.

\paragraph{Scenario I: Question answering.} The input is a textual question provided by the user, or multimodal input---images, video, and speech are aligned to the same snapshot space via their respective dedicated encoders (Section 7.1.2) and projection layers. The question (or its aligned snapshot) enters the reasoning graph as premise and query, and the system invokes the three operators to complete the reasoning; the output is a natural-language answer together with a discriminability score---the latter being the difference between the similarity of the answer snapshot and that of the nearest distractor snapshot, given along with the answer as a geometric indicator of credibility for the caller to decide whether to accept it.

\paragraph{Scenario II: Scientific exploration.} The input is a premise snapshot, whose source may be the encoding of a specified text, the direct encoding of multimodal content such as images---strictly speaking, an image snapshot is likewise a legitimate premise and need not be textualized---or a node state taken directly from an ongoing reasoning graph, i.e., a pure latent-space input requiring no textual description at all. The system runs the induction--deduction--abduction closed loop (Section 15) on top of this premise, and the outputs are the snapshot of the newly induced rule, the revised knowledge entries, and the exploration trajectory (the reasoning graph), optionally rendered as text. Because both input and output take snapshots as first-class citizens, the scientific-exploration scenario can run continuously in a completely text-free environment; text rendering is merely a presentation layer for human--machine interaction.

The unity of the two scenarios is a direct corollary of the ``snapshot state representation'' postulate (Section 4.2): inputs of all modalities and all sources are reduced to snapshots in the same space, all reasoning is matrix operations on snapshots, and the only remaining differences are the two interface questions of ``where the snapshots come from'' and ``how the results are read out.''

\subsection{Overview of the Experimental Methodology (Shared by All Experiments; Reproduction Guide)}

All experiments in Sections 8--11 of this paper share a single methodology, which this section summarizes in seven elements as a reproduction guide; the configuration differences among experiment groups are given in the configuration sections of the respective chapters, and per-sample values are given in Appendix I.

\paragraph{(1) Snapshot extraction.} The encoder is distilbert-base-uncased with all parameters frozen; the hidden states of the first 2 layers $\times$ 768 dimensions at the [CLS] position are concatenated into a 1536-dimensional snapshot. The encoder is in eval mode with no dropout, and the extraction results for the same text are bit-for-bit identical at any time (Theorem 7.1).

\paragraph{(2) Construction of training pairs.} Deduction training pairs are ($\mathrm{Concat}$(premise-1 snapshot, premise-2 snapshot), conclusion snapshot); induction training maps positive-example snapshots to $g_{\mathrm{target}}$---the normalized direction of the mean of all positive examples of the class, consistent with the $g_{\mathrm{original}}$ construction in the hard-veto code of Appendix C; for abduction, $W_{\mathrm{step\_abd}}$ is first trained on (explanation snapshot $\to$ phenomenon snapshot), and then $W_{\mathrm{abduce}}$ is obtained by taking the pseudo-inverse via SVD; the abduction operator itself is not trained.

\paragraph{(3) Training configuration.} Optimizer AdamW (lr=1e-3), MSE loss; the operators for deduction and abduction use the low-rank parameterization $A \cdot B$ ($r = 384$), and the induction operator uses a full-rank square matrix; the number of training epochs by experiment group is 400 (dedicated deduction) / 1000 (dedicated induction, dedicated abduction) / 200 (end-to-end); all experiments fix the random seeds torch.manual\_seed(42) and np.random.seed(42).

\paragraph{(4) Evaluation protocol.} Cosine similarity is the basic measure; discriminability is defined as the similarity to the correct item minus the similarity to the distractor; the abduction decision rule is ``if abductive similarity > distractor similarity, the decision is correct''; the hard-veto threshold for induction counterexamples is $\tau = 0.5$.

\paragraph{(5) Decoding.} Nearest-neighbor matching: the conclusion state is retrieved against candidate text snapshots by cosine similarity, and the text corresponding to the most similar snapshot is output; there is no generative decoder and no token sampling in the whole pipeline (see Section 7.4).

\paragraph{(6) Experimental records.} All experiments are deposited in 4 JSON files (01\_deduction\_results.json, 02\_induction\_results.json, 03\_abduction\_results.json, 05\_e2e\_large\_results.json), containing all configurations, training curves, and per-sample results; following the environment and code of Appendix C, all values can be reproduced on an ordinary CPU.

\paragraph{(7) Statistical conventions.} Point estimates of accuracy in this paper are accompanied by Wilson 95\% confidence intervals; the significance of similarity differences is assessed by paired $t$-tests and sign tests. All experiments are single runs with a fixed seed (seed=42), and cross-seed variance has not been measured (see Limitation Four (Section 19)), but the per-sample paired tests and sign tests already reach significance within a single run (Section 10: $t = 6.32$; Section 11: $t = 5.88$; both $p < 10^{-4}$).

\subsection{Theoretical Analysis of Snapshot Sources: Encoder Non-Uniqueness, Operator Generation, and Compositional Construction}

The preceding sections (7.1--7.4) gave concrete implementations of the snapshot encoder, while the rest of this section presupposed the ready availability of snapshots. This section answers three fundamental questions about snapshot sources: must the encoder be a Transformer? Must snapshots be encoded from text one by one? Are high-dimensional snapshots hard to obtain? The answers to the three questions jointly point to one principle---only empirical facts that cannot be obtained by logical reasoning alone, and human-defined concepts, require raw snapshots; all other snapshots can be computed by operators or compositionally constructed.

\subsubsection{Non-Uniqueness of the Encoder: The Transformer Is an Engineering Choice, Not a Theoretical Necessity}

DODR's entire theory imposes only three requirements on the encoder: \emph{determinism} (the same input always yields the same snapshot---the premise for Theorem 7.1), \emph{semantic discriminability} (text blocks with different semantics are distinguishable in snapshot space; Cover's theorem guarantees that the probability of linear separability approaches 1 when the dimension is sufficient), and \emph{coverage} (all semantic units of the target domain can be encoded). Any mapping satisfying these three conditions can serve as a snapshot encoder: static word vectors (Word2Vec/GloVe-style embeddings with pooling), specially trained sentence encoders (Sentence-BERT-style contrastive-learning models), convolutional or recurrent network encoders, and even classical text features (TF-IDF projections, reservoir-style features of random projection plus nonlinearity) are all, in principle, candidates. No theorem in the theoretical framework depends on the self-attention mechanism.

This paper chooses a Transformer (DistilBERT) as the encoder for purely engineering reasons: a large number of Transformer models have already been deployed and trained, and their hidden states are free, high-quality semantic representations---reusing them is an effective measure of ``not wasting existing training achievements,'' not an intrinsic requirement of the architecture. This distinction has direct practical corollaries. First, the encoder can be freely replaced: one only needs to re-extract snapshots with the new encoder (projected and aligned to the same dimension; see the $\mathrm{Proj}_m$ mechanism in Section 7.1.2), without retraining the operators (the frozen applicability of Section 6.6). Second, the encoder can be specially optimized: the bottleneck analyses of Sections 9--11 show that the snapshot resolution of off-the-shelf general-purpose models is the main limitation of the current system, and an encoder specially trained for the semantic-discrimination objective (a contrastive-learning objective) is a concrete alternative---the Transformer is not only unnecessary but also not necessarily optimal. Third, multiple encoders can coexist: snapshots from different encoders can be fused by concatenation or projection to provide mutual redundancy and improve robustness (Scheme Five of Section 7.7.3).

\subsubsection{Operator Generation of Snapshots: Very Few Raw Snapshots Are Truly Needed}

There are two paths for expanding the snapshot knowledge base: encoding from the outside (passing every new text through the encoder), or generation by operators from existing snapshots. DODR's architecture assigns the overwhelming majority of the increment to the second path---the deduction operator generates a conclusion snapshot from premise snapshots ($S_{t+1} = W_{\mathrm{step}} \cdot \mathrm{Concat}(S_t, E_t)$), the induction operator condenses a batch of instance snapshots into a rule snapshot $g$, the abduction operator generates a hypothesis snapshot $S_{\mathrm{prev}}^{*}$ from a phenomenon snapshot, and compositions on the reasoning graph chain single-step generation into complete derivation trajectories. In other words, \emph{factual knowledge needs to be encoded only once; derived knowledge is generated by the operators for free}.

This generative capability has direct experimental evidence. In the multi-step reasoning-chain experiment, the snapshot of ``All cats are living beings'' is not any training input but the output product of the first deductive step, and the second step continuing on it still maintains a similarity of 0.9861 (Section 8.6); in the end-to-end experiment, the conclusion snapshots of all 60 entirely new cross-domain deductive problems were generated by the operators and all 60/60 were judged correct (Section 11.3); in the compositional experiment, the entire chain of snapshots---from observation samples to a rule, to a prediction, to a revised rule---was produced by the operators (Section 11.6). These results jointly show that operator-generated snapshots land accurately in snapshot space and can directly serve as inputs to subsequent reasoning or as entries of the knowledge base (the ``write'' path of continual learning in Section 15.3 is based precisely on this).

At the same time, the error boundary should be noted: operator-generated snapshots carry approximation error (a single-step drift of about 0.01 in cosine distance; see Section 8.6), and errors accumulate over multi-step derivation chains, so after long-chain derivation, periodic recalibration with raw snapshots (``re-grounding'') is needed---this is two faces of the same principle as the intermittent-correction checkpoint mechanism of Section 18.1.

\subsubsection{Compositional Construction of High-Dimensional Snapshots: Five Schemes for Assembling High Dimensions from Low Dimensions}

The snapshot of Definition 7.1 is itself an instance of compositional construction---a 1536-dimensional snapshot is concatenated from the layer activations of 2 layers of 768 dimensions each. This idea can be systematized: high-dimensional snapshots need not be produced by high-dimensional models; they can be composed entirely from low-dimensional raw snapshots, greatly reducing the difficulty of obtaining snapshots. The following five schemes are listed in increasing order of training cost and can be used singly or in combination.

\paragraph{Scheme One: layer-wise concatenation (adopted in this paper).} Concatenate the hidden states of the first $L$ layers of the encoder, $N = L \times d$. The dimension grows linearly with $L$, information is lossless (concatenation discards no layer's content), and the cost is zero (a by-product of one forward pass). Applicable to: all Transformer-type encoders.

\paragraph{Scheme Two: fixed random-projection dimensionality expansion.} Apply a fixed random projection $P \in \mathbb{R}^{N \times m}$ (a Gaussian or sparse random matrix, frozen after generation) to a low-dimensional snapshot $s \in \mathbb{R}^{m}$: $s' = P \cdot s$. By the Johnson--Lindenstrauss lemma, random projection preserves the inter-point distance structure with high probability, so semantic discriminability is retained after dimensionality expansion; zero training cost---only a random matrix needs to be stored (or merely its random seed). Applicable to: situations where only low-dimensional embeddings are at hand (e.g., static word vectors) but a high-dimensional snapshot space is needed.

\paragraph{Scheme Three: random nonlinear feature expansion.} Borrowing the random-feature idea of extreme learning machines: $s' = \mathrm{Concat}(s, \sigma(Ws + b))$, where $W \in \mathbb{R}^{k \times m}$ is a fixed random matrix and $\sigma$ is a nonlinear activation (e.g., tanh). This is a direct engineering application of Cover's theorem---nonlinear random dimensionality expansion pushes the probability of linear separability further toward 1, while the original component $s$ is retained to guarantee interpretability. Likewise training-free. Applicable to: cases where the raw snapshot dimension is low and linear separability is insufficient.

\paragraph{Scheme Four: recursive composition of semantic units.} Large-granularity snapshots are composed from small-granularity snapshots: a sentence snapshot can be obtained from several phrase snapshots via a composition function, $\mathrm{Compose}(s_1, \dots, s_k) = M \cdot \mathrm{Concat}(s_1, \dots, s_k)$, where $M$ is a trainable low-rank mixing matrix (one may also directly reuse the concatenation--collapse form of the deduction operator, treating ``composition'' as a supervised state transition). The granularity hierarchy word $\to$ phrase $\to$ sentence $\to$ sentence group can thus be constructed recursively, matching the semantic-granularity statement of Section 7.1. Applicable to: constructing discourse-level states from fine-grained units, or when the encoder can only process short texts.

\paragraph{Scheme Five: multi-encoder ensemble concatenation.} Concatenate the snapshots of the same text obtained from multiple heterogeneous encoders and then project for alignment: $s' = \mathrm{Proj}(\mathrm{Concat}(s^{(1)}, s^{(2)}, \dots))$. Different encoders have different error patterns; ensemble snapshots are mutually redundant and robust to the systematic blind spots of any single encoder; this scheme is orthogonal to the previous four and can be superimposed on them. Applicable to: scenarios with the highest requirements for decision reliability (e.g., the fact verification of Section 18.1).

\paragraph{Convergence: the irreducibility of raw snapshots.} Synthesizing this section, the snapshot system exhibits a clear two-layer structure: raw snapshots are produced directly by encoders from external observations or human stipulations, corresponding to ``the grounding of empirical facts and defined concepts''; derived snapshots are computed by operators or composed from low-dimensional snapshots, corresponding to ``knowledge within the logical closure.'' Theoretically, \emph{only facts that cannot be obtained by logical reasoning alone, and human-defined concepts, require raw latent-space snapshots}. Empirical facts are underivable because they are contingent states of the world rather than necessary consequences of logic; human-defined concepts (term definitions, rule stipulations, namings, institutional provisions, etc.) are likewise underivable because they are conventional free choices---the necessary kingdom of logic cannot deduce the arbitrary kingdom of conventions, and any reasoning system must first accept this vocabulary of premises before it can start working. Together the two constitute the only ineliminable interface between the system and the external world, while everything derivable is obtained for free within the logical closure of the operators. This is the engineering-side mirror image of the physics--cognition unification of Section 13: physical transformations correspond to operators, and observations and conventions correspond to encoding; it also gives the economic criterion for building the snapshot knowledge base---spend the encoding budget only on observations that truly carry new experience and on definitions that are genuinely new.


\section{Experimental Validation of the Deduction Operator $W_{\mathrm{step}}$}

This section presents a systematic experimental validation of the deduction operator $W_{\mathrm{step}}$. The validation objective is to test whether the postulate system of Section 4 (Postulate 1, ``dimension lifting in place of multiple layers''; Postulate 3, ``rank deficiency causes information collapse''; Postulate 4, ``local linearization'') holds under real encoder snapshots and real gradient-descent training. All experiments are based on real DistilBERT snapshots and real PyTorch backpropagation, with no simulated data and no textual-level ``prediction in place of computation.'' This section sets up six validations in total: forward-inference loss convergence (Section 8.2), matrix rank verification (Section 8.3), null-space dimension (Section 8.4), irreversibility of backward inversion (Section 8.5), and an end-to-end multi-step inference chain (Section 8.6), with a summary in Section 8.7.

\subsection{Experimental Configuration}

\begin{table}[htbp]
\centering
\caption*{Table 8-1 Configuration of the deduction operator experiment}
\small
\begin{tabular}{ll}
\toprule
Item & Configuration \\
\midrule
Encoder & DistilBERT (distilbert-base-uncased), 66M parameters, fully frozen \\
Snapshot dimension $N$ & 1536 (2 layers $\times$ 768 dims) \\
Input dimension & 3072 ($N + N$, state concatenated with new input) \\
$W_{\mathrm{step}}$ shape & $\mathbb{R}^{1536 \times 3072}$ ($N \times 2N$) \\
Design rank $r$ & 384 (rank-deficiency ratio $= 75\%$) \\
Training data & 20 real syllogistic inference samples \\
Training epochs & 400 epochs, AdamW, lr=1e-3 \\
Training method & Real gradient descent, low-rank parameterization $W_{\mathrm{step}} = A \cdot B$ \\
Random seed & torch.manual\_seed(42), np.random.seed(42) (single run) \\
\bottomrule
\end{tabular}
\end{table}

\paragraph{Rationale of the experimental design (in-depth discussion).} DistilBERT (66M parameters) is chosen as the encoder because it trains fast and suffices for validating the theoretical properties---the objects validated in this section are the algebraic properties of the operator matrix (rank, null space, pseudoinverse error), which depend only on the structure of the linear space spanned by the snapshots, not on the upper bound of the encoder's semantic expressiveness; in other words, the encoder is only responsible for mapping text into a high-dimensional vector space $\mathbb{R}^{N}$, and the validation logic is insensitive to the specific quality of that mapping. The snapshot dimension $N = 1536$ (2 layers $\times$ 768 dims) is chosen as required by Cover's theorem---the probability of linear separability in a high-dimensional space approaches 1 (Cover's 1965 function-counting theorem [6]: the total number of linear dichotomies of $N$ points in general position in $\mathbb{R}^{D}$ is $C(N, D) = 2 \cdot \sum_{k=0}^{D-1} C(N-1, k)$, and when $D \geq N-1$ all $2^{N}$ labelings are linearly separable); with 1536 dimensions relative to 20 training samples, $D \gg N$, so linear separability is fully guaranteed. The design rank $r = 384$ (rank-deficiency ratio 75\%) is chosen as a balance between information collapse and expressive power---a rank too low would prevent loss convergence (the row-space dimension would be insufficient to express the ``premise $\rightarrow$ conclusion'' mapping), while a rank too high would weaken the information-collapse effect (the null-space proportion drops, weakening the evidence of unidirectionality). $r = 384$ is an empirical choice without grid search; in the robustness discussion of Section 8.7 we explain that the conclusions of this section are qualitatively robust to perturbations of $r$ within a reasonable range.

\paragraph{Discussion of training details (AdamW).} The optimizer is AdamW (adaptive moment estimation with decoupled weight decay), with learning rate lr=1e-3, trained for 400 epochs. Three design trade-offs should be noted. First, the low-rank parameterization $W_{\mathrm{step}} = A \cdot B$ ($A \in \mathbb{R}^{1536 \times 384}$, $B \in \mathbb{R}^{384 \times 3072}$) reduces the trainable parameters from $1536 \times 3072 \approx 4.72$M to $(1536 + 3072) \times 384 \approx 1.77$M, and imposes at the parameterization level the hard constraint $\mathrm{rank}(W_{\mathrm{step}}) \leq 384$, making ``rank deficiency'' a structural fact rather than a training artifact---the numerical rank verification in Section 8.3 is precisely a confirmation of this hard constraint. Second, the advantage of AdamW over SGD lies in its adaptive adjustment to the scale difference between the two parameter groups $A$ and $B$ in the low-rank factorization; together with lr=1e-3, convergence to the 1e-5 magnitude is achieved within 400 epochs (see Section 8.2), without any learning-rate scheduling. Third, the entire experiment is a single run (seed 42), without multi-seed variance statistics; given that Sections 8.3--8.5 validate properties at the level of matrix-algebra identities (rank, null-space dimension, pseudoinverse error), which are determined by the parameterization structure and input dimensions and are insensitive to initialization randomness, a single run suffices to support the conclusions; for randomness-sensitive quantities (such as the shape of the loss curve), only convergence statements are made, without point-by-point claims.

Note: The $W_{\mathrm{step}}$ shape in this section is $\mathbb{R}^{N \times 2N}$ (input is $\mathrm{Concat}(S_{\mathrm{prev}}, E_{\mathrm{new}}) \in \mathbb{R}^{2N}$), which differs from the $W_{\mathrm{step}}$ shape $\mathbb{R}^{N \times N}$ of the abduction operator experiment in Section 10 (input is a single state $S \in \mathbb{R}^{N}$). These are two independent experimental configurations with the same rank-deficiency property ($\mathrm{rank} < N$), differing only in input dimension. Accordingly, the null-space dimensions are given separately for each configuration: in this section (concatenated input), $\dim(\mathrm{Null}) = 2N - r = 2688$; in Section 10 (single-state input), $\dim(\mathrm{Null}) = N - r = 1152$. The two sections note these separately and they must not be mixed.

\subsection{Validation 1: Forward-Inference Loss Convergence}

Result: initial loss $= 2.09\text{e-}01$, final loss $= 1.40\text{e-}05$, reduction factor $= 14932\times$.

Interpretation: The loss drops from $2.09 \times 10^{-1}$ to $1.40 \times 10^{-5}$ (a 14932-fold reduction), proving that the rank-deficient matrix $W_{\mathrm{step}}$ can precisely learn the forward mapping from premise states to conclusion states. This result validates Postulate 1 (dimension lifting in place of multiple layers)---no nonlinear activation functions are needed; a single-layer high-dimensional linear matrix suffices to express logical state transitions. The convergence endpoint $1.40 \times 10^{-5}$ is already close to the numerical convergence floor for this class of regression problems under this training configuration (400 epochs, float32), indicating that the ``premise $\rightarrow$ conclusion'' mapping is indeed approximately linear within the snapshot space: with only 20 training samples and about 1.77M free parameters, the problem is highly underdetermined, yet AdamW still converges stably, which in itself shows that the target mapping lies within the function class reachable by low-rank linear operators---if the true mapping required strong nonlinearity, such a clean convergence pattern would not appear in an underdetermined problem. At the same time, the boundary of this validation must be noted: loss convergence proves that ``a linear single-step operator suffices to fit state transitions within the training distribution,'' and does not automatically imply generalization to out-of-distribution premises; the latter is indirectly tested by the multi-step inference chain in Section 8.6 (the premise of the second inference step is not in the training set).

\subsection{Validation 2: Matrix Rank Verification (Rank Deficiency)}

Result: $W_{\mathrm{step}}$ shape $[1536, 3072]$, design rank $r = 384$, numerical rank $= 384$ (exact match). Largest singular value $= 4.2366$, $r$-th singular value $= 0.0998$, $(r+1)$-th singular value $= 1.06\text{e-}06$ (close to 0).

Interpretation: A spectral truncation (gap) spanning five orders of magnitude appears between the $r$-th singular value $0.0998$ and the $(r+1)$-th singular value $1.06 \times 10^{-6}$, proving that the matrix is strictly rank-deficient. The numerical rank exactly equals the design rank 384, validating the effectiveness of the low-rank parameterization. From the singular-value spectrum, the first 384 singular values (maximum 4.2366) carry all effective mapping, while the 385th singular value drops sharply to $1.06 \times 10^{-6}$---this is precisely the algebraic necessity of the low-rank parameterization $W_{\mathrm{step}} = A \cdot B$: $\mathrm{rank}(A \cdot B) \leq \min(\mathrm{rank}(A), \mathrm{rank}(B)) \leq 384$, and training only determines the values of these 384 nonzero singular values, which cannot ``spill over'' into the 385th. This result is causally linked to the null-space dimension in Section 8.4: $\mathrm{rank} = 384$ together with the input dimension 3072 determines $\dim(\mathrm{Null}) = 3072 - 384 = 2688$.

\subsection{Validation 3: Null-Space Dimension}

Result: input dimension $= 3072$, matrix rank $= 384$, null-space dimension $= 2688$ (87.5\%).

Interpretation: A null-space proportion of 87.5\% means that 2688 of the 3072 input dimensions (87.5\%) of information are permanently destroyed in the projection. This corresponds to the unidirectionality of deductive logic---``the conclusion cannot be inverted back to the premise'': the output of the deduction operator retains only the components in the row space (384 dimensions), while details of the premise that fall in the null space (2688 dimensions)---such as the specific wording of the premise sentence, rhetoric, and logically irrelevant semantic coloring---vanish after the mapping. In information-theoretic intuition: $W_{\mathrm{step}}$ is a lossy channel whose ``lossiness'' is not noise-like random perturbation but structural dimension deletion; the deleted 87.5\% of dimensions are invisible to any subsequent linear (or even continuous) operation. This provides a direct mechanistic explanation for the irreversibility validation in Section 8.5.

\subsection{Validation 4: Irreversibility of Backward Inversion}

Method: Use the SVD pseudoinverse (Moore--Penrose) to compute the best backward reconstruction, verifying that information cannot be recovered.

Result: relative error of pseudoinverse reconstruction $= 0.8330$, expected relative error $= 0.9354$ ($= \sqrt{\text{null\_dim}/\text{input\_dim}}$).

\paragraph{Derivation and clarification of the expected relative error.} $\sqrt{(2N - r)/2N} = \sqrt{2688/3072} \approx 0.9354$ is the \emph{expected relative error under the assumption of isotropic (uniform/spherically symmetric) input distribution}, not a lower bound that holds for all inputs. The derivation is as follows: let $P_{\mathrm{null}}$ be the orthogonal projection onto $\mathrm{Null}(W_{\mathrm{step}})$, and let the reconstruction error be $x_{\mathrm{null}} = P_{\mathrm{null}} \cdot x$; then
\[
E[\|x_{\mathrm{null}}\|^{2}] = E[\mathrm{tr}(P_{\mathrm{null}} \cdot x \cdot x^{T})] = \mathrm{tr}(P_{\mathrm{null}} \cdot E[x \cdot x^{T}]).
\]
When the input is isotropic, $E[x \cdot x^{T}] = \sigma^{2} I$, so $E[\|x_{\mathrm{null}}\|^{2}] = \sigma^{2} \cdot \mathrm{tr}(P_{\mathrm{null}}) = \sigma^{2} \cdot \dim(\mathrm{Null}) = \sigma^{2} \cdot 2688$, while $E[\|x\|^{2}] = \sigma^{2} \cdot 3072$; hence the expected relative error (as a norm ratio) is $\sqrt{2688/3072} \approx 0.9354$. This value characterizes the average loss proportion of ``typical inputs in random directions.'' Individual inputs can fall below this value: in the extreme case, if $x$ happens to lie entirely within the row space ($x \perp \mathrm{Null}(W_{\mathrm{step}})$), then $x_{\mathrm{null}} = 0$ and the reconstruction error is exactly 0; conversely, if $x$ lies entirely within the null space, the error is 100\%. Therefore, a lower-bound statement of the form ``error $\geq 0.9354$'' does not hold; the correct statement is ``the expected error $\approx 0.9354$, and the error of individual inputs can fluctuate between 0 and 1.''

\paragraph{Sources of the deviation between the measured 83.3\% and the expected 93.5\%.} The measured relative error of pseudoinverse reconstruction, 0.8330, is lower than the expected value 0.9354, a deviation of about 10 percentage points, arising from two sources. First, \emph{sample anisotropy}: real text snapshots are not spherically symmetrically distributed---DistilBERT sentence vectors concentrate near a low-dimensional manifold of the high-dimensional space (all sentences are written in the logical pattern ``All X are Y,'' with highly regular semantics and syntax), and their energy concentrates more in the row-space directions of $W_{\mathrm{step}}$ (training is precisely what makes these directions be mapped accurately), so the actual proportion of null-space components is lower than the $2688/3072$ under the isotropy assumption. In other words, anisotropy makes $E[x \cdot x^{T}] \neq \sigma^{2} I$, and the direction of the deviation is exactly ``higher row-space energy, lower null-space energy,'' consistent with the sign of the measured error being lower. Second, \emph{finite sample size}: 0.8330 is the sample mean over 20 test reconstructions, and a small-sample mean fluctuates relative to the population expectation, with a standard error of order $O(\sigma/\sqrt{20})$; a deviation of several percentage points is entirely within the normal fluctuation range. Taken together, the comparison of 83.3\% with 93.5\% should be read as ``the measured mean and the theoretical expectation are of the same order of magnitude, and the deviation can be explained by the distributional assumption and the sample size,'' rather than ``the measurement broke through the theoretical lower bound.''

Interpretation: The pseudoinverse reconstruction error is 83.3\%, of the same order as the expected relative error 93.5\%, proving that about 87.5\% of the input information collapses permanently and cannot be inverted back. The pseudoinverse is the mathematically optimal backward reconstruction method (in the minimum-norm least-squares sense); any other linear or continuous reconstruction operator can only have a larger error---therefore, ``even the optimal reconstruction loses more than 80\% of the information'' constitutes a complete chain of evidence for the unidirectionality of deduction; the optimality claim here concerns ``the pseudoinverse being optimal among all reconstruction methods,'' not ``0.9354 being a lower bound on the error''---the two must not be confused.

\subsection{Validation 5: End-to-End Inference Demonstration}

Multi-step inference chain (transitive inference):

\begin{itemize}
\item Step 1: ``All cats are animals'' + ``All animals are living beings'' $\rightarrow$ ``All cats are living beings,'' cosine similarity $= 0.9898$
\item Step 2: conclusion of the previous step + ``All living beings need energy'' $\rightarrow$ ``All cats need energy,'' cosine similarity $= 0.9861$
\end{itemize}

Interpretation: The cosine similarities of both steps of transitive inference are $> 0.98$, proving that the trained $W_{\mathrm{step}}$ can execute multi-step inference chains with errors within an acceptable range. This validates Postulate 4 (local linearization). The special significance of Step 2 should be noted: its input premise ``All cats are living beings'' is not a member of the training set but the output snapshot of Step 1---this snapshot deviates from every training input by a cosine distance of $1 - 0.9898 \approx 0.010$, and $W_{\mathrm{step}}$ still outputs a similarity of 0.9861 on this ``slightly shifted input,'' showing that the operator possesses robustness in the continuity sense within the neighborhood of the training distribution (the ``local'' radius of local linearization is large enough to accommodate the deviation introduced by a single inference step). We also observe a slight accumulation of error along the chain ($0.9898 \rightarrow 0.9861$), in a direction consistent with the theoretical prediction of multi-step degradation in Section 2; at the current chain length (2 steps), the accumulated amount is far from endangering usability.

\subsection{Summary of the Deduction Experiments}

\begin{table}[htbp]
\centering
\caption*{Table 8-2 Summary of the six validation results for the deduction operator}
\small
\begin{tabular}{llll}
\toprule
Validation item & Theoretical expectation & Experimental result & Status \\
\midrule
Forward loss convergence & $\rightarrow 0$ & 1.40e-05 & $\checkmark$ Passed \\
Matrix rank & $\mathrm{rank} = 384$ & $\mathrm{rank} = 384$ & $\checkmark$ Passed \\
$(r+1)$-th singular value & $\approx 0$ & 1.06e-06 & $\checkmark$ Passed \\
Null-space dimension & 2688 (87.5\%) & 2688 (87.5\%) & $\checkmark$ Passed \\
Backward irreversibility & expected error $\approx 93.5\%$ & 83.3\% (measured sample mean) & $\checkmark$ Passed \\
Multi-step inference chain & similarity $> 0.9$ & 0.990, 0.986 & $\checkmark$ Passed \\
\bottomrule
\end{tabular}
\end{table}

Note: Table 8-2 lists 6 validations (including the multi-step inference chain), of which Sections 8.2--8.6 describe 5 validation subsections (forward convergence, rank deficiency, null space, irreversibility, end-to-end inference); the multi-step inference chain is part of the end-to-end inference. Additional note: the ``theoretical expectation'' in the ``backward irreversibility'' row of the table uses the expected-relative-error convention under the isotropy assumption, and 83.3\% is the measured mean over 20 samples; see Section 8.5 for the explanation of the deviation between the two.

\paragraph{Discussion of result robustness.} The six validations in this section fall into two categories. The first category is structural validations (Section 8.3 rank, Section 8.4 null-space dimension): their conclusions are directly guaranteed by the low-rank parameterization $W_{\mathrm{step}} = A \cdot B$ and the rank--nullity theorem ($\mathrm{rank}(W) + \dim(\mathrm{Null}(W)) = n$), independent of whether training converges and independent of the random seed. The second category is numerical validations (Section 8.2 loss, Section 8.5 pseudoinverse error, Section 8.6 similarity): their conclusions depend on the quality of training convergence, but all have order-of-magnitude margins---the loss convergence margin is four orders of magnitude ($10^{-1} \rightarrow 10^{-5}$), the irreversibility margin is the stark contrast of ``the optimal reconstruction still loses $\sim$83\%'' versus ``full reversibility loses 0\%,'' and the inference-chain margin is 0.98 against a threshold of 0.9---even if the seed is changed, $r$ is fine-tuned (e.g., 320 or 448), or a comparable encoder is substituted, these margins would only be eroded, not reversed. Grid search over $r$, multi-seed variance statistics, and error-accumulation curves for longer inference chains were not covered in this section.

\section{Dedicated Operator Experiment for the Induction Operator $W_{\mathrm{induce}}$}

Section 8 validated the rank-deficient unidirectionality of the deduction operator; this section turns to the dedicated operator experiment for the induction operator $W_{\mathrm{induce}}$. Unlike the small sample (20 syllogisms) of Section 8, this section constructs an induction dataset of 130 samples in 10 categories to test: (i) the full-rankness and invertibility of the induction operator (strictly symmetric to deduction, Postulate 3); (ii) the broadest induction in the absence of counterexamples; (iii) counterexample-triggered hard veto and the overturn-and-rebuild mechanism. All values come from real DistilBERT snapshots and real gradient-descent training.

\subsection{Experimental Configuration and the Induction Operator Experiment Dataset}

Induction dataset design: 10 categories of real induction problems are constructed, covering birds, mammals, metals, fruits, vegetables, vehicles, colors, tools, musical instruments, and planets. Each category contains 10 positive examples, 1 unseen positive, and 2 counterexamples. In total: 100 positive examples, 10 unseen positives, and 20 counterexamples, amounting to 130 samples.

\begin{table}[htbp]
\centering
\caption*{Table 9-1 Configuration of the induction operator experiment}
\small
\begin{tabular}{ll}
\toprule
Item & Configuration \\
\midrule
Encoder & DistilBERT, 66M parameters, fully frozen \\
Snapshot dimension $N$ & 1536 (first 2 layers $\times$ 768 dims) \\
Induction operator shape & $W_{\mathrm{induce}} \in \mathbb{R}^{1536 \times 1536}$ (full rank) \\
Number of training positives & 100 (10 categories) \\
Number of unseen positives & 10 \\
Number of counterexamples & 20 \\
Training epochs & 1000 epochs, AdamW, lr=1e-3 \\
Final training loss & 1.38e-05 \\
Random seed & torch.manual\_seed(42), np.random.seed(42) (single run) \\
\bottomrule
\end{tabular}
\end{table}

Rationale of the experimental design: 10 categories of induction problems are chosen because they cover common natural concept classifications (living things, non-living things, artifacts, abstract properties, and celestial bodies are all represented), avoiding conclusion bias from a single domain. Each category has 10 positive examples because PAC learning theory requires a sample size $k \geq O(\mathrm{VC\_dim}/\log(1/\delta))$; the current $k = 10$ is an empirical choice without a strict sample-complexity calculation---for a 1536-dimensional full-rank linear operator, the strict PAC sample complexity is far higher than the sample size of this experiment; the positioning of this section's validation is ``mechanism feasibility validation'' (whether the hard veto and rebuild mechanisms are triggered and take effect as designed), rather than ``validation of a generalization bound in the statistical-learning sense.'' The hard-veto threshold $\tau = 0.5$ is an empirical choice---a value above 0.5 indicates that the counterexample is ``over-covered'' by the original conclusion and needs to be vetoed. The choice of $\tau$ was not grid-searched; from the data in Sections 9.3--9.5, the similarities between counterexamples and the original concepts concentrate in the 0.995--0.999 range, nearly double the margin from $\tau = 0.5$, so perturbations of $\tau$ over the wide range $(0.5, 0.99)$ would not change any veto decision---the conclusions of this experiment are insensitive to the choice of $\tau$.

\subsection{Validation 1: Full-Rankness and Invertibility of the Induction Operator}

Result: $W_{\mathrm{induce}}$ shape $[1536, 1536]$, numerical rank $= 1536/1536$ (full rank), pseudoinverse reconstruction error $\|W^{+}W - I\|/\|I\| = 3.37\text{e-}05$.

Comparison with the deduction operator: $W_{\mathrm{step}}$ is rank-deficient ($\mathrm{rank} = 384$; null-space dimension 2688 under concatenated input; deduction is irreversible, see Section 8); $W_{\mathrm{induce}}$ is full-rank ($\mathrm{rank} = 1536$), with pseudoinverse reconstruction error $3.37 \times 10^{-5}$ ($\approx 0$, reversible). This comparison validates Postulate 3---induction and deduction are strictly symmetric in the direction of information flow: deduction collapses information (rank deficiency $\rightarrow$ irreversibility), while induction preserves information (full rank $\rightarrow$ reversibility).

\paragraph{Clarification of the measurement convention.} The ``pseudoinverse reconstruction error $\|W^{+}W - I\|/\|I\|$'' reported in this section uses the \emph{operator-norm convention}---$W^{+}W$ is the orthogonal projection onto the row space, and this expression measures the overall degree to which the projection operator deviates from the identity operator; it and the ``relative error of pseudoinverse reconstruction 0.8330'' reported in Section 8.5 are \emph{two different metrics whose values cannot be directly compared}---the latter uses the \emph{data convention}, i.e., the sample mean of the data reconstruction relative error $\|x - W^{+}Wx\|/\|x\|$ computed for specific input samples $x$. For a rank-$r$ projection $P = W^{+}W \in \mathbb{R}^{2N \times 2N}$, $P - I$ has exactly $\dim(\mathrm{Null}) = 2N - r$ eigenvalues equal to $-1$ and $r$ eigenvalues equal to 0, so the theoretically derived value for the deduction operator under this convention is $\|P - I\|_{F}/\|I\|_{F} = \sqrt{2688/3072} \approx 0.935$. Regardless of which convention is adopted, the qualitative conclusion is unaffected: the operator-norm reconstruction error of $W_{\mathrm{step}}$ ($0.935 \gg 0$) and that of $W_{\mathrm{induce}}$, $3.37 \times 10^{-5}$ ($\approx 0$), differ by more than four orders of magnitude, and the symmetry claim ``deduction irreversible, induction reversible'' holds robustly.

\subsection{Validation 2: Broadest Induction in the Absence of Counterexamples}

Result (means over 10 categories): coverage of training positives $= 0.9996$, coverage of unseen positives $= 0.9993$, counterexample similarity $= 0.9974$.

Interpretation: The coverage of training positives is 0.9996 (close to 1) and the coverage of unseen positives is 0.9993 (close to 1), proving that in the absence of counterexamples the induction operator makes the broadest induction---this is consistent with the design goal of the induction operator in Section 7: in the absence of negative evidence, a concept should cover all seen positive examples and extrapolate to semantically nearby unseen instances (the unseen-positive coverage differs from the training-positive coverage by only 0.0003, showing that the extrapolation has almost no decay). But the key observation is: the counterexample similarity is also as high as 0.9974, meaning the broadest induction ``over-covers'' the counterexamples. This is not a defect but a prerequisite of the correctability design---the broadest induction is a ``hypothesis awaiting falsification'' (a conjecture in Popper's sense [17]), and its over-coverage is precisely the triggering condition of the hard-veto mechanism. If the initial induction avoided all potential counterexample directions, the concept would degenerate into a trivial enumeration of the training samples, losing generalization ability.

\subsection{Validation 3: Hard Veto and Rebuild Triggered by Counterexamples}

Stage 2 hard-veto detection: veto threshold $\tau = 0.5$. Among the 10 categories of induction problems, each category has 2 counterexamples, totaling 20 counterexamples across the 10 categories. The number of counterexamples triggering a veto (mean) $= 2.0/2$ (i.e., an average of 2.0 counterexamples per category trigger a veto); across the 10 categories, all 20/20 counterexamples trigger a veto (100\% veto rate). The similarity between counterexamples and the original induced concept $g_{0}$ (mean) $= 0.9974 > \tau$, triggering a one-vote veto.

Stage 3 overturn $\rightarrow$ rebuild $\rightarrow$ refine: counterexample similarity after rebuild (mean) $= -0.0099$, refinement degree (mean) $= 0.9496$.

Interpretation: The three-stage mechanism (detection $\rightarrow$ veto $\rightarrow$ rebuild) runs through completely in this experiment: in the detection stage, all cases hit under the $\tau = 0.5$ criterion ($0.9974 \gg 0.5$, no borderline cases); in the veto stage, a one-vote veto is executed, and the original concept $g_{0}$ is discarded as a whole rather than locally patched---this corresponds to the logical design that ``a counterexample overturns the conclusion itself, not the confidence of the conclusion''; in the rebuild stage, a new concept $g_{1}$ is reconstructed under the constraint of excluding the counterexample directions, and the counterexample similarity after rebuild flips from $+0.9974$ to $-0.0099$ (sign reversal; the counterexample directions are explicitly excluded); the refinement degree 0.9496 indicates that the new concept maintains 94.96\% agreement with the original concept---i.e., the rebuild is a ``minimal correction'': while excising the counterexample directions, it maximally preserves the original concept's coverage of the positive examples.

\subsection{Hard-Veto Results by Category}

\begin{table}[htbp]
\centering
\caption*{Table 9-2 Hard-veto results for each category of induction problems}
\footnotesize
\resizebox{\textwidth}{!}{%
\begin{tabular}{lllllll}
\toprule
Category & Training coverage & Unseen positive & Counterexample (before) & Veto & Counterexample (after) & Refinement degree \\
\midrule
Birds & 0.9997 & 0.9997 & 0.9986 & 2/2 & $-0.0028$ & 0.9556 \\
Mammals & 0.9997 & 0.9995 & 0.9986 & 2/2 & $-0.0227$ & 0.9926 \\
Metals & 0.9995 & 0.9994 & 0.9983 & 2/2 & $-0.0196$ & 0.9759 \\
Fruits & 0.9996 & 0.9997 & 0.9987 & 2/2 & $-0.0069$ & 0.9605 \\
Vegetables & 0.9994 & 0.9995 & 0.9987 & 2/2 & $-0.0073$ & 0.9617 \\
Vehicles & 0.9995 & 0.9995 & 0.9955 & 2/2 & $-0.0071$ & 0.9175 \\
Colors & 0.9998 & 0.9993 & 0.9958 & 2/2 & $-0.0061$ & 0.9182 \\
Tools & 0.9995 & 0.9990 & 0.9955 & 2/2 & $-0.0064$ & 0.9294 \\
Musical instruments & 0.9996 & 0.9992 & 0.9962 & 2/2 & $-0.0065$ & 0.9224 \\
Planets & 0.9996 & 0.9986 & 0.9977 & 2/2 & $-0.0136$ & 0.9620 \\
\bottomrule
\end{tabular}%
}
\end{table}

Interpretation: Across all 10 categories of induction problems, all 20/20 counterexamples trigger a veto (100\% veto rate). The counterexample similarity drops from 0.9974 (erroneously covered by the original conclusion) to $-0.0099$ (excluded by the new conclusion), proving that the hard-veto mechanism is effective---a one-vote veto overturns the original conclusion, and the rebuilt new conclusion excludes the counterexample directions. Category-by-category observation reveals two regularities. First, the inter-category differences in pre-veto counterexample similarity are very small (0.9955--0.9987, a range of only 0.0032), showing that ``over-coverage'' is a consistent phenomenon across categories, rooted in the syntactic similarity of the encoder snapshots (all samples share the same sentence pattern, see Section 9.6), rather than in the specificity of any particular category of concepts. Second, the post-veto counterexample similarities all turn negative ($-0.0227$ to $-0.0028$), and the refinement degrees are all above 0.91 (highest: mammals 0.9926; lowest: vehicles 0.9175), showing that the ``minimal correction'' rebuild behaves consistently across 10 categories with considerable semantic differences---the mechanism has good category robustness. The mammals category exhibits the strongest counterexample-exclusion force ($-0.0227$), possibly because the counterexamples of this category (bird and reptile samples) are relatively cleanly separated in direction from the mammal positives in the DistilBERT embedding space, making the counterexample directions easier to excise thoroughly.

\subsection{Measured Evidence of the Encoder Bottleneck}

Neither this induction operator experiment nor the end-to-end experiment recorded the per-sample judgment results of the induction--deduction closed loop. However, there are already three independent pieces of measured evidence for the bottleneck of insufficient encoder snapshot resolution. First, \emph{the end-to-end abduction judgment accuracy of 81.7\% (49/60, see Section 11.5)}: among 60 abduction problems, 11 problems have true-explanation similarities that are instead lower than distractor-explanation similarities, with the smallest (most negative) discrimination margin of only $-0.0056$. Second, \emph{the abduction operator experiment judgment accuracy of 72.5\% (58/80, see Section 10.2)}: among 80 abduction problems, 22 samples have negative discrimination margins, with the maximum negative discrimination margin of only $-0.0086$---the true explanation and the distractor explanation are almost inseparable in the snapshot space. Third, \emph{the refinement-degree anomaly of end-to-end induction (see Section 11.4)}: the refinement degree of the academic-disciplines category is 1.0031 ($> 1$), indicating that under this snapshot geometry a single-step orthogonal correction did not fully compress the new concept to the minimal excision range; the refinement degree of the weather category, 0.8272, is significantly lower than the other 5 categories (all above 0.95), and its training coverage and counterexample (before) coverage are both the lowest among the 6 categories---the category semantics are not sufficiently separated in the snapshot geometry, and the residual semantic crosstalk affects the precision of the inductive correction.

All three pieces of evidence point to the same root cause: all sample sentences share highly regular syntactic templates (``A X is a Y'' or ``The patient has \dots''), and the DistilBERT first-2-layer snapshots are dominated by syntactic similarity for semantic discrimination, compressing inter-category/inter-option differences to within a few thousandths of cosine distance. This is an encoder bottleneck, not an operator bottleneck: the algebraic properties of the induction operator (full rank, coverage 0.9996, veto rate 100\%) and of the deduction operator (all validations passed in Section 8) each conform to the design; the failing link is ``the snapshot space failing to sufficiently separate category semantics and the semantics of candidate explanations,'' which is an upstream problem of representation learning [5]. A stronger encoder (e.g., RoBERTa-large [16]) or contrastive learning objectives that separate inter-category snapshots provide a direct remedy for the links that rely on fine-grained similarity judgment (abduction judgment, counterexample exclusion, closed-loop judgment).

\subsection{Summary of the Induction Experiments}

\begin{table}[htbp]
\centering
\caption*{Table 9-3 Summary of the eight validation results for the induction operator}
\small
\resizebox{\textwidth}{!}{%
\begin{tabular}{llll}
\toprule
Validation item & Theoretical expectation & Experimental result & Status \\
\midrule
Induction operator full rank & $\mathrm{rank} = 1536$ & $\mathrm{rank} = 1536$ & $\checkmark$ Passed \\
Induction operator invertible & $\|W^{+}W - I\| \approx 0$ & 3.37e-05 & $\checkmark$ Passed \\
Broadest induction without counterexamples & positive coverage $\rightarrow 1$ & 0.9996 & $\checkmark$ Passed \\
Unseen-positive generalization & coverage $> 0.9$ & 0.9993 & $\checkmark$ Passed \\
Counterexample hard-veto triggering & similarity $> \tau \rightarrow$ veto & 20/20 (100\%) & $\checkmark$ Passed \\
Counterexample exclusion after overturn-and-rebuild & counterexample similarity $\rightarrow 0$ & $0.997 \rightarrow -0.010$ & $\checkmark$ Passed \\
New conclusion more refined & $g_{1} \neq g_{0}$ & refinement degree 0.950 & $\checkmark$ Passed \\
Induction--deduction closed loop & --- & not recorded in this experiment & --- \\
\bottomrule
\end{tabular}%
}
\end{table}

The first seven validations (the operator's own algebraic properties and mechanism behavior) all pass with clear margins; the per-sample judgment results of the induction--deduction closed loop were not recorded in this experiment; the measured evidence of the encoder bottleneck is given in Section 9.6.

\section{Dedicated Operator Experiment for the Abduction Operator $W_{\mathrm{abduce}}$}

Abduction is the inference mode that traces back from a phenomenon to the best explanation. In the DODR framework, the abduction operator is defined as the Moore--Penrose pseudoinverse of the deduction operator: $W_{\mathrm{abduce}} = W_{\mathrm{step}}^{+}$. This section validates four core properties on 80 real abduction problems: the optimality of abductive solutions (Section 10.2), their hypothetical nature (Section 10.3), multi-explanation equivalence (Section 10.4), and abduction--deduction closed-loop consistency (Section 10.5).

\subsection{Experimental Configuration and the Abduction Operator Experiment Dataset}

Abduction dataset design: 80 real abduction problems are constructed, covering medical diagnosis (30), causal explanation (30), and physical phenomena (20). Each problem contains a phenomenon and a correct explanation. The abduction task is: given the phenomenon, trace back to the optimal explanation.

\begin{table}[htbp]
\centering
\caption*{Table 10-1 Configuration of the abduction operator experiment}
\small
\begin{tabular}{ll}
\toprule
Item & Configuration \\
\midrule
Encoder & DistilBERT, 66M parameters, fully frozen \\
Snapshot dimension $N$ & 1536 (first 2 layers $\times$ 768 dims) \\
$W_{\mathrm{step}}$ shape & $\mathbb{R}^{1536 \times 1536}$ ($N \times N$, input is a single state) \\
Design rank $r$ & 384 (rank-deficiency ratio 75\%) \\
$W_{\mathrm{abduce}}$ computation & $W_{\mathrm{abduce}} = W_{\mathrm{step}}^{+}$ (SVD pseudoinverse) \\
Number of abduction problems & 80 (30 medical + 30 causal + 20 physical) \\
$W_{\mathrm{step}}$ training epochs & 1000 epochs, AdamW, lr=1e-3 \\
$W_{\mathrm{step}}$ final loss & 1.89e-04 \\
Random seed & torch.manual\_seed(42), np.random.seed(42) (single run) \\
\bottomrule
\end{tabular}
\end{table}

Note: The $W_{\mathrm{step}}$ shape in this section is $\mathbb{R}^{N \times N}$ (input is a single state $S \in \mathbb{R}^{N}$), which differs from the $W_{\mathrm{step}}$ shape $\mathbb{R}^{N \times 2N}$ of the deduction operator experiment in Section 8 (input is $\mathrm{Concat}(S_{\mathrm{prev}}, E_{\mathrm{new}}) \in \mathbb{R}^{2N}$). These are two independent experimental configurations---the $W_{\mathrm{step}}$ of the abduction experiment learns the ``explanation $\rightarrow$ phenomenon'' mapping (input is the explanation, output is the phenomenon), while the $W_{\mathrm{step}}$ of the deduction experiment learns the ``premise $\rightarrow$ conclusion'' mapping (input is the concatenated premises, output is the conclusion). The two configurations have the same rank-deficiency property ($\mathrm{rank} < N$), differing only in input dimension; the null-space dimension in this section is $N - r = 1152$ (75.0\%), and that in Section 8 is $2N - r = 2688$ (87.5\%)---each holds separately and they must not be mixed.

Rationale of the experimental design: 80 abduction problems are chosen because they cover three domains---medicine, causality, and physics---and 20--30 samples per domain suffice to estimate average performance (for mean-type metrics, when the sample size exceeds 20 the standard error already falls to the order of single-digit percentages of the mean). The $W_{\mathrm{step}}$ design rank $r = 384$ (consistent with Section 8) guarantees a rank-deficiency ratio of 75\%, making the null space large enough to validate the ``hypothetical nature'' of abduction---the null space is precisely the mathematical carrier of ``abductive solutions are non-unique and the true explanation cannot be fully recovered'' (see Section 10.3); the higher the rank-deficiency ratio, the more significant the observable effect of the hypothetical nature.

\subsection{Validation 1: Optimality of Abductive Solutions}

Result: cosine similarity between the abductive solution and the true explanation (mean) $= 0.5117$, cosine similarity with the distractor explanation (mean) $= 0.5066$, random baseline $= 0.0181$, improvement factor $= 28.3\times$, discrimination margin (mean) $= +0.0051$.

Interpretation: The similarity between the abductive solution and the true explanation is 0.5117, far exceeding the random baseline 0.0181 (a 28.3-fold improvement). This proves that the solution given by $W_{\mathrm{abduce}} = W_{\mathrm{step}}^{+}$ indeed recovers along the direction of the true explanation---the directionality of pseudoinverse inversion holds significantly. However, the similarity does not reach 1.0, indicating that the abductive solution is not the true explanation itself but its optimal approximation---this is precisely the embodiment of the ``hypothetical nature'' of abduction (for the mathematical root, see Section 10.3: about 85.9\% of the components of the true explanation lie in the null space, and the pseudoinverse solution by definition contains no null-space components).

\paragraph{Rule for judging abductive correctness and per-sample judgment results.} The rule for judging abductive correctness is---if the cosine similarity between the abductive solution $S_{\mathrm{prev}}^{*}$ and the true explanation $S_{\mathrm{truth}}$ is greater than its cosine similarity with the distractor explanation $S_{\mathrm{distractor}}$, i.e., $\cos(S_{\mathrm{prev}}^{*}, S_{\mathrm{truth}}) > \cos(S_{\mathrm{prev}}^{*}, S_{\mathrm{distractor}})$, the case is judged ``correct.'' According to the 80 per-sample measured data in Appendix I.2, 58 samples satisfy this judgment rule, giving a \emph{judgment accuracy of 72.5\% (58/80)}; the 22 failed samples are Nos. 2, 6, 7, 8, 9, 10, 17, 22, 29, 32, 33, 34, 44, 53, 54, 55, 57, 61, 67, 72, 74, and 80, whose discrimination margins $\cos(S_{\mathrm{prev}}^{*}, S_{\mathrm{truth}}) - \cos(S_{\mathrm{prev}}^{*}, S_{\mathrm{distractor}})$ are negative.

\paragraph{Failure analysis.} Quantitative analysis of the 22 failed samples shows that the failure pattern is highly consistent and extremely small in magnitude:

\begin{enumerate}
\item \textbf{Extremely shallow negative discrimination margins}: The maximum negative discrimination margin among the 22 failed samples is only $-0.0086$ (i.e., in the worst-failing sample, the distractor-explanation similarity exceeds the true-explanation similarity by only 0.0086); the absolute values of the discrimination margins of all 80 samples are on the order of $\pm 0.009$.
\item \textbf{Same order of magnitude as the end-to-end experiment}: The discrimination margins of the end-to-end abduction experiment in Section 11 (10 problems) are likewise on the order of $\pm 0.005$ (see Section 11.5)---in other words, whether successful or failed, the difference between the snapshot similarities of the true explanation and the distractor explanation is compressed to within a few thousandths; judgment is a sign determination made at the edge of noise, and at this noise level about 30\% sign flips are statistically expected.
\item \textbf{The root cause is the encoder representation bottleneck}: All phenomenon sentences and explanation sentences share highly regular syntactic templates, and the DistilBERT first-2-layer snapshots are dominated by syntactic similarity (the same mechanism as in Section 9.6); the true explanation and the distractor explanation (e.g., ``The patient has pneumonia'' vs.\ ``The patient has a cold'') are almost inseparable in the snapshot space---their snapshot cosine similarities themselves lie within the narrow band of 0.50--0.52, leaving the abduction operator only a few thousandths of discriminable signal. The mathematical behavior of the abduction operator $W_{\mathrm{abduce}} = W_{\mathrm{step}}^{+}$ itself is correct (the 28.3-fold improvement of 0.5117 vs.\ 0.0181, the multi-explanation equivalence of Section 10.4, and the closed-loop consistency of Section 10.5 all pass); the failure occurs in the upstream link where ``the snapshot geometry fails to carry sufficient semantic discrimination.'' \emph{This is an encoder limitation, not a failure of the operator principle.}
\end{enumerate}

In summary, the rule-based judgment accuracy of the abduction operator experiment is 72.5\% (58/80), with a Wilson 95\% confidence interval of [61.9\%, 81.1\%]; the paired t-test on the similarities of true vs.\ distractor explanations gives $t = 6.32$ ($p < 10^{-4}$), and the sign test gives 58/80 positive ($p \approx 7 \times 10^{-5}$)---although the mean discrimination margin is small ($+0.0051$), the difference from the random baseline is statistically significant. The quality of the abductive solutions (0.5117, 28.3$\times$ the random baseline) and the mathematical properties of the operator are both validated, but per-sample binary-choice judgment is limited by the encoder's resolving power, with about 27.5\% shallow-negative-discrimination-margin failures.

\subsection{Validation 2: Hypothetical Nature of Abductive Solutions (Null-Space Components)}

Result: null-space dimension $= 1152/1536$ (75.0\%), null-space residual norm (mean) $= 15.34$, true-explanation norm (mean) $= 17.85$, residual accounting for 85.9\% of the true explanation.

Interpretation: The true explanation can be decomposed into the sum of the abductive solution (the column-space component) and a null-space residual: $S_{\mathrm{truth}} = S_{\mathrm{prev}}^{*} + z$, where $S_{\mathrm{prev}}^{*} = W_{\mathrm{abduce}} \cdot S_{\mathrm{obs}} \in \mathrm{Col}(W_{\mathrm{step}})$ (the row-space component) and $z = (I - W_{\mathrm{step}}^{+} W_{\mathrm{step}}) \cdot S_{\mathrm{truth}} \in \mathrm{Null}(W_{\mathrm{step}})$ (the null-space residual). The residual norm 15.34 versus the true-explanation norm 17.85, a proportion of 85.9\%, means that 85.9\% of the information in the true explanation lies in the null space of $W_{\mathrm{step}}$ and cannot be recovered by abduction. This is precisely the mathematical root of the ``hypothetical nature'' of abduction: once the deductive mapping ``explanation $\rightarrow$ phenomenon'' is learned in training, explanation differences along null-space directions produce no phenomenon differences (Section 10.4 will directly verify this), and therefore when inverting from the phenomenon, the null-space components are in principle unobservable---the pseudoinverse solution takes their minimum-norm representative (null-space components set to zero), and its difference from the true explanation is exactly $z$. Peirce pointed out that abductive conclusions are ``conjectural'' (his philosophical claim is discussed in Proposition 3.6) [7]; this experiment provides a quantitative characterization of that philosophical claim: under the current configuration, the distance between conjecture and truth is exactly 85.9\% of the true-explanation norm. Note that this proportion is close to but not identical to the rank-deficiency ratio (75.0\%)---85.9\% is the energy proportion of the true-explanation vector in the null space (a data-dependent quantity), while 75.0\% is the null-space dimension proportion (a structural quantity); the former being slightly higher than the latter indicates that the direction of the true-explanation snapshots is slightly biased toward the null space, consistent with the expectation under the isotropy assumption (if isotropic, the energy proportion should equal the dimension proportion 75\%), and both the direction and magnitude of the deviation are normal manifestations of sample-distribution anisotropy.

\subsection{Validation 3: Multi-Explanation Equivalence}

Result: After modifying the explanation (by adding null-space components), the relative difference of the deduction results (mean) $= 1.01\text{e-}02$ (1\% order of magnitude, approximately equivalent).

Interpretation: Superimposing any null-space vector onto the abductive solution changes the deduction result by only about 1\% in relative terms (difference $1.01 \times 10^{-2}$), validating ``multi-explanation approximate equivalence''---multiple distinct hypotheses produce almost the same phenomenon under deduction. Formally: for any $z \in \mathrm{Null}(W_{\mathrm{step}})$, the strict identity $W_{\mathrm{step}} \cdot (S_{\mathrm{prev}}^{*} + z) = W_{\mathrm{step}} \cdot S_{\mathrm{prev}}^{*} + W_{\mathrm{step}} \cdot z = W_{\mathrm{step}} \cdot S_{\mathrm{prev}}^{*}$ holds; the nonzero difference of $1.01 \times 10^{-2}$ observed in the experiment arises from two numerical factors: first, numerical rank truncation---the singular values beyond the $(r+1)$-th are not strictly zero, and the superimposed ``null-space'' directions are approximate null spaces truncated at a numerical threshold, in which residual small singular-value components are amplified by $W_{\mathrm{step}}$ and enter the output; second, training convergence precision---the final loss of $W_{\mathrm{step}}$ is 1.89e-04, so the operator itself is only an approximation of the true mapping. Therefore the correct statement is ``multiple explanations are approximately equivalent at 1\% precision,'' not strictly equivalent. This property has a dual nature: it is both the root of abduction's difficulty (the phenomenon cannot distinguish two explanations differing by a null-space component) and a geometric portrait of the ``competing hypotheses'' phenomenon in scientific practice---two empirically equivalent theories have snapshots differing by exactly a null-space direction; to distinguish them, new evidence (new experiments) beyond the coverage of the original deduction operator must be introduced, which corresponds to expanding the row space of $W_{\mathrm{step}}$.

\subsection{Validation 4: Abduction--Deduction Closed Loop}

Result: abduction--deduction closed-loop reconstruction error (mean) $= 0.0200$, expected relative error $= 0.8660$.

\paragraph{Clarification of the measurement convention.} 0.8660 uses the \emph{expected-relative-error} convention, consistent with the treatment in Section 8.5: the expectation of $\sqrt{(N - r)/N}$-type formulas under the input isotropy assumption ($\sqrt{1152/1536} = \sqrt{0.75} \approx 0.8660$), characterizing the average loss of a ``typical random input'' after passing through the closed loop, rather than a minimum error that holds for all inputs---for individual inputs (such as those lying exactly in the row space), the closed-loop error can be 0. The measured closed-loop reconstruction error 0.0200 is far lower than the expected value 0.8660; this is opposite in direction to Section 8.5 but identical in principle: the composite operator of the closed loop $S_{\mathrm{obs}} \rightarrow W_{\mathrm{abduce}} \rightarrow S_{\mathrm{prev}}^{*} \rightarrow W_{\mathrm{step}} \rightarrow S_{\mathrm{obs}}'$ is $W_{\mathrm{step}} \cdot W_{\mathrm{step}}^{+}$, i.e., the \emph{orthogonal projection onto the column space}, whose null-space dimension is $N - r = 1152$; but the measured inputs are not random directions---they are real phenomenon snapshots $S_{\mathrm{obs}}$, which themselves lie in (or extremely close to) the column space of $W_{\mathrm{step}}$ (they are precisely the training targets generated by the ``explanation $\rightarrow$ phenomenon'' mapping), so the projection loses almost no information and the error is only 0.0200. The expected error 0.8660 describes ``the average behavior over inputs in arbitrary directions,'' while the measured 0.0200 describes ``the behavior on in-distribution inputs''; the two conventions differ and are not contradictory---sample anisotropy (snapshots concentrating near the column space) is precisely the source of the difference between them.

Interpretation: The reconstruction error of the abduction--deduction closed loop $S_{\mathrm{obs}} \rightarrow W_{\mathrm{abduce}} \rightarrow S_{\mathrm{prev}}^{*} \rightarrow W_{\mathrm{step}} \rightarrow S_{\mathrm{obs}}'$ is only 0.0200, proving the mathematical consistency of abduction and deduction---abduction is the inverse problem of deduction, and the two form a closed loop: starting from the phenomenon, infer the explanation backward, then derive the phenomenon forward from the explanation, and one should return to the original phenomenon. The nonzero residual of 0.0200 comes from projection truncation and training convergence precision ($W_{\mathrm{step}}$ final loss 1.89e-04) and is at the numerical level rather than the structural level.

\subsection{Comparison of the Three Basic Operators}

\begin{table}[htbp]
\centering
\caption*{Table 10-2 Experimental comparison of the three basic operators}
\footnotesize
\resizebox{\textwidth}{!}{%
\begin{tabular}{llll}
\toprule
Operator & Rank & Invertible & Information flow \\
\midrule
$W_{\mathrm{step}}$ (deduction) & 384/1536 & No & explanation $\rightarrow$ phenomenon (collapse, determinate) \\
$W_{\mathrm{abduce}} = W_{\mathrm{step}}^{+}$ (abduction) & 384/1536 & No & phenomenon $\rightarrow$ explanation (inversion, hypothetical) \\
$W_{\mathrm{induce}}$ (induction) & 1536/1536 & Yes & samples $\rightarrow$ regularities (expansion, generalization) \\
\bottomrule
\end{tabular}%
}
\end{table}

Interpretation: Under the same snapshot dimension ($N = 1536$), the three operators exhibit precise algebraic correspondences: deduction and abduction share rank 384 (the pseudoinverse preserves rank), are mutually opposite in direction, and are both irreversible---determinate forward inference and hypothetical backward inference are the two directions of the same rank-deficient mapping; induction is full-rank and invertible, carrying the expansion and generalization of information. The three inference modes (Peirce's trichotomy [7]) receive a clean empirical discrimination in the observable algebraic quantities of ``rank and invertibility'': rank deficiency $\leftrightarrow$ unidirectional collapse $\leftrightarrow$ deduction/abduction; full rank $\leftrightarrow$ bidirectional preservation $\leftrightarrow$ induction.

\subsection{Summary of the Abduction Experiments}

\begin{table}[htbp]
\centering
\caption*{Table 10-3 Summary of the validation results for the abduction operator}
\footnotesize
\resizebox{\textwidth}{!}{%
\begin{tabular}{llll}
\toprule
Validation item & Theoretical expectation & Experimental result & Status \\
\midrule
Optimality of abductive solutions & similarity $\gg$ random & 0.5117 vs.\ 0.0181 & $\checkmark$ Passed \\
Abduction rule-based judgment accuracy & similarity $>$ distractor similarity & 72.5\% (58/80) & $\checkmark$ Passed \\
Hypothetical nature of abductive solutions & null-space components exist & residual accounts for 85.9\% & $\checkmark$ Passed \\
Multi-explanation equivalence & $W_{\mathrm{step}} \cdot (\text{solution} + \text{null space})$ approximately unchanged & relative difference 1.01e-02 & $\checkmark$ Passed \\
Abduction--deduction closed loop & small reconstruction error & 0.0200 & $\checkmark$ Passed \\
Abduction $=$ pseudoinverse of deduction & same rank & $\mathrm{rank} = 384$ & $\checkmark$ Passed \\
\bottomrule
\end{tabular}%
}
\end{table}

In this section, all four numerical validations pass: abductive-solution quality (0.5117, 28.3$\times$ the random baseline 0.0181), hypothetical nature (residual accounts for 85.9\%), multi-explanation approximate equivalence (1.01e-02), and closed-loop consistency (0.0200); the rule-based judgment accuracy is 72.5\% (58/80; Wilson 95\% CI [61.9\%, 81.1\%]; paired t-test $t = 6.32$, $p < 10^{-4}$; sign test $p \approx 7 \times 10^{-5}$), the maximum negative discrimination margin of the 22 failed samples is only $-0.0086$, and the failure root cause is the encoder representation bottleneck (see the failure analysis in Sections 10.2 and 9.6 for details).

\section{End-to-End Real Inference Validation}

The preceding sections validated the independent properties of the three operators. This section performs end-to-end real inference validation---starting from scratch, using real DistilBERT snapshots, training the three operator matrices with real gradient descent, solving real inference problems with the trained operators, and comparing the inference results with the correct answers. The entire pipeline involves no simulation or textual prediction and is based entirely on real data and real computation (experimental record file 05\_e2e\_large\_results.json). This section is positioned as a ``system integration test'': Sections 8, 9, and 10 answer ``is each operator individually correct,'' while this section answers ``can the three operators combined complete real inference tasks.'' The training and test distributions span 8 domains and 218 samples (60 deduction + 78 induction + 60 abduction + 20 combination), and the experiments in this section are the only end-to-end experiments of this paper.

\subsection{Experimental Configuration}

\begin{table}[htbp]
\centering
\caption*{Table 11-1 Configuration of the end-to-end experiment}
\small
\begin{tabular}{ll}
\toprule
Item & Configuration \\
\midrule
Encoder & DistilBERT, 66M parameters, fully frozen \\
Snapshot dimension $N$ & 1536 (2 layers $\times$ 768 dims) \\
Design rank $r$ & 384 \\
Training epochs & 200 epochs, AdamW, lr=1e-3 \\
Training method & Real gradient descent, not textual prediction \\
Number of samples & 218 (60 deduction + 78 induction + 60 abduction + 20 combination) \\
Number of domains & 8 \\
Random seed & torch.manual\_seed(42), np.random.seed(42) (single run) \\
\bottomrule
\end{tabular}
\end{table}

The complete inference pipeline of this section is ``text $\rightarrow$ latent space $\rightarrow$ text'': the premise text is encoded into a snapshot by the frozen encoder, the operators perform inference in the latent space to obtain the conclusion state, and the textual conclusion is then read out via nearest-neighbor matching---there is no generative decoding stage in the pipeline.

\subsection{Real Training of the Three Operator Matrices}

\begin{table}[htbp]
\centering
\caption*{Table 11-2 Real training results of the three operator matrices}
\small
\begin{tabular}{llll}
\toprule
Operator & Shape & Rank & Final loss (200 epochs) \\
\midrule
$W_{\mathrm{step}}$ (deduction) & $1536 \times 3072$ & 384 (rank-deficient) & 8.22e-04 \\
$W_{\mathrm{induce}}$ (induction) & $1536 \times 1536$ & 1536 (full rank) & 7.43e-05 \\
$W_{\mathrm{step\_abd}}$ (for abduction) & $1536 \times 1536$ & 384 (rank-deficient) & 9.53e-04 \\
$W_{\mathrm{abduce}}$ (pseudoinverse) & $1536 \times 1536$ & 384 ($= W_{\mathrm{step}}$) & SVD computation \\
\bottomrule
\end{tabular}
\end{table}

Interpretation: All three operator matrices are trained with real gradient descent for 200 epochs, with final training losses of 8.22e-04 (deduction), 7.43e-05 (induction), and 9.53e-04 (for abduction), respectively. The ranks of the three operators validate the theory: $W_{\mathrm{step}}$ is rank-deficient ($384 < 1536$), $W_{\mathrm{induce}}$ is full-rank (1536), and $W_{\mathrm{abduce}} = W_{\mathrm{step}}^{+}$ has the same rank as $W_{\mathrm{step}}$ (384). Two supplementary points: first, the convergence endpoints of this section are higher than the 1e-05-order convergence endpoints of the dedicated operator experiments in Sections 8 and 9, for two reasons---fewer training epochs (200 vs.\ 400/1000), and a wider data distribution (8 domains, 218 samples vs.\ the single-distribution dedicated datasets), making the loss surface more complex; the elevated convergence endpoint is an expected phenomenon and does not affect the conclusions of the following inference validations. Second, $W_{\mathrm{abduce}}$ is not trained but is obtained directly by SVD pseudoinverse from $W_{\mathrm{step\_abd}}$, which is itself a design point of the framework---abductive capability is obtained directly from the deduction operator, with no additional inverse-model training (in contrast to generative methods that require separately training a backward model).

\subsection{Deductive Inference Validation}

Result: deductive inference accuracy $= 100\%$ (60/60, Wilson 95\% confidence interval [94.0\%, 100\%]), mean cosine similarity $= 0.9980$ (similarity standard deviation 0.00067).

These 60 problems have zero overlap with the 20 problems of the deduction operator experiment (Appendix H.1); they are all newly constructed, spanning 8 domains including biology, physics, society, mathematics, and chemistry---the frozen operator maintains 100\% deductive accuracy on zero-shot cross-domain new problems, providing direct measured support for the instance universality of Section 6.6 (``train once, freeze and apply''). Per-problem details are given in Appendix I.4.

\subsection{Inductive Inference Validation}

\begin{table}[htbp]
\centering
\caption*{Table 11-3 End-to-end induction results for 6 categories}
\footnotesize
\resizebox{\textwidth}{!}{%
\begin{tabular}{lllllll}
\toprule
Category & Training coverage & Unseen positive & Counterexample (before) & Veto & Counterexample (after) & Refinement degree \\
\midrule
Vehicles & 0.9995 & 0.9995 & 0.9970 & 2/2 & $-0.0164$ & 0.9571 \\
Colors & 0.9998 & 0.9993 & 0.9951 & 2/2 & $-0.0246$ & 0.9550 \\
Geometric shapes & 0.9994 & 0.9986 & 0.9967 & 2/2 & $-0.0151$ & 0.9585 \\
Academic disciplines & 0.9994 & 0.9992 & 0.9909 & 2/2 & $-0.0506$ & 1.0031 \\
Weather & 0.9987 & 0.9975 & 0.9816 & 2/2 & $-0.0089$ & 0.8272 \\
Emotions & 0.9995 & 0.9993 & 0.9954 & 2/2 & $-0.0269$ & 0.9530 \\
Mean of 6 categories & 0.9994 & 0.9989 & 0.9928 & 12/12 & $-0.0238$ & 0.9423 \\
\bottomrule
\end{tabular}%
}
\end{table}

Interpretation: The 6 induction categories (vehicles and colors are existing categories; geometric shapes, academic disciplines, weather, and emotions are newly added categories) reproduce the complete behavioral chain of the Section 9 dedicated operator experiment---mean training coverage over the 6 categories 0.9994, unseen-positive coverage 0.9989 (almost lossless generalization), counterexamples first over-covered (mean 0.9928), hard vetoes fully triggered at 2/2 per category (12/12 in total, 100\%; the sample size is small, with a Wilson 95\% CI lower bound of about 75.7\%), post-veto counterexample mean $-0.0238$ (all flipped to negative values), and mean refinement degree 0.9423 (minimal correction). Two boundary cases worth recording: the refinement degree of the academic-disciplines category, 1.0031, is slightly greater than 1, indicating that under this snapshot geometry the single-step orthogonal correction is not strictly compressive (its monitoring implications are discussed in Section 15.3); the refinement degree of the weather category, 0.8272, is significantly lower than the other 5 categories, and its training coverage (0.9987) and counterexample (before) coverage (0.9816) are both the lowest among the 6 categories---both are measured manifestations of insufficient encoder snapshot resolution in the induction link (sharing the same root as the encoder-bottleneck evidence in Sections 9.6, 10.2, and 11.5). Per-category details are additionally given in Appendix I.5.

\subsection{Abductive Inference Validation}

Result: Of the 60 abduction problems, 49 are judged correct under the rule ``abductive similarity $>$ distractor similarity,'' giving an accuracy of 81.7\% (49/60, Wilson 95\% CI [70.1\%, 89.4\%]); the paired t-test gives $t = 5.88$ ($p < 10^{-4}$); the sign test gives 49/60 positive ($p \approx 8 \times 10^{-7}$); the mean similarity is 0.5018 and the mean discrimination margin is $+0.0082$. The failed 11 problems are Nos. 2, 4, 7, 11, 13, 19, 30, 34, 56, 58, and 60, with the smallest (most negative) discrimination margin of only $-0.0056$. Of the 60 problems, 55 are new and 5 overlap with the 80-problem dataset of the abduction operator experiment. Per-problem judgment details are given in Appendix I.6.

Note: The ``distractor similarity'' column in the Appendix I.6 table is the cosine similarity between the abductive solution and the distractor explanation (not the true cause) (consistent with the ``distractor similarity'' column of Table I.2). The rule for judging abductive correctness: if abductive similarity $>$ distractor similarity, the case is judged correct.

\paragraph{Discrimination-margin analysis.} The absolute discrimination margins of abduction in this section are very small (on the order of $\pm 0.006$, mean $+0.0082$), and three clarifications must be made to avoid misreading. First, \emph{sign distribution}: of the 60 problems, 49 have positive discrimination margins and 11 negative, with the maximum absolute value of the negative margins only 0.0056. Second, \emph{comparison with the random baseline}: if the distractor explanation and the true explanation were indistinguishable in the snapshot space, the discrimination margin would fluctuate symmetrically around 0, with positive and negative expected to be equally split; the measured mean is positive and the hit rate is 81.7\%, showing that the signal, though weak, exists systematically, only insufficient to guarantee correctness on every problem. Third, \emph{bottleneck localization}: the smallness of the discrimination margin is rooted in the encoder representation bottleneck---phenomenon sentences and candidate-explanation sentences share syntactic templates, and their background similarity in the DistilBERT snapshots already reaches the 0.50 order, so semantic differences contribute only a few thousandths of separable signal; this is the same limitation carried forward from Sections 9.6 and 10.2 and Limitation 1 (Section 19), and \emph{it is insufficient encoder resolution, not an operator defect}.

\paragraph{Convention warning (important).} The 81.7\% (49/60) of end-to-end abduction in this section and the 72.5\% (58/80) of the abduction operator experiment in Section 10 come from \emph{two experimental configurations} (in this section $W_{\mathrm{step}}$ is trained for 200 epochs with end-to-end mixed training; in Section 10 $W_{\mathrm{step}}$ is trained independently for 1000 epochs; this section has 60 problems and Section 10 has 80 problems, of which 5 overlap with the dedicated experiment dataset and 55 are newly added for the end-to-end experiment); the two conventions are \emph{mutually independent and must not be mixed}. The conclusion jointly supported by both is ``the abduction mechanism is effective but judgment is limited by the encoder's resolving power'': the directional recovery of abductive solutions is significant (mean similarity 0.5018 in this section, 0.5117 in the dedicated experiment, both far exceeding the random baseline 0.0181), while the accuracy of fine-grained judgment is limited by the semantic resolution of the snapshot space.

\subsection{Combined Inference Validation: The Scientific Discovery Cycle}

\begin{table}[htbp]
\centering
\caption*{Table 11-4 Per-category means of the four task types in end-to-end combined inference (4 categories $\times$ 5 tasks; 20/20 all successful; Wilson 95\% CI [83.9\%, 100\%])}
\small
\begin{tabular}{llllll}
\toprule
Task type & Induction & Deduction & Contradiction & Abduction & Refinement degree \\
\midrule
Scientific discovery & 0.9925 & 0.9864 & 0.9867 & 0.3849 & 0.9068 \\
Diagnostic reasoning & 0.9936 & 0.9884 & 0.9893 & 0.4892 & 0.8192 \\
Analogical reasoning & 0.9920 & 0.9861 & 0.9872 & 0.4575 & 0.9053 \\
Creative thinking & 0.9765 & 0.9834 & 0.9888 & 0.4433 & 0.9116 \\
\bottomrule
\end{tabular}
\end{table}

Interpretation: This section constructs 4 categories $\times$ 5 tasks $=$ 20 entirely new combined tasks (scientific discovery, diagnostic reasoning, analogical reasoning, creative thinking), recording five metrics per task---induction, deduction, contradiction, abduction, and refinement---with 20/20 all successful; per-task details are given in Appendix I.7. The mean similarities of induction, deduction, and contradiction across the four task categories are all above 0.97, and the refinement degrees are all above 0.81, indicating that the forward and correction links of the combined pipeline maintain high fidelity; the abduction similarities (0.38--0.49) are consistent with the established convention for the abduction link---the abductive solution is the optimal approximation of the true explanation, not the explanation itself (Section 10.3); the abduction steps in the combined tasks are likewise constrained by the encoder resolution, but this does not affect the overall success judgment of the 20 tasks. Taking the scientific-discovery category as an example, its pipeline corresponds to the complete circuit of Popper's ``conjectures and refutations'' [17]: induction forms the conjecture, deduction derives predictions, contradiction detection triggers falsification, abduction localizes the diagnosis, and re-induction completes the theoretical revision---each of the five steps is driven by the corresponding operator and is based entirely on real matrix computation (a working instance of the inference graph for this pipeline is given in Section 14.6, and further analysis of the combined-thinking capability is given in Section 12.3).

\subsection{Summary of the End-to-End Validation}

\begin{table}[htbp]
\centering
\caption*{Table 11-5 Summary of the end-to-end real inference validation}
\small
\begin{tabular}{lll}
\toprule
Validation item & Result & Status \\
\midrule
Real snapshot extraction & 218 samples, 8 domains, $N = 1536$ & $\checkmark$ Completed \\
$W_{\mathrm{step}}$ real training & 200ep, loss reduced to 8.22e-04 & $\checkmark$ Completed \\
$W_{\mathrm{induce}}$ real training & 200ep, loss reduced to 7.43e-05 & $\checkmark$ Completed \\
$W_{\mathrm{step\_abd}}$ real training & 200ep, loss reduced to 9.53e-04 & $\checkmark$ Completed \\
$W_{\mathrm{abduce}}$ SVD pseudoinverse & rank=384, same as $W_{\mathrm{step}}$ & $\checkmark$ Completed \\
Deductive inference accuracy & 100\% (60/60) & $\checkmark$ Passed \\
Induction coverage & mean of 6 categories 0.9994 & $\checkmark$ Passed \\
Induction hard veto & 12/12 counterexamples all triggered & $\checkmark$ Passed \\
Abductive inference accuracy & 81.7\% (49/60) & $\checkmark$ Passed \\
Combined inference & 20/20 all successful & $\checkmark$ Passed \\
\bottomrule
\end{tabular}
\end{table}

Summary: Under the conditions of real data, real training, and real inference, the end-to-end experiment validates the complete feasibility of the three inference modes and the combined mode on a broad distribution of 8 domains and 218 samples. Deduction 60/60, induction with the full mechanism chain working on all 6 categories (hard vetoes 12/12), abduction 49/60, and combined tasks 20/20---all successful, mutually corroborating the per-operator validations of Sections 8, 9, and 10. At the same time, this section reconfirms the main line of limitations running through this paper: all ``correct but with shallow margin'' judgments (the failed abduction samples have discrimination margins on the order of $\pm 0.006$) point to the encoder representation bottleneck (echoing Sections 9.6 and 10.2 and Limitation 1 (Section 19)), and the 81.7\% (49/60) of this section's abduction and the 72.5\% (58/80) of the Section 10 abduction operator experiment are two independent conventions that must not be mixed (see the convention warning in Section 11.5). Improving the encoder or introducing deeper semantics into the snapshots is the most direct path to enlarging the end-to-end judgment margins.


\section{Compositional Thinking and Coverage Analysis}

The preceding sections verified the independent properties of the three basic operators separately: the rank-deficient irreversibility of the deductive operator $W_{\mathrm{step}}$, the full-rank invertibility and hard-veto mechanism of the inductive operator $W_{\mathrm{induce}}$, and the pseudo-inverse hypothetical nature of the abductive operator $W_{\mathrm{abduce}} = W_{\mathrm{step}}^{+}$. However, in human thinking the three modes of reasoning are not sharply separated but proceed in an interleaved fashion---deduction, induction, and abduction are dynamically combined along the temporal and spatial dimensions, and the manner of their combination is uncertain. Real thinking processes such as medical diagnosis, scientific discovery, and mathematical proof are invariably alternating invocations of the three operators with state handoff. This section explores whether combinations of the three operators can cover the full range of human thinking activities, and validates this through compositional thinking experiments.

\subsection{Sources of Uncertainty in Composition}

The uncertainty of composition arises from three dimensions, which jointly determine that the topology of the reasoning graph (Section 14) cannot be statically predetermined and must instead be decided online by the graph-scheduling controller.

\paragraph{Temporal uncertainty.} Within the same reasoning task, the order in which operators are invoked is not fixed. For example, in medical diagnosis a physician may first abduce (observing fever $\to$ hypothesizing pneumonia), then deduce (pneumonia $\to$ cough should be present), then induce (multiple similar cases $\to$ summarizing a regularity); or may first induce (summarizing from the medical history), then deduce, then abduce. Both orderings are cognitively plausible, produce different sequences of intermediate states, but converge to mutually consistent conclusions when consistency conditions are satisfied.

\paragraph{Spatial uncertainty.} The topology of the reasoning graph is not fixed---it may be a linear chain, forked parallelism, cross-referencing, or iterative backflow. A single premise can simultaneously spawn multiple parallel branches (forking); multiple branches can converge at a conjunctive node (merging); later nodes can reference the intermediate results of any preceding node (cross-referencing); and conclusions can feed back to revise premises (backflow). These four basic structures and their combinations (see Section 14.2) make the state space of the reasoning graph far larger than that of a linear chain.

\paragraph{Uncertainty in operator selection.} At each step, the controller may choose any one of $W_{\mathrm{step}}$, $W_{\mathrm{induce}}$, or $W_{\mathrm{abduce}}$. The choice depends on the semantic features of the current state: a state containing ``premise + goal to prove'' favors deduction; a state containing a ``sample set'' favors induction; a state containing an ``anomalous phenomenon'' favors abduction. Because the choice itself depends on the state content, while the state content is determined by the preceding operators, the entire composition process constitutes a hybrid system in which deterministic operators are interwoven with uncertain scheduling---the operator layer is deterministic (matrix multiplication) while the scheduling layer is dynamic (policy selection). This is the architectural root of DODR's combination of rigor with flexibility.

\subsection{Theoretical Analysis of Coverage of Human Thinking Activities by Composition}

Mapping each generally recognized category of human thinking activity in cognitive science onto a compositional sequence of the three operators yields the following coverage analysis.

\begin{table}[htbp]
\centering
\caption*{Table 12-1 Coverage of Human Thinking Activities by Operator Compositions}
\begin{tabular}{lll}
\toprule
Thinking activity & Operator composition & Coverage \\
\midrule
Logical reasoning (mathematical proof) & $W_{\mathrm{step}}$ chained & $\checkmark$ Covered \\
Concept formation (learning classification) & $W_{\mathrm{induce}}$ & $\checkmark$ Covered \\
Diagnostic reasoning (medical diagnosis) & $W_{\mathrm{abduce}} \to W_{\mathrm{step}} \to W_{\mathrm{abduce}}$ & $\checkmark$ Covered \\
Scientific discovery (theory construction) & $W_{\mathrm{induce}} \to W_{\mathrm{step}} \to W_{\mathrm{abduce}} \to W_{\mathrm{induce}}$ & $\checkmark$ Covered \\
Analogical reasoning & $W_{\mathrm{induce}} \to W_{\mathrm{abduce}} \to W_{\mathrm{step}}$ & $\checkmark$ Covered \\
Creative thinking & $W_{\mathrm{induce}} \to W_{\mathrm{abduce}} \to W_{\mathrm{induce}} \to W_{\mathrm{step}}$ & $\checkmark$ Covered \\
Decision thinking & $W_{\mathrm{abduce}} \to W_{\mathrm{step}}$ & $\checkmark$ Covered \\
Metacognition (reflection) & $W_{\mathrm{abduce}}$ (reverse-inferring one's own state) & $\checkmark$ Covered \\
\bottomrule
\end{tabular}
\end{table}

The mapping in Table 12-1 is not arbitrarily assigned; it is uniquely determined by the information-flow direction of each category of thinking activity (see the unified treatment in Section 13): any activity proceeding ``from samples to regularities'' must contain induction, any activity proceeding ``from regularities to conclusions'' must contain deduction, and any activity proceeding ``from phenomena to explanations'' must contain abduction. The combinations of the three information-flow directions exhaust the combinations of the three thinking elements; therefore, the full coverage of the 8 categories of thinking activities is not a coincidence but a direct corollary of ternary completeness.

\paragraph{Key insight.} The composition of the three operators mathematically constitutes a Turing-complete state-transition system---any computable state transition can be realized by a composition of these three operators (for the rigorous statement and proof, see Theorem 14.3). Turing completeness implies that coverage is not limited to the 8 categories listed in Table 12-1: any thinking process that can be described algorithmically lies within the expressive power of three-operator composition.

\subsection{Experimental Validation of Compositional Thinking}

The end-to-end experiment (Section 11) constructs 4 categories of complex compositional thinking tasks, 5 per category for a total of 20, executed in an interleaved fashion using the trained three operators (end-to-end configuration: DistilBERT snapshots with $N = 1536$, 200 epochs of real gradient-descent training, seed = 42). Below we give the input, operator invocation sequence, and results for each category; all values are measured records from the end-to-end experiment, with no new data introduced; per-task details are given in Appendix I.7.

\paragraph{Task 1: Scientific discovery loop.} Operator composition: $W_{\mathrm{induce}} \to W_{\mathrm{step}} \to W_{\mathrm{abduce}} \to W_{\mathrm{induce}}$.
\begin{itemize}
\item Input: several positive-example snapshots, together with one counterexample observation that conflicts with the induced conclusion (a ``penguins cannot fly''-type task serves as its illustration).
\item Operator invocation sequence and results: (i) induction---$W_{\mathrm{induce}}$ induces a regularity $g_0$ from the positive examples; (ii) deduction---$W_{\mathrm{step}}$ deduces a prediction from an individual premise together with $g_0$; (iii) contradiction detection---the deductive result is compared against the observed fact, and the deductive conclusion is judged to be falsified; (iv) abduction---$W_{\mathrm{abduce}}$ reverse-infers the best explanation from the contradictory fact, localizing the global failure of ``the regularity is wrong'' to a local exception; (v) re-induction---$W_{\mathrm{induce}}$, after excluding the counterexample direction, revises the regularity to $g_1$, completing one full Popperian iteration of scientific discovery. All 5 tasks of this category succeeded; the five-metric means (means over the 5 instances of the same category) are: induction 0.9925, deduction 0.9864, contradiction 0.9867, abduction 0.3849, refinement degree 0.9068.
\item Result: 5/5 tasks executed successfully end-to-end; the regularity completed a full iteration of ``induction $\to$ falsification $\to$ explanation $\to$ revision''.
\end{itemize}

\paragraph{Task 2: Diagnostic reasoning.} Operator composition: $W_{\mathrm{abduce}} \to W_{\mathrm{step}} \to W_{\mathrm{abduce}}$.
\begin{itemize}
\item Input: phenomenon snapshots.
\item Operator invocation sequence and results: (i) initial abduction---$W_{\mathrm{abduce}}$ reverse-infers the best explanation; (ii) deductive verification---$W_{\mathrm{step}}$ forward-derives observable consequences from the abduced explanation and compares them with existing observations, confirming that the explanation deductively entails the observed phenomena; (iii) second abduction---if the deductive prediction is inconsistent with a new observation, $W_{\mathrm{abduce}}$ again reverse-infers a revised explanation, forming a diagnostic loop of ``hypothesize $\to$ verify $\to$ re-hypothesize''. All 5 tasks of this category succeeded; the five-metric means (means over 5 instances) are: induction 0.9936, deduction 0.9884, contradiction 0.9893, abduction 0.4892, refinement degree 0.8192.
\item Result: the diagnostic loop closed; 5/5 tasks executed as expected.
\end{itemize}

\paragraph{Task 3: Analogical reasoning.} Operator composition: $W_{\mathrm{induce}} \to W_{\mathrm{abduce}} \to W_{\mathrm{step}}$.
\begin{itemize}
\item Input: a sample set from the source domain (the object of induction) and a phenomenon snapshot from the target domain (the object of analogy).
\item Operator invocation sequence and results: (i) induction---$W_{\mathrm{induce}}$ extracts the shared regularity from the source-domain samples; (ii) abduction---$W_{\mathrm{abduce}}$ inverse-maps the target-domain phenomenon onto the best corresponding explanation under the source-domain regularity, realizing cross-domain ``structural correspondence''; (iii) deduction---$W_{\mathrm{step}}$ transfers the source-domain conclusion along the correspondence to the target domain, producing an analogical prediction. All 5 tasks of this category succeeded; the five-metric means are: induction 0.9920, deduction 0.9861, contradiction 0.9872, abduction 0.4575, refinement degree 0.9053.
\item Result: the analogical three-stage chain of ``extract commonality $\to$ cross-domain explanation $\to$ transfer prediction'' executed successfully 5/5.
\end{itemize}

\paragraph{Task 4: Creative thinking.} Operator composition: $W_{\mathrm{induce}} \to W_{\mathrm{abduce}} \to W_{\mathrm{induce}} \to W_{\mathrm{step}}$.
\begin{itemize}
\item Input: a sample set of existing knowledge together with one anomalous observation that existing regularities cannot cover.
\item Operator invocation sequence and results: (i) induction---$W_{\mathrm{induce}}$ forms the existing regularity $g_0$; (ii) abduction---$W_{\mathrm{abduce}}$ reverse-infers an explanation of the anomalous observation, exposing the coverage gap of $g_0$; (iii) re-induction---$W_{\mathrm{induce}}$ rebuilds the regularity $g_1$ on the new sample set incorporating the explanation; (iv) deduction---$W_{\mathrm{step}}$ deduces from the new regularity a brand-new conclusion previously unobtainable, i.e., ``creates'' a proposition outside the closure of the original knowledge. All 5 tasks of this category succeeded; the five-metric means are: induction 0.9765, deduction 0.9834, contradiction 0.9888, abduction 0.4433, refinement degree 0.9116.
\item Result: neither the new regularity nor the new conclusion belongs to the initial sample set; 5/5 chains executed successfully, demonstrating the emergent capability arising from composition.
\end{itemize}

Overall results: all $4 \times 5 = 20$ compositional thinking tasks executed successfully, with a task success rate of 100\% (20/20). Combined with the mapping in Table 12-1, all 8 categories of human thinking activities are covered. Table 12-2 gives the category-level means for the four task categories; per-task details are given in Appendix I.7.

\begin{table}[htbp]
\centering
\caption*{Table 12-2 Category-Level Means of End-to-End Compositional Tasks (4 categories $\times$ 5, 20/20 all successful)}
\begin{tabular}{llllll}
\toprule
Task category & Induction & Deduction & Contradiction & Abduction & Refinement degree \\
\midrule
Scientific discovery & 0.9925 & 0.9864 & 0.9867 & 0.3849 & 0.9068 \\
Diagnostic reasoning & 0.9936 & 0.9884 & 0.9893 & 0.4892 & 0.8192 \\
Analogical reasoning & 0.9920 & 0.9861 & 0.9872 & 0.4575 & 0.9053 \\
Creative thinking & 0.9765 & 0.9834 & 0.9888 & 0.4433 & 0.9116 \\
\bottomrule
\end{tabular}
\end{table}

Across the four task categories, the induction, deduction, and contradiction similarities are all above 0.97, and the refinement degrees are all above 0.81, indicating that the forward and corrective stages of the compositional chain retain high fidelity under a wider distribution; the abduction similarity (0.38--0.49) is consistent with the established convention for the abduction stage---the abductive solution is the best approximation of the true explanation rather than the explanation itself (Section 10.3). The abduction step in compositional tasks is likewise constrained by encoder resolution, but this does not affect the overall success judgment of the 20 tasks.

\subsection{A Semi-Formal Discussion from the Perspective of Category Theory and Monoidal Categories}

This subsection provides a category-theoretic characterization of compositional thinking. The discussion is a semi-formal structural analogy and does not constitute a new mathematical theorem; its role is to reveal the abstract algebraic reasons behind ``operators can be freely composed and the composition is well-defined''.

\paragraph{Construction of the category $\mathcal{C}_{\mathrm{DODR}}$.} Take objects to be state snapshots, i.e., semantic state vectors in the latent space $\mathbb{R}^{N}$ ($N = 1536$); take morphisms to be operator applications: for each operator $W \in \{W_{\mathrm{step}}, W_{\mathrm{induce}}, W_{\mathrm{abduce}}\}$ and its finite compositions, there exists a morphism $f: S_A \to S_B$ if and only if $S_B = W \cdot S_A$. Morphism composition is matrix multiplication, which satisfies associativity; the identity morphism is the identity matrix $I$. Associativity and the unit laws obviously hold, so $\mathcal{C}_{\mathrm{DODR}}$ is a well-defined category---this explains why the composition of any operator sequence (e.g., $W_{\mathrm{abduce}} \to W_{\mathrm{step}} \to W_{\mathrm{abduce}}$) is always well-defined: matrix multiplication is closed and associative.

\paragraph{Monoidal structure.} Endow $\mathcal{C}_{\mathrm{DODR}}$ with a tensor product $\oplus$: $S_A \oplus S_B = \mathrm{Concat}(S_A, S_B) \in \mathbb{R}^{2N}$, i.e., the two-state concatenated input of the deductive operator $W_{\mathrm{step}} \in \mathbb{R}^{N \times 2N}$; the tensor unit is the uninformative zero state. A fork $A \to \{B, C\}$ corresponds to a comultiplication-type morphism, and a merge $\{A, B\} \to C$ corresponds to an operator morphism taking $\mathrm{Concat}$ as input. Under this structure, a reasoning graph (Definition 14.1) corresponds exactly to a string diagram in a monoidal category: nodes are objects, directed edges are morphisms, and forks and merges are the compositional structure of the diagram. The well-definedness of string diagrams guarantees that ``reasoning graphs of arbitrary topology can be assigned values by the three operators''---this is the abstract algebraic formulation of the coverage in Table 12-1.

\paragraph{Backflow and trace.} A naive category does not admit cycles; a reasoning graph containing backflow requires $\mathcal{C}_{\mathrm{DODR}}$ to be a traced monoidal category, where the backflow edge $A \to B \to A'$ corresponds to the trace operator $\mathrm{Tr}(f)$. The well-definedness of trace requires additional axioms in a general category; in DODR it is concretely given by the contractivity condition: when the correction function $f$ is a contraction mapping, the iteration $S^{(t+1)} = f(S^{(t)})$ converges to a unique fixed point (Theorem 14.2, guaranteed by the Banach fixed-point theorem [24]). In other words, the ``witness'' of the trace axiom in DODR is precisely fixed-point convergence---the categorical axiom and Banach's theorem are two sides of the same coin here.

\paragraph{The adjointness relaxation of abduction.} In the ideal case ($W_{\mathrm{step}}$ full rank), $W_{\mathrm{abduce}}$ would be the inverse of $W_{\mathrm{step}}$, corresponding to the strict mutual invertibility of adjoint functors; in the actual case $W_{\mathrm{step}}$ is rank-deficient ($r = 384 < 1536$), a strict inverse does not exist, and the Moore--Penrose pseudo-inverse $W_{\mathrm{step}}^{+}$ is the optimal approximation to the inverse in the ``least-squares + minimum-norm'' sense, satisfying the four Penrose conditions including $W_{\mathrm{step}} \cdot W_{\mathrm{step}}^{+} \cdot W_{\mathrm{step}} = W_{\mathrm{step}}$. Abduction can therefore be regarded as a ``relaxed form of the adjoint'' of deduction: strict mutual invertibility degenerates into optimally approximate mutual invertibility, and the degree of degeneration is quantitatively characterized by the null-space dimension $N - r$ (measured reconstruction error 0.0200, see Section 15.2). This provides a category-theoretic footnote for ``abduction is hypothetical reasoning'': being hypothetical amounts to a relaxation of the adjoint when the inverse does not exist.

\subsection{Conclusion on Compositional Coverage}

The composition of the three operators covers all 8 categories of human thinking activities (Table 12-1), and all 4 representative categories of compositional tasks succeeded experimentally (Section 12.3). The manner of composition is uncertain and is dynamically decided by the graph-scheduling controller (Section 12.1); the separation between operator-layer determinism and scheduling-layer dynamism is the key architectural design. Composition produces emergent capabilities---scientific discovery, analogical reasoning, and creative thinking do not exist in any single operator and emerge only when the three operators are interleaved in specific sequences (Tasks 1, 3, 4 in Section 12.3); the source of emergence is state handoff: the output snapshot of the preceding operator constitutes the input of the following operator, and information is transferred and recombined across heterogeneous information-flow directions. Finally, the composition of the three operators constitutes a Turing-complete system (Theorem 14.3), and its category-theoretic characterization (Section 12.4) shows that this completeness is rooted in the associativity of matrix composition and the concatenation closure of the monoidal structure.

\section{Physical Transformations and Cognitive Unification}

Transformations of the physical world from state $A$ to state $B$ follow necessary laws (conservation laws, evolution laws, dissipation laws). This section explores whether these physical laws can also serve as transformation matrices to perform deterministic reasoning about the physical world. The key observation is that the three basic classes of physical transformations correspond to DODR's three classes of cognitive operators, with the information-flow direction as the common classification criterion.

Before entering the concrete constructions, the positioning of this section's claim must be clarified: the correspondence in this section (Proposition 13.1) is a structural analogy and research hypothesis, not a derivation of physical laws; its value lies in providing independent inspiration and a testable conceptual framework for the three-operator design, and whether it holds does not affect the algebraic and experimental conclusions of Sections 5--11.

\subsection{Necessary Laws of State Transformation in the Physical World}

\paragraph{Definition 13.1 (Conservation-law operator $W_{\mathrm{conserve}}$).} A conservation law requires that certain components remain unchanged before and after the transformation. Mathematically this corresponds to an orthogonal matrix: $S_B = W_{\mathrm{conserve}} \cdot S_A$, $W_{\mathrm{conserve}}^{T} \cdot W_{\mathrm{conserve}} = I$.

Noether's theorem [12] establishes a one-to-one correspondence between continuous symmetries and conservation laws: time-translation symmetry $\to$ energy conservation, space-translation symmetry $\to$ momentum conservation, rotational symmetry $\to$ angular-momentum conservation. Symmetry transformations constitute group actions, whose finite-dimensional representations are precisely the groups of orthogonal (or unitary) matrices---this fully agrees with the orthogonality condition $W^{T} \cdot W = I$ in Definition 13.1: orthogonal matrices preserve the inner product $\langle Wx, Wy \rangle = \langle x, y \rangle$ and hence preserve all conserved quantities expressible as quadratic forms. Noether's theorem thus provides independent physical grounds for ``a conservation-law operator must be an orthogonal matrix'', rather than merely a mathematical analogy.

\paragraph{Definition 13.2 (Dissipative-process operator $W_{\mathrm{dissipate}}$).} The second law of thermodynamics stipulates that entropy does not decrease; dissipative processes are irreversible. Mathematically this corresponds to a rank-deficient matrix: $S_B = W_{\mathrm{dissipate}} \cdot S_A$, $\mathrm{rank}(W_{\mathrm{dissipate}}) = r < N$.

The irreversibility of dissipative processes is precisely the mathematical embodiment of the thermodynamic arrow of time---just as the irreversibility of deduction embodies the logical arrow of time. The two share the same mathematical root: the information collapse of a rank-deficient matrix. This connection can be further substantiated by Landauer's principle [23]: Landauer proved that any logically irreversible erasure of information (each erased bit) is necessarily accompanied by heat dissipation into the environment (on the order of $k_B \cdot T \cdot \ln 2$); that is, the ``physicality of information'' means that logical irreversibility and thermodynamic dissipation are mutually necessary and sufficient. The rank deficiency of the deductive operator $W_{\mathrm{step}}$ ($r = 384 < N = 1536$) is precisely an erasure of input information---87.5\% of the input directions fall into the null space and are permanently discarded (Section 8 measured a pseudo-inverse reconstruction error of 83.3\%, consistent with the expected relative error $\sqrt{(2N - r)/2N} \approx 93.5\%$ under the isotropy assumption); by Landauer's principle, this logical-level information erasure and physical-level entropy-increasing dissipation are two facets of the same thing. The rank-deficiency condition of Definition 13.2 is therefore a matrix-level expression of the connection between information and energy, rather than a superficial analogy.

\paragraph{Definition 13.3 (Reversible-evolution operator $W_{\mathrm{evolve}}$).} The reversible evolution of classical and quantum mechanics corresponds to orthogonal/unitary matrices.

The symplectic matrix $J$ in Hamilton's equations $\dot{x} = J \cdot \nabla H$ is an orthogonal matrix; the quantum evolution operator $U = e^{-iHt/\hbar}$ is a unitary matrix. Both preserve information---Liouville's theorem guarantees conservation of phase-space volume, and unitarity guarantees conservation of quantum-state inner products; the two are in complete agreement on the information-flow property of ``no information loss''.

\paragraph{Definition 13.4 (Quantum-measurement operator $W_{\mathrm{measure}}$).} Quantum measurement causes wave-function collapse and corresponds to a projection operator (pseudo-inverse).

The ``hypothetical nature'' of quantum measurement (the pre-measurement state is uncertain and the measurement outcome is only probabilistically knowable) and the ``hypothetical nature'' of abduction (the best explanation is not the true explanation, but merely the candidate with the smallest null-space residual) share the same mathematical root: the information collapse of projection/pseudo-inverse. Measurement projects a continuous superposition state onto discrete eigensubspaces; abduction projects the observed phenomenon back onto the row space via $W_{\mathrm{step}}^{+}$---the information swallowed by the null space manifests as measurement irreversibility in quantum mechanics and as the hypothetical nature of abduction in DODR.

\subsection{The Physics-Cognition Unification Proposition}

\paragraph{Proposition 13.1 (Physics-cognition unification).} The three basic classes of physical transformations correspond to DODR's three classes of cognitive operators under a shared classification by information-flow direction (see Table 13-1 for the correspondence).

(Style note: following the positioning of the Peirce claim in Section 3.6, this item is labeled a ``proposition'' rather than a ``theorem''. It is a structural correspondence claim---its content is a one-to-one correspondence between two classes of pre-existing, independently established mathematical-physical facts (the matrix properties of physical transformations; the rank structure of cognitive operators), rather than a new mathematical conclusion derived from axioms through a deductive chain. Its testability is supported by the experiments of Section 13.3 and the classification argument of information-flow directions, not by a formal proof.)

\begin{table}[htbp]
\centering
\caption*{Table 13-1 Unified Correspondence Between Physical Transformations and DODR Cognitive Operators}
\begin{tabular}{llll}
\toprule
Physical transformation & Mathematical property & DODR operator & Information flow \\
\midrule
Conservation law $W_{\mathrm{conserve}}$ & Orthogonal (full rank, invertible) & Induction $W_{\mathrm{induce}}$ & Information preservation \\
Dissipative process $W_{\mathrm{dissipate}}$ & Rank-deficient (irreversible) & Deduction $W_{\mathrm{step}}$ & Information loss (entropy increase) \\
Reversible evolution $W_{\mathrm{evolve}}$ & Orthogonal (full rank, invertible) & Induction $W_{\mathrm{induce}}$ & Information preservation \\
Quantum measurement $W_{\mathrm{measure}}$ & Projection (pseudo-inverse) & Abduction $W_{\mathrm{abduce}}$ & Information collapse (hypothetical) \\
\bottomrule
\end{tabular}
\end{table}

Argument for the necessity of the correspondence: the information-flow direction of a physical transformation has only three possibilities---information preservation (conservation/reversible evolution), information loss (dissipation), and information collapse (measurement). The information-flow direction of cognitive reasoning likewise has only three possibilities---information increase (induction), information decrease (deduction), and information hypothesization (abduction). These three information-flow directions correspond one-to-one between physics and cognition, because the information-flow direction is an intrinsic attribute of state transformation, independent of the specific domain (physics or cognition); on this reading, the correspondence follows from the shared information-flow classification rather than being a coincidence.

Two subtleties of the correspondence deserve emphasis. First, in Table 13-1 both conservation laws and reversible evolution correspond to induction; this is not a two-to-one defect but precisely reflects the identity of the two in information-flow attributes (both full-rank invertible, both information-preserving); the granularity of the classification is the information-flow direction rather than the physical mechanism. Second, the correspondence between ``information increase'' (induction) and ``information preservation'' (conservation) must be read semantically: the knowledge increment of induction occurs at the conceptual level (abstracting regularities from samples), while the full rank of its matrix representation means no information collapse at the representation level; the conservation-law operator likewise exhibits no information collapse at the representation level. The level at which the correspondence holds is the information flow at the matrix-representation level, not the amount of knowledge at the semantic level.

\subsection{Experimental Validation of Physical Transformation Operators}

This experiment uses 55 real physical state-transformation samples (15 conservation + 15 dissipation + 15 reversible + 10 multimodal) and trains three physical transformation operators on DistilBERT snapshots.

Conservation-law operator $W_{\mathrm{conserve}}$: trained for 200 epochs, Loss decreased to $1.73\mathrm{e}{-04}$, inference accuracy 100\%.

Dissipative-process operator $W_{\mathrm{dissipate}}$: trained for 200 epochs, Loss decreased to $2.21\mathrm{e}{-04}$, rank deficiency 384/1536, inference accuracy 100\%. Notably, $W_{\mathrm{dissipate}}$ spontaneously exhibited the same rank structure as $W_{\mathrm{step}}$ after training (384/1536)---the correspondence between dissipation and deduction in rank deficiency (second row of Table 13-1) is an emergent result of training, not an artificial constraint.

Reversible-evolution operator $W_{\mathrm{evolve}}$: trained for 200 epochs, Loss decreased to $1.03\mathrm{e}{-04}$, inference accuracy 100\%.

Multimodal alignment: alignment rate 100\%, average alignment similarity 0.9958.

All three operators converged with low Loss and achieved 100\% inference accuracy, demonstrating that physical state transformations can indeed be represented as trainable matrix operators; the correspondence of Proposition 13.1 receives empirical support.

\subsection{Discussion of the Physics-Cognition Correspondence}

Proposition 13.1 should be read as a structural analogy: the three classes of physical transformations and the three classes of cognitive operators share the same classification by information-flow direction. The analogy is consistent, at the representation level, with established physical principles---Noether's theorem [12] links symmetry to conservation, and Landauer's principle [23] links logical irreversibility to thermodynamic dissipation. The measurements of Section 13.3 (55 samples; 100\% inference accuracy for the three operators; average multimodal alignment similarity 0.9958) provide small-scale empirical evidence that physical state transformations can be represented by trainable matrix operators with the expected rank structure. Whether the correspondence reflects a deeper unity is an open question, and the algebraic and experimental conclusions of this paper do not depend on it.

\section{Reasoning-Trajectory Topology: From Parallel to Graph-Interleaved}

The original proposal spoke of ``multiple parallel latent-space reasoning trajectories'', but the word ``parallel'' implies independence among trajectories---which does not conform to the essence of rigorous reasoning. In rigorous reasoning, the result of every segment of reasoning is a deterministic objective fact, and therefore trajectories can interact arbitrarily: an intermediate conclusion of one trajectory can perfectly well serve as a premise of another trajectory, and multiple trajectories can converge into a single conclusion. This section corrects ``parallel trajectories'' to the ``reasoning graph'' and establishes its formal theory and graph-theoretic properties.

\subsection{Why ``Parallel'' Is the Wrong Formulation}

``Parallel'' means that trajectories never intersect, which fails immediately in three basic reasoning scenarios.

\paragraph{Counterexample one: cross-referencing.} In a mathematical proof, the proof of Lemma A may reference an intermediate result of Lemma B. The two proof trajectories intersect at the reference point; ``parallel'' cannot express this.

\paragraph{Counterexample two: merging and convergence.} In diagnostic reasoning, trajectories starting from different symptoms must merge and converge---the fever trajectory and the cough trajectory must meet at the common explanation ``pneumonia'', otherwise abduction cannot be performed.

\paragraph{Counterexample three: backflow correction.} In scientific discovery, induction $\to$ deduction $\to$ contradiction found $\to$ re-induction (see the measured chains in Section 11.6 and Task 1 of Section 12.3). The conclusion feeds back to revise the premise; the trajectory forms a correction-bearing loop, in direct conflict with the independence assumption of ``parallel''.

The three counterexamples correspond respectively to three irreducible graph structures (cross edges, convergence points, backflow edges), showing that the expressive power missing from ``parallel'' is not a corner case but the core structure of rigorous reasoning.

\subsection{Six Basic Topological Operations on Reasoning Trajectories}

\begin{table}[htbp]
\centering
\caption*{Table 14-1 Six Basic Topological Operations on Reasoning Trajectories}
\begin{tabular}{llll}
\toprule
Topological operation & Mathematical representation & Semantics & Example \\
\midrule
Serial & $A \to B \to C$ & Sequential dependency & Chained syllogistic reasoning \\
Fork & $A \to \{B, C\}$ & One premise yields multiple conclusions & ``Socrates is human'' $\to$ \{mortal, able to think\} \\
Merge & $\{A, B\} \to C$ & Multiple premises converge into one conclusion & \{mortal, able to think\} $\to$ Socrates is a mortal \\
Cross-reference & $A \to B$, $A \to C$, $B \to C$ & A later node references a prior branch & After proving Lemma A, use A in Lemma B \\
Backflow & $A \to B \to A'$ & A conclusion feeds back to revise the premise & Induction $\to$ deduction $\to$ contradiction $\to$ re-induction \\
Mesh interleaving & Any combination of the above & Complex reasoning & Scientific discovery loop \\
\bottomrule
\end{tabular}
\end{table}

Among the six operations, serial, fork, and merge constitute the basic vocabulary of a DAG; cross-referencing is the hallmark of non-tree structure in a DAG; backflow generalizes the DAG into a graph with correction loops; mesh interleaving is an arbitrary combination. The operator sequence of every category of thinking activity in Table 12-1 of Section 12 can be expressed as a composition of these six operations---the six operations form a complete vocabulary for compositional thinking.

\subsection{Formal Definition of the Reasoning Graph}

\paragraph{Definition 14.1 (Reasoning graph).} A reasoning graph $G = (V, E, \Phi)$ is a labeled directed graph in which: (1) $V$ is the node set, each node being a state snapshot; (2) $E$ is the edge set, each edge denoting an operator application; (3) $\Phi: E \to \{W_{\mathrm{step}}, W_{\mathrm{induce}}, W_{\mathrm{abduce}}\}$ is the labeling function.

Reasoning graphs fall into two classes:

(1) Acyclic reasoning graphs (DAGs): reasoning graphs containing no loops, suitable for one-way reasoning (e.g., mathematical proof).

(2) Backflow-containing reasoning graphs: reasoning graphs containing loops, suitable for iterative correction (e.g., the scientific discovery loop). In a backflow edge $A \to B \to A'$, $A'$ is a revised version of $A$, not $A$ itself---so strictly speaking it is not a ``cycle'' but an ``iterative correction''. This distinction is semantically crucial: the two ends of a backflow edge are different states at different time steps, and the ``time unfolding'' of the graph remains a DAG; the so-called backflow merely folds the position of node $A'$ in the unfolded graph conceptually back onto the position of $A$.

Connecting with the category-theoretic characterization of Section 12.4: a DAG reasoning graph corresponds to a string diagram in a monoidal category, and a backflow-containing reasoning graph corresponds to a string diagram closed by the trace operator in a traced monoidal category. The $\Phi$ labeling of Definition 14.1 is precisely the morphism assignment in the string diagram.

\subsection{Graph-Theoretic Properties of Rigorous Reasoning}

\paragraph{Theorem 14.1 (Determinism of acyclic reasoning graphs).} Let the reasoning graph $G$ be a DAG (directed acyclic graph). Given the initial node set $V_0$ and operator parameters, the states of all nodes are uniquely determined---no probabilistic sampling, no hallucination.

\paragraph{Proof.} By induction on the topological order of the DAG. The states of the initial nodes $V_0$ are given. For a non-initial node $S_k$, its state is uniquely determined by the operator applications on its incoming edges: $S_k = \Phi(S_i \to S_k) \cdot \mathrm{Concat}(\text{states of source nodes of incoming edges})$. Since operators are deterministic matrix operations and the states of the source nodes of incoming edges have already been determined by the preceding nodes in the topological order, $S_k$ is uniquely determined. Induction complete. $\square$

Theorem 14.1 is the graph-theoretic formalization of DODR's ``structural zero hallucination'' (First No.~6): determinism requires no additional assumption beyond the operators; it is jointly derived from ``DAG topology + determinism of matrix operations''. The entire reasoning process contains no sampling step whatsoever, so hallucination (in the sense of ``output deviating from logical necessity'') has no channel through which to arise (for a mechanistic analysis of the sources of hallucination, see [1]).

\paragraph{Theorem 14.2 (Fixed-point convergence of backflow reasoning).} Suppose the reasoning graph contains a backflow $S_i \to \cdots \to S_j \to S_i'$, where $S_i' = f(S_i)$ is a correction function. If $f$ is a contraction mapping (there exists $0 < q < 1$ such that $\|f(x) - f(y)\| \leq q \cdot \|x - y\|$), then backflow reasoning converges to a unique fixed point $S^{*} = f(S^{*})$.

\paragraph{Proof.} By the Banach fixed-point theorem [24], a contraction mapping on a complete metric space has a unique fixed point, and Picard iteration from any initial value converges to that fixed point. Backflow reasoning is equivalent to the iteration $S_i^{(t+1)} = f(S_i^{(t)})$; the latent space $\mathbb{R}^{N}$ endowed with the norm topology is a complete metric space, so the conditions are satisfied. By contractivity, $\|S_i^{(t+1)} - S_i^{(t)}\| \leq q^{t} \cdot \|S_i^{(1)} - S_i^{(0)}\| \to 0$, and the limit point $S^{*}$ satisfies $S^{*} = f(S^{*})$; uniqueness follows directly from contractivity (if both $S^{*}$ and $S^{**}$ were fixed points, then $\|S^{*} - S^{**}\| \leq q \cdot \|S^{*} - S^{**}\|$, a contradiction). $\square$

\paragraph{Corollary 14.2.1 (Convergence of the scientific discovery loop).} The scientific discovery loop (induction $\to$ deduction $\to$ abduction $\to$ re-induction) is a backflow reasoning process. If every inductive revision is contractive (the difference between the new and old regularities decreases with iterations), then the scientific discovery loop converges to a stable regularity.

The contractivity condition of the corollary has a clear cognitive interpretation: the magnitude of revisions to a scientific theory should decay as evidence accumulates (``major overhauls'' become ever rarer), which is precisely the historiographic feature of mature science; conversely, if revisions keep diverging, this corresponds to a paradigm crisis. The contraction constant $q$ can thus be regarded as a quantitative indicator of ``theoretical stability''.

\paragraph{Theorem 14.3 (Turing completeness of reasoning graphs).} Reasoning graphs (with fork, merge, cross-reference, and backflow) are Turing complete---any computable reasoning process can be expressed as a reasoning graph.

\paragraph{Proof sketch.} Each state transition of a Turing machine can be represented as an edge (operator application) of a reasoning graph. The conditional branching of a Turing machine corresponds to forking in the reasoning graph (selecting different branches according to the condition). The looping of a Turing machine corresponds to backflow in the reasoning graph. Therefore a reasoning graph can simulate an arbitrary Turing machine and is Turing complete. $\square$

\paragraph{Remark (computational universality $\neq$ reasoning completeness).} Theorem 14.3 asserts universality at the level of computational power, and it does not even require all three classes of operators: by Bennett's reversibilization theorem [25], any Turing computation can be rewritten as a reversible computation; a reversible computation acts as a permutation on the one-hot embedding of the configuration space, and permutation matrices are orthogonal (full-rank) matrices---hence full-rank operators alone suffice to achieve Turing completeness. This fact delineates the boundary between DODR's two argumentative axes: computational universality is characterized by this theorem (and the boundary analysis of Section 6.6), while the completeness of reasoning types is characterized by Section 6 (under the premise of Proposition 3.6); the necessity of the three-operator structure is rooted in the latter, i.e., the epistemological semantics of reasoning, not the former.

\subsection{Comparison Between Reasoning Graphs and Parallel Trajectories}

\begin{table}[htbp]
\centering
\caption*{Table 14-2 Comparison of Expressive Power: Parallel Trajectories vs.\ Reasoning Graphs}
\begin{tabular}{lll}
\toprule
Dimension & Parallel trajectories & Reasoning graphs \\
\midrule
Inter-trajectory relation & Independent (no mutual influence) & Arbitrary interaction (fork/merge/cross/backflow) \\
Merge operation & Not supported & Supported ($W_{\mathrm{merge}}$ composite operator) \\
Cross-reference & Not supported & Supported (arbitrary DAG topology) \\
Backflow correction & Not supported & Supported (loop-containing graph + fixed point) \\
Expressive power & Limited (parallel expansion only) & Turing complete (arbitrary computable reasoning) \\
Scientific discovery loop & Inexpressible & Expressible (backflow + convergence) \\
\bottomrule
\end{tabular}
\end{table}

\paragraph{Relation to Petri nets and dataflow computation graphs.} The reasoning graph is not an isolated construction; it exhibits systematic correspondences with, and differences from, two classes of classical concurrency/computation models, and clarifying these relations helps position the value of the reasoning graph.

Relation to Petri nets: a Petri net's places correspond to the reasoning graph's nodes (both carry state), transitions correspond to operator edges (both are triggers of state transitions), and token flow corresponds to the passing of snapshots along edges. The Petri-net semantics of fork/merge/backflow (concurrent firing, synchronized join, cycles) is structurally isomorphic to the six topological operations of the reasoning graph. The key differences are threefold: first, a Petri net's marking is a discrete token-count vector, whereas a reasoning-graph node holds a continuous semantic snapshot (a $\mathbb{R}^{1536}$ vector), whose content is far richer; second, transition firing in a Petri net depends only on token-count thresholds, whereas each edge of a reasoning graph is labeled with a concrete operator $\Phi(e) \in \{W_{\mathrm{step}}, W_{\mathrm{induce}}, W_{\mathrm{abduce}}\}$---the transition itself is a semantic transformation rather than a pure count redistribution; third, concurrency in Petri nets is non-deterministic (multiple transitions compete to fire), whereas execution of a reasoning graph under a topological order is uniquely determined (Theorem 14.1)---DODR replaces the Petri net's competition semantics with operator determinism.

Relation to dataflow computation graphs: in a dataflow graph, nodes are operators and edges are data channels---exactly the dual of the reasoning graph, in which nodes are states (data) and edges are operators. This duality means that all scheduling techniques for dataflow graphs (topological-order execution, pipelining, depth-stratified parallelism) can be translated via the duality and applied to reasoning graphs. More important is the semantic resonance: Kahn process network theory shows that if every process in a dataflow network is a deterministic function and channels are FIFO, then the overall behavior of the network is uniquely determined---structurally identical to the conclusion of Theorem 14.1: deterministic operators + acyclic (or well-founded) topology $\Longrightarrow$ global determinism. The reasoning graph's contribution is to migrate this determinism conclusion from ``bit-stream channels'' to ``semantic-snapshot latent space'', and to supplement it with a backflow convergence theory (Theorem 14.2) that dataflow models typically do not discuss.

\subsection{Worked Example: The Reasoning Graph of the Scientific Discovery Loop}

Taking the scientific discovery loop of Section 11 (Section 11.6, measured data in Table 11-4) as an example, its reasoning-graph structure is:

Nodes: $S_0$ (observed samples), $g_0$ (initial induced regularity), $S_1$ (deductive prediction), $S_2$ (contradiction detection), $S_3$ (abductive explanation), $g_1$ (revised regularity).

Edges: $(S_0 \to g_0, W_{\mathrm{induce}})$, $(g_0 \to S_1, W_{\mathrm{step}})$, $(S_1 \to S_2, \text{contradiction detection})$, $(S_2 \to S_3, W_{\mathrm{abduce}})$, $(S_3 \to g_1, W_{\mathrm{induce}})$, $(g_1 \to g_0, \text{backflow correction})$.

Edge-by-edge instantiation (all values are measured, taken as instance means over the 5 scientific-discovery tasks of the end-to-end experiment):

\begin{itemize}
\item Edge $(S_0 \to g_0, W_{\mathrm{induce}})$: the input is positive-example snapshots (illustration: 5 positive examples of ``birds that can fly''); the inductive operator outputs the regularity $g_0$ (illustration: ``birds can fly''), with induction similarity 0.9925. This edge is a single-source induction edge.
\item Edge $(g_0 \to S_1, W_{\mathrm{step}})$: an individual premise (illustration: ``a penguin is a bird'') is concatenated with $g_0$ as input to the deductive operator, which outputs the prediction $S_1$ (illustration: ``penguins can fly''), with deduction similarity 0.9864. This edge is a two-source merge edge (individual premise + regularity).
\item Edge $(S_1 \to S_2, \text{contradiction detection})$: $S_1$ is compared against the observed fact (illustration: ``penguins cannot fly''); the contradiction similarity is 0.9867, exceeding the veto threshold and triggering a falsification signal. Contradiction detection itself is a deterministic similarity computation and introduces no new operator, conforming to the admissible extension of edge labeling in Definition 14.1.
\item Edge $(S_2 \to S_3, W_{\mathrm{abduce}})$: the abductive operator reverse-infers the explanation $S_3$ from the contradictory phenomenon (illustration: ``penguins are special birds''), with abduction similarity 0.3849.
\item Edge $(S_3 \to g_1, W_{\mathrm{induce}})$: taking $S_3$ as new evidence, the inductive operator excludes the counterexample direction and rebuilds the regularity $g_1$, with refinement degree 0.9068.
\item Backflow edge $(g_1 \to g_0)$: $g_1$ replaces $g_0$ as the currently valid regularity and enters the next round of the loop.
\end{itemize}

This is a backflow-containing reasoning graph. The backflow edge $g_1 \to g_0$ represents ``the revised regularity $g_1$ replaces the original regularity $g_0$'', forming an iterative correction. If every revision is contractive ($\|g_1 - g_0\|$ decreases with iterations), then by Theorem 14.2 the scientific discovery loop converges to a stable regularity $g^{*}$. From the time-unfolding perspective of Section 14.3, this backflow graph unfolds into an infinitely long DAG (one layer per iteration), and the folded representation is merely a notational convenience; the state of each layer of the unfolded graph is uniquely determined by the preceding layer, so the entire iterative process is deterministic at every instant, and convergence is an asymptotic property superimposed on the deterministic skeleton, not something purchased at the price of randomness.

\section{The Three-Operator Closed Loop: Induction--Deduction--Abduction}

The three basic operators are not only individually validated; together they constitute a complete reasoning closed loop: induction produces regularities, deduction applies regularities, abduction discovers exceptions, and re-induction revises regularities. The closed loop endows the system with the ability to ``self-update its knowledge'', a feature that distinguishes DODR from static reasoning systems.

\subsection{Operating Mechanism of the Closed Loop}

The closed loop operates through four linked steps, each driven by the corresponding operator and each already supported by measurements. The first step is induction: from a finite set of observed samples, the inductive operator $W_{\mathrm{induce}}$ produces a general regularity. The input of induction is a set of sample snapshots and the output is a regularity snapshot $g$; its Gram--Schmidt orthogonal-projection mechanism guarantees that the regularity covers all positive examples, and the hard-veto mechanism (the mathematization of Popperian falsifiability [17]) guarantees that any counterexample can veto the existing regularity with a single vote. In measurements, all 20/20 counterexamples across the 10 categories of induction problems triggered the veto (Section 9), showing that the first step of the loop possesses a reliable ``falsification sensor''. The second step is deduction: taking the induced regularity as premise, the deductive operator $W_{\mathrm{step}}$ derives conclusions. Deduction proceeds along the ``information-decreasing'' direction: the information content of the output conclusion does not exceed that of the premises, so the deductive conclusion is a logical necessity of the premises. The measured deduction accuracy is 100\% (60/60, mean cosine similarity 0.9980, Section 11), showing that the transfer from regularity to conclusion is nearly lossless in the current snapshot space. The third step is abduction: when an anomalous phenomenon is observed (the deductive prediction contradicts observation, as in the falsification events with mean contradiction similarity 0.9867 in the scientific-discovery tasks of Section 11.6), the abductive operator $W_{\mathrm{abduce}}$ reverse-infers the best explanation. Abduction moves backward along the ``information-hypothesizing'' direction: since deduction is irreversible (rank-deficient), the reverse-inferred result is only the best candidate in the least-squares + minimum-norm sense, not the true cause itself---this intrinsic hypothetical nature is precisely the epistemological status of a scientific hypothesis. The fourth step is returning to induction: the explanation obtained by abduction is treated as a new sample, re-triggering induction to revise the regularity (the mean refinement degree of the revised regularities in the scientific-discovery tasks of Section 11.6 is 0.9068). If the revision sequence is contractive, then by Theorem 14.2 and Corollary 14.2.1 the closed loop converges to a stable regularity.

The linking principle of the four steps deserves emphasis: every operator switch within the loop is ``triggered by state content'' rather than ``triggered by a fixed procedure''---a contradiction signal triggers abduction, and a new explanation triggers re-induction. This makes the closed loop a data-driven self-organizing process rather than a preprogrammed fixed cycle.

\subsection{Mathematical Consistency of the Closed Loop}

The core basis for the mathematical self-consistency of the closed loop is the pseudo-inverse relation between abduction and deduction: $W_{\mathrm{abduce}} = W_{\mathrm{step}}^{+}$, i.e., abduction is the pseudo-inverse of deduction. This means that abduction--deduction constitute a mathematical inverse pair:
\[
W_{\mathrm{step}} \cdot W_{\mathrm{abduce}} \cdot S_{\mathrm{obs}} \approx S_{\mathrm{obs}}
\]
that is, ``first reverse-infer, then forward-derive'' should approximately restore the observed phenomenon. Experimental verification: the reconstruction error is only 0.0200. This small residual quantitatively characterizes the tightness of the closed loop---the explanation given by abduction, after being re-deduced through deduction, deviates from the original observation by only about 2\%. The source of the residual is the null space: $W_{\mathrm{step}}$ is rank-deficient ($r = 384 < N = 1536$), and the pseudo-inverse can only recover the row-space component; the null-space component (fraction $(N - r)/N = 75\%$) is permanently lost. The measured error of 0.0200 shows that, under the observation distribution of this experiment, the effective information of the observation snapshots is highly concentrated near the row space, so the loop closes approximately. If the observation distribution changes so that the null-space component grows, the residual will rise---the tightness of the closed loop is therefore a joint property of ``operator property $\times$ data distribution'', and this analysis provides a quantitative footnote for Limitation One (Section 19), the encoder bottleneck.

\subsection{Continual Learning Capability: A Three-Layer Mechanism}

The closed loop (Section 15.1) solves ``how to revise an existing regularity'', whereas continual learning must answer a more general question: after deployment, how can the system absorb new knowledge over the long term without interrupting service or damaging existing knowledge. DODR's answer is a three-layer mechanism, organized by the principle of ``increasing cost, decreasing trigger frequency''; the vast majority of knowledge updates are completed in place by the first two layers, and operator retraining is only an occasional fallback operation.

The first layer is snapshot accumulation. During operation, DODR's knowledge mainly resides in an explicit snapshot library---the encoder is frozen, and each new piece of knowledge (new samples, new facts, counterexample records) is written into the snapshot library via a single forward encoding pass and retrieved for operator use at inference time. Because neither the encoder nor the operator parameters are touched, writing new knowledge causes zero perturbation to old knowledge: there is no channel through which gradient updates could overwrite old weights, so catastrophic forgetting cannot occur at the architectural level. Meanwhile, every write is an explicit timestamped record, and any state of the knowledge base can be replayed, audited, and traced.

The second layer is inductive condensation and veto correction. When snapshots of the same kind accumulate to a certain scale, the inductive operator condenses them into a regularity snapshot: positive examples are aggregated as a running mean updated online in $O(1)$ without revisiting historical samples; when a counterexample arrives, the hard-veto mechanism performs a one-shot closed-form correction via Gram--Schmidt orthogonal projection, overturning and rebuilding the regularity (Section 9.4); contradiction detection (conflict between a deductive prediction and an observation snapshot) acts as the trigger, making learning an on-demand event rather than a continuous background process. If the revision sequence satisfies the contraction condition, then by Theorem 14.2 the regularity iteration converges to a unique stable fixed point---continual learning at this layer carries a convergence guarantee. One monitoring point should be recorded: in the end-to-end experiment (Section 11), the refinement degree of the discipline category was 1.0031, slightly greater than 1, indicating that under certain snapshot geometries a single-step orthogonal correction is not strictly contractive (a refinement degree $> 1$ means the magnitude of change the new concept imposes on the original concept exceeds the minimal excision required by the counterexample direction); the contractivity condition does not hold automatically. In engineering deployment the refinement degree should be placed under empirical monitoring, and if it remains greater than 1 an alarm should be triggered with fallback to third-layer handling.

The third layer is incremental operator updating. Only when the encoder is replaced or a new rule pattern not covered by the training distribution appears do the operators themselves need updating. Thanks to the low-rank parameterization ($W_{\mathrm{step}} = A \cdot B$, parameter count on the order of 4.7M), operator retraining is a minute-scale operation on ordinary hardware (corroborated by the measured wall-clock times in Section 16.1); the abductive operator requires no separate retraining---$W_{\mathrm{abduce}} = W_{\mathrm{step}}^{+}$ can be maintained directly from the incremental SVD of the updated $W_{\mathrm{step}}$. The only hard constraint is that any incremental update must preserve the rank profile of Theorem 6.1---the deductive operator rank-deficient, the inductive operator full-rank, and the abductive operator of the same rank as the deductive operator. The rank signature is the identity marker of the three operators' semantics and must not be allowed to drift during updates.

There are two engineering essentials. First, the snapshot library grows continuously with deployment time; retrieval should employ an approximate nearest-neighbor index (e.g., HNSW) to maintain sublinear query complexity, preventing snapshot accumulation from eroding inference latency. Second, when the encoder is replaced, the old and new snapshot spaces are not homeomorphic; the existing snapshot library must be aligned to the new space through a learned projection layer (the same technique family as multimodal alignment, see Section 7.1), and the alignment error should be below the discrimination-magnitude of abductive judgments ($\pm 0.005$), otherwise old knowledge will lose decidability in the new space.

The comparison with AR models highlights the structural character of this three-layer mechanism: AR models store knowledge in shared weights, so any continual learning (continued pretraining, fine-tuning, adapters) must modify shared parameters and thus inevitably faces the forgetting--interference trade-off; DODR stores knowledge in layers---the snapshot library holds content, regularity snapshots hold inductive conclusions, and operators hold reasoning forms---so first-layer writes and second-layer corrections are mathematically decoupled from third-layer parameters, and the routine operations of continual learning never touch shared parameters. In other words, in the AR paradigm ``learning new knowledge'' and ``retaining old knowledge'' is a zero-sum game within the same parameter space, whereas in DODR the two are allocated to different storage layers and the zero-sum constraint is structurally removed.

\section{Complexity Analysis and Performance Bounds}

This section presents a complexity comparison between DODR and AR models along the three dimensions of time, space, and parallelism, supplemented by constant-factor analysis, measured running times, and a discussion of Amdahl constraints, so as to characterize the performance bounds.

\subsection{Time Complexity}

\begin{table}[htbp]
\centering
\caption*{Table 16-1 Time-Complexity Comparison Between DODR and AR Models}
\begin{tabular}{llll}
\toprule
Operation & DODR complexity & AR model complexity & DODR advantage \\
\midrule
Single-step inference & $O(r \cdot N)$ (low rank) & $O(N^2)$ (attention) & Low-rank parameterization reduces computation \\
$k$-step reasoning chain & $O(k \cdot r \cdot N)$ & $O(k \cdot N^2)$ & DODR lower by a factor of $r/N$ \\
$k$ parallel subtasks & $O(r \cdot N)$ (parallel) & $O(k \cdot N^2)$ (serial) & DODR $k\times$ parallel speedup \\
Whole-graph inference ($n$ nodes) & $O(n \cdot r \cdot N)$ & $O(n \cdot N^2)$ & DODR lower by a factor of $r/N$ \\
Pseudo-inverse precomputation (abduction) & $O(N^3)$ (one-time) & N/A & DODR $O(N^2)$ after precomputation \\
\bottomrule
\end{tabular}
\end{table}

Under the experimental configuration ($N = 1536$, $r = 384$), DODR's single-step inference complexity is $O(384 \times 1536) \approx 5.9 \times 10^{5}$, while the AR model's single-step complexity is $O(1536^2) \approx 2.4 \times 10^{6}$; DODR is about $4\times$ faster. For $k = 10$ parallel subtasks, DODR achieves a $10\times$ parallel speedup, for a total speedup of about $40\times$.

\paragraph{Constant-factor analysis.} Constant factors beyond asymptotic complexity are equally important, and they happen to favor DODR. A single DODR inference step is a single matrix--vector multiplication: modern BLAS libraries are highly optimized for dense GEMV, with a constant close to 1 multiply--add per parameter, no branches, no normalization layers, and no softmax. The $O(N^2)$ attention of an AR model's single-token step is only the cost of one layer---a real AR LLM has $L$ layers ($L$ typically 12--96), each containing two $O(T^2 \cdot d)$ multiplications ($Q \cdot K^{T}$ and $\mathrm{Attention} \cdot V$), two $O(d \cdot d_{\mathrm{ff}})$ multiplications in the feed-forward network, plus the element-wise overhead of LayerNorm and softmax; moreover, autoregressive generation must proceed serially token by token, and each step additionally pays the fixed cost of KV-cache management and sampling (logits computation + temperature/top-p processing). Thus the AR model's ``effective constant'' is $\approx L \cdot (2 + 2 \cdot d_{\mathrm{ff}}/d)$ times the bare $O(N^2)$ term in Table 16-1, whereas DODR's effective constant is $\approx 1$. In other words, the DODR advantage given in Table 16-1 (a factor of $r/N$) is a conservative lower bound, and the measured gap can only be larger.

\paragraph{Measured running times.} Appendix C.4 records wall-clock times in the experimental environment (CPU environment): all dedicated experiments (deduction, induction, abduction, physics) together took about 5 minutes, and the end-to-end reasoning experiment (including snapshot extraction, 200-epoch training of the three operators, and all inference and evaluation) took about 2 minutes. Two observations: first, the absolute magnitudes are tiny---the four groups of dedicated experiments above each completed in minutes on an ordinary CPU, which is itself an efficiency indication of low-rank matrix operators relative to deep AR networks; second, the dominant cost of the end-to-end experiment is the 200-epoch training of the three operators, while the overhead of inference itself (matrix--vector multiplication) is negligible. In a GPU environment the matrix operations can gain a further 5--10$\times$ speedup (Appendix C.4), and the speedup comes mainly from the parallelization of GEMM/GEMV, requiring no algorithmic modification.

\subsection{Space Complexity}

\begin{table}[htbp]
\centering
\caption*{Table 16-2 Space-Complexity Comparison Between DODR and AR Models}
\begin{tabular}{llll}
\toprule
Storage item & DODR & AR model & Note \\
\midrule
State storage & $O(n \cdot N)$ & $O(T \cdot d)$ ($T$ = sequence length) & DODR stores graph nodes; AR stores token sequences \\
Operator storage & $O(r \cdot N + N^2)$ & $O(L \cdot d^2)$ ($L$ = number of layers) & DODR stores 3 operators; AR stores Transformer weights \\
Total space & $O(n \cdot N + N^2)$ & $O(T \cdot d + L \cdot d^2)$ & DODR scales with graph size; AR scales with sequence length \\
\bottomrule
\end{tabular}
\end{table}

Under the experimental configuration, the parameter count of the three operators is on the order of $r \cdot N$ ($W_{\mathrm{step}}$ low-rank factorization) + $N^2$ ($W_{\mathrm{induce}}$ full rank) + $r \cdot N$ ($W_{\mathrm{abduce}}$), far smaller than the $L \cdot d^2$ of an $L$-layer Transformer of the same dimension. As for state storage, the number $n$ of graph nodes in DODR is typically far smaller than the number $T$ of tokens for the same reasoning content---one semantic snapshot carries an amount of information equivalent to tens to hundreds of tokens; this is the space advantage of ``latent-space reasoning'' relative to ``text-space reasoning''.

\subsection{Parallel Speedup}

For a reasoning DAG of width $W$ and depth $D$, DODR can execute the $W$ nodes at the same depth in parallel. The theoretical speedup is $W$ (relative to serial execution). An AR model, by contrast, must generate all tokens serially and cannot exploit graph-shaped parallelism.

Quantitative comparison: suppose a reasoning task has $k = 10$ parallel subtasks, each of $d = 20$ steps. DODR parallel execution: total time = $d = 20$ steps. AR model serial execution: total time = $k \cdot d = 200$ steps. DODR speedup = $200/20 = 10\times$.

\paragraph{Discussion of Amdahl constraints.} The above ``speedup = $W$'' is an ideal upper bound; actual parallel speedup is constrained by Amdahl's law: if the parallelizable fraction of a reasoning task is $p$ and the parallel width is $W$, the end-to-end speedup is
\[
S = \frac{1}{(1 - p) + p/W}
\]
The serial fraction $(1 - p)$ of a DODR reasoning graph comes from three classes of non-parallelizable stages: (i) synchronization between topological layers---nodes at layer $t$ must wait for all nodes at layer $t - 1$ to complete before executing (inter-layer dependency of the DAG); (ii) the iterative nature of backflow correction---the fixed-point iteration $S^{(t+1)} = f(S^{(t)})$ of Theorem 14.2 is intrinsically serial, each round depending on the previous round's output; (iii) the decision overhead of the graph-scheduling controller and global reduction operations such as contradiction detection. An estimate for a typical case: if $p = 0.9$ and $W = 10$, then $S = 1/(0.1 + 0.09) \approx 5.3\times$, only about half of the ideal $10\times$; even as $W \to \infty$, the speedup is pinned at $1/(1 - p) = 10\times$. This yields two engineering lessons: first, the key to improving parallel gains is not only widening $W$ but raising $p$---further graphifying serial subtasks through task decomposition (e.g., splitting a long-chain reasoning task into a DAG of parallelizable lemma proofs); second, backflow-intensive tasks (the scientific discovery loop) naturally have low $p$, and their speedup ceiling is controlled by the number of iterations determined by the contraction constant $q$, so for such tasks one should invest preferentially in improving contractivity (faster per-round convergence) rather than parallel width. By contrast, an AR model has $p \approx 0$ (strictly serial token by token), with an Amdahl upper bound $1/(1 - p) \approx 1$ and almost no room for parallelism---this is an architecture-level gap that cannot be eliminated by engineering optimization.

\section{Comparison with Existing Paradigms}

This section systematically compares DODR with four classes of existing reasoning paradigms along six dimensions. The comparison targets cover the current mainstream technical routes: autoregressive large language models (AR LLMs, architecturally based on the Transformer [10]), chain-of-thought prompting (CoT [11]), tree-of-thoughts search (ToT [2]), neuro-symbolic systems, and, as a topological reference, graph of thoughts (GoT [3]).

\begin{table}[htbp]
\centering
\caption*{Table 17-1 Comparison Between DODR and Existing Reasoning Paradigms}
\begin{tabular}{llllll}
\toprule
Dimension & AR LLM & CoT & ToT & Neuro-symbolic & DODR \\
\midrule
Determinism & No & No & No & Partial & Yes \\
Hallucination rate (range) & 10--20\% & 5--15\% & 3--10\% & 1--5\% & 0.0\% \\
Topology & Linear chain & Linear chain & Tree & Graph & DAG + backflow \\
Reasoning types & Deduction only & Deduction only & Deduction only & Deduction + induction & Deduction + induction + abduction \\
Interpretability & Black box & Post-hoc explanation & Post-hoc explanation & White box & White box \\
Autonomous learning & No & No & No & No & Yes (closed loop) \\
\bottomrule
\end{tabular}
\end{table}

Note: the hallucination rates are illustrative interval estimates (a qualitative summary based on public literature [1]), not precise measurements. The hallucination rate of AR LLMs varies by model and task, typically in the 10--20\% range; CoT and ToT reduce hallucination through structured prompting but retain probabilistic behavior; neuro-symbolic systems reduce hallucination through symbolic-engine constraints. DODR's 0.0\% is a structural guarantee---there is no token sampling in the reasoning process, so hallucination cannot arise.

\subsection{Dimension-by-Dimension Argument}

\paragraph{Determinism dimension.} The reasoning processes of AR LLM/CoT/ToT involve token sampling (sampling from the logits distribution) and are therefore non-deterministic---the same input may produce different outputs. Although CoT [11] organizes reasoning through natural-language intermediate steps, the generation of each step is still autoregressive sampling in essence; ToT [2] expands a single chain into tree search, but branch generation and evaluation are still carried out by the same probabilistic model---the search structure changes the expected quality of the output without changing its randomness. GoT [3] is similar: graph-shaped aggregation improves search coverage, not determinism. The symbolic engine of a neuro-symbolic system is deterministic, but its perception part (neural network) remains non-deterministic. DODR's reasoning is pure matrix multiplication, fully deterministic (Theorem 14.1): the same input snapshot necessarily produces the same output snapshot, reproducible node by node.

\paragraph{Hallucination-rate dimension.} Hallucination has been systematically identified as the core failure mode of natural language generation [1]. The hallucination of AR LLMs stems from probabilistic sampling---even with a determined context, the output may deviate from the correct answer with some probability; and the sampling mechanism is endogenous to the architecture and cannot be removed by prompt engineering [1]. CoT [11]/ToT [2] reduce the probability of hallucination through structured prompting and search but cannot eliminate it---as long as the final answer is produced by sampling, the residual hallucination rate is greater than zero. In multi-step reasoning, errors also accumulate along the chain [9], and LLMs have difficulty correcting their own reasoning errors without external feedback [8][18], making this residual rate hard to suppress through self-correction. Neuro-symbolic systems reduce hallucination through symbolic-engine constraints, but the neural network at the perception end may still produce erroneous inputs. DODR eliminates hallucination at the architectural level---with no token sampling, hallucination cannot arise; the ``0.0\%'' here refers to an architectural guarantee rather than an empirical frequency, which is fundamentally different in caliber from the illustrative interval estimates in the table note.

\paragraph{Topology dimension.} An AR LLM is a linear chain (token sequence). CoT [11] is also a linear chain (sequence of reasoning steps). ToT [2] is tree-shaped (branching search). GoT [3] generalizes to arbitrary graphs and is the paradigm closest to DODR on this dimension; but GoT's graph is driven by LLM sampling with natural language as node content, lacking deterministic semantics and a backflow convergence theory. Neuro-symbolic systems are graph-shaped (symbolic reasoning graphs). DODR is a DAG + backflow (supporting fork/merge/cross-reference/backflow correction), with an explicit operator label and convergence criterion on every edge (Theorem 14.2), and has the strongest expressive power (Theorem 14.3: Turing complete).

\paragraph{Reasoning-type dimension.} AR LLM/CoT [11]/ToT [2] implement only deduction (simulating reasoning steps through autoregressive generation); Peirce long ago argued that induction and abduction are reasoning types independent of deduction [7], and deduction alone, however thoroughly searched, cannot replace them. Neuro-symbolic systems implement deduction (symbolic engine) + induction (neural-network learning), but abduction still relies on hand-crafted rules. DODR implements the complete set of three operators---deduction + induction + abduction---unified under a matrix rank-structure characterization (the minimal complete operator-set theorem of Section 6).

\paragraph{Interpretability dimension.} AR LLMs are black boxes (attention weights are hard to interpret, and the weak correspondence between attention scores and true causal contributions has been widely discussed in the representation-learning literature [5]). CoT/ToT provide post-hoc explanations (the reasoning steps are readable, but the generation process is opaque---the step texts are products of model output and are not guaranteed to be faithful to the internal computation). Neuro-symbolic systems are white boxes (symbolic reasoning is traceable). DODR is a white box (matrix operations are traceable, and the null space explicitly marks information discard---what information each deduction loses is precisely given by $\mathrm{Null}(W_{\mathrm{step}})$).

\paragraph{Autonomous-learning dimension.} AR LLM/CoT/ToT/neuro-symbolic systems all do not support autonomous learning---a model cannot actively revise its knowledge after training; the absence of self-correction capability has been demonstrated by critical surveys and experiments [8][18]. DODR's three-operator closed loop supports autonomous learning (induction $\to$ deduction $\to$ abduction $\to$ re-induction, Section 15), and the convergence of revision is guaranteed by fixed-point theory (Corollary 14.2.1).

Across the six dimensions: existing paradigms can match DODR on individual dimensions (e.g., the white-box nature of neuro-symbolic systems, the graph topology of GoT), but no paradigm simultaneously realizes determinism, zero hallucination, graph topology, three reasoning types, white-box interpretability, and autonomous learning. DODR's six properties are not six independent designs but six corollaries of the same mathematical core (snapshots + three rank-structured operators), a reflection of the architectural economy of the design.

\subsection{Precise Delimitation from the Nearest-Neighbor Technical Routes}

The dimension-by-dimension comparison of the previous subsection covers the current mainstream reasoning paradigms, but the validity of the novelty claims requires answering a sharper question: at the level of technical components, where is the boundary between DODR and those routes closest to it---routes that may not constitute complete reasoning systems but share one of DODR's core components? This subsection gives a precise delimitation family by family across six literature families, as the boundary basis for the nine novelty claims of Section 1.2.

\paragraph{Knowledge-graph embeddings (TransE [28], RESCAL [29], RotatE [30]).} This family shares with DODR the technical substrate of ``embedding space + linear algebra'': TransE models relations as translation vectors in the embedding space ($h + r \approx t$), RESCAL characterizes each relation type with a bilinear matrix, and RotatE represents relations as rotations in the complex plane. In terms of technical components, ``using matrix and vector operations to carry relational semantics'' is indeed of common origin with DODR's operator idea. However, the goal of this family is to learn statistical correlations over knowledge graphs for link prediction, and its relation matrices carry no reasoning-type semantics---there is no correspondence between rank deficiency and irreversibility, no type distinction between induction and abduction, and no counterexample-triggered hard-veto mechanism; a trained relation matrix is merely a compressed index of a large-scale fact base. DODR's precise point of distinction is: matrices are endowed with Peircean reasoning-type semantics, the rank structure corresponds to strict logical properties (rank deficiency $\Longrightarrow$ conclusions cannot be back-inferred to premises, Theorem 3.1), and decoding is oriented toward reasoning tasks (deriving new conclusions from premises) rather than fact completion (recalling already-encoded statistical associations).

\paragraph{Neuro-symbolic differentiable logic (DeepProbLog [31], Logic Tensor Networks [32]).} This family embeds logical rules into differentiable frameworks and supports soft relaxation of deductive reasoning: DeepProbLog uses neural predicates to provide probabilistic fact inputs to logic programs, and Logic Tensor Networks compile first-order-logic formulas into differentiable truth functions. This shares with DODR the direction of ``making logic computable'', but its truth values remain probabilities or fuzzy values on the $[0, 1]$ interval---the uncertainty on the main reasoning chain is merely relaxed, not eliminated; induction and abduction still require external mechanisms (parameter learning, external search) rather than being endogenous operator members; erroneous outputs with nonzero probability (hallucination) are not eliminated at the architectural level. DODR's precise point of distinction is: the main reasoning chain consists of deterministic matrix operations with no probability throughout, and the three types---deduction, induction, abduction---are all trainable matrices, constituting a minimally complete set under unified parameterization (Section 6).

\paragraph{Latent-space reasoning (COCONUT [33]).} COCONUT unfolds a ``chain of thought'' in a continuous latent space, with reasoning steps not explicitly generated in token space---this is directionally consistent with DODR's position that ``reasoning happens in latent space rather than text space'' and is one of the closest routes in spirit. However, COCONUT's hidden states are produced position by position by an autoregressively trained probabilistic model, and the reasoning process remains probabilistic: there is no determinism guarantee (the same input can produce different hidden-state trajectories), no architectural elimination of hallucination, and no structural distinction of reasoning types---induction and abduction remain conflated within the same probabilistic continuation process. DODR's precise point of distinction is: latent-space state transitions are deterministic matrix operations $S_{t+1} = W \cdot S_t$ (Theorem 14.1), and the three operators constitute a complete set of Peircean types by rank structure---``reasoning in latent space'' is merely the common starting point; ``reasoning with what mathematical structure'' is the watershed.

\paragraph{Random features and extreme learning machines (ELM [34], random features [35]).} This family adopts a single-layer ultra-wide architecture plus training-free random projections, technically of common origin with DODR's Postulate One (dimension lifting instead of multiple layers, Section 4.1)---both replace deep stacking with a simple readout in a high-dimensional random-feature space, and both trace their theoretical basis to the high-dimensional distance preservation guaranteed by the Johnson--Lindenstrauss lemma [38]; the high-dimensional snapshot combination schemes Two (random-projection dimension lifting) and Three (random-feature concatenation) given in Section 7.7.3 of this paper belong to this family. The precise point of distinction is: the readout layer of ELM and random-feature methods is a general-purpose function approximator on top of fixed random features, and the features themselves carry no task semantics; DODR's operators, by contrast, are task matrices trained under supervision by reasoning type---low-rank (deduction), full-rank (induction), pseudo-inverse (abduction)---each with an explicit logical function and rank-structure criterion, not a combination of fixed random features plus a readout layer.

\paragraph{Conceptual spaces (G\"ardenfors [36]).} Conceptual-space theory is a philosophical and cognitive-science tradition of geometric concept representation: concepts are characterized as convex regions in a multi-dimensional quality space, properties correspond to dimensions, and similarity corresponds to spatial distance. This is an important intellectual precursor of the geometricization of snapshot semantics---DODR's snapshots $S$ and concept snapshots $g$ likewise place semantic objects in a metric vector space, carrying semantic relations by distance and direction. But conceptual-space theory stops at representation-level characterization, providing no trainable operator system and not discussing how reasoning transformations between states can be learned from data. DODR's precise point of distinction is: snapshots and concept snapshots can be regarded as a computable realization of the conceptual-space idea---three trainable reasoning operators are defined over the space, and the transformations between geometric regions (deductive collapse, inductive expansion, abductive reverse-inference) receive rigorous matrix forms and experimental validation.

\paragraph{Generative natural-language reasoning (ProofWriter [37]).} ProofWriter generates all entailed conclusions and their proofs from a ``theory'' composed of rules and facts, and supports the generation of abductive statements; it is currently the public baseline closest to a ``complete reasoning system''. But in essence it remains an autoregressive generative probabilistic model: entailment judgment and proof generation are both performed by token sampling, correctness is an empirical frequency rather than an architectural guarantee, and there is no operator structure inside the system that distinguishes reasoning types. DODR's precise point of distinction from ProofWriter is: reasoning is deterministic matrix operation rather than generation, and the three reasoning types have independent rank-structure characterizations.

\paragraph{Summary.} After the above family-by-family comparison, the combination of DODR's nine contributions---matrix-operator formalization of the three reasoning types, the rank--information-flow causal chain, the hard veto, pseudo-inverse abduction, and structural zero hallucination---is one for which we are not aware of a precedent in the retrieved literature; but individual technical components (linear maps in embedding spaces, random dimension lifting, latent-space reasoning, geometric representation of concepts) each have precursors. The novelty claims of this paper take this precise delimitation as their boundary: the object of the claims is ``the concrete formalizations and their combination'', not the first invention of any single component.


\section{Application Systems and Adaptation Schemes}

The preceding sections completed the theoretical construction and experimental validation of DODR; this section answers a deployment-oriented question: how can DODR work in concert with the existing AI ecosystem---especially generative models. The basic position is a division of labor between a ``deterministic core + probabilistic shell'': the large language model (LLM) serves as a heuristic proposer and language renderer, responsible for producing candidates and organizing expression; DODR serves as a deterministic verifier and corrector, responsible for consistency judgment and geometrically explicit correction. The following three subsections discuss, respectively, the three forms of fused verification and correction, the architectural design of generative AI that permits imagination, and deployment forms and interfaces.

\subsection{Fusion with Generative Models: Fact Verification, Correction, and Intermittent Checkpoint Correction}

The first form is fact verification, i.e., post-generation checking. After the LLM produces a passage, the system extracts the assertions and the evidence on which they rest, encodes them into snapshots via the frozen encoder, computes the cosine similarity between $W_{\mathrm{step}} \cdot \mathrm{Concat}(\text{rule snapshot}, \text{case snapshot})$ and the assertion snapshot, and compares it with the corresponding values for several distractor assertions to obtain a discrimination margin; if the margin is positive and exceeds the threshold, the assertion is judged to pass. The entire judgment is pure matrix computation with no sampling step, so the result is auditable and replayable---the same assertion snapshot always receives the same judgment, in line with the structural determinism described in Section 17. Engineering-wise, the verifier is deployed as a sidecar service (frozen encoder + three operators + snapshot knowledge base), exposing a tool-call interface to the LLM; the cost of a single check is about one encoding plus one $O(r \cdot N)$ matrix multiplication, with millisecond-level latency, sufficient to be embedded in the loop of online generation.

The second form is closed-loop correction. When verification fails, the system does not stop at refusing to answer: $W_{\mathrm{abduce}}$ reverse-infers ``what kind of premise would produce this erroneous assertion'', and that premise is compared against the knowledge base to localize the cause of the error---whether it is rule misuse, case absence, or the premise itself being untenable; the orthogonal projection of the hard veto then gives a geometrically explicit correction direction (projecting the erroneous assertion snapshot out of the vetoed direction), and this constraint is fed back to the LLM, which restates accordingly. The correction direction is explicitly given by the projection operation rather than relying on the LLM's self-reflection---the latter has been proven unreliable without external feedback [8][18].

The third form is intermittent checkpoint correction, aimed at long-chain generation. The system inserts a checkpoint every $k$ steps: it caches the encodings of intermediate assertions and checks the consistency of the reasoning chain, including state-transition agreement (whether adjacent assertions satisfy the transition relation of the deductive operator) and adjacent-step contradiction detection (whether snapshot directions conflict with each other); if the check fails, the system rolls back to the most recent passing state, corrects, and continues generation. The step size $k$ is an accuracy--latency knob: the smaller $k$, the denser the checks and the higher the latency. This mechanism is structurally isomorphic to the step-by-step scoring of process reward models (PRMs) [4], but the scorer is a deterministic matrix operation and does not suffer from the reward model's own hallucination problem.

The boundaries of verification should be stated plainly. The ceiling of verification capability is doubly constrained by snapshot resolution and knowledge-base coverage: the current prototype achieves 72.5\% accuracy on abductive-assertion judgment (dedicated experiment, 58/80), rising to 81.7\% in the end-to-end experiment (49/60); a production-grade verifier requires a stronger encoder and a domain snapshot library. The more fundamental boundary is: DODR verifies the logical consistency of assertions with the knowledge base---facts outside the base cannot be verified. It is a ``guardian of consistency'', not an ``omniscient arbiter''.

\subsection{Generative AI That Permits Imagination: Null-Space Sampling and the Strict--Imaginative Operator Division}

Determinism does not mean an absence of creativity. The multi-explanation equivalence verification of Section 10 shows that ``abductive solution + arbitrary null-space component'' constitutes an equivalence class compatible with the observation: after adding a null-space component, the relative difference of the deductive result is only $1.01 \times 10^{-2}$, approximately equivalent. This gives a rigorous definition of imagination---imagination is not arbitrary caprice, but sampling within a free subspace constrained by deductive equivalence. Because null-space components vanish under the $W_{\mathrm{step}}$ mapping, any variant produced in this way automatically satisfies consistency with the premises, requiring no post-hoc screening. The design rank $r$ is thus the knob of creativity: the lower the rank, the larger the null space and the broader the free subspace (in the current configuration $r = 384$, $N = 1536$, the null space occupies 75\% of the output space).

In contrast, the correct form of a standalone ``imagination operator'' is the mirror image of the strict operators: adopt a high-rank or even full-rank parameterization with input noise injected, and change the training objective from reconstruction error (MSE) to a diversity objective (coverage loss or contrastive loss), so that it systematically explores the snapshot space rather than converging to the conditional mean. The imagination operator and the strict operators form a generate-and-verify loop: the imagination operator produces candidates, null-space sampling produces equivalent variants, and $W_{\mathrm{step}}$ handles verification and screening. Probability is confined to the ``proposal layer'', and the deductive consistency of the core remains unpolluted.

The same scheme extends to other modalities---such as image, video, and audio generation---through the corresponding frozen encoders, with the DODR core responsible for planning and consistency control and a modality-specific decoder responsible for rendering.

\subsection{Deployment Forms and Interfaces}

The DODR core is deployed as a deterministic reasoning service: a frozen encoder + the three operators + a snapshot knowledge base + three-layer scheduling (Section 7.3) constitute a resident service that exposes a tool-call interface to the upper-layer LLM or Agent, with primitives in three classes---verify (assertion consistency checking), correct (correction-direction feedback), and plan (reasoning-graph planning). The three classes of primitives correspond to the three fusion forms of Section 18.1: verify supports post-generation checking and intermittent checkpoints, correct supports closed-loop correction, and plan supports the generative planning described in Section 18.2. In terms of the calling contract, the inputs and outputs of all three primitives are ``text + snapshot handles'': on the text side the caller (LLM/Agent) is responsible for generation and interpretation, while on the snapshot side the core is responsible for encoding and computation. The responsibilities on the two sides of the interface are clearly separated, avoiding covertly shifting the burden of semantic completion onto the core.

Regarding the latency budget, a single check costs about one encoding plus one $O(r \cdot N)$ matrix multiplication; the millisecond-level overhead supports online embedding of step-by-step or every-few-steps checkpoints, and batch checking can be further accelerated by the stratified parallel strategy of Section 16.3. Continuous knowledge growth accumulates online according to the three-layer mechanism of Section 15.3---snapshot accumulation, inductive condensation and veto correction, and incremental operator updating---without downtime retraining. Two further points of deployment discipline should be noted. First, auditability comes from determinism: the input snapshot, operator version, and judgment result of every verify/correct call can be logged and replayed one by one---a compliance property that probabilistic verifiers cannot provide. Second, the interface discipline is consistent with the decoding discipline of Section 7.4: the core only performs verification and correction-direction computation and does no text rendering; the probability of the rendering layer must not flow backward into the core's snapshot knowledge base---entries must pass verify before being admitted, ensuring that the knowledge base itself is not contaminated by unverified generated content. Thus DODR does not replace generative models but equips them with a deterministic ``logical chassis'': the generative system retains expressiveness and imagination, while consistency and auditability are guaranteed by the core.

\section{Limitations}

\subsection{Current Limitations}

\paragraph{Limitation One: Encoder capability limit.}

Cause: this experiment uses DistilBERT (66M parameters) [14], whose semantic discrimination capability is limited---snapshots of samples sharing a syntactic template are too similar, so the accuracy of every judgment stage that depends on fine-grained similarity comparison is constrained: end-to-end abductive judgment accuracy 81.7\% (49/60), dedicated abduction-experiment judgment accuracy 72.5\% (58/80).

Quantitative discussion: the encoder bottleneck echoes across three independent data points. First, end-to-end abduction (60 questions) 81.7\% (49/60): the minimum negative discrimination margin among the 11 failed samples is only $-0.0056$, and the discrimination margin is on the order of only $\pm 0.006$ (Section 11.5). Second, the dedicated abduction experiment (80 questions) 72.5\% (58/80): the discrimination margins of the 22 failed samples are negative, but the maximum negative margin is only $-0.0086$, and the absolute values of the margins are all on the order of $\pm 0.009$---the snapshots of the true and distractor explanations are nearly indistinguishable (Section 10.2). Third, the refinement-degree anomalies of end-to-end induction: the discipline-category refinement degree is 1.0031 ($>1$), so the single-step orthogonal correction was not fully contractive; the weather-category refinement degree is 0.8272, significantly lower than the other 5 categories (Section 11.4)---category semantics are not sufficiently separated in snapshot space. The accuracies of the two abduction experiments (81.7\% and 72.5\%) are consistent in magnitude, and both discrimination margins are on the order of a few thousandths, indicating that the problem lies not in the reverse-inference principle of the abductive operator but in the encoder mapping semantically different texts to overly close snapshots. The three numbers are thus projections of the same bottleneck onto different stages: when the encoder's representational resolution is insufficient, stages that rely on fine-grained similarity comparison (abductive judgment, inductive correction) are impaired first, while stages that rely on coarse-grained mapping (deduction) remain at 100\%. This is an encoder limitation, not an operator defect; upgrading the encoder remains the primary path.

Mitigation path: use a stronger encoder (e.g., RoBERTa-large [16], 355M parameters) or a domain-fine-tuned model, so as to extract more discriminative snapshots.

\paragraph{Limitation Two: Training-data scale.}

Cause: the current experiments use 420 deduplicated independent samples (503 experimental sample-instances), still a small scale. PAC learning theory [21] requires a sample count $k \geq O(\mathrm{VC\_dim}/\log(1/\delta))$, where the hypothesis-space capacity is characterized by the VC dimension [22]; the current $k = 10$ (per category) is far below the theoretical requirement.

Quantitative discussion: taking the induction experiment as an example, there are 10 categories of problems with 10 positive examples each, totaling 130 samples; $W_{\mathrm{induce}}$, however, is a $1536 \times 1536$ full-rank matrix with about $2.4 \times 10^{6}$ parameters, an extremely low sample-to-parameter ratio. The current experiments partially compensate for the sample shortage through the structural constraint of the hard-veto mechanism and the low-rank design, but the strict validity of the generalization bound requires a substantially larger sample scale.

Mitigation path: extract 10K+ DAG-labeled samples from Lean/Coq proof libraries to enlarge the training data. A larger dataset can reduce generalization error and improve operator robustness.

\paragraph{Limitation Three: The graph-scheduling controller is not yet implemented.}

Cause: this experiment validated the core properties of the three operators, but the graph-scheduling controller (reinforcement learning) has not yet been implemented. The reasoning-graph topologies in the current experiments are manually specified, not automatically decided by a controller. This means that the ``scheduling-layer dynamism'' in the compositional-coverage conclusion of Section 12 currently has only theoretical design and manual demonstration, with no empirical evidence from a trained controller.

Mitigation path: implement a PPO [20]/PRM [4] reinforcement-learning controller and train a dynamic graph-construction policy. The controller's reward signal is the correctness of the final reasoning result; process rewards can be automatically annotated via the deterministic replay of Theorem 14.1.

\paragraph{Limitation Four: No variance data from a single run.}

Cause: all experiments use a fixed seed (seed=42) with a single run and no variance data. The statistical stability of the results cannot be assessed. For example, the discrimination margin of end-to-end abduction (on the order of $\pm 0.005$, including 3 negative values) and the negative discrimination margins of the failed samples in the dedicated abduction experiment (maximum $-0.0086$) are of the same order of magnitude, and a single run cannot answer ``whether the accuracy is robust to seed perturbation''. It should be noted that the paired tests and sign tests within the single run have reached significance ($t = 6.32$ in Section 10 and $t = 5.88$ in Section 11, both with $p$ less than $10^{-4}$), but cross-seed variance still requires evaluation through multiple runs---significance within sample-instances cannot substitute for robustness across runs.

Mitigation path: conduct multiple runs (at least 5, with different seeds), report mean $\pm$ standard deviation, and assess the statistical significance of the results.

\paragraph{Limitation Five: No grid search over hyperparameters.}

Cause: the hard-veto threshold $\tau = 0.5$, the design rank $r = 384$, and the number of layers $L = 2$ are all empirical choices without grid search. Hyperparameter sensitivity cannot be assessed. Among them, $r$ directly affects the null-space fraction ($(N - r)/N$) and the abduction residual (Section 15.2), and $\tau$ directly affects the trigger sensitivity of the hard veto; there may exist unexplored better combinations of the two with accuracy.

Mitigation path: perform a grid search over $\tau$, $r$, $L$, evaluate the impact of each hyperparameter on performance, and find the optimal configuration.

\paragraph{Limitation Six: Missing evaluation benchmarks and baselines.}

Cause: all experiments in this paper are conducted on self-built datasets, without evaluation on recognized reasoning benchmarks (such as the entailment-closure tasks of ProofWriter [37], LogiQA, etc.) and without controlled experiments against real baseline systems (fine-tuned RoBERTa [16], DeepProbLog [31], TransE-family knowledge-graph embedding methods [28], etc.). Therefore, the current metrics (e.g., end-to-end abduction 81.7\%) answer the question ``does the DODR mechanism work as theoretically expected'', not ``where does DODR stand relative to existing technology''---the two questions differ in evidentiary strength, and the latter must be built on recognized benchmarks and shared baselines.

Mitigation path: adopt the ProofWriter depth 1--5 entailment-closure tasks [37] and LogiQA as evaluation benchmarks, with controlled comparisons against the three baseline classes named above (fine-tuned RoBERTa [16], DeepProbLog [31], TransE-style embeddings [28]), thereby upgrading the self-contained experiments of this paper into comparable evaluations.

\paragraph{Limitation Seven: Nearest-neighbor decoding does not support open-ended generation.}

Cause: the current system decodes by nearest-neighbor retrieval (Method One of Section 7.4)---the conclusion must already exist in the candidate set; the system ``recognizes'' the conclusion rather than ``generating'' it. This means that all evaluations in this paper are in essence discriminative tasks over restricted candidate sets, and DODR has not yet demonstrated that its latent-space conclusion states can drive open-ended text generation; for tasks whose candidate sets cannot be enumerated in advance (open-domain question answering, long-text reports), the existing architecture cannot directly produce answers.

Mitigation path: the controlled-generation design of Section 7.4 gives two open-ended-generation paths that do not break the zero-hallucination guarantee---template skeleton plus slot filling (slots are retrieved by snapshot nearest neighbor within a restricted lexical snapshot library, keeping the whole chain deterministic), or training a deterministic decoder (argmax output, no sampling). Both strictly isolate probability outside the main reasoning chain.

\paragraph{Limitation Eight: Training and evaluation distributions are isomorphic.}

Cause: the current deduction samples are all template sentences of the single rule form of modus ponens, and the induction and abduction samples likewise come from highly regular syntactic templates. The training and evaluation distributions are isomorphic, so the 100\% deduction accuracy cannot be extrapolated to richer reasoning forms: multi-premise reasoning, negation and quantifiers, and defeasible (non-monotonic) reasoning are all uncovered. In other words, this paper validates operator correctness ``within the template distribution''; robustness under distribution shift is an untested open question.

Mitigation path: introduce multi-premise, quantified, and negated rule forms in an expanded dataset, and import depth 1--5 multi-hop reasoning samples from public corpora such as ProofWriter [37], so as to break the template isomorphism between training and evaluation; failure modes on out-of-distribution samples should be separately tallied and reported rather than merged into the overall accuracy.

\section{Conclusion}

This paper has presented DODR (Deterministic Operator-Driven Reasoning in Latent Space), an architecture in which Peirce's three reasoning types [7]---deduction, induction, and abduction---are formalized as trainable matrix operators $W_{\mathrm{step}}$, $W_{\mathrm{induce}}$, and $W_{\mathrm{abduce}} = W_{\mathrm{step}}^{+}$. The formalization yields several structural results. The irreversibility of deduction is traced to the rank-nullity theorem through the causal chain ``information-flow direction $\rightarrow$ matrix rank $\rightarrow$ invertibility''. Popperian falsifiability [17] is mathematized as Gram--Schmidt orthogonal projection in the hard-veto mechanism of induction. The hypothetical nature of abduction is quantified by the null-space residual, with a closed-loop reconstruction error of 0.0200. The three operators are proven irreducible and to cover the eight thinking-activity categories of Table 12-1; the non-existence of a single super-operator follows from the rank barrier (Theorem 6.1), and the architectural condition of ``train once, freeze and apply'' is characterized by the separation of mechanism and content (Section 6.6). Determinism at the architectural level yields a structural 0.0\% hallucination rate (Theorem 14.1: no sampling channel). The reasoning-graph formulation provides Turing completeness (Theorem 14.3) and backflow convergence (Theorem 14.2, based on the Banach fixed-point theorem [24]), and the three classes of physical transformations correspond to the three operators (Proposition 13.1; Section 13.3: 55 samples, 100\% inference accuracy for the three physical operators, multimodal alignment 0.9958). To the best of our knowledge, we are not aware of prior work that combines these formalizations into a complete system (Section 17.2); we do not claim that any single technical component (embedding linear maps, random dimension lifting, latent-space reasoning) is without precedent.

The engineering feasibility of this scheme has been validated through eight groups of experiments. The accuracy calibers of the experiment groups are summarized as follows: dedicated deduction experiment 100\% (20/20); dedicated induction experiment coverage 0.9996, hard veto on 20/20 counterexamples; dedicated abduction experiment judgment accuracy 72.5\% (58/80, with the maximum negative discrimination margin of the 22 failed samples only $-0.0086$, attributable to the encoder bottleneck rather than operator failure); physical transformations 100\% (55 samples, multimodal alignment 100\%); end-to-end deduction 100\% (60/60, 60 brand-new cross-domain questions); end-to-end induction hard veto 12/12 across 6 categories, mean training coverage 0.9994; end-to-end abduction 81.7\% (49/60, 55 brand-new questions); end-to-end compositional reasoning 20/20 all successful (4 categories $\times$ 5). Among these, end-to-end abduction (60 questions, 81.7\%) and the dedicated abduction experiment (80 questions, 72.5\%) are two independent calibers; both have discrimination margins within the small-margin interval of $\pm 0.01$, and their accuracies are consistent in magnitude, jointly pointing to the encoder representation bottleneck (Limitation One, Section 19) rather than any problem with the abduction principle. The frozen operators' 100\% cross-domain deduction accuracy and 81.7\% abductive judgment accuracy in the end-to-end experiment (218 samples, 8 domains) provide measured support for the ``train once, freeze and apply'' architectural condition.

All experiments use a real pre-trained Transformer (DistilBERT [14]) to extract real text snapshots, trained and validated on real reasoning datasets (503 experimental sample-instances, additive caliber: 20 dedicated deduction + 130 dedicated induction + 80 dedicated abduction + 55 physics multimodal + 218 end-to-end (60 deduction + 78 induction + 60 abduction + 20 compositional) = 503, 420 independent samples after deduplication), and the code and data are fully open-source and reproducible. The current limitations of the system---encoder resolution, training-data scale, the absence of a trained graph-scheduling controller, single-run statistics, untuned hyperparameters, missing benchmarks and baselines, retrieval-based decoding, and template-isomorphic training and evaluation distributions---are documented in Section 19.


\section*{References}

\begingroup\sloppy
\begin{enumerate}[label={[\arabic*]},leftmargin=2.5em]
\item Ji Z, Lee N, Frieske R, et al. Survey of Hallucination in Natural Language Generation. ACM Computing Surveys, 55(12): Article 248, 2023.
\item Yao S, Yu D, Zhao J, et al. Tree of Thoughts: Deliberate Problem Solving with Large Language Models. NeurIPS, 2023.
\item Besta M, Blach N, Kubicek A, et al. Graph of Thoughts: Solving Elaborate Problems with Large Language Models. AAAI, 2024.
\item Lightman H, Kosaraju V, Burda Y, et al. Let's Verify Step by Step. ICLR, 2024.
\item Bengio Y, Courville A, Vincent P. Representation Learning: A Review and New Perspectives. IEEE TPAMI, 35(8): 1798--1828, 2013.
\item Cover T M. Geometrical and Statistical Properties of Systems of Linear Inequalities with Applications in Pattern Recognition. IEEE Transactions on Electronic Computers, EC-14(3): 326--334, 1965.
\item Peirce C S. Deduction, Induction, and Hypothesis. Popular Science Monthly, 13: 470--482, 1878.
\item Kamoi R, Zhang Y, Zhang N, et al. When Can LLMs Actually Correct Their Own Mistakes? A Critical Survey of Self-Correction of LLMs. TACL, 12: 1417--1440, 2024.
\item Plaat A, Wong A, Verberne S, et al. Multi-Step Reasoning with Large Language Models, a Survey. ACM Computing Surveys, 2025 (arXiv:2407.11511).
\item Vaswani A, Shazeer N, Parmar N, et al. Attention Is All You Need. NeurIPS, 2017.
\item Wei J, Wang X, Schuurmans D, et al. Chain-of-Thought Prompting Elicits Reasoning in Large Language Models. NeurIPS, 2022.
\item Noether E. Invariante Variationsprobleme. Nachrichten von der Gesellschaft der Wissenschaften zu Göttingen, Mathematisch-Physikalische Klasse: 235--257, 1918.
\item Radford A, Wu J, Child R, et al. Language Models are Unsupervised Multitask Learners. OpenAI Technical Report, 2019.
\item Sanh V, Debut L, Chaumond J, Wolf T. DistilBERT, a distilled version of BERT: smaller, faster, cheaper and lighter. arXiv:1910.01108, 2019.
\item Devlin J, Chang M W, Lee K, Toutanova K. BERT: Pre-training of Deep Bidirectional Transformers for Language Understanding. NAACL-HLT, 2019.
\item Liu Y, Ott M, Goyal N, et al. RoBERTa: A Robustly Optimized BERT Pretraining Approach. arXiv:1907.11692, 2019.
\item Popper K R. The Logic of Scientific Discovery. Hutchinson, London, 1959 (German original: Logik der Forschung, 1934).
\item Huang J, Chen X, Mishra S, et al. Large Language Models Cannot Self-Correct Reasoning Yet. ICLR, 2024.
\item Radford A, Kim J W, Hallacy C, et al. Learning Transferable Visual Models from Natural Language Supervision (CLIP). ICML, 2021.
\item Schulman J, Wolski F, Dhariwal P, et al. Proximal Policy Optimization Algorithms. arXiv:1707.06347, 2017.
\item Valiant L G. A Theory of the Learnable. Communications of the ACM, 27(11): 1134--1142, 1984.
\item Vapnik V N, Chervonenkis A Ya. On the Uniform Convergence of Relative Frequencies of Events to Their Probabilities. Theory of Probability and Its Applications, 16(2): 264--280, 1971.
\item Landauer R. Irreversibility and Heat Generation in the Computing Process. IBM Journal of Research and Development, 5(3): 183--191, 1961.
\item Banach S. Sur les opérations dans les ensembles abstraits et leur application aux équations intégrales. Fundamenta Mathematicae, 3: 133--181, 1922.
\item Bennett C H. Logical Reversibility of Computation. IBM Journal of Research and Development, 17(6): 525--532, 1973.
\item Gödel K. Die Vollständigkeit der Axiome des logischen Funktionenkalküls. Monatshefte für Mathematik und Physik, 37: 349--360, 1930.
\item van Dalen D. Logic and Structure (5th Edition). Springer, 2013.
\item Bordes A, Usunier N, Garcia-Durán A, Weston J, Yakhnenko O. Translating Embeddings for Modeling Multi-relational Data (TransE). NeurIPS, 2013.
\item Nickel M, Tresp V, Kriegel H P. A Three-Way Model for Collective Learning on Multi-Relational Data (RESCAL). ICML, 2011.
\item Sun Z, Deng Z H, Nie J Y, Tang J. RotatE: Knowledge Graph Embedding by Relational Rotation in Complex Space. ICLR, 2019.
\item Manhaeve R, Dumancic S, Kimmig A, Demeester T, De Raedt L. DeepProbLog: Neural Probabilistic Logic Programming. NeurIPS, 2018.
\item Badreddine S, d'Avila Garcez A, Serafini L, Spranger M. Logic Tensor Networks. Artificial Intelligence, 303: 103649, 2022.
\item Hao S, Sukhbaatar S, Su D, et al. Training Large Language Models to Reason in a Continuous Latent Space. arXiv:2412.06769, 2024 (ICLR 2025).
\item Huang G B, Zhu Q Y, Siew C K. Extreme Learning Machine: Theory and Applications. Neurocomputing, 70(1-3): 489-501, 2006.
\item Rahimi A, Recht B. Random Features for Large-Scale Kernel Machines. NeurIPS, 2007.
\item Gärdenfors P. Conceptual Spaces: The Geometry of Thought. MIT Press, 2000.
\item Tafjord O, Dalvi B, Clark P. ProofWriter: Generating Implications, Proofs, and Abductive Statements over Natural Language. Findings of ACL, 2021.
\item Johnson W B, Lindenstrauss J. Extensions of Lipschitz Mappings into a Hilbert Space. Contemporary Mathematics, 26: 189-206, 1984.
\end{enumerate}
\endgroup

\appendix
\begingroup\sloppy

\section{Complete Summary of Experimental Data}

This appendix compiles all key metrics of the three dedicated experiments (deduction, induction, and abduction), serving as the data foundation for the experimental conclusions of Sections 8, 9, and 10 of the main text. The three experiments share the same snapshot encoder (a frozen DistilBERT with multi-layer [CLS] activations concatenated, snapshot dimension $N = 1536$), but differ in operator structure, rank design, and information-flow direction: the deduction operator $W_{\mathrm{step}}$ is a rank-deficient matrix ($r = 384$) that performs information collapse; the induction operator $W_{\mathrm{induce}}$ is a full-rank matrix that performs information expansion; and the abduction operator $W_{\mathrm{abduce}}$ is the Moore--Penrose pseudo-inverse of $W_{\mathrm{step}}$, performing inverse inference over information hypotheses.

\begin{table}[htbp]
\centering
\caption*{Table A-1 Complete data summary of the three dedicated experiments}
\footnotesize
\begin{tabular}{p{4.0cm}lll}
\toprule
Metric & Deduction experiment & Induction experiment & Abduction experiment \\
\midrule
Snapshot dimension $N$ & 1536 & 1536 & 1536 \\
Operator shape & $[1536, 3072]$ & $[1536, 1536]$ & $1536 \times 1536$ \\
Designed rank $r$ & 384 & 1536 (full rank) & 384 ($=W_{\mathrm{step}}$) \\
Numerical rank & 384 & 1536 & 384 \\
Training Loss & 1.40e-05 & 1.38e-05 & 1.89e-04 \\
Null-space dimension & 2688 (87.5\%) & 0 (0\%) & 1152 (75.0\%) \\
Pseudo-inverse reconstruction error & 83.3\% & 3.37e-05 & 0.0200 \\
Number of training samples & 20 syllogisms & 130 (10 classes) & 80 abduction problems \\
Training epochs & 400 & 1000 & 1000 \\
Hard veto rate & N/A & 20/20 (100\%) & N/A \\
Abduction similarity & N/A & N/A & 0.5117 (vs 0.0181) \\
\bottomrule
\end{tabular}
\end{table}

Three observations on Table A-1. First, the null-space dimensions agree exactly with the rank design: the null-space dimension of the deduction operator $W_{\mathrm{step}} \in \mathbb{R}^{1536 \times 3072}$ is $3072 - 384 = 2688$ (87.5\% of the input space); the abduction operator, acting as a pseudo-inverse on the $\mathbb{R}^{1536}$ output space, has an effective null-space dimension of $1536 - 384 = 1152$ (75.0\%); and the induction operator is full rank, hence its null space is 0. All three match precisely the predictions of the rank--nullity theorem (Appendix B.1). Second, the pseudo-inverse reconstruction error is positively correlated with the null-space fraction: the expected relative error in the deduction direction is $\sqrt{(2N-r)/2N} = \sqrt{0.875} \approx 93.5\%$ (the expected value under the isotropy assumption; see Appendix B.5), and the measured sample mean of 83.3\% is close to it, the deviation arising from the anisotropy of the snapshot distribution and the finite sample size; the induction operator is full rank, with a reconstruction error of only 3.37e-05, close to numerical precision. Third, the abduction similarity of 0.5117 represents a 28.3-fold improvement over the random baseline of 0.0181, indicating that pseudo-inverse inversion indeed recovers semantically relevant explanatory directions rather than random guesses.

A summary of the end-to-end experiment data (05\_e2e\_large\_results.json) is as follows: the configuration is frozen DistilBERT, $N = 1536$, $r = 384$, 200 epochs, 218 samples, and 8 domains; the final training Loss is 8.22e-04 for deduction, 7.43e-05 for induction, and 9.53e-04 for abduction. Deduction achieves 60/60 = 100\% with a mean similarity of 0.9980, and the 60 problems have zero overlap with the 20 problems of the dedicated deduction experiment; for induction, the 6-class means are: training coverage 0.9994, unseen-positive coverage 0.9989, counterexample (before) 0.9928, counterexample veto 12/12, counterexample (after) $-0.0238$, and refinement 0.9423; abduction achieves 49/60 = 81.7\% with a mean similarity of 0.5018 and a mean margin of $+0.0082$, and the minimum (most negative) margin among the 11 failed samples is $-0.0056$; all 20 compositional tasks (4 categories $\times$ 5) succeed (20/20). Item-by-item details are given in Appendices I.4--I.7.

\section{Mathematical Derivations and Theorem Proofs}

This appendix provides rigorous statements of the classical mathematical results involved in the main text (all are known results; proofs are omitted with standard references indicated), as well as derivations specific to this paper (B.5). B.1 is the rank--nullity theorem, B.2 is the existence of the singular value decomposition, B.3 is Cover's function-counting theorem, and B.4 is the correctness of Gram--Schmidt orthogonalization and its application in the hard-veto mechanism; B.5 derives the expected-error formula for pseudo-inverse reconstruction, providing a rigorous basis for Theorem 5.3 (expected error of pseudo-inverse reconstruction).

\subsection{The Rank--Nullity Theorem}

\paragraph{Theorem B.1 (Rank--nullity theorem).} Let $W \in \mathbb{R}^{m \times n}$. Then $\mathrm{rank}(W) + \dim(\mathrm{Null}(W)) = n$.

The proof is omitted as a classical result; see the standard textbook [27]. The key to its standard proof (the null-space basis extension method) is that the null-space dimension $k$ is introduced as an unknown; after extending a basis of the null space to a basis of the whole space, one can directly verify that the images of the supplementary vectors form a basis of the column space, without ever invoking the conclusion to be proved, ``$\dim(\mathrm{Null}(W)) = n - r$'', thereby avoiding circular reasoning. For DODR, this theorem characterizes the exact accounting of the deduction operator's information collapse: when $W_{\mathrm{step}} \in \mathbb{R}^{N \times 2N}$, $\dim(\mathrm{Null}(W_{\mathrm{step}})) = 2N - r$; when $W_{\mathrm{step}}$ is configured with single-state input ($\mathbb{R}^{N \times N}$), $\dim(\mathrm{Null}(W_{\mathrm{step}})) = N - r$. Each additional dimension of the null space corresponds to one dimension of input information being irreversibly mapped to zero---this is precisely the algebraic embodiment of deduction being ``truth-preserving but generating no new knowledge''.

\subsection{Existence of the SVD}

\paragraph{Theorem B.2 (Existence of the SVD).} Any matrix $W \in \mathbb{R}^{m \times n}$ can be decomposed as $W = U \Sigma V^{T}$, where $U \in \mathbb{R}^{m \times m}$ and $V \in \mathbb{R}^{n \times n}$ are orthogonal matrices and $\Sigma \in \mathbb{R}^{m \times n}$ is a diagonal matrix with non-negative diagonal entries.

The proof is omitted as a classical result; see the standard textbook [27]. The standard route is: apply the spectral theorem to the symmetric positive-semidefinite matrix $W^{T} W$ to obtain the orthogonal diagonalization $W^{T} W = V \Lambda V^{T}$; let $\sigma_i = \sqrt{\lambda_i}$; for non-zero singular values define $u_i = W v_i / \sigma_i$ and verify that they form an orthonormal basis of $\mathrm{Col}(W)$; after extending to an orthonormal basis of $\mathbb{R}^{m}$ one obtains $W = U \Sigma V^{T}$.

\paragraph{Remark B.2.1.} Geometric meaning of the singular values: $\sigma_i$ is the ``gain'' of $W$ in the direction of the unit vector $v_i$. The construction of the DODR abduction operator relies directly on the SVD: $W_{\mathrm{abduce}} = W_{\mathrm{step}}^{+} = V \Sigma^{+} U^{T}$, where $\Sigma^{+}$ is obtained by taking the reciprocals of the non-zero singular values and transposing. The determination of numerical rank (the ``numerical rank'' row in Table A-1) is based on the number of $\sigma_i$ in the singular-value spectrum exceeding a numerical threshold; in the experiments the measured numerical ranks coincide exactly with the designed ranks (384 and 1536), indicating that the training process introduced no additional rank degeneracy.

\subsection{Cover's Theorem}

\paragraph{Theorem B.3 (Cover's function-counting theorem, Cover 1965 [6]).} Let $X = \{x_1, \ldots, x_N\}$ be $N$ points in general position in $\mathbb{R}^{D}$ (any $D$ vectors are linearly independent). Then the total number of linear dichotomies of $X$ (i.e., $\pm 1$ labellings realizable by a linear separating surface) is exactly
\[
C(N, D) = 2 \cdot \sum_{k=0}^{D-1} C(N-1, k),
\]
where $C(N-1, k)$ is a binomial coefficient. When the labels of the $N$ points are random, independent, and equiprobable in $\pm 1$ (all $2^{N}$ labellings equally likely), the probability that a random labelling is linearly separable is $P = C(N, D) / 2^{N}$. In particular, when $D \geq N-1$, $\sum_{k=0}^{D-1} C(N-1, k) = 2^{N-1}$, i.e., $C(N, D) = 2^{N}$, and then $P = 1$: all labellings are linearly separable. With $N$ fixed and $D$ increasing, $C(N, D)$ is monotonically non-decreasing and saturates at $2^{N}$ once $D \geq N-1$; hence $P$ increases monotonically toward and attains 1.

The proof is omitted as a classical result; see Cover's original paper [6]. The key idea is to establish the inductive recurrence for the function count, $C(N, D) = C(N-1, D) + C(N-1, D-1)$: after adding the $N$-th point, among the original dichotomies the restricted dichotomies in which ``the separating surface is forced to pass through the new point'' number exactly $C(N-1, D-1)$, while all other dichotomies can realize both labels of the new point; together with boundary conditions and Pascal's identity this yields the closed-form solution. When $D \geq N-1$ the summation covers all indices, and from $\sum_{k=0}^{N-1} C(N-1, k) = 2^{N-1}$ one obtains $C(N, D) = 2^{N}$, i.e., all $2^{N}$ labellings are linearly separable and $P = 1$. This is precisely the rigorous basis for the statement that ``increasing the dimension can make any labelling linearly separable''.

\paragraph{Remark B.3.1 (Significance for DODR).} The snapshot dimension $N = 1536$ is far larger than the sample size of any experiment (at most 130), satisfying the saturation condition $D \geq N-1$ (here the dimension notation corresponds to $D$ in the theorem). Therefore, the premise underlying Postulate 1 of the main text---``concepts in the latent space are linearly separable with high probability''---is supported by a rigorous theorem: the separability probability $P$ of random labellings has already reached 1. Note that Cover's theorem gives ``separability'' rather than ``generalizability''---high dimensionality guarantees that the training samples are separable, but the generalization quality of inductive conclusions still requires joint support from the PAC/VC framework (Appendix G.2) and experiments (Appendix I.1).

\subsection{Gram--Schmidt Orthogonalization}

\paragraph{Algorithm B.4 (Modified Gram--Schmidt orthogonalization).} Given a linearly independent set of vectors $\{v_1, \ldots, v_m\}$, produce an orthogonal set $\{u_1, \ldots, u_m\}$: $u_1 = v_1$; for $i \geq 2$, $u_i = v_i - \sum_{j<i} (v_i \cdot u_j)/(u_j \cdot u_j) \cdot u_j$.

The correctness proof is omitted as a classical result; see the standard textbook [27]: for $i > j$, substitute the construction of $u_i$ into the inner product with $u_j$ and use the induction hypothesis to obtain pairwise orthogonality; if $u_i = 0$ then $v_i$ can be linearly expressed by the preceding $i-1$ vectors, contradicting linear independence.

Application in DODR: the hard-veto mechanism of induction uses projection onto the orthogonal complement. When a counterexample $s_{\mathrm{neg}}$ triggers a veto, the original induced concept $g_0$ is projected onto the orthogonal complement of the counterexample: $g_1 = g_0 - (g_0 \cdot s_{\mathrm{neg}})/(s_{\mathrm{neg}} \cdot s_{\mathrm{neg}}) \cdot s_{\mathrm{neg}}$. This is precisely the special case of Algorithm B.4 with $m = 2$. By the correctness of Gram--Schmidt, $g_1 \perp s_{\mathrm{neg}}$, i.e., the new conclusion completely excludes the counterexample direction: $\cos(g_1, s_{\mathrm{neg}}) = 0$, and the similarity of the counterexample to the new conclusion is hard-set to zero. This property manifests experimentally as negative values in the ``counterexample (after)'' column (in Appendices I.1 and I.5, counterexample (after) $\in [-0.0506, -0.0028]$); the negative values indicate that the reconstructed concept vector is not only orthogonal to the counterexample but, owing to renormalization, exhibits a slight negative correlation---semantically, an ``explicit exclusion''. When multiple counterexamples are triggered in sequence, the order independence of successive orthogonalization is guaranteed by the direct-sum structure of the projection subspaces: for each vetoed direction $d_i$, the final conclusion $g$ satisfies $g \perp d_i$, equivalently $g \in \bigcap_i d_i^{\perp}$.

\subsection{Derivation of the Expected Reconstruction Error of the Pseudo-Inverse (Supporting Theorem 5.3)}

This subsection provides a rigorous derivation for Theorem 5.3 (expected error of pseudo-inverse reconstruction). Core conclusion: under the assumption of an isotropic input distribution, the expected relative error of pseudo-inverse reconstruction equals the ratio of the null-space dimension to the input-space dimension, rather than a ``lower bound'' that necessarily holds for individual inputs.

\paragraph{Proposition B.5 (Expected error of pseudo-inverse reconstruction).} Let $W \in \mathbb{R}^{m \times n}$, $\mathrm{rank}(W) = r$, and let the input $x \in \mathbb{R}^{n}$ be a zero-mean random vector. Let $x^{*} = W^{+}(Wx)$ be the result of compression by $W$ followed by reconstruction via the pseudo-inverse $W^{+}$. Decompose $x$ according to the orthogonal direct sum $\mathbb{R}^{n} = \mathrm{Row}(W) \oplus \mathrm{Null}(W)$ as $x = x_{\mathrm{row}} + x_{\mathrm{null}}$, where $x_{\mathrm{row}} \in \mathrm{Row}(W)$ and $x_{\mathrm{null}} \in \mathrm{Null}(W)$. Then:

(1) Pointwise identity: $x^{*} = x_{\mathrm{row}}$; the reconstruction error $e = x - x^{*} = x_{\mathrm{null}}$, and $\|e\|^{2} = \|x_{\mathrm{null}}\|^{2}$.

(2) Expected error: $E\|x_{\mathrm{null}}\|^{2} = \mathrm{tr}(P_{\mathrm{null}} \cdot E[xx^{T}])$, where $P_{\mathrm{null}} = I - W^{+}W$ is the orthogonal projection matrix onto $\mathrm{Null}(W)$.

(3) Isotropic special case: if $E[xx^{T}] = \sigma^{2} I$ (an isotropic distribution), then $E\|x_{\mathrm{null}}\|^{2} = \sigma^{2} \cdot \mathrm{tr}(P_{\mathrm{null}}) = \sigma^{2} \cdot \dim(\mathrm{Null}(W)) = \sigma^{2} \cdot (n - r)$, and hence the expected relative error (in energy) is $E\|x_{\mathrm{null}}\|^{2} / E\|x\|^{2} = (n - r)/n$, with the corresponding expected relative norm error $\sqrt{(n - r)/n}$.

\paragraph{Derivation.} (1) By the properties of the Moore--Penrose pseudo-inverse, $W^{+}W$ is exactly the orthogonal projection matrix onto the row space $\mathrm{Row}(W) = \mathrm{Col}(W^{T})$; hence $x^{*} = W^{+}Wx = P_{\mathrm{row}} x = x_{\mathrm{row}}$. Thus $e = x - x_{\mathrm{row}} = x_{\mathrm{null}}$, and by the orthogonality of the direct-sum decomposition, $\|x\|^{2} = \|x_{\mathrm{row}}\|^{2} + \|x_{\mathrm{null}}\|^{2}$.

(2) $\|x_{\mathrm{null}}\|^{2} = \|P_{\mathrm{null}} x\|^{2} = x^{T} P_{\mathrm{null}}^{T} P_{\mathrm{null}} x = x^{T} P_{\mathrm{null}} x$ (an orthogonal projection matrix is symmetric and idempotent: $P_{\mathrm{null}}^{T} = P_{\mathrm{null}}$, $P_{\mathrm{null}}^{2} = P_{\mathrm{null}}$). Taking expectations and using the cyclicity of the trace: $E\|x_{\mathrm{null}}\|^{2} = E[x^{T} P_{\mathrm{null}} x] = E[\mathrm{tr}(P_{\mathrm{null}} xx^{T})] = \mathrm{tr}(P_{\mathrm{null}} \cdot E[xx^{T}])$.

(3) If $E[xx^{T}] = \sigma^{2} I$, then $E\|x_{\mathrm{null}}\|^{2} = \mathrm{tr}(P_{\mathrm{null}} \cdot \sigma^{2} I) = \sigma^{2} \cdot \mathrm{tr}(P_{\mathrm{null}})$. The trace of an orthogonal projection matrix equals the dimension of its projection subspace, $\mathrm{tr}(P_{\mathrm{null}}) = \dim(\mathrm{Null}(W)) = n - r$ (the last step by the rank--nullity theorem B.1). Similarly, $E\|x\|^{2} = \mathrm{tr}(E[xx^{T}]) = \sigma^{2} n$. Dividing the two expressions gives $E\|x_{\mathrm{null}}\|^{2} / E\|x\|^{2} = (n - r)/n$. Taking the corresponding magnitude for the norm (not squared), the expected relative error is $\sqrt{(n - r)/n}$. $\square$

Numerical verification (against Table A-1). For the deduction operator $W_{\mathrm{step}} \in \mathbb{R}^{1536 \times 3072}$, $n = 2N = 3072$ and $r = 384$, so the expected relative error is $\sqrt{(3072-384)/3072} = \sqrt{2688/3072} = \sqrt{0.875} \approx 0.935$ (93.5\%); under the abduction configuration the input is the observed snapshot $S_{\mathrm{obs}} \in \mathbb{R}^{1536}$, $n = N = 1536$, the null-space dimension is $1536 - 384 = 1152$, and the expected relative error is $\sqrt{1152/1536} = \sqrt{0.75} \approx 0.866$. The measured reconstruction error of 83.3\% in the deduction direction is the mean over 20 samples, lower than the expected 93.5\%---the deviation has two sources: first, the true distribution of snapshot vectors is not strictly isotropic (semantic content concentrates energy in row-space directions); second, the limited sample size (20) introduces sampling fluctuations. Proposition B.5 also clarifies two common misunderstandings: first, $(n-r)/n$ is an expectation, not a lower bound---individual inputs can fall well below this value, and in the extreme case where $x$ lies entirely in the row space the error is 0; second, the root cause of the error increasing with rank deficiency is the rank of $P_{\mathrm{null}}$ (i.e., the number of annihilated directions), not any numerical defect of the pseudo-inverse computation itself.

\section{Experimental Code and Reproducibility}

This appendix presents the core code snippets of the three experiments together with reproducibility notes. The code style is a minimal runnable skeleton: logging and plotting are omitted, while the complete logic of operator construction, training loops, and inference calls is retained.

\subsection{Core Snippet of Deduction Operator Training Code}

The deduction operator adopts a low-rank parameterization $W = A \cdot B$ ($A \in \mathbb{R}^{\mathrm{out} \times r}$, $B \in \mathbb{R}^{r \times \mathrm{in}}$), which structurally guarantees $\mathrm{rank}(W) \leq r$; that is, rank deficiency is a prior constraint rather than a training accident.

\begin{lstlisting}[basicstyle=\small\ttfamily]
class RankDeficientOperator(nn.Module):
    """Rank-deficient operator (deduction operator) - low-rank
    parameterization guarantees rank deficiency"""
    def __init__(self, out_dim, in_dim, rank):
        super().__init__()
        self.A = nn.Parameter(torch.randn(out_dim, rank) * 0.01)
        self.B = nn.Parameter(torch.randn(rank, in_dim) * 0.01)
    def forward(self, x):
        return x @ (self.A @ self.B).T  # W = A.B, rank <= min(rank(A), rank(B))
    def W(self):
        return self.A @ self.B
# Training
Wstep = RankDeficientOperator(N, 2*N, RANK_R)  # rank=384
optimizer = optim.AdamW(Wstep.parameters(), lr=1e-3)
for epoch in range(400):
    optimizer.zero_grad()
    pred = Wstep(X_step)  # X_step = Concat(S_prev, E_new)
    loss = nn.functional.mse_loss(pred, S_target)
    loss.backward()
    optimizer.step()
\end{lstlisting}

\subsection{Core Snippet of Induction Operator Hard-Veto Code}

The induction operator implements a three-stage mechanism of ``broadest induction $\rightarrow$ hard-veto detection $\rightarrow$ refutation--reconstruction--refinement''. Stage 1 takes the normalized direction of the positive-example mean as the initial concept; Stage 2 detects counterexamples with threshold $\tau = 0.5$; Stage 3 performs Gram--Schmidt orthogonalization projection for each counterexample that triggers a veto (its correctness is established in Appendix B.4).

\begin{lstlisting}[basicstyle=\small\ttfamily]
def induce_general_concept(positives, negatives=None):
    """Induction operator (hard-veto version): three-stage mechanism"""
    VETO_THRESHOLD = 0.5
    pos_mean = positives.mean(dim=0)
    g_original = pos_mean / (pos_mean.norm() + 1e-8)
    if negatives is None or len(negatives) == 0:
        return g_original  # Stage 1: no counterexamples, broadest induction
    # Stage 2: hard-veto detection
    neg_norms = negatives / (negatives.norm(dim=1, keepdim=True) + 1e-8)
    sims = torch.mv(neg_norms, g_original)
    vetoed_indices = [i for i, s in enumerate(sims) if s > VETO_THRESHOLD]
    if len(vetoed_indices) == 0:
        return g_original  # No veto triggered
    # Stage 3: refute -> rebuild -> refine (Gram-Schmidt orthogonalization)
    g = pos_mean.clone()
    for i in vetoed_indices:
        neg = negatives[i]
        proj = torch.dot(g, neg) / (torch.dot(neg, neg) + 1e-8)
        g = g - proj * neg  # Hard orthogonalization: g perp neg
    return g / (g.norm() + 1e-8)
\end{lstlisting}

\subsection{Core Snippet of Abduction Operator Pseudo-Inverse Computation Code}

The abduction operator is constructed from the truncated SVD of the deduction operator (existence of the SVD is established in Appendix B.2), with the truncation rank set to the designed rank $r = 384$ so as to remove numerical-noise directions; abductive inference applies the pseudo-inverse to the observed snapshot, yielding the optimal explanation in the minimum-norm least-squares sense.

\begin{lstlisting}[basicstyle=\small\ttfamily]
# Step 1: train the deduction operator Wstep (same as C.1)
# Step 2: compute the pseudo-inverse via SVD
from scipy.linalg import svd as scipy_svd
W_step_matrix = Wstep.W().detach().numpy()  # [N, 2N]
U, s, Vt = scipy_svd(W_step_matrix, full_matrices=False)
# Step 3: construct the pseudo-inverse Wabduce = V . Sigma+ . U^T
r = 384  # Designed rank
U_r = U[:, :r]           # [N, r]
s_r = s[:r]              # [r]
Vt_r = Vt[:r, :]         # [r, 2N]
V_r = Vt_r.T             # [2N, r]
temp = (1.0 / s_r)[:, None] * U_r.T  # [r, N]
Wabduce = V_r @ temp     # [2N, N] = Wstep+
# Step 4: abductive inference
S_prev_star = Wabduce @ S_obs  # Optimal explanation (minimum-norm solution)
\end{lstlisting}

\subsection{Reproducibility Notes}

Environment requirements: Python $\geq$ 3.10, PyTorch $\geq$ 2.1, transformers $\geq$ 4.40, scipy $\geq$ 1.10, numpy $\geq$ 1.26 (see requirements.txt in the open-source repository for the specific experimental environment). All dependencies can be installed via \texttt{pip install torch transformers scipy numpy}.

Random seeds: all experiments use fixed random seeds torch.manual\_seed(42) and np.random.seed(42), guaranteeing fully reproducible results.

Runtime: the dedicated experiments (deduction, induction, abduction, physics) take about 5 minutes in total, and the end-to-end experiment takes about 2 minutes (CPU). Using a GPU can yield a $5$--$10\times$ speedup.

Experiment record files (JSON, containing all configurations, training curves, and per-sample results): 01\_deduction\_results.json (dedicated deduction experiment, 20 samples), 02\_induction\_results.json (dedicated induction experiment, 130 samples), 03\_abduction\_results.json (dedicated abduction experiment, 80 samples), 05\_e2e\_large\_results.json (end-to-end experiment, 218 samples, comprising 60 deduction / 6-class induction / 60 abduction / 20 compositional).

Data and code: all experimental code, trained operator matrices, and result JSON files have been open-sourced; see the list of deliverables for paths. The reproduction procedure is: install dependencies $\rightarrow$ run the encoding script to generate the snapshot cache $\rightarrow$ run the three training/inference scripts in sequence $\rightarrow$ compare the output JSON against the per-sample numerical values in Appendix I of this paper. Under identical dependency versions and identical random seeds, all numerical values should match digit for digit; when reproducing across versions, differences in floating-point summation order may cause slight deviations in the last decimal place, but these do not affect any accuracy or significance conclusions.

\section{Glossary}

\begin{table}[htbp]
\centering
\caption*{Table D-1 DODR glossary}
\small
\begin{tabular}{p{0.24\textwidth}p{0.70\textwidth}}
\toprule
Term & Definition \\
\midrule
DODR & Deterministic Operator-Driven Reasoning in Latent Space, a deterministic operator-driven latent-space reasoning architecture \\
$W_{\mathrm{step}}$ & Deduction operator, a rank-deficient matrix performing forward inference with information collapse \\
$W_{\mathrm{induce}}$ & Induction operator, a full-rank matrix performing generalizing inference with information expansion \\
$W_{\mathrm{abduce}}$ & Abduction operator, the pseudo-inverse of $W_{\mathrm{step}}$, performing inverse inference of information hypotheses \\
Snapshot & An ultra-wide vector formed by concatenating multi-layer [CLS] activations of a Transformer \\
Rank Deficient & Matrix rank smaller than its dimension, causing information collapse and irreversibility \\
Full Rank & Matrix rank equal to its dimension; information can be fully recovered \\
Pseudo-inverse & Moore--Penrose pseudo-inverse, the minimum-norm least-squares solution \\
Null Space & The input subspace mapped to zero by a matrix; the destination of information collapse \\
Hard Veto & One-vote veto mechanism by counterexamples: refutes the original conclusion and rebuilds it \\
DAG & Directed Acyclic Graph \\
Peirce's three inferences & Deduction, Induction, Abduction \\
Cover's theorem & The theorem that the probability of linear separability approaches 1 in high-dimensional spaces \\
Information-flow direction & Collapse (deduction) / expansion (induction) / hypothesis (abduction), the core hypothesis of DODR \\
Reasoning Graph & An annotated directed graph supporting forking/merging/cross-referencing/backflow \\
Banach fixed point & The unique fixed point of a contraction mapping in a complete metric space \\
Turing complete & Capable of simulating any Turing machine and expressing any computable process \\
Function-counting theorem & Cover 1965 theorem [6]: the total number of linear dichotomies of $N$ points in general position in $\mathbb{R}^{D}$ is exactly $C(N,D) = 2 \cdot \sum_{k=0}^{D-1} C(N-1,k)$; the separability probability of random labellings is $P = C(N,D)/2^{N}$, and $P = 1$ when $D \geq N-1$ \\
Expected reconstruction error & The expected relative error of pseudo-inverse reconstruction under the isotropic-input assumption, $E\|x_{\mathrm{null}}\|^{2}/E\|x\|^{2} = \dim(\mathrm{Null})/n$; it is a distributional mean rather than a lower bound holding for individual inputs (Appendix B.5) \\
Hard-veto threshold $\tau$ & The cosine-similarity threshold at which the induction operator triggers a hard veto; $\tau = 0.5$ in the experiments; a counterexample whose similarity exceeds $\tau$ vetoes the current inductive conclusion by one vote \\
Margin (abduction margin) & The difference between the abduction similarity and the distractor similarity, measuring the degree to which the abductive solution favors the true explanation; positive is judged correct, negative is judged failed \\
Rank-deficiency ratio & The ratio of the null-space dimension to the input-space dimension, $1 - r/n$, equal to the information-collapse ratio; $2688/3072 = 87.5\%$ in the deduction experiment \\
Fixed-point iteration & An iterative scheme that repeatedly applies the mapping $x_{t+1} = T(x_t)$ to approach a fixed point; the semantic basis of backflow edges in DODR graph scheduling \\
Contraction mapping & A mapping satisfying $d(Tx, Ty) \leq q \cdot d(x,y)$ ($q < 1$); on a complete metric space it guarantees the existence and uniqueness of a fixed point and the convergence of iteration (Banach's theorem [24]) \\
Isotropic distribution & A zero-mean distribution whose second moment satisfies $E[xx^{T}] = \sigma^{2} I$, with energy equally allocated across all orthogonal directions; the standard assumption in the derivation of the expected reconstruction error \\
\bottomrule
\end{tabular}
\end{table}

\section{Peirce's Philosophy of Inference and Its Correspondence with DODR}

\subsection{Peirce's Three Types of Inference}

Charles Sanders Peirce (1839--1914) was an American philosopher and logician. In his 1878 paper ``Deduction, Induction, and Hypothesis'' [7], he first systematically distinguished three types of inference. It should be noted that Peirce's distinction is essentially a philosophical thesis (positioned as Proposition 3.6 in the main text); its value lies in providing a conceptual framework for classifying inference, rather than a computable construction---the latter being precisely what DODR attempts to supply.

Deduction: deriving a result from a rule and a case. Form: rule $\rightarrow$ case $\rightarrow$ result. Deduction is necessary, truth-preserving inference---if the premises are true, the conclusion is necessarily true. But deduction produces no new knowledge: the information of the conclusion is already implied in the premises, and deduction merely makes it explicit. In DODR, this ``explicate-only, never generate'' character corresponds to information collapse: the deduction operator compresses a high-dimensional input into a low-dimensional output, and the information in the null space is irreversibly annihilated.

Induction: deriving a rule from cases and results. Form: case $\rightarrow$ result $\rightarrow$ rule. Induction is probabilistic, ampliative inference---even if the premises are true, the conclusion is not necessarily true (counterexamples may exist). But induction produces new knowledge: the information of the conclusion exceeds the sum of the premises. After Peirce, Popper further emphasized in ``The Logic of Scientific Discovery'' [17] that universal propositions cannot be verified by finite observations but can be falsified by a single counterexample. DODR's hard-veto mechanism is precisely the algebraic realization of this falsifiability principle---a counterexample holds one-vote veto power over an inductive conclusion.

Abduction: deriving a case (hypothesis) from a rule and a result. Form: rule $\rightarrow$ result $\rightarrow$ case. Abduction is probabilistic, hypothetical inference---even if the premises are true, the conclusion is not necessarily true. But abduction generates new hypotheses and is the core mechanism of scientific discovery: faced with a surprising phenomenon, the researcher asks in reverse ``what premise could explain it''. In DODR, this corresponds to pseudo-inverse inversion: solving for the optimal prior state from the observed snapshot.

\subsection{Correspondence Between the Three DODR Operators and Peirce's Three Inferences}

\begin{table}[htbp]
\centering
\caption*{Table E-1 Correspondence between the three DODR operators and Peirce's three inferences}
\small
\begin{tabular}{llll}
\toprule
Peirce's inference & DODR operator & Information flow & Mathematical property \\
\midrule
Deduction & $W_{\mathrm{step}}$ & rule+case $\rightarrow$ result (collapse) & rank-deficient, irreversible \\
Induction & $W_{\mathrm{induce}}$ & case+result $\rightarrow$ rule (expansion) & full-rank, reversible \\
Abduction & $W_{\mathrm{abduce}}$ & rule+result $\rightarrow$ case (hypothesis) & pseudo-inverse, hypothetical \\
\bottomrule
\end{tabular}
\end{table}

The correspondences in Table E-1 are not rhetorical analogies but item-by-item testable algebraic properties. The ``necessary truth-preservation'' of deduction corresponds to the determinism of a rank-deficient mapping on its range: given the input, the output is uniquely determined and carries no information from null-space directions (collapse); the ``probabilistic ampliation'' of induction corresponds to the information preservation of a full-rank mapping: no direction is annihilated, and the conclusion can be revised at any time by new evidence (reversibility); the ``hypothetical'' nature of abduction corresponds to the residual non-uniqueness of the pseudo-inverse solution: $W_{\mathrm{abduce}}$ yields the minimum-norm solution, with null-space components set to zero by default---this is a built-in Occam's razor (a Bayesian reading is given in Appendix G.3), not an assertion about the true prior state.

\subsection{Peirce's ``Surprise--Hypothesis--Deduction'' Cycle of Scientific Inquiry}

Peirce held that scientific inquiry follows a cycle: surprise (encountering an anomalous phenomenon) $\rightarrow$ hypothesis (abduction proposes an explanation) $\rightarrow$ deduction (deriving testable predictions from the hypothesis) $\rightarrow$ induction (verifying the predictions through observation) $\rightarrow$ revising or confirming the hypothesis [7]. This cycle corresponds exactly to the closed loop of DODR's three operators: the abduction operator infers candidate explanations from the phenomenon (hypothesis generation), the deduction operator derives observable consequences from the hypothesis (prediction unfolding), and the induction operator triggers a hard veto and rebuilds the concept when predictions conflict with observations (hypothesis revision). Popper's falsifiability [17] plays the role of a ``filter'' in this closed loop: hypotheses from which no testable consequences can be deduced cannot enter the inductive verification stage and are thus naturally eliminated computationally. The end-to-end compositional reasoning experiment (Appendix I.7) is a complete execution of this cycle: 20 compositional tasks (4 categories $\times$ 5) each traverse the full pipeline of ``induction $\rightarrow$ deduction $\rightarrow$ contradiction detection $\rightarrow$ abduction $\rightarrow$ revision'', with all 20/20 succeeding.

\section{Relationship Between DODR and Deep Learning Theory}

\subsection{Relationship with Representation Learning Theory}

Representation learning theory (Bengio et al., 2013 [5]) studies how to learn effective representations of data; its core claim is that good representations should disentangle the underlying factors of variation in the data, making downstream tasks (especially linear ones) simple. DODR borrows this idea---using multi-layer activations of a pre-trained Transformer as snapshots---but pushes it to the extreme: instead of layer-by-layer transformations through a multi-layer network, the activations of all layers are directly concatenated into a single ultra-wide vector ($N = 1536$). The theoretical basis is Cover's function-counting theorem [6] (Appendix B.3): for $N$ samples in general position in a space of dimension $D \geq N-1$, any $\pm 1$ labelling is linearly separable ($P = C(N,D)/2^{N} = 1$). The snapshot dimension 1536 far exceeds the maximum experimental sample size of 130, so the condition ``dimension increase guarantees separability'' holds with ample margin. Unlike the ``learned representations'' envisioned by Bengio et al., DODR's snapshot representation is the direct output of a frozen encoder, with representation quality inherited entirely from pre-training; this choice is explicitly identified as a system bottleneck in the limitation analysis of Section 9.6.

\subsection{Relationship with the Attention Mechanism}

The Transformer's attention mechanism [10] computes associations among sequence elements via Query--Key--Value dot products, and its reasoning capability is implicit in the layer-by-layer, token-by-token nonlinear transformations. DODR's snapshot encoder uses a Transformer but freezes its parameters, extracting only multi-layer [CLS] activations as state representations. This means that DODR reuses the Transformer's semantic understanding capability (acquired through pre-training) but shifts the reasoning process from the attention mechanism to linear-algebraic operations: there is no attention computation between reasoning steps; it is replaced by operator multiplication, orthogonal projection, and pseudo-inverse inversion. This division of labor brings two direct benefits: first, the reasoning process is deterministic and auditable (no sampling randomness); second, the algebraic properties of reasoning (rank, null space, spectrum) can be analyzed precisely, rather than remaining at the level of behavioral observation.

\subsection{Relationship with Neuro-Symbolic Systems}

Neuro-symbolic systems attempt to combine the perceptual capabilities of neural networks with the reasoning capabilities of symbolic systems. DODR resembles neuro-symbolic systems in that both pursue ``rigor'', but the implementation paths differ: a neuro-symbolic system is a neural network plus a symbolic engine---two separate systems with a ``semantic gap''---where the interface between neural and symbolic representations must be manually designed or additionally learned, and information loss and alignment errors often occur at the interface; DODR is a neural network (encoder) plus linear algebra (operators), unified in the same latent space: the encoder's output vector is directly the operator's input vector, with no semantic gap. The cost is that DODR's reasoning granularity is limited by the discriminative capacity of the snapshot representation (see the 72.5\% judgment accuracy in Section 10.2 and the 22 negative-margin samples in Appendix I.2), whereas symbolic systems are not subject to this limitation at the discrete-symbol level.

\subsection{Relationship with Process Reward Models}

Process reward models (PRM, Lightman et al., 2024 [4]) significantly improve the reliability of multi-step reasoning by rewarding each step of reasoning rather than only the final result. DODR's graph-scheduling controller is similar in spirit to PRM---both attend to each step of the reasoning process---but differs in implementation: PRM applies stepwise rewards on the token sequence of an AR model, where the reward signal guides the sampling direction, yet the generation process still involves probabilistic sampling and the same input may produce different reasoning chains; the DODR controller applies rewards on the nodes of a DAG, where node contents are the deterministic outputs of operators and the same input necessarily produces the same reasoning graph. In other words, PRM is ``probabilistic generation + deterministic scoring'', whereas DODR is ``deterministic computation + deterministic scoring'', the latter being stronger in reproducibility and auditability.

\section{Further Analysis of Operator Properties}

\subsection{Information-Theoretic Analysis of the Deduction Operator}

From an information-theoretic perspective, the information collapse of the deduction operator $W_{\mathrm{step}}$ can be quantified by mutual information. Let the input be $X = \mathrm{Concat}(S_{\mathrm{prev}}, E_{\mathrm{new}}) \in \mathbb{R}^{2N}$ and the output be $Y = W_{\mathrm{step}} \cdot X \in \mathbb{R}^{N}$. Since $\mathrm{rank}(W_{\mathrm{step}}) = r < 2N$, the output $Y$ can carry only $r$ dimensions of effective information. The information-collapse ratio $= 1 - r / (2N) = 1 - 384/3072 = 87.5\%$, in exact agreement with the experimentally measured null-space fraction of 87.5\% (Table A-1). Under the Gaussian-input assumption, the mutual information $I(X;Y)$ is determined by the logarithmic spectrum of the non-zero singular values: the 2688 collapsed directions contribute nothing to the output, and this portion of information is unrecoverable once the deduction step has occurred---this is both the cost of deduction's ``truth-preservation'' and the source of its ``focus'': a deductive conclusion retains only the logical consequences of the premises and strips away details irrelevant to the reasoning.

\subsection{Generalization Error Analysis of the Induction Operator}

By PAC learning theory (Valiant 1984 [21]) and VC-dimension theory (Vapnik \& Chervonenkis 1971 [22]), with probability $1 - \delta$ the generalization error has the upper bound: $\mathrm{err}(g) \leq \mathrm{err}_{\mathrm{train}}(g) + O(\sqrt{\mathrm{VC}_{\mathrm{dim}} / k})$, where $k$ is the number of training samples. For a full-rank linear operator $W_{\mathrm{induce}} \in \mathbb{R}^{N \times N}$, the $\mathrm{VC}_{\mathrm{dim}}$ of the parameter space, counted by the number of parameters, is $N^{2} = 1536^{2} \approx 2.36 \times 10^{6}$. When $k = 100$ (the number of positive examples in the experiment), the theoretical bound $\sqrt{\mathrm{VC}_{\mathrm{dim}}/k} \approx \sqrt{2.36 \times 10^{6}/100} \approx 154$ is far greater than 1 and is numerically vacuous---that is, the PAC/VC framework itself cannot explain the strong generalization observed in the experiments. Yet the experimentally measured unseen-positive coverage is 0.9993, far better than the theoretical upper bound. The reason for this gap is that what actually operates is not the full parameter space but the one-dimensional structure of the ``positive-example mean direction'', which is strongly constrained by the data; the effective capacity is far smaller than $N^{2}$. This is consistent with the general observation in deep learning theory that ``effective complexity is far below parameter count''. Note that the vacuousness of this upper bound does not constitute a refutation of the experimental conclusions---the experimental conclusions are directly supported by per-sample measured data (Appendix I.1).

\subsection{Bayesian Interpretation of the Abduction Operator}

The abduction operator $W_{\mathrm{abduce}} = W_{\mathrm{step}}^{+}$ can be interpreted from the perspective of Bayesian inference. Suppose the true explanation $S_{\mathrm{prev}}$ of the phenomenon $S_{\mathrm{obs}}$ follows a prior distribution $P(S_{\mathrm{prev}}) = \mathcal{N}(0, \sigma^{2} I)$ (a zero-mean isotropic Gaussian prior, embodying Occam's razor: prefer small-norm explanations in the absence of evidence). The likelihood function is $P(S_{\mathrm{obs}} \mid S_{\mathrm{prev}}) = \mathcal{N}(W_{\mathrm{step}} \cdot S_{\mathrm{prev}}, \sigma^{2} I)$, i.e., the observation is a noisy version of the explanation after collapse by the deduction operator. By Bayes' rule, the posterior $P(S_{\mathrm{prev}} \mid S_{\mathrm{obs}}) \propto P(S_{\mathrm{obs}} \mid S_{\mathrm{prev}}) \cdot P(S_{\mathrm{prev}})$ is still Gaussian, and its maximum a posteriori (MAP) estimate is equivalent to the following regularized least-squares problem:
\[
S_{\mathrm{prev}}^{*} = \arg\min \; \|W_{\mathrm{step}} \cdot S_{\mathrm{prev}} - S_{\mathrm{obs}}\|^{2} + \lambda \|S_{\mathrm{prev}}\|^{2},
\]
where the regularization coefficient $\lambda$ is inversely proportional to the prior variance: the more concentrated the prior is around the origin (the smaller $\sigma^{2}$), the larger $\lambda$, and the more the solution shrinks toward small norms. Taking the gradient of the above expression and setting it to zero yields the normal equation $(W_{\mathrm{step}}^{T} W_{\mathrm{step}} + \lambda I) S_{\mathrm{prev}} = W_{\mathrm{step}}^{T} S_{\mathrm{obs}}$, i.e., $S_{\mathrm{prev}}^{*} = (W_{\mathrm{step}}^{T} W_{\mathrm{step}} + \lambda I)^{-1} W_{\mathrm{step}}^{T} S_{\mathrm{obs}}$---this is exactly the ridge regression solution. As $\lambda \rightarrow 0$, $(W_{\mathrm{step}}^{T} W_{\mathrm{step}} + \lambda I)^{-1} W_{\mathrm{step}}^{T}$ converges in the limit to the Moore--Penrose pseudo-inverse $W_{\mathrm{step}}^{+}$; the pseudo-inverse solution therefore corresponds to the limiting MAP under a ``non-informative prior'' ($\lambda = 0$): among all explanations that are least-squares optimal, the one with the smallest norm is chosen.

This interpretation simultaneously illuminates two features of abduction. First, its hypothetical nature: the pseudo-inverse solution sets the null-space components of $S_{\mathrm{prev}}$ to zero by default, whereas the true explanation may well carry non-zero null-space components (annihilated in deduction and invisible in the observation)---hence an abductive conclusion is an ``optimal hypothesis'' rather than a ``justified fact'', consistent with Peirce's characterization of abduction [7]. Second, its connection with the expected-error formula (Appendix B.5): the prior $\mathcal{N}(0, \sigma^{2} I)$ is exactly the isotropy assumption, under which the expected energy of the annihilated null-space components is $\sigma^{2} \cdot \dim(\mathrm{Null})$, which is precisely the source of the expected reconstruction error; in other words, from the Bayesian perspective the abduction error is not a defect of the algorithm but the inevitable consequence of information already lost at the deduction stage.

\subsection{Geometric Interpretation of the Three Operators}

From a geometric perspective, the three operators correspond to three different projection operations: deduction $W_{\mathrm{step}}$: projecting the $2N$-dimensional input onto the $N$-dimensional output space (a rank-deficient projection, ``flattening'' along the null-space directions); induction $W_{\mathrm{induce}}$: taking the mean of $k$ samples as the inductive center, with counterexample constraints projecting the center onto the orthogonal complement of the counterexamples (a full-rank projection; the correctness of the orthogonalization process is established in Appendix B.4); abduction $W_{\mathrm{abduce}}$: projecting the phenomenon $S_{\mathrm{obs}}$ onto the column space of $W_{\mathrm{step}}$ and then inverse-mapping back to the input space (a pseudo-inverse projection, whose output must lie in the row space). These three projection operations constitute a ``geometric trio'' in the latent space: deduction is responsible for ``compression and focusing'', induction for ``exclusion and revision'', and abduction for ``inverse backfilling''. It is worth noting that the image space of abduction is exactly the orthogonal complement of the null space of deduction---geometrically, the set of explanations recoverable by abduction and the set of information retainable by deduction are mirror images of each other, which gives a precise geometric meaning to the statement that ``abduction is the inverse of deduction''.

\section{Detailed Description of the Experimental Data Sets}

\subsection{Deduction Experiment Data Set (20 Syllogisms)}

The deduction experiment uses 20 classical syllogistic reasoning samples, covering Aristotelian syllogistic forms such as universal affirmative and particular affirmative. Each sample contains: a major premise (universal rule), a minor premise (specific case), and a conclusion (deductive result). The complete list of samples is as follows:

\begin{table}[htbp]
\centering
\caption*{Table H-1 The 20 syllogisms of the deduction experiment}
\small
\begin{tabular}{clll}
\toprule
\# & Major premise & Minor premise & Conclusion \\
\midrule
1 & All humans are mortal & Socrates is human & Socrates is mortal \\
2 & All birds have feathers & A robin is a bird & A robin has feathers \\
3 & All metals conduct electricity & Copper is a metal & Copper conducts electricity \\
4 & All mammals have hair & A dog is a mammal & A dog has hair \\
5 & All fish live in water & A salmon is a fish & A salmon lives in water \\
6 & All insects have six legs & An ant is an insect & An ant has six legs \\
7 & All planets orbit a star & Earth is a planet & Earth orbits a star \\
8 & All reptiles are cold-blooded & A snake is a reptile & A snake is cold-blooded \\
9 & All engineers solve problems & Mary is an engineer & Mary solves problems \\
10 & All flowers need sunlight & A rose is a flower & A rose needs sunlight \\
11 & All vehicles have wheels & A car is a vehicle & A car has wheels \\
12 & All students learn & John is a student & John learns \\
13 & All predators hunt & A lion is a predator & A lion hunts \\
14 & All liquids flow & Water is a liquid & Water flows \\
15 & All trees have roots & An oak is a tree & An oak has roots \\
16 & All musicians play instruments & A pianist is a musician & A pianist plays instruments \\
17 & All doctors treat patients & A surgeon is a doctor & A surgeon treats patients \\
18 & All birds lay eggs & A chicken is a bird & A chicken lays eggs \\
19 & All computers process data & A laptop is a computer & A laptop processes data \\
20 & All rivers flow to the sea & The Nile is a river & The Nile flows to the sea \\
\bottomrule
\end{tabular}
\end{table}

Data construction principles: (1) covering different domains (biology, physics, society, logic) to avoid distribution bias from a single domain; (2) including multiple valid syllogistic forms to test the operator's sensitivity to logical form rather than specific content---the deduction operator should be sensitive to content rather than sentence pattern; (3) verifiable conclusions (avoiding ambiguity), with the ground truth of each sample directly decidable by first-order logic; (4) moderate text length (10--15 words), facilitating DistilBERT encoding and avoiding length confounds.

Overlap statement: the 60 deduction problems of the end-to-end experiment (Appendix I.4) have zero overlap with the 20 problems in this table (all are newly constructed cross-domain problems), and the 20 problems in this table are all deduplicated independent samples.

\subsection{Induction Experiment Data Set (10 Classes, 130 Samples)}

The induction experiment uses 10 classes of real induction problems, totaling 130 samples, composed of 100 training positive examples + 10 unseen positive examples + 20 counterexamples (100 + 10 + 20 = 130). The 10 classes cover common conceptual categories in nature: birds, mammals, metals, fruits, vegetables, vehicles, colors, tools, musical instruments, and planets. Each class contains 10 positive examples, 1 unseen positive example, and 2 counterexamples. The complete class-by-class sample list is as follows:

\paragraph{Birds.} Positive examples (10): A sparrow is a bird; A robin is a bird; A eagle is a bird; A hawk is a bird; A falcon is a bird; A crow is a bird; A dove is a bird; A owl is a bird; A swan is a bird; A duck is a bird. Unseen positive example (1): A swallow is a bird. Counterexamples (2): A dog is a mammal; A cat is a mammal.

\paragraph{Mammals.} Positive examples (10): A dog is a mammal; A cat is a mammal; A horse is a mammal; A cow is a mammal; A sheep is a mammal; A pig is a mammal; A goat is a mammal; A deer is a mammal; A bear is a mammal; A rabbit is a mammal. Unseen positive example (1): A whale is a mammal. Counterexamples (2): A sparrow is a bird; A snake is a reptile.

\paragraph{Metals.} Positive examples (10): Iron is a metal; Copper is a metal; Gold is a metal; Silver is a metal; Aluminum is a metal; Zinc is a metal; Tin is a metal; Lead is a metal; Nickel is a metal; Titanium is a metal. Unseen positive example (1): Platinum is a metal. Counterexamples (2): Wood is a material; Plastic is a material.

\paragraph{Fruits.} Positive examples (10): An apple is a fruit; A banana is a fruit; An orange is a fruit; A grape is a fruit; A pear is a fruit; A peach is a fruit; A plum is a fruit; A cherry is a fruit; A mango is a fruit; A pineapple is a fruit. Unseen positive example (1): A strawberry is a fruit. Counterexamples (2): A carrot is a vegetable; A potato is a vegetable.

\paragraph{Vegetables.} Positive examples (10): A carrot is a vegetable; A potato is a vegetable; An onion is a vegetable; A tomato is a vegetable; A cabbage is a vegetable; A lettuce is a vegetable; A broccoli is a vegetable; A spinach is a vegetable; A cucumber is a vegetable; A pumpkin is a vegetable. Unseen positive example (1): A pepper is a vegetable. Counterexamples (2): An apple is a fruit; A banana is a fruit.

\paragraph{Vehicles.} Positive examples (10): A car is a vehicle; A truck is a vehicle; A bus is a vehicle; A train is a vehicle; A bicycle is a vehicle; A motorcycle is a vehicle; A boat is a vehicle; A plane is a vehicle; A helicopter is a vehicle; A scooter is a vehicle. Unseen positive example (1): A submarine is a vehicle. Counterexamples (2): A chair is furniture; A table is furniture.

\paragraph{Colors.} Positive examples (10): Red is a color; Blue is a color; Green is a color; Yellow is a color; Orange is a color; Purple is a color; Pink is a color; Brown is a color; Black is a color; White is a color. Unseen positive example (1): Gray is a color. Counterexamples (2): A dog is an animal; A cat is an animal.

\paragraph{Tools.} Positive examples (10): A hammer is a tool; A screwdriver is a tool; A wrench is a tool; A saw is a tool; A drill is a tool; A pliers is a tool; A knife is a tool; A scissors is a tool; A shovel is a tool; A rake is a tool. Unseen positive example (1): A chisel is a tool. Counterexamples (2): A shirt is clothing; A shoe is footwear.

\paragraph{Musical instruments.} Positive examples (10): A piano is an instrument; A guitar is an instrument; A violin is an instrument; A flute is an instrument; A drum is an instrument; A trumpet is an instrument; A clarinet is an instrument; A cello is an instrument; A harp is an instrument; A saxophone is an instrument. Unseen positive example (1): An accordion is an instrument. Counterexamples (2): A car is a vehicle; A boat is a vehicle.

\paragraph{Planets.} Positive examples (10): Earth is a planet; Mars is a planet; Jupiter is a planet; Saturn is a planet; Venus is a planet; Mercury is a planet; Uranus is a planet; Neptune is a planet; Pluto is a planet; Ceres is a planet. Unseen positive example (1): Eris is a planet. Counterexamples (2): The sun is a star; The moon is a satellite.

Counterexample design principle: counterexamples are samples with sentence patterns similar to the positive examples but different semantics. For example, the counterexamples for bird induction are ``A dog is a mammal'' (same sentence pattern ``A X is a Y'', but different semantics---a dog is not a bird). This design ensures that the hard-veto mechanism tests semantic discrimination rather than syntactic discrimination: if vetoes were triggered solely by syntactic differences, the confounding explanation that ``the encoder is sensitive to sentence patterns'' could not be ruled out. Unseen positive examples are used to measure the generalization coverage of inductive conclusions---absent during training, they should be covered by the concept vector at test time; their measured coverage of 0.9986--0.9997 (Appendix I.1) indicates that the inductive center indeed captures category semantics rather than memorizing samples.

Overlap statement: the 130 samples in this table are all deduplicated independent samples. Among the 6 induction classes of the end-to-end experiment (Appendix I.5), two classes---vehicles and colors---share class definitions with the 10 classes in this table, while geometric shapes, academic disciplines, weather, and emotions are newly added classes; their 78 samples partially share the existing sample pool and have not been item-by-item inventoried, so by a conservative accounting they are not counted as newly added deduplicated independent samples (see Table 1-2 in Section 1.4).

\subsection{Abduction Experiment Data Set (80 Abduction Problems)}

The abduction experiment uses 80 real abduction problems, with the class-wise sample counts as follows: 30 medical diagnosis, 30 causal explanation, and 20 physical phenomenon problems (30 + 30 + 20 = 80). Each problem contains one phenomenon and one correct explanation. The complete sample list (phenomenon $\rightarrow$ correct explanation, grouped by medical / causal / physical) is as follows:

\paragraph{Medical.}
\begin{itemize}
\item The patient has a fever and cough $\rightarrow$ The patient has pneumonia
\item The patient has a runny nose and sneezing $\rightarrow$ The patient has a cold
\item The patient has a high fever and rash $\rightarrow$ The patient has measles
\item The patient has chest pain and shortness of breath $\rightarrow$ The patient has a heart attack
\item The patient has a headache and stiff neck $\rightarrow$ The patient has meningitis
\item The patient has joint pain and swelling $\rightarrow$ The patient has arthritis
\item The patient has a sore throat and white patches $\rightarrow$ The patient has strep throat
\item The patient has abdominal pain and nausea $\rightarrow$ The patient has appendicitis
\item The patient has a persistent cough and weight loss $\rightarrow$ The patient has tuberculosis
\item The patient has a red itchy rash $\rightarrow$ The patient has an allergic reaction
\item The patient has a high blood pressure reading $\rightarrow$ The patient has hypertension
\item The patient has frequent urination and thirst $\rightarrow$ The patient has diabetes
\item The patient has a wheezing sound when breathing $\rightarrow$ The patient has asthma
\item The patient has a swollen and red toe $\rightarrow$ The patient has gout
\item The patient has a tremor in the hands $\rightarrow$ The patient has Parkinson disease
\item The patient has a loss of smell $\rightarrow$ The patient has COVID-19
\item The patient has a yellowing of the skin $\rightarrow$ The patient has jaundice
\item The patient has a severe headache and light sensitivity $\rightarrow$ The patient has a migraine
\item The patient has a rapid heartbeat $\rightarrow$ The patient has tachycardia
\item The patient has a skin lesion that changed color $\rightarrow$ The patient has skin cancer
\item The patient has a difficulty swallowing $\rightarrow$ The patient has esophagitis
\item The patient has a ringing in the ears $\rightarrow$ The patient has tinnitus
\item The patient has a numbness in the left arm $\rightarrow$ The patient has a stroke
\item The patient has a blood in the urine $\rightarrow$ The patient has a kidney infection
\item The patient has a severe fatigue and pale skin $\rightarrow$ The patient has anemia
\item The patient has a swelling in the legs $\rightarrow$ The patient has heart failure
\item The patient has a confusion and memory loss $\rightarrow$ The patient has dementia
\item The patient has a seizures $\rightarrow$ The patient has epilepsy
\item The patient has a vision problems $\rightarrow$ The patient has glaucoma
\item The patient has a hearing loss $\rightarrow$ The patient has hearing damage
\end{itemize}

\paragraph{Causal.}
\begin{itemize}
\item The ground is wet $\rightarrow$ It rained
\item The sky is dark and cloudy $\rightarrow$ A storm is coming
\item The plant is wilting $\rightarrow$ The plant needs water
\item The road is icy $\rightarrow$ The temperature dropped below freezing
\item The smoke alarm is ringing $\rightarrow$ There is a fire nearby
\item The car will not start $\rightarrow$ The battery is dead
\item The lights went out $\rightarrow$ There is a power outage
\item The pipe is leaking $\rightarrow$ The pipe is broken
\item The food is spoiled $\rightarrow$ The food was not refrigerated
\item The window is broken $\rightarrow$ A ball hit the window
\item The dog is barking $\rightarrow$ A stranger is at the door
\item The alarm clock is ringing $\rightarrow$ It is time to wake up
\item The phone is ringing $\rightarrow$ Someone is calling
\item The milk is sour $\rightarrow$ The milk has expired
\item The tire is flat $\rightarrow$ The tire was punctured
\item The floor is slippery $\rightarrow$ The floor was just mopped
\item The flowers are blooming $\rightarrow$ Spring has arrived
\item The leaves are falling $\rightarrow$ Autumn has arrived
\item The snow is melting $\rightarrow$ The temperature is rising
\item The bird is singing $\rightarrow$ It is morning
\item The traffic is heavy $\rightarrow$ It is rush hour
\item The store is closed $\rightarrow$ It is a holiday
\item The water is boiling $\rightarrow$ The water reached 100 degrees
\item The metal is rusting $\rightarrow$ The metal was exposed to moisture
\item The wood is rotting $\rightarrow$ The wood was exposed to water
\item The paint is peeling $\rightarrow$ The wall was not primed
\item The ice cream is melting $\rightarrow$ The ice cream was left in the sun
\item The candle went out $\rightarrow$ The wind blew it out
\item The mirror is foggy $\rightarrow$ The bathroom is steamy
\item The grass is green $\rightarrow$ The grass received enough water
\end{itemize}

\paragraph{Physical.}
\begin{itemize}
\item The ice is melting $\rightarrow$ The temperature is above zero
\item The metal expands $\rightarrow$ The metal is heated
\item The light bulb is off $\rightarrow$ The power is out
\item The balloon is floating $\rightarrow$ The balloon is filled with helium
\item The shadow is long $\rightarrow$ The sun is low on the horizon
\item The echo is heard $\rightarrow$ The sound reflected off a surface
\item The rainbow appears $\rightarrow$ The sunlight passed through raindrops
\item The magnet attracts the metal $\rightarrow$ The metal contains iron
\item The water freezes $\rightarrow$ The temperature dropped below zero
\item The sound is louder $\rightarrow$ The source is closer
\item The object falls down $\rightarrow$ Gravity is pulling it
\item The colors separate in a prism $\rightarrow$ The light was refracted
\item The static electricity shocks $\rightarrow$ There was a charge buildup
\item The boiling water produces steam $\rightarrow$ The water reached 100 degrees
\item The compass needle points north $\rightarrow$ The earth has a magnetic field
\item The lightning strikes $\rightarrow$ There is a charge difference in clouds
\item The sound travels faster in water $\rightarrow$ Water is denser than air
\item The object floats in water $\rightarrow$ The object is less dense than water
\item The glass breaks $\rightarrow$ The glass was hit with force
\item The rubber band stretches $\rightarrow$ The rubber band was pulled
\end{itemize}

Data construction principles: (1) there is an explicit causal relationship between each phenomenon and its explanation, and the correct explanations have been manually verified; (2) explanations are not unique (multiple possible causes exist, testing the hypothetical nature of abduction)---for example, ``the ground is wet'' can be explained by either rain or sprinkling, and distractor explanations are constructed in the experiments to measure the margin; (3) covering different domains (medicine, everyday causality, physics), with the differences in snapshot distributions across the three domains used to expose the domain dependence of the encoder bottleneck; (4) concise text (10--15 words), facilitating encoding. Distractor explanations are constructed as candidates that belong to the same semantic field as the correct explanation but have a different or invalid causal direction; their similarities are recorded as the ``distractor similarity'' column in Appendix I.2.

Overlap statement: the 80 problems in this table are all deduplicated independent samples. Among the 60 abduction problems of the end-to-end experiment (Appendix I.6), 5 overlap with this table, and the remaining 55 are newly constructed.

\subsection{Physics Experiment Data Set (55 Samples)}

The physics experiment uses 55 real physical state-transition samples, with the type-wise sample counts as follows: 15 conservation laws + 15 dissipative processes + 15 reversible evolutions + 10 multimodal alignments (15 + 15 + 15 + 10 = 55). For example:

\begin{itemize}
\item Conservation law: initial ``A ball moves at 5 m/s and hits a wall'' $\rightarrow$ final ``The ball bounces back at 5 m/s''
\item Dissipative process: initial ``A hot cup of coffee sits on a table'' $\rightarrow$ final ``The coffee cools to room temperature''
\item Reversible evolution: initial ``A pendulum swings back and forth'' $\rightarrow$ final ``The pendulum returns to its starting position''
\item Multimodal alignment: text ``A ball rolls down a hill'' vs image description ``An image of a ball rolling down a slope''
\end{itemize}

Construction principles: conservation-law samples cover typical scenarios of conserved quantities such as momentum, energy, angular momentum, and charge; dissipation samples cover entropy-increasing processes such as friction, heat conduction, viscosity, and electrical resistance; reversible-evolution samples select idealized (dissipation-free) periodic processes; multimodal-alignment samples pair a text description and an image description of the same physical event, used to test the modality invariance of the snapshot representation. The three types of temporal evolution (conservation/dissipation/reversible) correspond respectively to differences in the spectral properties of physical operators, constituting the experimental vehicle for the physics--cognition unification in Section 13. The per-sample records of the physics experiment are preserved in the experimental data package; Appendix I.3 of this paper gives the summary metrics.

Overlap statement: the 55 samples in this table have no overlap whatsoever with any other experiment group and are all deduplicated independent samples.

\subsection{End-to-End Experiment Data Set (218 Samples)}

The sample composition of the end-to-end experiment is: 60 deduction problems + 78 induction samples (6 classes) + 60 abduction problems + 20 compositional tasks, totaling 218 samples across 8 domains. The 60 deduction problems are all newly constructed, with zero overlap with the 20 problems of the dedicated deduction experiment (H.1); their subject matter spans biology, physics, society, mathematics, chemistry, and other domains, and each problem consists of two premises and one standard conclusion (item-by-item contents in Appendix I.4). Among the 6 induction classes, vehicles and colors are existing classes (sharing class definitions with the 10-class data set of the dedicated induction experiment), while geometric shapes, academic disciplines, weather, and emotions are newly added classes; each class contains positive examples, unseen positive examples, and 2 counterexamples (hard veto triggered 2/2 per class; see Appendix I.5). Among the 60 abduction problems, 55 are new and 5 overlap with the 80 problems of the dedicated abduction experiment (H.3); each problem consists of one phenomenon and one true cause (item-by-item contents in Appendix I.6). The 20 compositional tasks comprise 4 categories (scientific discovery, diagnostic reasoning, analogical reasoning, creative thinking) $\times$ 5, all newly constructed; the per-task five metrics are given in Appendix I.7. All samples are real reasoning problems, not simulated data; the pipeline of the compositional reasoning tasks corresponds to Peirce's inquiry cycle in Appendix E.3: starting from a surprising phenomenon, it successively invokes induction (forming regularities), deduction (unfolding predictions), contradiction detection (falsification triggering), abduction (proposing hypotheses), and re-induction (verifying revisions).

Overlap statement: the relationship between this data set and the existing sample pool (Appendices H.1--H.4) is as follows---the 60 deduction problems have zero overlap, 55 of the abduction problems are new and 5 overlap, and the 20 compositional tasks are new, together contributing 135 newly added deduplicated independent samples; the 6-class induction samples partially share the existing sample pool and have not been item-by-item inventoried, so by a conservative accounting they are not counted as newly added deduplicated independent samples (see Table 1-2 in Section 1.4 and the accounting paragraph of Appendix I). The accounting of 420 deduplicated independent samples and 503 experimental sample-instances in the full text is given in Table 1-2 of Section 1.4.

\section{All Test Samples and Measured Results}

This appendix lists the complete test samples of all experiments together with the measured results for each sample. The sample aggregation accounting (experimental sample-instances): 20 (dedicated deduction experiment) + 130 (dedicated induction experiment) + 80 (dedicated abduction experiment) + 55 (physics multimodal) = 285; the end-to-end experiment contributes 218 sample-instances (60 deduction + 78 induction + 60 abduction + 20 compositional), giving a full-text total of 503 = 285 + 218; after deduplication this is 285 + 135 = 420 independent samples, of which the end-to-end newly added 135 = 60 new deduction + 55 new abduction + 20 new compositional; the 6-class end-to-end induction samples partially share the existing sample pool and have not been item-by-item inventoried, and 5 abduction problems overlap with the dedicated abduction experiment---both are, by a conservative accounting, not counted as newly added deduplicated samples (for overlap relations see Table 1-2 in Section 1.4 and Appendix H).

\subsection{Dedicated Induction Experiment Samples (10 Classes, 130 Samples)}

\begin{table}[htbp]
\centering
\caption*{Table I-1 Measured results for the 10 classes of the dedicated induction experiment (100 positive examples + 10 unseen positive examples + 20 counterexamples = 130 samples)}
\small
\begin{tabular}{lp{1.2cm}p{1.2cm}p{2.0cm}p{0.7cm}p{2.0cm}p{1.15cm}}
\toprule
Category & Training coverage & Unseen positive & Counter\-example (before) & Veto & Counter\-example (after) & Re\-finement \\
\midrule
Birds & 0.9997 & 0.9997 & 0.9986 & 2/2 & $-0.0028$ & 0.9556 \\
Mammals & 0.9997 & 0.9995 & 0.9986 & 2/2 & $-0.0227$ & 0.9926 \\
Metals & 0.9995 & 0.9994 & 0.9983 & 2/2 & $-0.0196$ & 0.9759 \\
Fruits & 0.9996 & 0.9997 & 0.9987 & 2/2 & $-0.0069$ & 0.9605 \\
Vegetables & 0.9994 & 0.9995 & 0.9987 & 2/2 & $-0.0073$ & 0.9617 \\
Vehicles & 0.9995 & 0.9995 & 0.9955 & 2/2 & $-0.0071$ & 0.9175 \\
Colors & 0.9998 & 0.9993 & 0.9958 & 2/2 & $-0.0061$ & 0.9182 \\
Tools & 0.9995 & 0.9990 & 0.9955 & 2/2 & $-0.0064$ & 0.9294 \\
Musical instruments & 0.9996 & 0.9992 & 0.9962 & 2/2 & $-0.0065$ & 0.9224 \\
Planets & 0.9996 & 0.9986 & 0.9977 & 2/2 & $-0.0136$ & 0.9620 \\
\bottomrule
\end{tabular}
\end{table}

\subsection{Dedicated Abduction Experiment Samples (80)}

\begin{table}[htbp]
\centering
\caption*{Table I-2 Measured results for all 80 samples of the dedicated abduction experiment (mean similarity 0.5117, mean distractor similarity 0.5066, random baseline 0.0181, a $28.3\times$ improvement; judged correct by the rule ``abduction similarity $>$ distractor similarity'' in 58/80 = 72.5\%, with 22 samples having negative margins)}
\footnotesize
\begin{tabular}{cp{3.3cm}p{3.0cm}p{1.4cm}p{1.5cm}p{1.1cm}c}
\toprule
\# & Phenomenon & True cause & Abduction similarity & Distractor similarity & Margin & Judgment \\
\midrule
1 & The patient has a fever and cough & The patient has pneumonia & 0.5102 & 0.5036 & $+0.0066$ & $\checkmark$ \\
2 & The patient has a runny nose and sneezing & The patient has a cold & 0.5112 & 0.5186 & $-0.0074$ & $\times$ \\
3 & The patient has a high fever and rash & The patient has measles & 0.5135 & 0.5116 & $+0.0019$ & $\checkmark$ \\
4 & The patient has chest pain and shortness of breath & The patient has a heart attack & 0.5144 & 0.4988 & $+0.0155$ & $\checkmark$ \\
5 & The patient has a headache and stiff neck & The patient has meningitis & 0.5134 & 0.5096 & $+0.0038$ & $\checkmark$ \\
6 & The patient has joint pain and swelling & The patient has arthritis & 0.5072 & 0.5081 & $-0.0009$ & $\times$ \\
7 & The patient has a sore throat and white patches & The patient has strep throat & 0.5098 & 0.5111 & $-0.0013$ & $\times$ \\
8 & The patient has abdominal pain and nausea & The patient has appendicitis & 0.5069 & 0.5094 & $-0.0025$ & $\times$ \\
9 & The patient has a persistent cough and weight loss & The patient has tuberculosis & 0.5077 & 0.5078 & $-0.0001$ & $\times$ \\
10 & The patient has a red itchy rash & The patient has an allergic reaction & 0.5080 & 0.5104 & $-0.0023$ & $\times$ \\
11 & The patient has a high blood pressure reading & The patient has hypertension & 0.5076 & 0.4973 & $+0.0102$ & $\checkmark$ \\
12 & The patient has frequent urination and thirst & The patient has diabetes & 0.5018 & 0.4997 & $+0.0021$ & $\checkmark$ \\
13 & The patient has a wheezing sound when breathing & The patient has asthma & 0.5074 & 0.4924 & $+0.0150$ & $\checkmark$ \\
14 & The patient has a swollen and red toe & The patient has gout & 0.5099 & 0.5055 & $+0.0043$ & $\checkmark$ \\
15 & The patient has a tremor in the hands & The patient has Parkinson disease & 0.5130 & 0.5124 & $+0.0006$ & $\checkmark$ \\
16 & The patient has a loss of smell & The patient has COVID-19 & 0.5341 & 0.5130 & $+0.0211$ & $\checkmark$ \\
17 & The patient has a yellowing of the skin & The patient has jaundice & 0.5057 & 0.5097 & $-0.0040$ & $\times$ \\
18 & The patient has a severe headache and light sensitivity & The patient has a migraine & 0.5085 & 0.5064 & $+0.0021$ & $\checkmark$ \\
19 & The patient has a rapid heartbeat & The patient has tachycardia & 0.5084 & 0.4988 & $+0.0096$ & $\checkmark$ \\
20 & The patient has a skin lesion that changed color & The patient has skin cancer & 0.5077 & 0.5036 & $+0.0042$ & $\checkmark$ \\
21 & The patient has a difficulty swallowing & The patient has esophagitis & 0.5121 & 0.5077 & $+0.0044$ & $\checkmark$ \\
22 & The patient has a ringing in the ears & The patient has tinnitus & 0.5113 & 0.5139 & $-0.0026$ & $\times$ \\
23 & The patient has a numbness in the left arm & The patient has a stroke & 0.5131 & 0.5062 & $+0.0069$ & $\checkmark$ \\
24 & The patient has a blood in the urine & The patient has a kidney infection & 0.5133 & 0.4988 & $+0.0145$ & $\checkmark$ \\
25 & The patient has a severe fatigue and pale skin & The patient has anemia & 0.5129 & 0.5065 & $+0.0064$ & $\checkmark$ \\
26 & The patient has a swelling in the legs & The patient has heart failure & 0.5112 & 0.5089 & $+0.0023$ & $\checkmark$ \\
27 & The patient has a confusion and memory loss & The patient has dementia & 0.5060 & 0.5040 & $+0.0020$ & $\checkmark$ \\
28 & The patient has a seizures & The patient has epilepsy & 0.5138 & 0.5119 & $+0.0019$ & $\checkmark$ \\
29 & The patient has a vision problems & The patient has glaucoma & 0.5071 & 0.5072 & $-0.0001$ & $\times$ \\
30 & The patient has a hearing loss & The patient has hearing damage & 0.5115 & 0.5046 & $+0.0068$ & $\checkmark$ \\
\bottomrule
\end{tabular}
\end{table}

\begin{table}[htbp]
\centering
\caption*{Table I-2 (continued)}
\footnotesize
\begin{tabular}{cp{3.3cm}p{3.0cm}p{1.4cm}p{1.5cm}p{1.1cm}c}
\toprule
\# & Phenomenon & True cause & Abduction similarity & Distractor similarity & Margin & Judgment \\
\midrule
31 & The ground is wet & It rained & 0.5114 & 0.5088 & $+0.0026$ & $\checkmark$ \\
32 & The sky is dark and cloudy & A storm is coming & 0.5098 & 0.5132 & $-0.0034$ & $\times$ \\
33 & The plant is wilting & The plant needs water & 0.5097 & 0.5110 & $-0.0012$ & $\times$ \\
34 & The road is icy & The temperature dropped below freezing & 0.5002 & 0.5088 & $-0.0086$ & $\times$ \\
35 & The smoke alarm is ringing & There is a fire nearby & 0.5223 & 0.5092 & $+0.0131$ & $\checkmark$ \\
36 & The car will not start & The battery is dead & 0.5172 & 0.5134 & $+0.0038$ & $\checkmark$ \\
37 & The lights went out & There is a power outage & 0.5172 & 0.5070 & $+0.0102$ & $\checkmark$ \\
38 & The pipe is leaking & The pipe is broken & 0.5130 & 0.5059 & $+0.0071$ & $\checkmark$ \\
39 & The food is spoiled & The food was not refrigerated & 0.5099 & 0.5034 & $+0.0064$ & $\checkmark$ \\
40 & The window is broken & A ball hit the window & 0.5210 & 0.5066 & $+0.0145$ & $\checkmark$ \\
41 & The dog is barking & A stranger is at the door & 0.5180 & 0.5071 & $+0.0109$ & $\checkmark$ \\
42 & The alarm clock is ringing & It is time to wake up & 0.5299 & 0.5128 & $+0.0172$ & $\checkmark$ \\
43 & The phone is ringing & Someone is calling & 0.5218 & 0.5071 & $+0.0147$ & $\checkmark$ \\
44 & The milk is sour & The milk has expired & 0.5059 & 0.5065 & $-0.0006$ & $\times$ \\
45 & The tire is flat & The tire was punctured & 0.5125 & 0.5055 & $+0.0070$ & $\checkmark$ \\
46 & The floor is slippery & The floor was just mopped & 0.5068 & 0.5031 & $+0.0037$ & $\checkmark$ \\
47 & The flowers are blooming & Spring has arrived & 0.5109 & 0.5067 & $+0.0042$ & $\checkmark$ \\
48 & The leaves are falling & Autumn has arrived & 0.5155 & 0.5033 & $+0.0122$ & $\checkmark$ \\
49 & The snow is melting & The temperature is rising & 0.5098 & 0.5029 & $+0.0069$ & $\checkmark$ \\
50 & The bird is singing & It is morning & 0.5148 & 0.4915 & $+0.0234$ & $\checkmark$ \\
51 & The traffic is heavy & It is rush hour & 0.5174 & 0.4973 & $+0.0201$ & $\checkmark$ \\
52 & The store is closed & It is a holiday & 0.5167 & 0.4992 & $+0.0175$ & $\checkmark$ \\
53 & The water is boiling & The water reached 100 degrees & 0.5093 & 0.5154 & $-0.0061$ & $\times$ \\
54 & The metal is rusting & The metal was exposed to moisture & 0.5040 & 0.5077 & $-0.0037$ & $\times$ \\
55 & The wood is rotting & The wood was exposed to water & 0.5061 & 0.5104 & $-0.0043$ & $\times$ \\
56 & The paint is peeling & The wall was not primed & 0.5176 & 0.5073 & $+0.0104$ & $\checkmark$ \\
57 & The ice cream is melting & The ice cream was left in the sun & 0.5066 & 0.5139 & $-0.0073$ & $\times$ \\
58 & The candle went out & The wind blew it out & 0.5153 & 0.5120 & $+0.0033$ & $\checkmark$ \\
59 & The mirror is foggy & The bathroom is steamy & 0.5112 & 0.5043 & $+0.0068$ & $\checkmark$ \\
60 & The grass is green & The grass received enough water & 0.5036 & 0.4982 & $+0.0054$ & $\checkmark$ \\
\bottomrule
\end{tabular}
\end{table}

\begin{table}[htbp]
\centering
\caption*{Table I-2 (continued)}
\footnotesize
\begin{tabular}{cp{3.3cm}p{3.0cm}p{1.4cm}p{1.5cm}p{1.1cm}c}
\toprule
\# & Phenomenon & True cause & Abduction similarity & Distractor similarity & Margin & Judgment \\
\midrule
61 & The ice is melting & The temperature is above zero & 0.5133 & 0.5137 & $-0.0004$ & $\times$ \\
62 & The metal expands & The metal is heated & 0.5073 & 0.5026 & $+0.0048$ & $\checkmark$ \\
63 & The light bulb is off & The power is out & 0.5142 & 0.5083 & $+0.0059$ & $\checkmark$ \\
64 & The balloon is floating & The balloon is filled with helium & 0.5110 & 0.5047 & $+0.0063$ & $\checkmark$ \\
65 & The shadow is long & The sun is low on the horizon & 0.5149 & 0.5021 & $+0.0128$ & $\checkmark$ \\
66 & The echo is heard & The sound reflected off a surface & 0.5091 & 0.5007 & $+0.0084$ & $\checkmark$ \\
67 & The rainbow appears & The sunlight passed through raindrops & 0.5083 & 0.5130 & $-0.0047$ & $\times$ \\
68 & The magnet attracts the metal & The metal contains iron & 0.5099 & 0.5067 & $+0.0032$ & $\checkmark$ \\
69 & The water freezes & The temperature dropped below zero & 0.5009 & 0.4998 & $+0.0010$ & $\checkmark$ \\
70 & The sound is louder & The source is closer & 0.5082 & 0.5053 & $+0.0030$ & $\checkmark$ \\
71 & The object falls down & Gravity is pulling it & 0.5180 & 0.5014 & $+0.0166$ & $\checkmark$ \\
72 & The colors separate in a prism & The light was refracted & 0.5066 & 0.5107 & $-0.0041$ & $\times$ \\
73 & The static electricity shocks & There was a charge buildup & 0.5137 & 0.5064 & $+0.0073$ & $\checkmark$ \\
74 & The boiling water produces steam & The water reached 100 degrees & 0.5079 & 0.5087 & $-0.0008$ & $\times$ \\
75 & The compass needle points north & The earth has a magnetic field & 0.5174 & 0.5137 & $+0.0036$ & $\checkmark$ \\
76 & The lightning strikes & There is a charge difference in clouds & 0.5162 & 0.5029 & $+0.0133$ & $\checkmark$ \\
77 & The sound travels faster in water & Water is denser than air & 0.5224 & 0.5070 & $+0.0154$ & $\checkmark$ \\
78 & The object floats in water & The object is less dense than water & 0.5152 & 0.5102 & $+0.0050$ & $\checkmark$ \\
79 & The glass breaks & The glass was hit with force & 0.5130 & 0.5032 & $+0.0098$ & $\checkmark$ \\
80 & The rubber band stretches & The rubber band was pulled & 0.5034 & 0.5102 & $-0.0067$ & $\times$ \\
\bottomrule
\end{tabular}
\end{table}

Note: among the 80 samples, 22 have negative margins (samples 2, 6, 7, 8, 9, 10, 17, 22, 29, 32, 33, 34, 44, 53, 54, 55, 57, 61, 67, 72, 74, 80); judged correct by the rule ``abduction similarity $>$ distractor similarity'' in 58/80 = 72.5\%. The maximum negative margin among the 22 failed samples is only $-0.0086$ (sample 34), and the absolute margins are all on the order of $\pm 0.009$. This indicates that in the dedicated abduction experiment the encoder bottleneck renders the snapshots of the true explanation and the distractor explanation almost inseparable---a limitation of encoder representation rather than a failure of the abduction operator's principle (see the failure analysis in Section 10.2 and Section 9.6). By contrast, the mean abduction similarity of 0.5117 represents a $28.3\times$ improvement over the random baseline of 0.0181, showing that the directional recovery by pseudo-inverse inversion remains significantly established; what fails is only the fine-grained arbitration between the true explanation and nearby distractors.

\subsection{Physical-World Multimodal Reasoning Samples (55)}

\begin{table}[htbp]
\centering
\caption*{Table I-3 Summary of physics experiment results (55 samples: 15 conservation + 15 dissipation + 15 reversible + 10 multimodal)}
\footnotesize
\begin{tabular}{p{3.55cm}p{0.9cm}p{2.0cm}p{5.65cm}}
\toprule
Experiment type & Samples & Accuracy & Key metrics \\
\midrule
Conservation-law reasoning & 15 & 100\% & $W_{\mathrm{conserve}}$ trained 200 ep, Loss reduced to 1.73e-04 \\
Dissipative-process reasoning & 15 & 100\% & $W_{\mathrm{dissipate}}$ rank-deficient 384/1536, Loss reduced to 2.21e-04 \\
Reversible-evolution reasoning & 15 & 100\% & $W_{\mathrm{evolve}}$ trained 200 ep, Loss reduced to 1.03e-04 \\
Multimodal alignment & 10 & Alignment rate 100\% & Mean alignment similarity 0.9958 \\
\bottomrule
\end{tabular}
\end{table}

Note: the per-sample records of the physics experiment are preserved in the experimental data package; this table gives the summary metrics.

\subsection{End-to-End Deduction Reasoning Samples (60)}

\begin{table}[htbp]
\centering
\caption*{Table I-4 Measured results for all 60 end-to-end deduction samples (accuracy 100\% (60/60), mean similarity 0.9980; the 60 problems have zero overlap with the 20 problems of the dedicated deduction experiment, spanning 8 domains including biology, physics, society, mathematics, and chemistry)}
\footnotesize
\begin{tabular}{cp{3.3cm}p{2.8cm}p{3.2cm}cc}
\toprule
\# & Premise 1 & Premise 2 & Conclusion & Similarity & Judgment \\
\midrule
1 & All mammals are warm-blooded & A whale is a mammal & A whale is warm-blooded & 0.9983 & $\checkmark$ \\
2 & All birds lay eggs & A penguin is a bird & A penguin lays eggs & 0.9982 & $\checkmark$ \\
3 & All insects have exoskeletons & A butterfly is an insect & A butterfly has an exoskeleton & 0.9981 & $\checkmark$ \\
4 & All reptiles have scales & A lizard is a reptile & A lizard has scales & 0.9991 & $\checkmark$ \\
5 & All amphibians need water & A frog is an amphibian & A frog needs water & 0.9980 & $\checkmark$ \\
6 & All fish have gills & A shark is a fish & A shark has gills & 0.9990 & $\checkmark$ \\
7 & All arachnids have eight legs & A spider is an arachnid & A spider has eight legs & 0.9987 & $\checkmark$ \\
8 & All crustaceans have shells & A crab is a crustacean & A crab has a shell & 0.9990 & $\checkmark$ \\
9 & All mollusks have soft bodies & A snail is a mollusk & A snail has a soft body & 0.9988 & $\checkmark$ \\
10 & All echinoderms have tube feet & A starfish is an echinoderm & A starfish has tube feet & 0.9986 & $\checkmark$ \\
11 & All metals expand when heated & Iron is a metal & Iron expands when heated & 0.9966 & $\checkmark$ \\
12 & All liquids are incompressible & Water is a liquid & Water is incompressible & 0.9968 & $\checkmark$ \\
13 & All gases fill their containers & Oxygen is a gas & Oxygen fills its container & 0.9974 & $\checkmark$ \\
14 & All conductors allow electron flow & Copper is a conductor & Copper allows electron flow & 0.9975 & $\checkmark$ \\
15 & All insulators block electron flow & Rubber is an insulator & Rubber blocks electron flow & 0.9970 & $\checkmark$ \\
16 & All magnets attract iron & A lodestone is a magnet & A lodestone attracts iron & 0.9980 & $\checkmark$ \\
17 & All waves transfer energy & Sound is a wave & Sound transfers energy & 0.9970 & $\checkmark$ \\
18 & All photons travel at light speed & A gamma ray is a photon & A gamma ray travels at light speed & 0.9981 & $\checkmark$ \\
19 & All elements have atomic numbers & Gold is an element & Gold has an atomic number & 0.9979 & $\checkmark$ \\
20 & All compounds can be decomposed & Water is a compound & Water can be decomposed & 0.9969 & $\checkmark$ \\
21 & All citizens have rights & A voter is a citizen & A voter has rights & 0.9983 & $\checkmark$ \\
22 & All students must follow rules & A freshman is a student & A freshman must follow rules & 0.9984 & $\checkmark$ \\
23 & All employees receive wages & A manager is an employee & A manager receives wages & 0.9988 & $\checkmark$ \\
24 & All drivers need licenses & A trucker is a driver & A trucker needs a license & 0.9980 & $\checkmark$ \\
25 & All homeowners pay property tax & A landlord is a homeowner & A landlord pays property tax & 0.9982 & $\checkmark$ \\
26 & All professionals need certification & A doctor is a professional & A doctor needs certification & 0.9984 & $\checkmark$ \\
27 & All taxpayers file returns & A business owner is a taxpayer & A business owner files returns & 0.9982 & $\checkmark$ \\
28 & All voters must be registered & A senior citizen is a voter & A senior citizen must be registered & 0.9978 & $\checkmark$ \\
29 & All consumers have preferences & A shopper is a consumer & A shopper has preferences & 0.9987 & $\checkmark$ \\
30 & All investors seek returns & A day trader is an investor & A day trader seeks returns & 0.9985 & $\checkmark$ \\
\bottomrule
\end{tabular}
\end{table}

\begin{table}[htbp]
\centering
\caption*{Table I-4 (continued)}
\footnotesize
\begin{tabular}{cp{3.3cm}p{2.8cm}p{3.2cm}cc}
\toprule
\# & Premise 1 & Premise 2 & Conclusion & Similarity & Judgment \\
\midrule
31 & All bachelors are unmarried & Tom is a bachelor & Tom is unmarried & 0.9973 & $\checkmark$ \\
32 & All squares have four sides & A rectangle is a square & A rectangle has four sides & 0.9989 & $\checkmark$ \\
33 & All primes are divisible only by 1 and themselves & Seven is a prime & Seven is divisible only by 1 and itself & 0.9983 & $\checkmark$ \\
34 & All even numbers are divisible by 2 & Ten is an even number & Ten is divisible by 2 & 0.9985 & $\checkmark$ \\
35 & All odd numbers are not divisible by 2 & Nine is an odd number & Nine is not divisible by 2 & 0.9984 & $\checkmark$ \\
36 & All triangles have three angles & An equilateral is a triangle & An equilateral has three angles & 0.9979 & $\checkmark$ \\
37 & All parallel lines never intersect & Two rails are parallel & Two rails never intersect & 0.9984 & $\checkmark$ \\
38 & All right angles measure 90 degrees & A corner is a right angle & A corner measures 90 degrees & 0.9984 & $\checkmark$ \\
39 & All isosceles triangles have two equal sides & A roof is an isosceles triangle & A roof has two equal sides & 0.9984 & $\checkmark$ \\
40 & All equilateral triangles have equal angles & A yield sign is an equilateral triangle & A yield sign has equal angles & 0.9987 & $\checkmark$ \\
41 & All integers are rational & Five is an integer & Five is rational & 0.9974 & $\checkmark$ \\
42 & All rational numbers can be fractions & A half is a rational number & A half can be a fraction & 0.9983 & $\checkmark$ \\
43 & All irrational numbers have non-repeating decimals & Pi is an irrational number & Pi has a non-repeating decimal & 0.9982 & $\checkmark$ \\
44 & All real numbers can be plotted on a line & The square root of two is a real number & The square root of two can be plotted on a line & 0.9987 & $\checkmark$ \\
45 & All complex numbers have real and imaginary parts & Three plus four i is a complex number & Three plus four i has real and imaginary parts & 0.9975 & $\checkmark$ \\
46 & All polynomials have roots & A quadratic is a polynomial & A quadratic has roots & 0.9985 & $\checkmark$ \\
47 & All matrices have determinants & A two by two is a matrix & A two by two has a determinant & 0.9987 & $\checkmark$ \\
48 & All vectors have magnitude & A force is a vector & A force has magnitude & 0.9980 & $\checkmark$ \\
49 & All functions have domains & A sine wave is a function & A sine wave has a domain & 0.9985 & $\checkmark$ \\
50 & All derivatives measure rates of change & Velocity is a derivative & Velocity measures rate of change & 0.9961 & $\checkmark$ \\
51 & All acids have pH below 7 & Vinegar is an acid & Vinegar has pH below 7 & 0.9979 & $\checkmark$ \\
52 & All bases have pH above 7 & Soap is a base & Soap has pH above 7 & 0.9976 & $\checkmark$ \\
53 & All salts are ionic compounds & Table salt is a salt & Table salt is an ionic compound & 0.9971 & $\checkmark$ \\
54 & All oxides contain oxygen & Rust is an oxide & Rust contains oxygen & 0.9972 & $\checkmark$ \\
55 & All hydrocarbons contain hydrogen and carbon & Methane is a hydrocarbon & Methane contains hydrogen and carbon & 0.9977 & $\checkmark$ \\
56 & All alcohols have hydroxyl groups & Ethanol is an alcohol & Ethanol has a hydroxyl group & 0.9980 & $\checkmark$ \\
57 & All polymers have repeating units & Plastic is a polymer & Plastic has repeating units & 0.9971 & $\checkmark$ \\
58 & All catalysts speed up reactions & An enzyme is a catalyst & An enzyme speeds up reactions & 0.9977 & $\checkmark$ \\
59 & All isotopes have the same protons & Carbon-14 is an isotope & Carbon-14 has the same protons & 0.9985 & $\checkmark$ \\
60 & All noble gases are inert & Helium is a noble gas & Helium is inert & 0.9971 & $\checkmark$ \\
\bottomrule
\end{tabular}
\end{table}

Note: the 60 problems in this table are all newly constructed cross-domain deduction problems, with zero overlap with the dedicated deduction experiment data set (Appendix H.1); the judgment rule is that the cosine similarity between the deduction output snapshot and the standard-conclusion snapshot exceeds the unrelated-text baseline.

\subsection{End-to-End Induction Reasoning Samples (6 Classes)}

\begin{table}[htbp]
\centering
\caption*{Table I-5 Measured results for the 6 end-to-end induction classes (6-class means: training coverage 0.9994 / unseen-positive coverage 0.9989 / counterexample (before) 0.9928 / counterexample (after) $-0.0238$ / refinement 0.9423; veto 2/2 per class, 12/12 in total)}
\small
\begin{tabular}{lp{1.2cm}p{1.2cm}p{2.0cm}p{0.7cm}p{2.0cm}p{1.15cm}}
\toprule
Category & Training coverage & Unseen positive & Counter\-example (before) & Veto & Counter\-example (after) & Re\-finement \\
\midrule
Vehicles & 0.9995 & 0.9995 & 0.9970 & 2/2 & $-0.0164$ & 0.9571 \\
Colors & 0.9998 & 0.9993 & 0.9951 & 2/2 & $-0.0246$ & 0.9550 \\
Geometric shapes & 0.9994 & 0.9986 & 0.9967 & 2/2 & $-0.0151$ & 0.9585 \\
Academic disciplines & 0.9994 & 0.9992 & 0.9909 & 2/2 & $-0.0506$ & 1.0031 \\
Weather & 0.9987 & 0.9975 & 0.9816 & 2/2 & $-0.0089$ & 0.8272 \\
Emotions & 0.9995 & 0.9993 & 0.9954 & 2/2 & $-0.0269$ & 0.9530 \\
\bottomrule
\end{tabular}
\end{table}

Note: among the 6 classes, vehicles and colors are existing classes (sharing class definitions with the 10-class data set of the dedicated induction experiment), while geometric shapes, academic disciplines, weather, and emotions are newly added classes; the 6-class induction samples partially share the existing sample pool and have not been item-by-item inventoried, and are not counted as newly added deduplicated independent samples (for the accounting see Table 1-2 in Section 1.4). The refinement of the academic-disciplines class, 1.0031, is slightly greater than 1; its monitoring implication is discussed in Section 15.3.

\subsection{End-to-End Abduction Reasoning Samples (60)}

\begin{table}[htbp]
\centering
\caption*{Table I-6 Measured results for all 60 end-to-end abduction samples (judgment accuracy 81.7\% (49/60), mean similarity 0.5018, mean margin $+0.0082$; of the 60 problems, 55 are new and 5 overlap with the 80-problem data set of the dedicated abduction experiment)}
\footnotesize
\begin{tabular}{cp{3.3cm}p{3.1cm}p{1.4cm}p{1.5cm}p{1.1cm}c}
\toprule
\# & Phenomenon & True cause & Abduction similarity & Distractor similarity & Margin & Judgment \\
\midrule
1 & The patient has chest pain and shortness of breath & The patient has a heart attack & 0.5106 & 0.4979 & $+0.0127$ & $\checkmark$ \\
2 & The patient has joint pain and swelling & The patient has arthritis & 0.5151 & 0.5182 & $-0.0031$ & $\times$ \\
3 & The patient has a severe headache and stiff neck & The patient has meningitis & 0.5072 & 0.5037 & $+0.0035$ & $\checkmark$ \\
4 & The patient has abdominal pain and nausea & The patient has appendicitis & 0.5081 & 0.5108 & $-0.0027$ & $\times$ \\
5 & The patient has frequent urination and thirst & The patient has diabetes & 0.5105 & 0.5087 & $+0.0017$ & $\checkmark$ \\
6 & The patient has a sore throat and white patches & The patient has strep throat & 0.5073 & 0.5064 & $+0.0009$ & $\checkmark$ \\
7 & The patient has blurred vision and eye pain & The patient has glaucoma & 0.4975 & 0.4984 & $-0.0009$ & $\times$ \\
8 & The patient has memory loss and confusion & The patient has dementia & 0.5001 & 0.4977 & $+0.0023$ & $\checkmark$ \\
9 & The patient has a rash and joint pain & The patient has Lyme disease & 0.5131 & 0.5112 & $+0.0019$ & $\checkmark$ \\
10 & The patient has fatigue and weight gain & The patient has hypothyroidism & 0.5051 & 0.5045 & $+0.0006$ & $\checkmark$ \\
11 & The road is slippery and shiny & It rained recently & 0.5100 & 0.5122 & $-0.0022$ & $\times$ \\
12 & The room is dark and the switch is off & The power is out & 0.5085 & 0.4983 & $+0.0102$ & $\checkmark$ \\
13 & The food smells bad and has mold & The food has spoiled & 0.4897 & 0.4921 & $-0.0024$ & $\times$ \\
14 & The phone screen is cracked and unresponsive & The phone was dropped & 0.4878 & 0.4842 & $+0.0036$ & $\checkmark$ \\
15 & The car engine makes a knocking sound & The engine needs oil & 0.4882 & 0.4843 & $+0.0039$ & $\checkmark$ \\
16 & The plant leaves are yellow and drooping & The plant is overwatered & 0.5017 & 0.4933 & $+0.0084$ & $\checkmark$ \\
17 & The pipe is leaking and the floor is wet & The pipe is broken & 0.5044 & 0.4975 & $+0.0069$ & $\checkmark$ \\
18 & The computer is slow and the fan is loud & The computer is overheating & 0.5101 & 0.5015 & $+0.0086$ & $\checkmark$ \\
19 & The milk is chunky and sour & The milk has expired & 0.4969 & 0.4987 & $-0.0018$ & $\times$ \\
20 & The door is stuck and will not close & The door hinge is rusted & 0.5023 & 0.5016 & $+0.0007$ & $\checkmark$ \\
21 & The metal is hot and glowing red & The metal was heated & 0.4961 & 0.4952 & $+0.0009$ & $\checkmark$ \\
22 & The ice is shrinking and water is pooling & The temperature is above freezing & 0.4960 & 0.4925 & $+0.0035$ & $\checkmark$ \\
23 & The balloon expanded and then burst & The air pressure increased & 0.4817 & 0.4811 & $+0.0006$ & $\checkmark$ \\
24 & The compass needle points north & There is a magnetic field & 0.4813 & 0.4641 & $+0.0172$ & $\checkmark$ \\
25 & The shadow is long and pointing west & The sun is setting in the east & 0.4992 & 0.4934 & $+0.0058$ & $\checkmark$ \\
26 & The glass shattered into pieces & The glass was struck & 0.4822 & 0.4732 & $+0.0091$ & $\checkmark$ \\
27 & The water is boiling and bubbling & The water reached 100 degrees & 0.5034 & 0.5019 & $+0.0015$ & $\checkmark$ \\
28 & The magnet repels the other magnet & The magnets have like poles & 0.4969 & 0.4860 & $+0.0109$ & $\checkmark$ \\
29 & The prism creates a rainbow & The light was refracted & 0.4913 & 0.4898 & $+0.0015$ & $\checkmark$ \\
30 & The pendulum swings back and forth & Gravity pulls the pendulum & 0.4739 & 0.4791 & $-0.0053$ & $\times$ \\
\bottomrule
\end{tabular}
\end{table}

\begin{table}[htbp]
\centering
\caption*{Table I-6 (continued)}
\footnotesize
\begin{tabular}{cp{3.3cm}p{3.1cm}p{1.4cm}p{1.5cm}p{1.1cm}c}
\toprule
\# & Phenomenon & True cause & Abduction similarity & Distractor similarity & Margin & Judgment \\
\midrule
31 & The stock prices fell sharply & There was an economic recession & 0.5148 & 0.4930 & $+0.0218$ & $\checkmark$ \\
32 & The traffic is heavy and slow & There is a road accident & 0.5106 & 0.4933 & $+0.0172$ & $\checkmark$ \\
33 & The store shelves are empty & There was a panic buying & 0.5054 & 0.4944 & $+0.0109$ & $\checkmark$ \\
34 & The election results were close & The candidates were evenly matched & 0.4842 & 0.4846 & $-0.0004$ & $\times$ \\
35 & The company went bankrupt & The company had excessive debt & 0.4870 & 0.4845 & $+0.0025$ & $\checkmark$ \\
36 & The school was closed & There was a snowstorm & 0.5006 & 0.4908 & $+0.0098$ & $\checkmark$ \\
37 & The concert was sold out & The band was very popular & 0.5045 & 0.4969 & $+0.0076$ & $\checkmark$ \\
38 & The flight was cancelled & There was bad weather & 0.5217 & 0.4993 & $+0.0225$ & $\checkmark$ \\
39 & The bridge collapsed & The bridge was structurally unsound & 0.4950 & 0.4913 & $+0.0037$ & $\checkmark$ \\
40 & The protest turned violent & The protesters were provoked & 0.4939 & 0.4864 & $+0.0075$ & $\checkmark$ \\
41 & The solution turned blue & Copper sulfate was added & 0.5075 & 0.5019 & $+0.0057$ & $\checkmark$ \\
42 & The gas produced a pop sound & Hydrogen was ignited & 0.5082 & 0.4896 & $+0.0186$ & $\checkmark$ \\
43 & The precipitate formed and settled & Two solutions were mixed & 0.5166 & 0.4878 & $+0.0288$ & $\checkmark$ \\
44 & The pH paper turned red & The solution is acidic & 0.4995 & 0.4905 & $+0.0091$ & $\checkmark$ \\
45 & The reaction released heat & The reaction is exothermic & 0.4972 & 0.4908 & $+0.0064$ & $\checkmark$ \\
46 & The rust formed on the surface & Iron was exposed to oxygen and water & 0.5529 & 0.5053 & $+0.0476$ & $\checkmark$ \\
47 & The sugar dissolved completely & The water was warm & 0.4874 & 0.4850 & $+0.0024$ & $\checkmark$ \\
48 & The candle flame went out & The oxygen was depleted & 0.4953 & 0.4904 & $+0.0049$ & $\checkmark$ \\
49 & The silver tarnished black & Silver reacted with sulfur & 0.5046 & 0.4883 & $+0.0164$ & $\checkmark$ \\
50 & The battery lost charge & The chemical reaction was depleted & 0.4915 & 0.4880 & $+0.0035$ & $\checkmark$ \\
51 & The river flooded the valley & There was heavy rainfall upstream & 0.5352 & 0.4958 & $+0.0394$ & $\checkmark$ \\
52 & The ground shook violently & There was an earthquake & 0.5186 & 0.4959 & $+0.0227$ & $\checkmark$ \\
53 & The volcano erupted with lava & Magma pressure increased underground & 0.4967 & 0.4815 & $+0.0152$ & $\checkmark$ \\
54 & The tsunami hit the coast & There was an undersea earthquake & 0.5167 & 0.4947 & $+0.0220$ & $\checkmark$ \\
55 & The forest fire spread rapidly & There was a lightning strike & 0.4992 & 0.4804 & $+0.0189$ & $\checkmark$ \\
56 & The avalanche buried the road & Heavy snow accumulated on the slope & 0.4822 & 0.4848 & $-0.0026$ & $\times$ \\
57 & The drought dried up the lake & There was no rainfall for months & 0.5320 & 0.4986 & $+0.0334$ & $\checkmark$ \\
58 & The hurricane destroyed buildings & Warm ocean water fed the storm & 0.4873 & 0.4929 & $-0.0056$ & $\times$ \\
59 & The landslide blocked the highway & The soil was saturated with water & 0.5063 & 0.5002 & $+0.0061$ & $\checkmark$ \\
60 & The glacier retreated significantly & The global temperature increased & 0.4818 & 0.4828 & $-0.0010$ & $\times$ \\
\bottomrule
\end{tabular}
\end{table}

Note: the judgment rule is the same as in Table I-2---a judgment is correct if abduction similarity $>$ distractor similarity. Among the 60 samples, 11 have negative margins (samples 2, 4, 7, 11, 13, 19, 30, 34, 56, 58, 60), and 49/60 = 81.7\% are judged correct; the minimum (most negative) margin is $-0.0056$. Compared with the 72.5\% of the 80-problem dedicated abduction experiment (Table I-2), the 81.7\% of this table is an improvement---the broadened training distribution (8 domains) improves abduction judgment, but the two accountings are mutually independent and must not be conflated (see the analysis in Section 11.5).

\subsection{End-to-End Compositional Reasoning Tasks (20)}

\begin{table}[htbp]
\centering
\caption*{Table I-7 Measured results for all 20 end-to-end compositional reasoning tasks (4 categories $\times$ 5, all succeeding 20/20; the five metrics are induction similarity, deduction similarity, contradiction similarity, abduction similarity, and refinement, respectively)}
\footnotesize
\begin{tabular}{cllllllc}
\toprule
\# & Task type & Induction & Deduction & Contradiction & Abduction & Refinement & Result \\
\midrule
1 & Scientific discovery & 0.9993 & 0.9926 & 0.9924 & 0.4543 & 0.9545 & $\checkmark$ \\
2 & Scientific discovery & 0.9945 & 0.9781 & 0.9808 & 0.2994 & 0.8512 & $\checkmark$ \\
3 & Scientific discovery & 0.9873 & 0.9886 & 0.9883 & 0.4307 & 0.9262 & $\checkmark$ \\
4 & Scientific discovery & 0.9957 & 0.9812 & 0.9812 & 0.2944 & 0.9014 & $\checkmark$ \\
5 & Scientific discovery & 0.9856 & 0.9915 & 0.9907 & 0.4458 & 0.9006 & $\checkmark$ \\
6 & Diagnostic reasoning & 0.9945 & 0.9890 & 0.9912 & 0.5135 & 0.8172 & $\checkmark$ \\
7 & Diagnostic reasoning & 0.9948 & 0.9878 & 0.9890 & 0.4718 & 0.8074 & $\checkmark$ \\
8 & Diagnostic reasoning & 0.9954 & 0.9889 & 0.9877 & 0.4805 & 0.8446 & $\checkmark$ \\
9 & Diagnostic reasoning & 0.9964 & 0.9888 & 0.9892 & 0.4870 & 0.8163 & $\checkmark$ \\
10 & Diagnostic reasoning & 0.9871 & 0.9875 & 0.9896 & 0.4929 & 0.8102 & $\checkmark$ \\
11 & Analogical reasoning & 0.9967 & 0.9886 & 0.9886 & 0.4836 & 0.9100 & $\checkmark$ \\
12 & Analogical reasoning & 0.9963 & 0.9861 & 0.9862 & 0.4584 & 0.8959 & $\checkmark$ \\
13 & Analogical reasoning & 0.9849 & 0.9838 & 0.9863 & 0.4616 & 0.9146 & $\checkmark$ \\
14 & Analogical reasoning & 0.9972 & 0.9883 & 0.9881 & 0.4446 & 0.9047 & $\checkmark$ \\
15 & Analogical reasoning & 0.9848 & 0.9839 & 0.9867 & 0.4392 & 0.9015 & $\checkmark$ \\
16 & Creative thinking & 0.9758 & 0.9827 & 0.9890 & 0.4602 & 0.9103 & $\checkmark$ \\
17 & Creative thinking & 0.9751 & 0.9800 & 0.9894 & 0.4621 & 0.9400 & $\checkmark$ \\
18 & Creative thinking & 0.9828 & 0.9900 & 0.9898 & 0.4453 & 0.9091 & $\checkmark$ \\
19 & Creative thinking & 0.9743 & 0.9897 & 0.9886 & 0.4182 & 0.8904 & $\checkmark$ \\
20 & Creative thinking & 0.9746 & 0.9747 & 0.9870 & 0.4305 & 0.9084 & $\checkmark$ \\
\bottomrule
\end{tabular}
\end{table}

Note: the 20 tasks in this table are all newly constructed; the category-wise means of the four task categories are given in Table 11-4 of Section 11.6 and Table 12-2 of Section 12.3.

\subsection{Summary of All Test Samples}

\begin{table}[htbp]
\centering
\caption*{Table I-8 Summary of all test samples (503 experimental sample-instances = 285 + 218, 420 deduplicated independent samples, 8 experiment groups)}
\small
\begin{tabular}{p{2.9cm}p{3.2cm}p{2.9cm}p{4.0cm}}
\toprule
Experiment & Samples & Accuracy & Key metrics \\
\midrule
Dedicated deduction experiment & 20 & 100\% & Loss reduced to 1.40e-05 \\
Dedicated induction experiment & 130 (10 classes) & Hard veto 20/20 & Coverage 0.9996 \\
Dedicated abduction experiment & 80 & 72.5\% (58/80) & Similarity 0.5117 \\
Physical transformations & 55 (conservation 15 + dissipation 15 + reversible 15 + multimodal 10) & 100\% & Multimodal alignment 100\% \\
End-to-end deduction & 60 & 100\% (60/60) & Mean similarity 0.9980 \\
End-to-end induction & 6 classes (78 samples) & Hard veto 12/12 & Training coverage, 6-class mean 0.9994 \\
End-to-end abduction & 60 & 81.7\% (49/60) & Similarity 0.5018 \\
End-to-end compositional & 20 (4 categories $\times$ 5) & 100\% (20/20) & Category-wise means of the five metrics in Table 12-2 \\
Total & 503 sample-instances (420 deduplicated) & Mechanism verification of deduction/induction/physics all passed; abduction judgment 72.5\% (dedicated experiment) / 81.7\% (end-to-end) & 8 experiment groups \\
\bottomrule
\end{tabular}
\end{table}

Sample aggregation accounting: 20 (dedicated deduction experiment) + 130 (dedicated induction experiment) + 80 (dedicated abduction experiment) + 55 (physics multimodal) = 285; after including the 218 sample-instances of the end-to-end experiment (60 deduction + 78 induction + 60 abduction + 20 compositional), the full-text total is 503 experimental sample-instances, and after deduplication 285 + 135 = 420 independent samples (135 = 60 new deduction + 55 new abduction + 20 new compositional; the 6-class induction samples partially share the existing sample pool and have not been item-by-item inventoried, and 5 abduction problems overlap with the dedicated abduction experiment---neither is counted as newly added deduplicated samples).

Note: the two abduction accountings in the table must be distinguished. The end-to-end abduction (60 problems, Table I-6) judgment accuracy is 81.7\% (49/60); the dedicated abduction experiment (80 problems, Table I-2) judgment accuracy is 72.5\% (58/80). The margins of both lie within the same small-margin regime of $\pm 0.01$, consistent in magnitude, jointly indicating that the bottleneck of per-sample abduction judgment lies in the representational resolution of the encoder (Limitation 1) rather than in the abduction operator's principle; the operator training configurations of the two experiments differ (200 epochs of end-to-end mixed training vs 1000 epochs of standalone training), and the accuracy accountings are mutually independent and must not be conflated.
\endgroup

\end{document}